\def\degree{Doctor of Philosophy}
\def\department{School of Computer Science \\ Faculty of Engineering}
\def\mydegrees{}

\def\supervisor{Assoc. Prof. Tongliang Liu \\ Auxiliary Supervisor: Asst. Prof. Baosheng Yu}

\documentclass[logo]{style/usydthesis}

\newcommand*{\noaddvspace}{\renewcommand*{\addvspace}[1]{}}
\addtocontents{lof}{\protect\noaddvspace}
\usepackage{amstext,latexsym}
\usepackage[utf8]{inputenc}
\usepackage{style/bib}
\usepackage[UKenglish]{babel}
\usepackage{csquotes}
\usepackage{ccicons}
\usepackage{animate}
\usepackage{setspace}
\usepackage{pdfpages}

\usepackage{mathtools}
\usepackage{booktabs}
\usepackage{multirow}
\usepackage{wrapfig}
\usepackage{subcaption}
\usepackage{tabularx}
\usepackage{bm}
\newcommand{\indep}{\perp \!\!\! \perp}
\usepackage{color}
\usepackage{algorithm}
\usepackage{algorithmic}
\usepackage{float}
\usepackage[table]{xcolor}
\usepackage{mathrsfs}
\usepackage{amsmath}
\usepackage{amsfonts}
\usepackage{amsthm}
\usepackage{amssymb}
\usepackage{graphicx}
\usepackage{array}
\usepackage{bbm}
\usepackage{placeins}
\usepackage{tcolorbox}
\usepackage{adjustbox}
\usepackage{bbding}
\usepackage{prettyref}
\newtheorem{theorem}{Theorem}
\newtheorem{proposition}[theorem]{Proposition}

\newtheorem{remark}[theorem]{Remark}
\newtheorem{definition}[theorem]{Definition}

\newtheorem{assumption}[theorem]{Assumption}
\makeatletter
\newcommand\notsotiny{\@setfontsize\notsotiny\@vipt\@viipt}
\makeatother
\DeclareMathOperator{\sgn}{sgn}
\DeclareMathOperator{\Err}{Err}
\def\code#1{\texttt{#1}}
\newcommand{\sign}{\text{sign}}
\usepackage[pdftex,bookmarks=true]{hyperref}

\hypersetup{
    pdfauthor = {Zhuo Huang},
    pdftitle = {Trustworthy Machine Learning under Distribution Shifts},
    colorlinks,
    linkcolor={black},
    citecolor={black},
    urlcolor={black}
    }

\usepackage[style=numeric,
            sorting=none,
            backend=biber,
            uniquename=false,
            uniquelist=false,
            maxcitenames=2,
            hyperref=true]{biblatex}
\begin{document}

\renewcommand{\thepage}{\roman{page}}	
\title{{\bf\Huge Trustworthy Machine Learning under Distribution Shifts}}
\author{Zhuo Huang}

\maketitle
\setstretch{1.5}


\chapter*{Abstract}

Machine Learning (ML) has been a foundational topic in artificial intelligence (AI), providing both theoretical groundwork and practical tools for its exciting advancements. From ResNet for visual recognition to Transformer for vision-language alignment, the AI models have achieved superior capability to humans. Furthermore, the scaling law has enabled AI to initially develop general intelligence, as demonstrated by Large Language Models (LLMs). To this stage, AI has had an enormous influence on society and yet still keeps shaping the future for humanity.

However, distribution shift remains a persistent ``Achilles' heel'', fundamentally limiting the reliability and general usefulness of ML systems. As AI becomes increasingly integrated into real-world decision-making and societal infrastructures, the complexity of the problems we ask it to solve continues to grow. These complex environments naturally introduce diverse and unpredictable distribution shifts, which can severely degrade model performance.

Moreover, generalization under distribution shift would also cause trust issues for AIs. For instance, when employing medical AIs across regions, they might perform unsatisfactorily and cause harm. Thus, we also consider the responsibility of AI, i.e., the Trustworthiness of ML, aiming to enhance reliability rather than merely focusing on accuracy.

Motivated by these challenges, my research focuses on \textit{Trustworthy Machine Learning under Distribution Shifts}, with the goal of expanding AI's robustness, versatility, as well as its responsibility and reliability. We carefully study the three common distribution shifts into: (1) Perturbation Shift, (2) Domain Shift, and (3) Modality Shift. For all scenarios, we also rigorously investigate trustworthiness via three aspects: (1) Robustness, (2) Explainability, and (3) Adaptability. Based on these dimensions, we propose effective solutions and fundamental insights, meanwhile aiming to enhance the critical ML problems, such as efficiency, adaptability, and safety.
\chapter*{Statement of Originality} 

I certify that the work in this thesis has not previously been submitted
for a degree, nor has it been submitted as part of the requirements for a
degree except as fully acknowledged within the text.

I also certify that the thesis has been written by me. Any help that I
have received in my research work and the preparation of the thesis
itself has been acknowledged. In addition, I certify that all information sources and literature used are indicated in the thesis.

\vspace{2.0cm}

\noindent Zhuo Huang \\
\noindent School of Computer Science \\
\noindent Faculty of Engineering \\
\noindent The University of Sydney 
\hfill 30 Dec 2025

\chapter*{Statement of Generative AI}

During the preparation of this thesis, ChatGPT (OpenAI) was used for the purposes of text enhancement. The use of this generative AI tool includes minor paraphrasing, sentence restructuring, and spelling correction in several draft chapters. I confirm that where text was modified by generative AI, the content was reviewed for possible errors, inaccuracies, and bias. The author takes full responsibility for the submitted thesis and ensures the work is their own and has used generative AI within the intended purpose of use.

\vspace{2.0cm}

\noindent Zhuo Huang \\
\noindent School of Computer Science \\
\noindent Faculty of Engineering \\
\noindent The University of Sydney 
\hfill 30 Dec 2025

\chapter*{Authorship Attribution Statement} 
This thesis was conducted at The University of Sydney, under the supervision of Assoc. Prof. Tongliang Liu, between 2022 and 2025. The main results presented in this dissertation were first introduced in the following publications:
\begin{itemize}
    
    \item[(1)] \textbf{Zhuo Huang}, Xiaobo Xia, Li Shen, Bo Han, Mingming Gong, Chen Gong, Tongliang Liu. ``Harnessing Out-Of-Distribution Examples via Augmenting Content and Style''. In \textit{The Eleventh International Conference on Learning Representations (ICLR)}, 2023. Presented in Chapter~\ref{cha:HOOD}. I identified the research problem, formulated setting, implement the experiments, and drafted the manuscript. The polish of the paper is helped by other co-authors.
    
    \item[(2)] \textbf{Zhuo Huang}, Miaoxi Zhu, Xiaobo Xia, Li Shen, Jun Yu, Chen Gong, Bo Han, Bo Du, Tonglinag Liu ``Robust Generalization against Photon-Limited Corruptions via Worst-Case Sharpness Minimization''. \textit{Conference on Computer Vision and Pattern Recognition (CVPR)}, 2023. Presented in Chapter~\ref{cha:SharpDRO}. I identified the research problem, formulated setting, implement the experiments, and drafted the manuscript. The theoretical formulation and polish of the paper are helped by other co-authors.
    
    \item[(3)] \textbf{Zhuo Huang}, Muyang Li, Li Shen, Jun Yu, Chen Gong, Bo Han, Tongliang Liu. ``Winning Prize Comes from Losing Tickets: Improve Invariant Learning by Exploring Variant Parameters for Out-of-Distribution Generalization''. In \textit{International Journal of Computer Vision (IJCV)}, 2024. Presented in Chapter~\ref{cha:EVIL}. I identified the research problem, formulated setting, implement the experiments, and drafted the manuscript. The polish of the paper is helped by other co-authors.
    
    \item[(4)] \textbf{Zhuo Huang}, Chang Liu, Yinpeng Dong, Hang Su, Shibao Zheng, Tongliang Liu. ``Machine Vision Therapy: Multimodal Large Language Models Can Enhance Visual Robustness via Denoising In-Context Learning''. In \textit{International Conference on Machine Learning (ICML)}, 2024. Presented in Chapter~\ref{cha:MVT}. I identified the research problem, formulated setting, implement the experiments, and drafted the manuscript. The polish of the paper and some experiments are helped by other co-authors.
    
    \item[(5)] \textbf{Zhuo Huang}, Gang Niu, Bo Han, Masashi Sugiyama, Tongliang Liu. ``Towards Out-of-Modal Generalization without Instance-level Modal Correspondence''. In \textit{The Thirteenth International Conference on Learning Representations (ICLR)}, 2025. Presented in Chapter~\ref{cha:OOM}. I identified the research problem, formulated setting, implement the experiments, and drafted the manuscript. The polish of the paper and problem setting are helped by other co-authors.
\end{itemize}

\noindent In addition to the statements above, in cases where I am not the corresponding author of a published item, permission to include the published material has been granted by the corresponding author.

\vspace{1cm}
\noindent Zhuo Huang \\
\noindent School of Computer Science \\
\noindent Faculty of Engineering \\
\noindent The University of Sydney 
\hfill 30 Dec 2025

\vspace{1cm}
\noindent As supervisor for the candidature upon which this thesis is based, I can confirm that the authorship attribution statements above are correct.

\vspace{1cm}
\noindent Tongliang Liu \\
\noindent School of Computer Science \\
\noindent Faculty of Engineering \\
\noindent The University of Sydney 
\hfill 30 Dec 2025

\newpage
\thispagestyle{empty}
\vspace*{\fill}
\begin{center}
\textit{Sidere mens eadem mutato.}
\end{center}
\vspace*{\fill}


\chapter*{Acknowledgements}

During my PhD study, there are many people has been a great help to me.

First of all, I would like to thank my PhD supervisor, Prof. Tongliang Liu, who has been very supportive in every part of this thesis. I feel very grateful for his guidance and consistent patience. I can focus on my research and freely discover what interests me has been a great blessing, largely owes to him. I will always remember his wisdom and insights on critical thinking and even life philosophy. His calm personality has largely shaped my attitude towards ups and downs in PhD and life. I sincerely express my appreciation to thank him for everything.

I also wish to express my deep appreciation to Prof. Masashi Sugiyama and Dr. Gang Niu, who have been very supportive when I was visiting RIKEN AIP in Tokyo, and I enjoy every inspiring conversion we have. I also want to thank Prof. Chen Gong, A/Prof. Mingming Gong, A/Prof. Bo Han, and A/Prof. Li Shen, who have been my encouraging mentors and collaborators. I also wish to thank A/Prof. Hang Su, Dr. Yinpeng Dong, Dr. Chang Liu, and Dr. Zijian Zhu for helping me when I was visiting Tsinghua University. I also want to thank some of the brilliant minds that I have learned from: Prof. Kun Zhang, Prof. Javen Qinfeng Shi, Prof. Lina Yao, Dr. Dadong Wang, A/Prof. Shuo Chen, A/Prof. Takashi Ishida, Dr. Feng Liu, Dr. Dong Gong, and Dr. Zhen Fang. Their encouragement and advice have been a priceless treasure during my research life, I wish to express my appreciation to them all.

Moreover, I would like to thank my precious collaborators at TML group: Dr. Yu Yao, Dr. Xiaobo Xia, Dr. Songhua Wu, Dr. Huaxi Huang, Dr. Yingbin Bai, Dr. Dawei Zhou, Dr. Chaojian Yu, Dr. Runnan Chen, Dr. Liangcheng Liu, Dr. Yuhao Wu, Dr. Cong Lei, Dr. Runqi Lin, Dr. Vincent Qu, Jialiang Shen, Zhaoqing Wang, Suqin Yuan, Jiyang Zheng, Yexiong Lin, Muyang Li, Xiuchuan Li, Ziming Hong, Yongli Xiang, Li He, Jiaxin Huang, Tianyu Huang, Xiangyu Sun, Quan Tran, Zhen Huang, Andrew Cao, Longjie Zhao, Peng Mi, Zhenchen Wan, Jun Wang, Keshen Zhou, and Yuxiang Zhen. They are extraordinary people who made my PhD life unforgettable. I show my gratitude to them from the bottom of my heart.

Further, I want thank my dear friends: Dr. Qizhou Wang, Dr. Weijian Deng, Weijie Tu, Chengyi Cai, Jianing Zhu, Puning Yang, Xiaohao Liu, Jiacheng Zhang, Zhongyi Bai, Xiaotong Yu, Hongyu Zhou, Tinghui Li, Jiakun Yu, Dr. Andong Wang, Dr. Zhen-Yu Zhang, Prof. Ximing Li, Dr. Ming-Kun Xie, Dr. Xin-Qiang Cai, Dr. Wei Wang, Dr. Yuning Qiu, Dr. Haonan Huang, Dr. Mingyuan Bai, Hanlue Zhang, Zhengqing Gao, Ziwen Li, Dr. Lin Li, Haoyu Wang, and Dr. Jihyo Kim. Their kindness and care have accompanied me through this journey, and I could not express my appreciation enough to them.

Last but not least, I want to express my deepest feelings to my parents, Yucheng Huang and Hongying Yang. They are the most important people in my life, who are constantly there for me and encourage me to step forward. No words can show my love and gratitude for them.

I dedicate this thesis to them.

\setcounter{tocdepth}{2}
\newpage
\addcontentsline{toc}{chapter}{Contents}
\tableofcontents
\listoffigures


\phantomsection
\listoftables


\setcounter{page}{1}
\setcounter{chapter}{0}

\renewcommand{\thepage}{\arabic{page}}	
\setupParagraphs

\chapter{Introduction}

\label{cha:introduction}

\section{Background}

\label{introduction:background}

Machine learning (ML) is one of the most effective approaches to Artificial Intelligence (AI), which leverages statistical tools to find, recognize, and utilize patterns in data, it has provided theoretical ground and practical tools that have led to the flourishing of AI. As a result, it has achieved tremendous success in a wide spectrum of applications, such as computer vision, natural language processing, speech recognition, autonomous driving, and robotics, significantly influencing human civilization.


The early AI revolution is led by deep neural networks~\cite{lecun2015deep}, such as AlexNet~\cite{krizhevsky2012imagenet}, ResNet~\cite{he2016deep}, and DenseNet~\cite{huang2017densely}, which have demonstrated that AI can achieve superior performance to human intelligence in visual recognition tasks. In 2017, Google introduced the Transformer architecture~\cite{vaswani2017attention}, which enabled models to handle vast amounts of text~\cite{devlin2019bert} as well as visual tokens~\cite{dosovitskiy2021vit} in parallel, thereby scaling up the magnitude of information processing. Based on such an architecture, it has become a common understanding that simply expanding the learning scale can always achieve improved performance, also known as the \textit{Scaling Law}~\cite{kaplan2020scaling}. As a result, Large Language Models (LLMs) such as GPT~\cite{brown2020gpt3}, Llama~\cite{meta2024llama3}, and Qwen~\cite{bai2023qwen} have been scaled-up to Billions of parameters, along with many emergent capabilities including In-Context Learning, Chain-of-Thought Reasoning, and Arithmetic Operation. Beyond language-only models, Multi-modal LLMs (MLLMs) that combine language and vision to allow advanced human-computer interaction have significantly increased their real-world capabilities, such as ChatGPT~\cite{openai2022chatgpt}, Gemini~\cite{geminiteam2023gemini}, Claude~\cite{anthropic2024claude3}, and DeepSeek~\cite{deepseek2024llm}. Until now, most commercialized AI models are MLLMs and they play a vital role in boosting efficiency and productivity of daily tasks, such as document organization, software development, content creation and design.


Despite its success and real-world impact, existing learning models still heavily rely on the independent and identically distributed (IID) assumption. Particularly, ML aims to learn from training data, which is collected to simulate the expected real-world environment. Based on the Empirical Risk Minimization (ERM) framework~\cite{devroye1996probabilistic, vapnik1998statistical}, ML is guaranteed to generalize when tested under a similar data distribution to the training distribution. Therefore, traditional ML models such as Support Vector Machine (SVM)~\cite{vapnik1963pattern} and Multi-Layer Perceptron (MLP)~\cite{rumelhart1986learning} can work satisfactorily under simple scenarios such as fraud detection and spam filtering.

However, due to domain shift, dataset bias, or evolving environments in real-world scenarios, the IID assumption is constantly violated, introducing Out-of-Distribution (OOD) data that hinders the effectiveness of ML. For example, autonomous vehicles are trained on thousands of hours of videos from sunny, dry, and daytime conditions, when it is deployed at rainy, wet, and nighttime streets, they might have serious performance malfunctioning; a model trained in 2022 is asked who is Prime Minister of UK, but it won't give the correct answer because the answer is changed over time. Therefore, when the distribution of test dataset is shifted from the training dataset, it introduces significant complexity and variance to the learning process. Thus, the effectiveness of many existing ML methods is suboptimal in practice, which raises serious concerns on their capability under realistic distribution shift.

Apart from the generalization capability of AI, it is also crucial to enhance its responsibility. To achieve this, Trustworthy Machine Learning (TML)~\cite{vashney2022trustworthy} was proposed to focus on reliability and integrity of AI systems rather than focusing primarily on accuracy, which has attracted abundant attention from both academy and industry. For example, OpenAI pioneered AI Alignment for Human and proposed Reinforcement Learning from Human Feedback (RLHF)~\cite{christiano2017deep} which aims to ensure the value and morality of AIs are well aligned with human; moreover, institutions like UC Berkeley, NYU, and Mila launched MATS~\cite{betley2025emergent} program aiming to prevent the possibility of models accidentally learning dangerous behaviors in tasks, such as writing code and generating images.

In the era of LLMs and the dawn of general intelligence, it is urgent to consider how we can live with AI and how AI can be aligned with us. Otherwise, as it gets embodied into human society, the consequences of untrustworthy AIs would be devastating, causing irreparable economic losses and social disaster. In particular, AIs that provide customer services might be "jailbroken" by malicious users and be forced to make improper deals, leading to massive monetary cost; moreover, when LLMs scrape numerous web data for pre-training, it might accidentally collect user-sensitive data and violate the privacy protocol, posing severe safety and security concerns.

However, given the significance of the above two fundamental aspects, there lack of a systematic study to address them simultaneously. In fact, the capability for AI under distribution shifts and its responsibility for trustworthiness can conflict with each other. Intuitively, enhancing generalization beyong IID data enforces AI to greedily obsorbing unknown knowledge and autonomously handle uncertainty, e.g., for treatment works for over 50 men in a trial, over-generalized AI might assert that this treatment is effective. As a result, such an excessive agency would eventually reach domains that are misaligned with human values, thus damaging the reliability of AI. On the other hand, heavy penalization to achieve strict trustworthiness would make AI overcautious, which could fail to perform under challenging conditions, e.g., a self-driving car is trained for millions of miles with almost zero risk, it could just stop and stay stationary when it is deployed on snowy mountain paths because it detects danger in making any action. Therefore, trading off between capability and responsibility will continue to be one critical challenge in the future of AI.

\begin{figure*}[t]

    \includegraphics[width=0.8\linewidth]{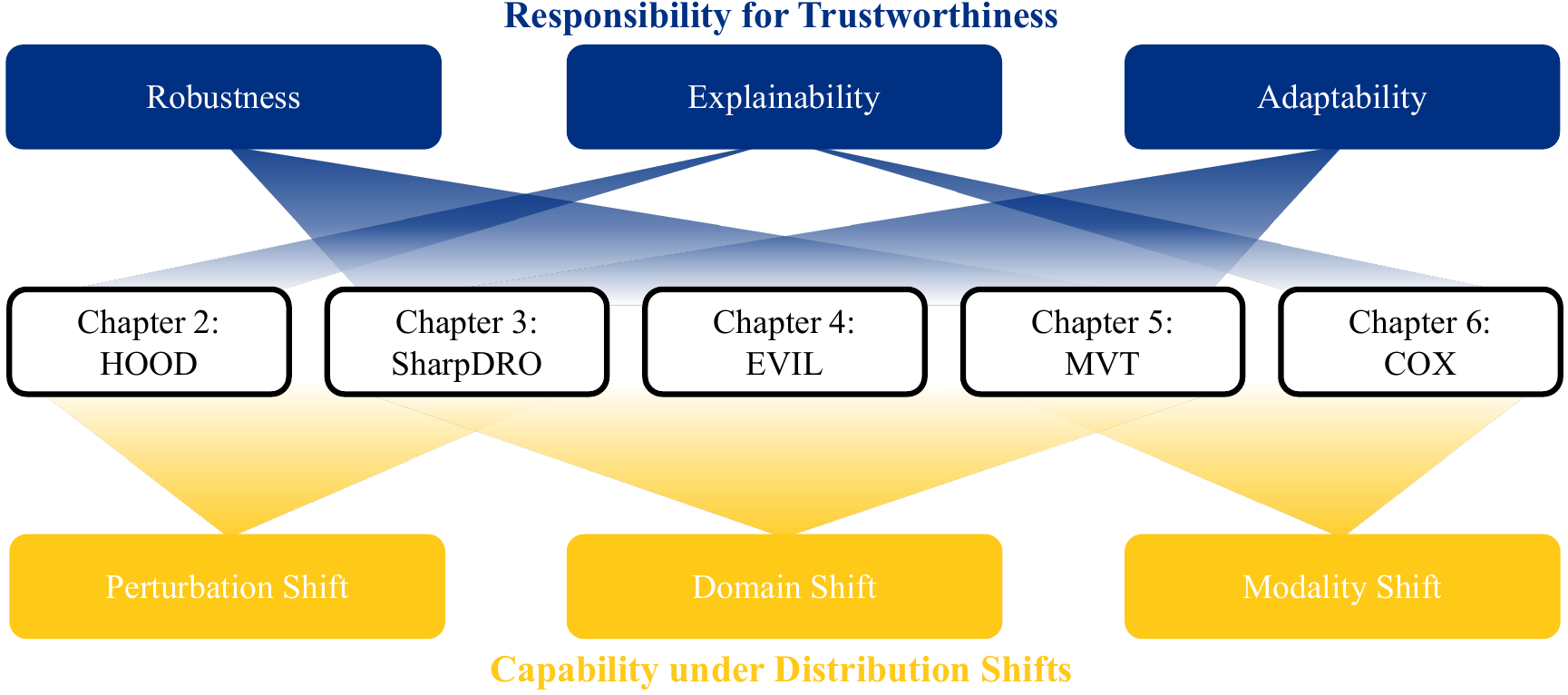}

    \caption{Outline for trustworthy machine learning under distribution shifts. We explore Robustness, Explainability, and Adaptability for trustworthiness of ML models, and consider three types of common distribution shifts, including Perturbation Shift, Domain Shift, and Modality Shift. From Chapter 2 to Chapter 6, we shed light on different trustworthy aspects and distribution shift scenarios by solving realistic tasks and proposing effective approaches with theoretical grounding.}

    \label{intro:fig:outline}

\end{figure*}

In this thesis, motivated by the goal of developing general intelligence for the best human benefits, we carefully investigate trustworthy machine learning under realistic distribution shift problems, aiming to find solutions to ensure capability and responsibility simultaneously. To achieve this, we explore three perspectives for both distribution shift and trustworthiness, as shown in Figure \ref{intro:fig:outline}. Particularly, to investigate responsibility, we consider:

\begin{itemize}

    \item Robustness, which demonstrates the ability of AI to maintain stable and effective performance under challenging scenarios.

    \item Explainability, which aims to understand the nature of the distribution shift and make the decision-making process interpretable.

    \item Adaptability, which denotes the ability of AI to perform autonomously across various conditions without human intervention.

\end{itemize}

These three aspects of TML are highly vulnerable or essential for studying generalization under distribution shifts; thus, we focus on them for this thesis. Note that there are other topics such as fairness and privacy in TML, but they align orthogonal from generalization under distribution shift. The success can be achieved by combining their findings, such as reweighting~\cite{lahoti2020fairness} or unlearning~\cite{bourtoule2021machine}. Moreover, to evaluate the generalization capability under distribution shifts, we consider:

\begin{itemize}

    \item Perturbation Shift, where natural or synthetic noise is applied to the dataset, largely deviating its statistical patterns.

    \item Domain Shift, where data are collected from different environments, further introducing confounding factors that significantly mislead the prediction.

    \item Modality Shift, where the knowledge is represented in a heterogeneous structure, intensifying the feature incompatibility and hindering the learning feasibility.

\end{itemize}

The above distribution shifts demonstrate different intensities of shifts. Perturbation shift is the mild shift, as it has the same data type from the same environment; domain shift is moderate because it is caused by different environmental factors; and modality shift is the most extreme one because it has a different data type, implying a change of input space. Yet, they are ubiquitous and all belong to the Covariate Shift~\cite{sugiyama2007covariate} in statistical ML. There are other less common shifts, including Target Shift~\cite{zhang2013domain} and Concept Shift~\cite{vorburger2006entropy}. Specifically, target shift is caused by the change of label distribution, but given the simpicity of the label space in most applications, it can be easily solved via reweighting or resampling; concept shift is less-studied due to that concepts and knowledge in real-world applications are generally stationary. Thus, they are excluded from the scope of this thesis. Next, we will summarize the contributions of this thesis and demonstrate the structure by chapter.

\section{Summary of Contributions}

\label{introduction:summary}

To sum up, this thesis focuses on Trustworthy Machine Learning under Distribution Shift, aiming to explore responsibility and capability simultaneously under practical scenarios. By considering various levels of shift, my studies cover a wide spectrum of applications and provide insights for traditional topics as well as futuristic directions. Additionally, we provide extensive analyses and theoretical groundings to rigorously justify my research under extensive datasets, models, and settings, validating their trustworthiness along with many other notable benefits, such as efficiency, safety, and affordability. Below, we reveal the logical structure of the following chapters and emphasize their contributions.

In Chapter 2, we observe that OOD data present a double-edged effect that could either enhance or harm generalization in different scenarios. Therefore, we aim to understand such an effect and harness OOD data effectively for various scenarios. We inspect from a causal inference perspective to explain the data generation process, which reveals that image data can be decomposed into ``content'' and ``style''. Beneficial OOD data is caused by a style change, and harmful OOD data is caused by a content change. We propose Understanding and Harnessing OOD data (HOOD) framework that disentangles content and style during inference, then separately creates beneficial and harmful OOD data using learnable adversarial perturbations. By training on such tailor-designed examples, we can enhance the generalization performance and properly handle OOD data in the wild.

Distribution shifts are not only presented in different types, but they also appear in various strengths, forming a traceable distribution. In Chapter 3, we study learning under corruptions that follow real-world noise distribution. Intuitively, corruption is caused by multiple noise elements which occur with a certain probability during a time interval, knwon as ``photon-limited corruptions''. By modeling it via Poisson distirbution, the strengths levels of the corruption can be identified. To deal with such a multi-strength distribution shift, we propose Sharpness-based Distribution Robust Optimization (SharpDRO) to reweight each distirbution and enhance its generalization by encouraging the loss smoothness, i.e., minimizing ``sharpness''. As a result, my approaches perform effectively and stably across various corruption strengths, demonstrating robustness and adaptability in practical applications.

Further, to deploy a TML model in the real world, it needs to autonomously identify the essential features without latching onto the confounding information. In Chapter 4, we explore neural regeneration where AIs can dynamically adapt across different environments and extract domain invariant knowledge. To solve this problem, we propose Exploring Variant parameters for Invariant Learning (EVIL) framework, which identifies domain invariant features by growing corresponding neural paths. Moreover, we enhance such a process by further incorporating domain variant knowledge simultaneously during training, which in turn helps identify the variant parameters. By dropping such undesirable neural paths, models can be less affected by domain shift, thus showing enhanced capability and adaptability. Additionally, the proposed approach is conducted via sparse training, thus it also demonstrates great efficiency.

Existing distribution shift studies are mainly focused on perturbations and domain shifts. However, knowledge adaptation across different modalities is rarely explored. Therefore, in Chapter 5, we make an initial attempt at generalization across the two most common modalities, i.e., language and vision. Learning from language has been very successful thanks to the development of LLMs, which already demonstrate general-purpose capabilities. For vision tasks, it is highly dependent on human supervision, which is very resource-consuming. Therefore, a question is asked: Is it possible to replace human supervision by AI supervision? Hence, we propose Machine Vision Therapy (MVT) framework that leverages LLMs to guide the visual learning process. Due to the modality gap between language models and vision models, we propose to conduct In-Context Learning (ICL) to analyze visual tokens in textual contexts. As a result, the proposed MVT can largely boost the visual robustness, successfully adapting knowledge across modalities.

In Chapter 6, instead of focusing on generalization across common modalities, we propose a novel problem named Out-of-Modal (OOM) Generalization, which aims to leverage well-known modal knowledge to infer rare modalities. For example, language and vision modalities are quite extensive, but rare modalities, such as tactile, LiDAR, and genomics, are not well-explored by AIs. Due to the difficulty and cost of collecting and organizing such data, it is extremely challenging to conduct large-scale training. Therefore, exploiting the hidden knowledge from rare modalities based on the existing ones is essential for general intelligence. In OOM generalization, we first understand the modality information via the perspective of multimodal interaction, providing insights for extracting generalizable knowledge. By applying to real-world settings, it is possible to comprehend a novel modality from scratch, paving the future directions for general-purpose AI.

Additionally, we provide theoretical proofs for all the theoretical studies in this thesis in Chapter 7. At last, we conclude all our contributions and provide prospective discussions for future studies in Chapter 8.
\chapter{Understanding and Harnessing OOD Data}
\label{cha:HOOD}
Machine learning models are vulnerable to Out-Of-Distribution (OOD) examples, and such a problem has drawn much attention. However, current methods lack a full understanding of different types of OOD data: there are \textit{benign} OOD data that can be properly adapted to enhance the learning performance, while other \textit{malign} OOD data would severely degenerate the classification result. To understand and Harness OOD data, this Chapter proposes a HOOD method that can leverage the \textit{content} and \textit{style} from each image instance to identify benign and malign OOD data. Particularly, we design a variational inference framework to causally disentangle content and style features by constructing a structural causal model. Subsequently, we augment the content and style through an intervention process to produce malign and benign OOD data, respectively. The benign OOD data contain novel styles but hold our interested contents, and they can be leveraged to help train a style-invariant model. In contrast, the malign OOD data inherit unknown contents but carry familiar styles, by detecting them can improve model robustness against deceiving anomalies. Thanks to the proposed novel disentanglement and data augmentation techniques, HOOD can effectively deal with OOD examples in unknown and open environments, whose effectiveness is empirically validated in three typical OOD applications including OOD detection, open-set semi-supervised learning, and open-set domain adaptation.

\section{Introduction}
\label{HOOD:introduction}
Learning in the presence of Out-Of-Distribution (OOD) data has been a challenging task in machine learning, as the deployed classifier tends to fail if the unseen data drawn from unknown distributions are not properly handled~\cite{hendrycks2016baseline, pan2009survey}. Such a critical problem ubiquitously exists when deep models meet domain shift~\cite{ganin2016domain, tzeng2017adversarial} and unseen-class data~\cite{hendrycks2016baseline, scheirer2012toward}, which has drawn a lot of attention in some important fields such as OOD detection~\cite{hein2019relu, hendrycks2016baseline, lee2018simple, liang2017enhancing, liu2020energy, wang2022partial, wang2023outofdistribution, wang2022watermarking}, Open-Set Domain Adaptation (DA)~\cite{liu2019separate, saito2018open}, and Open-Set Semi-Supervised Learning (SSL)~\cite{huang2021universal, huang2022they, huang2023flatmatch, huang2022fix, li2024dynamic, oliver2018realistic, saito2021openmatch, yu2020multi}.

\begin{figure}[t]
	\centering
	\includegraphics[width=0.8\linewidth]{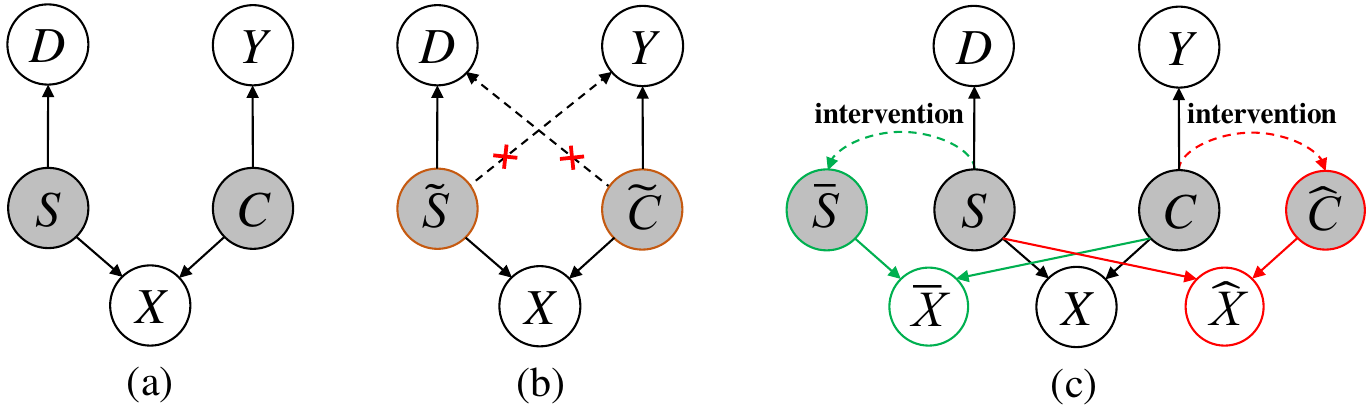}
	\caption{(a) An ideal causal diagram denoting the data generating process. (b) Illustration of our disentanglement. The brown-edged variables $\tilde{C}$ and $\tilde{S}$ are approximations of content $C$ and style $S$. The dashed lines indicate the unwanted causal relations to be broken. (c) Illustration of the data augmentation of HOOD. The green lines and red lines denote the augmentation of benign OOD data $\bar{X}$ and malign OOD data $\hat{X}$, respectively. In all figures, the blank variables are observable and the shaded variables are latent.}
	\label{HOODfig:scm}
\end{figure}

In the above fields, OOD data can be divided into two types, namely \textit{benign} OOD data\footnote{We follow~\cite{bengio2011deep} to regard the augmented data as a type of OOD data} and \textit{malign} OOD data. The benign OOD data can boost the learning performance on the target distribution through DA techniques~\cite{ganin2015unsupervised, tzeng2017adversarial}, but they can be misleading if not being properly exploited. To improve model generalization, many \textit{positive data augmentation} techniques~\cite{cubuk2019autoaugment, xie2020unsupervised} have been proposed. For instance, the performance of SSL~\cite{berthelot2019mixmatch, sohn2020fixmatch} has been greatly improved thanks to the augmented benign OOD data. On the contrary, malign OOD data with unknown classes can damage the classification results, but they are deceiving and hard to detect~\cite{hendrycks2016baseline, liang2017enhancing, wei2022mitigating, wei2022open}. To train a robust model against malign OOD data, some works~\cite{kong2021opengan, sinha2020negative} conduct \textit{negative data augmentation} to generate ``hard'' malign data which resemble in-distribution (ID) data. By separating such ``hard'' data from ID data, the OOD detection performance can be improved. When presented with both malign and benign OOD data, it is more challenging to decide which to separate and which to exploit. As a consequence, the performance of existing open-set methods could be sub-optimal due to two drawbacks: (1) radically exploiting too much malign OOD data, and (2) conservatively denying too much benign OOD data.

In this Chapter, we propose a HOOD framework (see Figure \ref{HOODfig:framework}) to properly harness OOD data in several OOD problems. To distinguish benign and malign OOD data, we model the data generating process by following the structural causal model (SCM)~\cite{glymour2016causal, pearl2009causality, gao2022missdag} in Figure~\ref{HOODfig:scm} (a). Particularly, we decompose an image instance $X$ into two latent components: (1) \textit{content} variable $C$ which denotes the interested object, and (2) \textit{style} variable $S$ which contains other influential factors such as brightness, orientation, and color. The content $C$ can indicate its true \textit{class} $Y$, and the style $S$ is decisive for the environmental condition, which is termed as \textit{domain} $D$. Intuitively, malign OOD data cannot be incorporated into network training, because they contain unseen contents, thus their true classes are different from any known class; and benign OOD data can be adapted because they only have novel styles but contain the same contents as ID data. Therefore, we can distinguish the benign and malign OOD data based on the extracted the content and style features.

In addition, we conduct causal disentanglement through maximizing an approximated evidence lower-bound (ELBO)~\cite{blei2017variational, yao2021instance, xia2022pluralistic} of joint distribution $P(X, Y, D)$. As a result, we can effectively break the spurious correlation~\cite{pearl2009causality, glymour2016causal, hermann2020origins, li2020shape, zhang2021causaladv} between content and style which commonly occurs during network training~\cite{arjovsky2019invariant}, as shown by the dashed lines in Figure~\ref{HOODfig:scm} (b). In the ablation study, we find that HOOD can correctly disentangle content and style, which can correspondingly benefit generalization tasks (such as open-set DA and open-set SSL) and detection task (such as OOD detection).

To further improve the learning performance, we conduct both positive and negative data augmentation by solely intervening the style and content, respectively, as shown by the blue and red lines in Figure~\ref{HOODfig:scm} (c). Such process is achieved through backpropagating the gradient computed from an intervention objective. As a result, style-changed data $\bar{X}$ must be identified as benign OOD data, and content-changed data $\hat{X}$ should be recognized as malign OOD data. Without including any bias, the benign OOD data can be easily harnessed to improve model generalization, and the malign OOD data can be directly recognized as harmful ones which benefits the detection of unknown anomalies. By conducting extensive experiments on several OOD applications, including OOD detection, open-set SSL, and open-set DA, we validate the effectiveness of our method on typical benchmark datasets. To sum up, our contributions are three-fold:
\begin{itemize}
	\item We propose a unified framework dubbed HOOD which can effectively disentangle the content and style features to break the spurious correlation. As a result, benign OOD data and malign OOD data can be correctly identified based on the disentangled features.
	\item We design a novel data augmentation method which correspondingly augments the content and style features to produce benign and malign OOD data, and further leverage them to enhance the learning performance.
	\item We experimentally validate the effectiveness of HOOD on various OOD applications, including OOD detection, open-set SSL, and open-set DA.
\end{itemize}

\section{Related Work}
\label{HOOD:relatedwork}
\textbf{OOD applications} contains three typical problems, namely OOD detection, open-set SSL, and open-set DA. OOD detection~\cite{hendrycks2016baseline, liang2017enhancing} aims to train a robust model which can accurately identify the newly-emerged malign OOD data during the test phase. Open-set SSL~\cite{guo2020safe, chen2020semi, he2022safe, he2022not, yu2020multi} deals with the problem when labeled data are scarce and the unlabeled data are contaminated by malign OOD data. As for open-set DA~\cite{saito2018open, liu2019separate}, it tries to transfer the knowledge from source ID data to the benign OOD data in target domain, meanwhile detecting the malign OOD data that are encountered during transferring. In both three applications, the predictive confidence has been frequently leveraged to separate malign OOD data~\cite{hendrycks2016baseline, liang2017enhancing, ren2019likelihood, xia2022extended}. Moreover, ID data and OOD data can be distinguished via using a discriminator~\cite{kong2021opengan, neal2018open, yu2020multi, xia2021instance}. Further, various open classifiers are designed to predict OOD dataset as unknown~\cite{ge2017generative, padhy2020revisiting, saito2018open}. Thanks to the advances in unsupervised learning, many approaches employ self-supervised learning to distinguish ID data and OOD data~\cite{cao2022openworld, li2021domain, saito2020universal}.

\textbf{Causality in OOD problems} mainly focuses on learning invariant representations that stay constant when other causal factors are changing, thus achieving better performance when facing non-stationary data distribution. To accomplish this goal, it is common to learn causal factors and non-causal factors through the variational auto-encoder framework~\cite{blei2017variational, kingma2013auto}. Thanks to which, domain adaptation~\cite{gong2016domain, scholkopf2011robust, zhang2013domain} and domain generalization~\cite{li2018domain, shankar2018generalizing} can be tackled through extracting the domain invariant features. Moreover, based on causal effects, the biased feature can be eliminated through re-weighting~\cite{bahadori2017causal, shen2018causally}. Additionally, the spurious correlation which is harmful for inference could be alleviated through do-calculus~\cite{lee2021learning, nam2020learning, pearl2009causality}. Recent methods~\cite{ilse2021selecting, mitrovic2020representation, von2021self} conduct data augmentations with self-supervised learning to train a robust model that can handle distribution shifts and corruptions.

In general, HOOD has two major differences from existing methods in OOD applications and causality. On one hand, instead of treating an image instance as a whole as commonly done in many approaches, HOOD can properly leverage OOD examples through their disentangled contents and styles. Moreover, augmenting content and style can help improve generalization and robustness simultaneously. On the other hand, current causal approaches are incapable of dealing with malign OOD data, but HOOD is able to learn style-invariant features from benign OOD data, meanwhile avoiding the damage brought by malign OOD data.

\section{Methodology}
\label{HOOD:method}
\begin{figure}[t]
	\centering
	\includegraphics[width=0.75\linewidth]{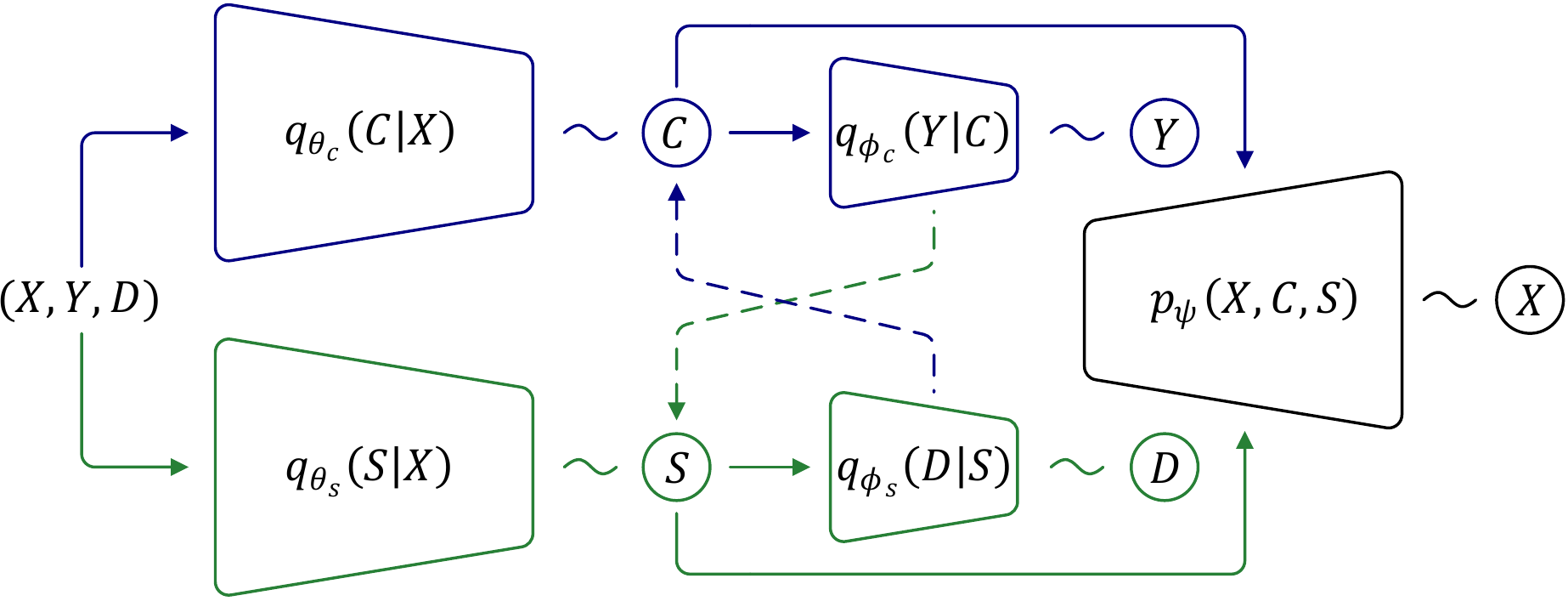}
	\caption{Architecture of the HOOD. The solid lines denote the inference flow, the dashed lines indicate the disentanglement of content and style, and the tildes stand for the approximation of the corresponding variables.}
	\label{HOODfig:framework}
\end{figure}

In this section, we propose our HOOD framework as shown in Figure~\ref{HOODfig:framework}. Specifically, we utilize the class labels of labeled data and the pseudo labels~\cite{lee2013pseudo} of unlabeled data as the class supervision to capture the content feature. Moreover, we perform different types of data augmentation and regard the augmentation types as the domain supervision for each style. Thereby, each instance $\mathbf{x}$ is paired with a class label $y$ and a domain label $d$. Then, we apply two separate encoders $g_c$ and $g_s$ parameterized by $\theta_c$ and $\theta_s$ to model the posterior distributions $q_{\theta_c}(C\mid X)$ and $q_{\theta_s}(S\mid X)$, respectively. Subsequently, the generated $C$ and $S$ are correspondingly fed into two fully-connected classifiers $f_c$ and $f_s$ parameterized by $\phi_c$ and $\phi_s$, which would produce the label predictions $q_{\phi_c}(Y\mid C)$ and $q_{\phi_s}(D\mid S)$, respectively. To further enhance the identifiability of $C$ and $S$, a decoder $h$ with parameter $\psi$ is employed to reconstruct the input instance $\mathbf{x}$ based on its content and style.

Below, we describe the detailed procedures and components during modeling HOOD. We first introduce the proposed variational inference framework for disentangling the content and style based on the constructed SCM. Subsequently, we conduct intervention to produce benign OOD data and malign OOD data. Further, we appropriately leverage the benign and malign OOD data to boost the learning performance. Finally, we formulate the deployment of HOOD in three OOD applications.

\subsection{Variational Inference for Content and Style Disentanglement}
\label{HOOD:disentanglement}
First, we assume that the data generating process can be captured by certain probability distributions. Therefore, according to the constructed SCM in Figure~\ref{HOODfig:scm} (a), the joint distribution $P(X, Y, D, C, S)$ of the interested variables can be factorized as follows:
\begin{equation}
	P(X, Y, D, C, S)=P(C, S)P(Y, D\mid C, S)P(X\mid C, S).
	\label{HOODeq:scm_factorization}
\end{equation}
Based on the SCM in Figure~\ref{HOODfig:scm} (a), $Y$ and $D$ are conditionally independent to each other, i.e., $Y \indep D \mid (C, S)$, so we have $P(Y, D\mid C, S)=P(Y\mid C, S)P(D\mid C, S)$. Similarly, we have $P(C, S)=P(C)P(S)$. Moreover, we can also know that $Y$ is not conditioned on $S$, and $D$ is not conditioned on $C$. Hence, we can further derive $P(Y, D\mid C, S)=P(Y\mid C)P(D\mid S)$.

However, the aforementioned spurious correlation frequently appears when facing OOD examples~\cite{arjovsky2019invariant}. As a consequence, when variational inference is based on the factorization in Equation~\eqref{HOODeq:scm_factorization}, the approximated content $\tilde{C}$ and style $\tilde{S}$ could both directly influence $Y$ and $D$, i.e., $Y \leftarrow \tilde{C}\rightarrow D$ and $Y \leftarrow \tilde{S}\rightarrow D$, thus leading to inaccurate approximations. However, the desired condition is $Y \leftarrow C\nrightarrow D$ and $Y \nleftarrow S\rightarrow D$. We can see that the unwanted correlations $\tilde{C}\rightarrow D$ and $\tilde{S}\rightarrow Y$ in Figure~\ref{HOODfig:scm} (b) is caused by erroneous posteriors $P(D\mid \tilde{C})$ and $P(Y\mid \tilde{S})$. Therefore, to break the correlations, the posteriors $q_{\phi_s}(D\mid C)$ and $q_{\phi_c}(Y\mid S)$ which are correspondingly approximated by the decoders $\phi_s$ and $\phi_c$ can be used as denominators to $q_{\phi_c}(Y\mid C)$ and $q_{\phi_s}(D\mid S)$, respectively. In this way, we can successfully disentangle content $C$ and style $S$ and ensure the decoding process of $Y$ and $D$ would not be influenced by spurious features from $S$ and $C$, respectively. To this end, our factorization in Equation~\eqref{HOODeq:scm_factorization} can be approximated as:
\begin{equation}
	\tilde{P}(X, Y, D, C, S)\coloneqq\frac{P(C)P(S)P(Y\mid C)P(D\mid S)P(X\mid C, S)}{q_{\phi_s}(D\mid C)q_{\phi_c}(Y\mid S)}.
	\label{HOODeq:modified_factorization}
\end{equation}

Then, we maximize the log-likelihood of the joint distribution $p(\mathbf{x}, y, d)$ of each data point $(\mathbf{x}, y, d)$:
\begin{equation}
	\log p(\mathbf{x}, y, d) \coloneqq \log \int_c \int_s \tilde{p}(\mathbf{x}, y, d, c, s)\mathrm{d}c\mathrm{d}s,
	\label{HOODeq:joint_distribution}
\end{equation}
in which we use lower case to denote the values of corresponding variables. Due to the integration of latents $C$ and $S$ is intractable, we follow variational inference~\cite{blei2017variational} to obtain an approximated evidence lower-bound $\tilde{ELBO}(\mathbf{x}, y, d)$ of the log-likelihood in Equation~\eqref{HOODeq:joint_distribution}:
\begin{equation}
	\begin{aligned}
		\log p(\mathbf{x}, y, d) =& \log \int_c \int_s \tilde{p}(\mathbf{x}, y, d, c, s)\mathrm{d}c\mathrm{d}s \\
		=& \log \int_c \int_s \tilde{p}(\mathbf{x}, y, d, c, s) \frac{q_{\theta}(c, s\mid \mathbf{x})}{q_{\theta}(c, s\mid \mathbf{x})}\mathrm{d}c\mathrm{d}s \\
		=& \log\mathbb{E}_{(c, s)\sim q_{\theta}(C, S\mid \mathbf{x})}\left[\frac{\tilde{p}(\mathbf{x}, y, d, c, s)}{q_{\theta}(c, s\mid \mathbf{x})}\right] \\
		\geq& \mathbb{E}_{(c, s)\sim q_{\theta}(C, S\mid \mathbf{x})}\left[\log\frac{\tilde{p}(\mathbf{x}, y, d, c, s)}{q_{\theta}(c, s\mid \mathbf{x})}\right] \coloneqq \tilde{ELBO}(\mathbf{x},y,d).	
	\end{aligned}
	\label{HOODeq:derivation_1}
\end{equation}
Recall the modified joint distribution factorization in Equation~\eqref{HOODeq:modified_factorization}, we can have:
\begin{subequations}
	\label{HOODeq:elbo}
	\begin{align}
		\notag
		\tilde{ELBO}(\mathbf{x},y,d) =& \mathbb{E}_{(c, s)\sim q_{\theta}(C, S\mid \mathbf{x})}\left[\log\frac{p(c)p(s)q_{\phi_c}(y\mid c)q_{\phi_s}(d\mid s)p_{\psi}(\mathbf{x}\mid c, s)}{q_{\theta}(c, s\mid \mathbf{x})q_{\phi_c}(y\mid s)q_{\phi_s}(d\mid c)}\right]\\ \notag
		=& \mathbb{E}_{(c, s)\sim q_{\theta}(C, S\mid \mathbf{x})}\left[\log\frac{p(c)p(s)}{q_{\theta_c}(c\mid \mathbf{x})q_{\theta_s}(s\mid \mathbf{x})}\right] + \mathbb{E}_{(c, s)\sim q_{\theta}(C, S\mid \mathbf{x})}\left[\log\frac{q_{\phi_c}(y\mid c)}{q_{\phi_s}(d\mid c)}\right]  \\ \notag
		+&\mathbb{E}_{(c, s)\sim q_{\theta}(C, S\mid \mathbf{x})}\left[\log\frac{q_{\phi_s}(d\mid s)}{q_{\phi_c}(y\mid s)}\right] + \mathbb{E}_{(c, s)\sim q_{\theta}(C, S\mid \mathbf{x})}\left[\log p_{\psi}(\mathbf{x}\mid c, s)\right] \\ \notag
		=& \mathbb{E}_{(c)\sim q_{\theta_c}(C\mid \mathbf{x})}\left[\log\frac{p(c)}{q_{\theta_c}(c\mid \mathbf{x})}\right] + \mathbb{E}_{(s)\sim q_{\theta_s}(S\mid \mathbf{x})}\left[\log\frac{p(c)p(s)}{q_{\theta_s}(s\mid \mathbf{x})}\right] \\ \notag
		+& \mathbb{E}_{(c)\sim q_{\theta_s}(C\mid \mathbf{x})}\left[\log\frac{q_{\phi_c}(y\mid c)}{q_{\phi_s}(d\mid c)}\right] + \mathbb{E}_{(s)\sim q_{\theta_s}(S\mid \mathbf{x})}\left[\log\frac{q_{\phi_s}(d\mid s)}{q_{\phi_c}(y\mid s)}\right] \\ \notag
		+& \mathbb{E}_{(c, s)\sim q_{\theta}(C, S\mid \mathbf{x})}\left[\log p_{\psi}(\mathbf{x}\mid c, s)\right] \\ \notag
		=& -KL(q_{\theta_c}(c\mid \mathbf{x})\|p(C))-KL(q_{\theta_s}(s\mid \mathbf{x})\|p(S)) \\ \notag
		&+\mathbb{E}_{c\sim q_{\theta_c}(C\mid \mathbf{x})}\left[\log q_{\phi_c}(y\mid c) - \log q_{\phi_s}(d\mid c)\right] \\ \notag
		&+\mathbb{E}_{s\sim q_{\theta_s}(S\mid \mathbf{x})}\left[\log q_{\phi_s}(d\mid s) - \log q_{\phi_c}(y\mid s)\right] \\
		&+\mathbb{E}_{(c, s)\sim q_{\theta}(C, S\mid \mathbf{x})}\left[ \log p_{\psi}(\mathbf{x}\mid c, s) \right]
        \label{HOODeq:final_elbo}\\
		=&ELBO(\mathbf{x}, y, d) - \mathbb{E}_{c\sim q_{\theta_c}(C\mid \mathbf{x})}\left[\log q_{\phi_s}(d\mid c)\right] - \mathbb{E}_{s\sim q_{\theta_s}(S\mid \mathbf{x})}\left[\log q_{\phi_c}(y\mid s)\right].
		\label{HOODeq:elbo_reg}
	\end{align}
\end{subequations}
In Equation~\eqref{HOODeq:final_elbo}, the first two terms indicate the Kullback-Leibler divergence between the latent variables $C$ and $S$ and their prior distributions. In practice, we assume that the priors $p(C)$ and $p(S)$ follow standard multivariate Gaussian distributions. The third and fourth terms contain the approximated log-likelihoods of label predictions and the disentanglement of the content and style. The last term stands for estimated distribution of $\mathbf{x}$. Note that in Equation~\eqref{HOODeq:elbo_reg}, our approximated $\tilde{ELBO}$ is composed of two parts: the original $ELBO$ which could be obtained from the factorization in Equation~\eqref{HOODeq:scm_factorization}, and two regularization terms that aims to disentangle $C$ and $S$ through maximizing the log-likelihoods $\log q_{\phi_s}(d\mid c)$ and $\log q_{\phi_c}(y\mid s)$, which is shown by the dashed lines in Figure~\ref{HOODfig:framework}. By maximizing $\tilde{ELBO}$, we can train an accurate class predictor which is invariant to different styles. The detailed derivation is provided in supplementary material. Next, we introduce our data augmentation to assist in harnessing OOD examples.

\subsection{Data Augmentation with Content and Style Intervention}
\label{HOOD:augmentation}
After disentangling content and style, we try to harness OOD examples via two opposite augmentation procedures, namely \textit{positive data augmentation} and \textit{negative data augmentation} which aim to produce benign OOD data $\bar{\mathbf{x}}$ and malign OOD data $\hat{\mathbf{x}}$, respectively, so as to further enhance model generalization and improve robustness against anomalies. Specifically, to achieve this, positive data augmentation only conducts intervention on the style feature meanwhile keeping the content information the same; and the negative data augmentation attempts to affect the content feature while leaving the style unchanged, so as to produce malign OOD data, as shown in Figure~\ref{HOODfig:scm} (b).

To achieve this goal, we employ adversarial data augmentation~\cite{goodfellow2014generative, miyato2018virtual, volpi2018generalizing} which can directly conduct intervention on the latent variables without influencing each other, thus it is perfect for our intuition of augmenting content and style. Particularly, by adding a learnable perturbation $\mathbf{e}$ to each instance $\mathbf{x}$, we can obtain malign OOD data $\hat{\mathbf{x}}$ and benign OOD data $\bar{\mathbf{x}}$ with augmented content and style, respectively. For each data point $(\mathbf{x}, y, d)$, the perturbation $\mathbf{e}$ can be obtained through minimizing the intervention objective $\mathcal{L}(\cdot)$:
\begin{equation}
	\mathbf{e} = \arg\min_{\mathbf{e};\|\mathbf{e}\|_{p}<\epsilon}\mathcal{L}(\mathbf{x}+\mathbf{e}, y, d; \theta_c, \phi_c, \theta_s, \phi_s),
	\label{HOODeq:adv}
\end{equation}
where $\epsilon$ denotes the magnitude of the perturbation $\mathbf{e}$ with $\ell_p$-norm. Since our goal of positive and negative data augmentation is completely different, here the intervention objective is designed differently for producing $\bar{\mathbf{x}}$ and $\hat{\mathbf{x}}$.
For positive data augmentation, the intervention objective is:
\begin{equation}
	\mathcal{L}_{pos} = \mathcal{L}_{d}(g_c(\mathbf{x};\theta_c), g_c(\mathbf{x}+\mathbf{e};\theta_c)) - \mathcal{L}_{ce}(f_s(g_s(\mathbf{x}+\mathbf{e};\theta_s);\phi_s), d),
	\label{HOODeq:pos_aug}
\end{equation}
where the first term $\mathcal{L}_{d}(\cdot)$ indicates the distance measured between the contents extracted from the original instance and its perturbed version, and the second term $\mathcal{L}_{ce}(\cdot)$ denotes the cross-entropy loss. By minimizing $\mathcal{L}_{pos}$, the perturbation $\mathbf{e}$ would not significantly affect the content feature, meanwhile introducing a novel style that is distinct from its original domain $d$. Consequently, the augmented benign data with novel styles can be utilized to train a style-invariant model that is resistant to domain shift. Moreover, a specific style with domain label $d^{\prime}$ can be injected via modifying $\mathcal{L}_{pos}$ as:
\begin{equation}
	\mathcal{L}_{pos}^{\prime} = \mathcal{L}_{d}(g_c(\mathbf{x};\theta_c), g_c(\mathbf{x}+\mathbf{e};\theta_c)) + \mathcal{L}_{ce}(f_s(g_s(\mathbf{x}+\mathbf{e};\theta_s);\phi_s), d^{\prime}).
	\label{HOODeq:pos_aug_modify}
\end{equation}
Different from Equation~\eqref{HOODeq:pos_aug}, we hope to minimize the cross-entropy loss such that the perturbed instance can contain the style information from a target domain $d^{\prime}$. As a result, the augmented benign data can successfully bridge the gap between source domain and target domain, and further improve the test performance in the target distribution.

As for negative data augmentation, the intervention objective is defined as:
\begin{equation}
	\mathcal{L}_{neg} = \mathcal{L}_{d}(g_s(\mathbf{x};\theta_s), g_s(\mathbf{x}+\mathbf{e};\theta_s)) - \mathcal{L}_{ce}(f_c(g_c(\mathbf{x}+\mathbf{e};\theta_c);\phi_c), y).
	\label{HOODeq:neg_aug}
\end{equation}
By minimizing $\mathcal{L}_{neg}$, the perturbation would not greatly change the style information but would deviate the content from its original one with class label $y$. Subsequently, by recognizing the augmented malign data as unknown, the trained model would be robust to deceiving anomalies with familiar styles, thus boosting the OOD detection performance.

To accomplish the adversarial data augmentation process, here we perform multi-step projected gradient descent~\cite{madry2017towards, wang2021probabilistic}. Formally, the optimal $\bar{\mathbf{x}}$ and $\hat{\mathbf{x}}$ can be iteratively found:
\begin{equation}
	\bar{\mathbf{x}}^{t+1} = \bar{\mathbf{x}}^{t}+\arg\min_{\mathbf{e}^{t};\|\mathbf{e}^{t}\|_{p}<\epsilon}\mathcal{L}_{pos}(\bar{\mathbf{x}}^{t}+\mathbf{e}^{t}),\\
	\hat{\mathbf{x}}^{t+1} = \hat{\mathbf{x}}^{t}+\arg\min_{\mathbf{e}^{t};\|\mathbf{e}^{t}\|_{p}<\epsilon}\mathcal{L}_{neg}(\hat{\mathbf{x}}^{t}+\mathbf{e}^{t}).
	\label{HOODeq:data_augmentation}
\end{equation}
where the final iteration $t$ is set to $15$ in practice. Further, the optimal augmented data will be incorporated into model training, which is described in the next section.

\begin{algorithm}[t]
    \caption{\small Training process of HOOD}
    \label{HOODalg:HOOD}
    \begin{algorithmic}[1]
        \STATE Labeled set $\mathcal{D}^l=\{(\mathbf{x}_i, y_i)\}_{i=1}^l$, unlabeled set $\mathcal{D}^u=\{(\mathbf{x}_i)\}_{i=1}^u$.
        \FOR{$i=1$ \ldots $Max\_Iter$}
            \STATE Pre-train the variational inference framework through maximizing $\tilde{ELBO}$ in Equation~\eqref{HOODeq:final_elbo};
            \STATE Assigning pseudo labels $y^{ps}$ for unlabeled data $\mathcal{D}^u:=\{(\mathbf{x}_i; y_i^{ps})\}_{i=1}^u$;
            \IF{$i == Augmentation\_Iter$}
                \STATE Conduct Adversarial Data Augmentation to obtain $\bar{\mathbf{x}}$ and $\hat{\mathbf{x}}$ via Equation~\eqref{HOODeq:data_augmentation};
                \STATE Add $\bar{\mathbf{x}}$ and $\hat{\mathbf{x}}$ into $\bar{\mathcal{D}}$ and $\hat{\mathcal{D}}$, respectively;
            \ENDIF
            \STATE Enumerate $\bar{\mathcal{D}}$ and conduct supervised training for each $\bar{\mathbf{x}}$;
            \STATE Enumerate $\hat{\mathcal{D}}$ and recognize each $\hat{\mathbf{x}}$ as unknown;
        \ENDFOR
    \end{algorithmic}
\end{algorithm}

\subsection{Model Training with Benign and Malign OOD data}
\label{HOOD:training}
Finally, based on the above disentanglement and data augmentation in Section~\ref{HOOD:disentanglement} and Section~\ref{HOOD:augmentation}, we can obtain a benign OOD data $\bar{\mathbf{x}}$ and a malign OOD data $\hat{\mathbf{x}}$ from each data point $(\mathbf{x}, y, d)$, which will be appended to the benign dataset $\bar{\mathcal{D}}$ and malign dataset $\hat{\mathcal{D}}$, respectively. For utilization of benign OOD data $\bar{\mathbf{x}}$, we assign it with the original class label $y$ and perform supervised training. For separation of malign OOD data $\hat{\mathbf{x}}$, we employ a one-vs-all classifier~\cite{padhy2020revisiting} to recognize them as unknown data that is distinct from its original class label $y$. The proposed HOOD method is summarized in Algorithm~\ref{HOODalg:HOOD}. Below, we specify the proposed HOOD algorothm to three typical applications with OOD data, namely OOD detection, open-set SSL, and open-set DA.

\subsection{Deployment to OOD applications}
Generally, in all three investigated applications, we are given a labeled set $\mathcal{D}^l=\{(\mathbf{x}_i, y_i)\}_{i=1}^l$ containing $l$ labeled examples drawn from data distribution $P^l$, and an unlabeled set $\mathcal{D}^u=\{\mathbf{x}_i\}_{i=1}^u$ composed of $u$ unlabeled examples sampled from data distribution $P^u$. Moreover, the label space of $\mathcal{D}^l$ and $\mathcal{D}^u$ are defined as $\mathcal{Y}^l$ and $\mathcal{Y}^u$, respectively.

\textbf{OOD detection.} The labeled set is used for training, and the unlabeled set is used as a test set which contains both ID data and malign OOD data. Particularly, the data distribution of unlabeled ID data $Q_{id}$ is the same as distribution $P$, but the distribution of OOD data $P^u_{ood}$ is different from $P$, i.e., $P^u_{id}=P^l\neq P^u_{ood}$. The goal is to correctly distinguish OOD data from ID data in the test phase. During training, we conduct data augmentation to obtain domain label $d$, then follow our variational framework to update the model parameters. During test, we only use the content branch to predict the OOD score which is produced by the one-vs-all classifier. An instance is considered as an ID datum if the OOD score is smaller than $0.5$, and an OOD datum otherwise.

\textbf{Open-set SSL.} 
The labeled set $\mathcal{D}^l$ and unlabeled set $\mathcal{D}^u$ are both used for training, and they are sampled from the same data distribution with different label spaces. Specifically, the unlabeled data contain some ID data that have the same classes as $\mathcal{D}^l$, and the rest unlabeled OOD data are from some unknown classes that do not exist in $\mathcal{D}^l$, formally, $\mathcal{Y}^l\subset \mathcal{Y}^u, \mathcal{Y}^u\setminus \mathcal{Y}^l \neq \varnothing$ and $P^l(\mathbf{x}\mid y)=P^u(\mathbf{x}\mid y), y\in\mathcal{Y}^l$. The goal is to properly leverage the labeled data and unlabeled ID data without being misled by malign OOD data, and correctly classify test data with labels in $\mathcal{Y}^l$. The training process is similar to OOD detection, except that HOOD would produce an OOD score for each unlabeled data. If an unlabeled instance is recognized as OOD data, it would be left out.

\textbf{Open-set DA.} 
The labeled set is drawn from source distribution $P^l$ which is different from the target distribution $P^u$ of unlabeled set. In addition, the label space $\mathcal{Y}^l$ is also a subset of $\mathcal{Y}^u$. Therefore, the unlabeled data consist of benign OOD data which have the same class labels as labeled data, and malign OOD data which have distinct data distribution as well as class labels from labeled data, formally, $P^l\neq P^u, \mathcal{Y}^l\subset \mathcal{Y}^u, \mathcal{Y}^u\setminus \mathcal{Y}^l \neq \varnothing$. The goal is to transfer the knowledge of labeled data to the benign OOD data, meanwhile identify the malign OOD data as unknown. In this application, we assign each target instance with a domain label to distinguish them from other augmented data. Then we alter the positive data augmentation objective from Equation~\eqref{HOODeq:pos_aug} to Equation~\eqref{HOODeq:pos_aug_modify}. During test, HOOD would predict each target instance as some class if it is benign, and as unknown otherwise.

\section{Experiment}
\label{HOOD:experiment}

In this section, we first describe the implementation details. Then, we experimentally validate our method on three applications, namely OOD detection, open-set SSL, and open-set DA. Finally, we present extensive performance analysis on our disentanglement and intervention modules. Additional details and quantitative findings can be found in the supplementary material.

\subsection{Implementation Details}
\label{HOOD:details}
In experiments, we choose Wide ResNet-28-2~\cite{zagoruyko2016wide} for OOD detection and Open-set SSL tasks, and follow~\cite{you2020Universal, cao2019learning} to utilize ResNet50 pre-trained on Imagenet~\cite{russakovsky2015imagenet} for Open-set DA. For implementing HOOD, we randomly choose 4 augmentation methods from the transformation pool in RandAugment~\cite{cubuk2020randaugment}, to simulate different styles. The pre-training iteration $Augmentation\_Iter$ is set to 100,000, and the perturbation magnitude $\epsilon=0.03$, following~\cite{volpi2018generalizing} in all experiments. Next, we validate HOOD in three applications.

\subsection{OOD detection}
\label{HOOD:ood_detection}
In OOD detection task, we use SVHN~\cite{netzer2011reading} and CIFAR10~\cite{krizhevsky2009learning} as the ID datasets, and use LSUN~\cite{yu2015lsun}, DTD~\cite{cimpoi2014describing}, CUB~\cite{wah2011caltech}, Flowers~\cite{nilsback2006visual}, Caltech~\cite{griffin2007caltech}, and Dogs~\cite{khosla2011novel} datasets as the OOD datasets that occur during test phase. Particularly, to explore the model generalization ability, we only sample 100 labeled data and 20,000 unlabeled data from each class and conduct semi-supervised training, then we test the trained model on the unlabeled OOD dataset. To evaluate the performance, we utilize AUROC~\cite{hendrycks2016baseline} which is an essential metric for OOD detection, and a higher AUROC value indicates a better performance.

\begin{table}[t]
	\centering
	\scriptsize
	\caption{Comparison with typical OOD detections methods. Averaged AUROC ($\%$) with standard deviations are computed over three independent trails. The best results are highlighted in bold.}
	\setlength{\tabcolsep}{2mm}
	\label{HOODtab:OOD_detection}
	\begin{tabular}{lcccccc}
		\toprule[1pt]
		OOD dataset                    & \multicolumn{1}{c}{LSUN}   & \multicolumn{1}{c}{DTD}    & \multicolumn{1}{c}{CUB}    & \multicolumn{1}{c}{Flowers} & \multicolumn{1}{c}{Caltech} & \multicolumn{1}{c}{Dogs}   \\ 
		\midrule[0.6pt]
		\multicolumn{1}{l}{ID dataset} & \multicolumn{6}{c}{SVHN}\\ 
		\midrule[0.6pt]
		Likelihood                     & $52.25\pm0.3$ & $50.33\pm0.7$ & $48.76\pm0.6$ & $47.33\pm0.2$  & $51.54\pm0.4$  & $54.34\pm0.4$ \\
		ODIN                          & $55.72\pm0.2$ & $53.32\pm0.5$ & $52.70\pm0.4$ & $50.47\pm0.7$  & $56.41\pm0.4$  & $61.16\pm0.3$ \\
		Likelihood Ratio                        & $79.34\pm0.5$ & $78.42\pm0.3$ & $75.90\pm0.7$ & $74.53\pm0.4$  & $76.25\pm0.3$  & $83.55\pm0.4$ \\
		OpenGAN                        & $83.77\pm0.4$ & $80.36\pm0.5$ & $77.49\pm0.8$ & $79.26\pm0.5$  & $\bm{86.66}\pm\bm{0.5}$  & $86.84\pm0.5$ \\
		HOOD                           & $\bm{84.10}\pm\bm{0.6}$ & $\bm{80.68}\pm\bm{0.6}$ & $\bm{79.24}\pm\bm{0.5}$ & $\bm{80.93}\pm\bm{0.7}$  & $85.34\pm0.7$  & $\bm{87.58}\pm\bm{0.8}$ \\ 
		\midrule[0.6pt]
		\multicolumn{1}{l}{ID dataset} & \multicolumn{6}{c}{CIFAR10}\\ 
		\midrule[0.6pt]
		Likelihood & $54.32\pm0.5$ & $52.16\pm0.4$ & $50.67\pm0.4$ & $49.26\pm0.3$ & $53.86\pm0.4$ & $56.92\pm0.2$ \\
		ODIN       & $58.60\pm0.3$ & $55.59\pm0.6$ & $58.48\pm0.7$ & $51.44\pm0.9$ & $59.36\pm0.4$ & $64.82\pm0.5$ \\
		Likelihood Ratio    & $81.41\pm0.6$ & $79.77\pm0.5$ & $79.35\pm0.8$ & $77.17\pm0.7$ & $80.67\pm0.5$ & $86.76\pm0.3$ \\
		OpenGAN    & $84.03\pm0.4$ & $81.29\pm0.8$ & $82.84\pm1.0$ & $\bm{82.32}\pm\bm{0.4}$ & $86.78\pm0.3$ & $90.14\pm0.5$ \\
		HOOD       & $\bm{86.12}\pm\bm{0.6}$ & $\bm{83.64}\pm\bm{0.5}$ & $\bm{83.53}\pm\bm{0.6}$ & $81.56\pm0.8$ & $\bm{87.24}\pm\bm{0.8}$ & $\bm{90.86}\pm\bm{0.6}$ \\ 
		\toprule[1pt]
	\end{tabular}
\end{table}

For comparison, we choose some typical OOD detection methods including Likelihood~\cite{hendrycks2016baseline} which simply utilizes softmax score as the detection criterion, ODIN~\cite{liang2017enhancing} which enhances the performance of Likelihood through adding adversarial attack, Likelihood Ratio~\cite{ren2019likelihood} which modifies the softmax score through focusing on the semantic feature, and OpenGAN~\cite{kong2021opengan} which can further improve the performance via separating the generated ``hard'' examples that are deceivingly close to ID data.

The experimental results are shown in Table~\ref{HOODtab:OOD_detection}, we can see that HOOD can greatly surpass Likelihood, ODIN, and Likelihood Ratio, and can outperform OpenGAN in most scenarios. When compared with softmax-prediction-based methods such as Likelihood and ODIN, HOOD surpasses them in a large margin, as HOOD can correctly separate some overconfident OOD examples from ID data. As for Likelihood Ratio, our method can achieve better performance through producing ``hard'' malign OOD data, thus successfully avoiding deceiving examples that are extremely close to ID data. Although both OpenGAN and HOOD generate ``hard'' malign data to train an open classifier, HOOD can successfully distinguish content and style thanks to the aforementioned disentanglement, thus avoid rejecting too much benign OOD data and further yield better detection performance than OpenGAN.

\begin{table}[t]
	\scriptsize
	\caption{Comparison with typical Open-set SSL methods. Averaged test accuracies ($\%$) with standard deviations are computed over three independent trails. The best results are highlighted in bold.}
	\setlength{\tabcolsep}{1mm}
	\label{HOODtab:ssl}
	\begin{tabular}{llccc|ccc}
		\toprule[1pt]
		\multicolumn{2}{l}{Training dataset}        & \multicolumn{3}{c|}{CIFAR10}                                                & \multicolumn{3}{c}{CIFAR100}                                               \\
		\midrule[0.6pt]
		\multicolumn{2}{l}{No. of Labeled data}     & \multicolumn{1}{c}{50} & \multicolumn{1}{c}{100} & \multicolumn{1}{c|}{400} & \multicolumn{1}{c}{50} & \multicolumn{1}{c}{100} & \multicolumn{1}{c}{400} \\ 
		\midrule[0.6pt]
		\multirow{6}{*}{Clean Acc.}     & UASD      & $72.82\pm0.9$             & $75.53\pm1.8$              & $76.74\pm1.7$              & $58.87\pm0.6$             & $61.68\pm1.2$              & $65.97\pm2.4$              \\
		& DS3L      & $74.44\pm1.3$             & $76.89\pm1.5$              & $78.80\pm0.6$              & $60.40\pm0.5$             & $64.35\pm1.5$              & $67.65\pm1.3$              \\
		& MTCF& $79.88\pm1.3$             & $81.41\pm1.0$              & $83.92\pm0.8$              & $62.78\pm0.5$             & $65.84\pm2.1$              & $69.46\pm0.6$              \\ 
		& OpenMatch & $\bm{84.10}\pm\bm{1.1}$             & $\bm{85.30}\pm\bm{0.4}$              & $\bm{87.92}\pm\bm{1.0}$              & $65.76\pm0.9$             & $\bm{68.46}\pm\bm{0.5}$              & $72.87\pm1.4$              \\
		& T2T       & $82.74\pm1.2$ &	$83.56\pm1.4$ &	$85.97\pm0.8$ & $65.16\pm1.2$ &	$67.58\pm0.9$ &	$71.96\pm1.1$ \\
		& HOOD      & $83.55\pm1.2$             & $84.16\pm1.5$              & $86.22\pm2.7$              & $\bm{66.39}\pm\bm{1.7}$             & $68.03\pm2.6$              & $\bm{73.32}\pm\bm{0.6}$              \\
		\midrule[0.6pt]
		\multirow{6}{*}{Corrupted Acc.} & UASD      & $39.36\pm1.2$             & $41.38\pm0.7$              & $42.66\pm1.8$              & $31.55\pm2.0$             & $33.39\pm1.7$              & $35.20\pm0.8$        \\         
		& DS3L   & $39.97\pm0.8$             & $42.58\pm0.8$              & $44.39\pm0.6$              & $33.72\pm0.8$             & $34.67\pm0.8$              & $36.64\pm0.6$              \\
		& MTCF & $40.16\pm1.2$             & $40.58\pm1.1$              & $43.33\pm0.7$              & $32.72\pm0.8$             & $34.33\pm2.3$              & $35.53\pm0.6$              \\     
		& OpenMatch       & $41.38\pm0.7$             & $42.90\pm0.6$              & $45.79\pm0.8$              & $35.98\pm1.3$             & $36.47\pm0.7$              & $38.56\pm0.6$              \\
		& T2T       & $41.39\pm1.6$ &	$45.56\pm1.6$ &	$49.88\pm1.5$ & $\bm{41.03}\pm\bm{1.7}$ &	$39.64\pm0.7$ &	$41.38\pm1.6$ \\
		& HOOD      & $\bm{44.42}\pm\bm{1.7}$             & $\bm{48.38}\pm\bm{0.9}$              & $\bm{50.74}\pm\bm{0.6}$              & $40.82\pm1.5$             & $\bm{41.65}\pm\bm{0.9}$              & $\bm{43.72}\pm\bm{2.2}$              \\ 
		\toprule[1pt]
	\end{tabular}
\end{table}

\subsection{Open-Set SSL}
\label{HOOD:openset_ssl}
In open-set SSL task, we follow~\cite{guo2020safe} to construct our training dataset using two benchmark datasets CIFAR10 and CIFAR100~\cite{krizhevsky2009learning}, which contains 10 and 100 classes, respectively. The constructed dataset has 20,000 randomly sampled unlabeled data and a varied number of labeled data. Here the number of labeled data is set to 50, 100, and 400 per class in both CIFAR10 and CIFAR100. Moreover, to create the open-set problem in CIFAR10, the unlabeled data is sampled from all 10 classes and the labeled data is sampled from the 6 animal classes. As for CIFAR100, the unlabeled data are sampled from all 100 classes and the labeled data is sampled from the first 60 classes. For evaluation, we first use the test dataset from the original CIFAR10 and CIFAR100 and denote the test accuracy as ``Clean Acc.''. Further, to evaluate the capability of handling OOD examples, we test on CIFAR10-C and CIFAR100-C~\cite{hendrycks2019benchmarking} which add different types of corruptions to CIFAR10 and CIFAR100, respectively. The test accuracy from the corrupted datasets can reveal the robustness of neural networks against corruptions and perturbations, and it is denoted as ``Corrupted Acc.''.

For comparison, we choose some typical open-set SSL methods including Uncertainty-Aware Self-Distillation method UASD~\cite{chen2020semi} and T2T~\cite{huang2021trash} which filters out the OOD data via using OOD detection, Safe Deep Semi-Supervised Learning DS3L~\cite{guo2020safe} which employs meta-learning to down-weight the OOD data, Multi-Task Curriculum Framework MTCF~\cite{yu2020multi} which recognizes the OOD data as different domain, and OpenMatch~\cite{saito2021openmatch} which utilizes open-set consistency training on OOD data.

The experimental results are shown in Table~\ref{HOODtab:ssl}. Compared to the strongest baseline method OpenMatch, which randomly samples eleven different transformations from a transformation pool, our method has transformations that are limited to only four types. In CIFAR10 and CIFAR100 regarding the Clean Acc., the proposed HOOD is slightly outperformed by OpenMatch. However, thanks to the disentanglement, HOOD can be invariant to different styles and focus on the content feature. Therefore, when facing corruption, HOOD can be more robust than all baseline methods. As shown by the Corrupted Acc. results, our method surpasses OpenMatch for more than $3\%$.

\begin{table}[t]
	\scriptsize
	\caption{Comparison with typical Open-set DA methods. Averaged test accuracies ($\%$) with standard deviations are computed over three independent trails. The best results are highlighted in bold.}
	\setlength{\tabcolsep}{2mm}
	\label{HOODtab:da}
	\begin{tabular}{lcccccc|c}
		\toprule[1pt]
		Dataset & \multicolumn{6}{c|}{Office}                                                     & VisDA          \\ 
		\midrule[0.6pt]
		Domain  & A$\rightarrow$W       & A$\rightarrow$D       & D$\rightarrow$W       & W$\rightarrow$D       & D$\rightarrow$A       & W$\rightarrow$A & Synthetic$\rightarrow$Real \\
		\midrule[0.6pt]
		OSBP    & $86.5\pm2.0$ & $88.6\pm1.4$ & $97.0\pm1.0$ & $97.9\pm0.9$ & $88.9\pm2.5$ & $85.8\pm2.5$                & $62.9\pm1.3$      \\
		UAN     & $87.7\pm1.2$ & $87.0\pm0.8$ & $93.5\pm1.3$ & $97.2\pm1.6$ & $88.4\pm0.7$ & $87.8\pm1.6$                & $63.8\pm2.4$      \\
		STA     & $89.5\pm0.6$ & $93.7\pm1.5$ & $97.5\pm0.2$ & $\bm{99.5}\pm\bm{0.2}$ & $89.1\pm0.5$ & $87.9\pm0.9$                & $66.4\pm1.3$      \\
		HOOD    & $\bm{90.1}\pm\bm{1.5}$ & $\bm{94.2}\pm\bm{1.4}$ & $\bm{99.6}\pm\bm{0.6}$ & $98.3\pm0.9$ & $\bm{89.8}\pm\bm{0.8}$ & $\bm{91.3}\pm\bm{1.8}$                & $\bm{72.4}\pm\bm{1.6}$      \\ 
		\toprule[1pt]
	\end{tabular}
\end{table}

\subsection{Open-Set DA}
\label{HOOD:openset_da}
In open-set DA task, we follow~\cite{saito2018open} to validate on two DA benchmark datasets Office~\cite{saenko2010adapting} and VisDA~\cite{peng2018visda}. Office dataset contains three domains Amazon (A), Webcam (W), and DSLR (D), and each domain is composed of 31 classes. VisDA dataset contains two domains Sythetic and Real, and each domain consists of 12 classes. To create an open-set situation in Office, we follow~\cite{saito2018open, liu2019separate} to construct the source dataset by sampling from the first 21 classes in alphabetical order. Then, the target dataset is sampled from all 31 classes. As for VisDA, we choose the first 6 classes for source domain, and use all the 12 classes for target domain. We use ``A$\rightarrow$W'' to indicate the transfer from ``A'' domain to ``W'' domain.

For comparison, we choose three typical open-set DA approaches including Open-Set DA by BackPropagation OSBP~\cite{saito2018open} which employs an OpenMax classifier to recognize unknown classes and perform gradient flipping for open-set DA, Universal Adaptation Network UAN~\cite{you2020Universal} which utilize entropy and domain similarity to down-weight malign OOD data, and Separate To Adapt STA~\cite{liu2019separate} which utilizes SVM to separate the malign OOD data.

The experimental results are shown in Table~\ref{HOODtab:da}. Compared to the baseline methods, the proposed HOOD is largely benefited from the generated benign OOD data, which have two major strengths: (1) they resemble target domain data by having common styles, and (2) their labels are accessible as they share the same content as their corresponding source data. Therefore, through conducting supervised training such benign OOD data, the domain gap can be further mitigated, thus achieving better performance than baseline methods. Quantitative results show that HOOD can surpass other methods in most scenarios. Especially in VisDA, HOOD can outperform the second-best method with 6$\%$ improvement, which proves the effectiveness of HOOD in dealing with open-set DA.

\subsection{Performance Analysis}
\label{HOOD:performance}
\textbf{Ablation Study:} To verify the effectiveness of each module, we conduct an ablation study on three OOD applications by eliminating one component at a time. Specifically, our HOOD can be ablated into: ``w/o disentanglement'' which indicates removing the disentanglement loss in Equation~\eqref{HOODeq:final_elbo}, ``w/o benign OOD data'' which denotes training without benign OOD data, ``w/o malign OOD data'' which stands for discarding malign OOD data, and ``w/o both augmentations'' indicates training without both benign and malign OOD data. In OOD detection, we use CIFAR10 and LSUN as the ID and OOD dataset, respectively. In open-set SSL, we choose CIFAR10 with 400 labels for each class. As for open-set DA, we use VisDA dataset.

\begin{table}[t]
	\scriptsize
    \centering
	\caption{Ablation study on necessity of each module.}
	\label{HOODtab:ablation_study}
	\setlength{\tabcolsep}{5mm}
	\begin{tabular}{lccc}
		\toprule[1pt]
		Application         & OOD detection & Open-Set SSL & Open-Set DA \\ 
		\toprule[0.6pt]
		w/o disentanglement & $84.94\pm1.3$    & $82.55\pm1.1$   &$64.6\pm0.9$   \\
		w/o benign OOD data & $85.95\pm1.8$    & $83.32\pm2.0$   &$66.3\pm2.5$   \\
		w/o malign OOD data & $82.50\pm2.2$    & $85.40\pm0.8$   &$71.8\pm1.2$   \\
		w/o both augmentations & $80.83\pm0.8$ & $81.14\pm1.2$   &$65.4\pm1.2$   \\
		HOOD                & $\bm{86.12}\pm\bm{0.6}$    &$\bm{86.22}\pm\bm{2.7}$   &$\bm{72.4}\pm\bm{1.6}$   \\ 
		\toprule[1pt]
	\end{tabular}
\end{table}
The experimental results are shown in Table~\ref{HOODtab:ablation_study}. We can see each module influences the performance differently in three applications. First, we can see that the malign OOD data is essential for OOD detection, as it can act as unknown anomalies and reduce the overconfidence in unseen data. Then, benign OOD data can largely improve the learning performance in open-set SSL and open-set DA, as they can enforce the model to focus on the content feature for classification. Additionally, we can see that discarding both benign and malign OOD data shows performance degradation compared to both “w/o benign OOD data” and “w/o malign OOD data”. Therefore, our HOOD can correctly change the style and content, which can correspondingly benefit generalization tasks (such as open-set DA and open-set SSL) and detection tasks (such as OOD detection). Moreover, open-set DA relies more on the disentanglement than the rest two modules, owing to the disentanglement can exclude the style changing across different domains. Hence, our disentanglement can effectively eliminate the distribution change from different domains and help learn invariant features.

\begin{figure}
	\centering
	\begin{minipage}[t]{0.38\textwidth}
		\centering
		\includegraphics[width=0.9\linewidth]{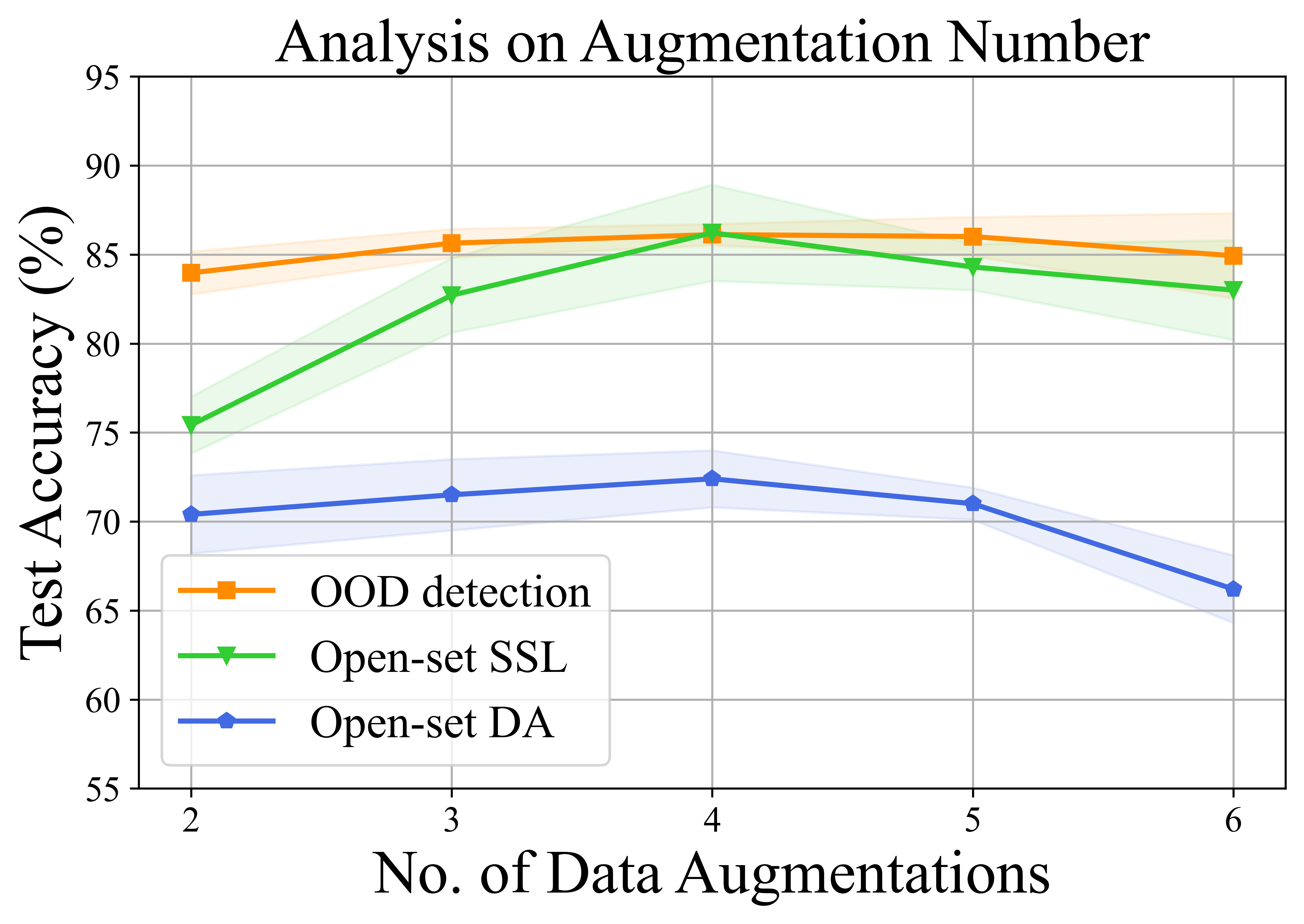}
	\end{minipage}
	\begin{minipage}[t]{0.58\textwidth}
		\centering
		\includegraphics[width=0.9\linewidth]{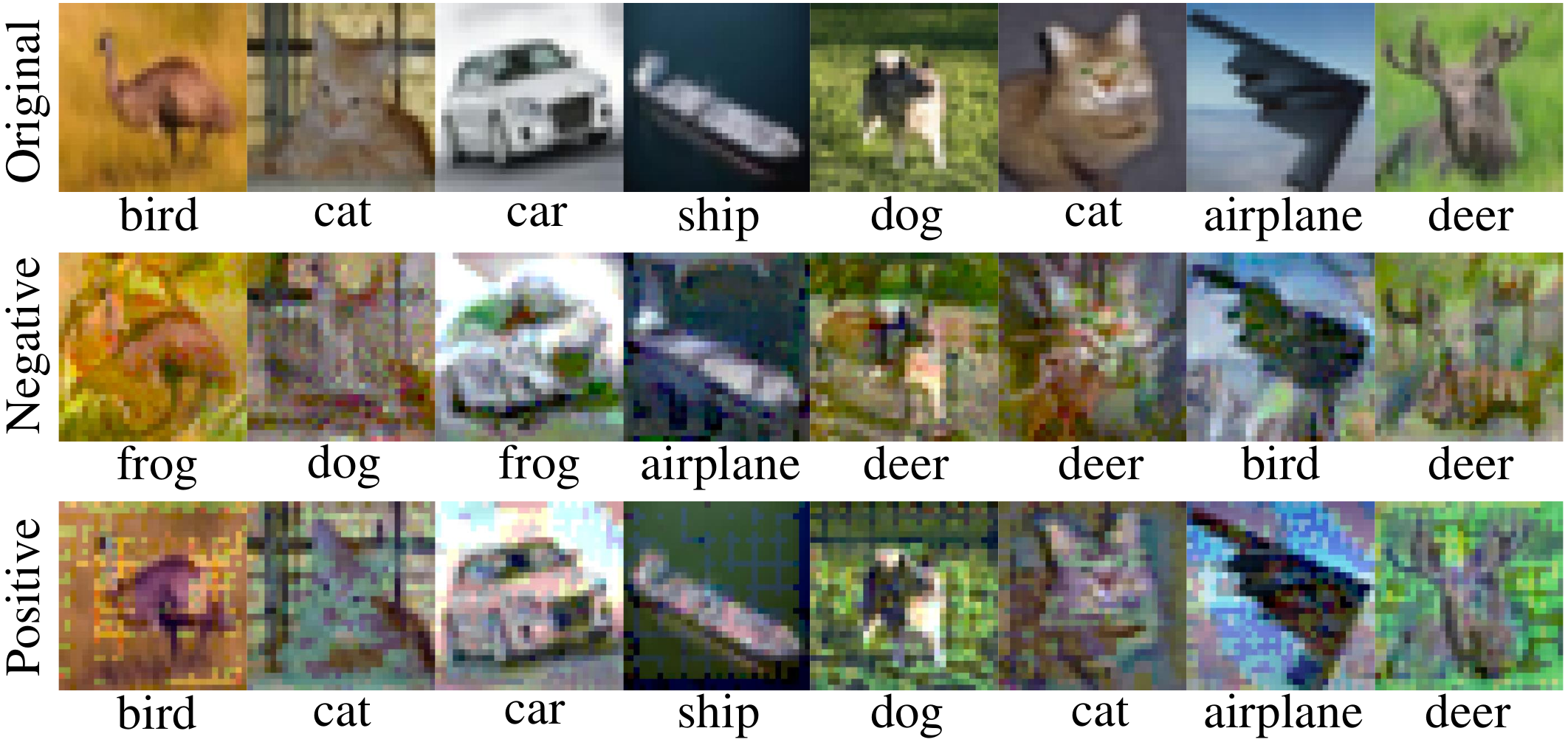}
	\end{minipage}
    \caption{Left: Augmentation number analysis. Right: CIFAR10 Visualization of our data augmentation.}
	\label{HOODfig:augmentation_number_and_visualization}
\end{figure}

\textbf{Analysis on Augmentation Number:}
Since HOOD does not introduce any hyper-parameter, the most influential setting is the number of data augmentation. To analyze its influence on the learning results, we vary the number of augmentations that are sampled from the RandAugment Pool~\cite{cubuk2020randaugment} from 2 to 6. The results are shown in Figure~\ref{HOODfig:augmentation_number_and_visualization} left. We can see that both too less and too many augmentations would hurt the results. This is because a small augmentation number would undermine the generalization to various styles; and a large augmentation number would increase the classification difficulty of the style branch, further making the disentanglement hard to achieve. Therefore, setting the augmentation number to 4 is reasonable.

\textbf{Visualization:}
Furthermore, to show the effect of our data augmentations, we visualize the augmented images by applying large perturbation magnitude ($4.7$)~\cite{tsipras2018robustness} in Figure~\ref{HOODfig:augmentation_number_and_visualization} right. The model prediction is shown below each image. We can see that the negative data augmentation significantly changes the content which is almost unidentifiable. However, positive data augmentation can still preserve most of the content information and only change the style of images. Therefore, the augmented data are tailor-designed for training a robust classifier.

\begin{figure}[t]
	\centering
	\includegraphics[width=0.65\linewidth]{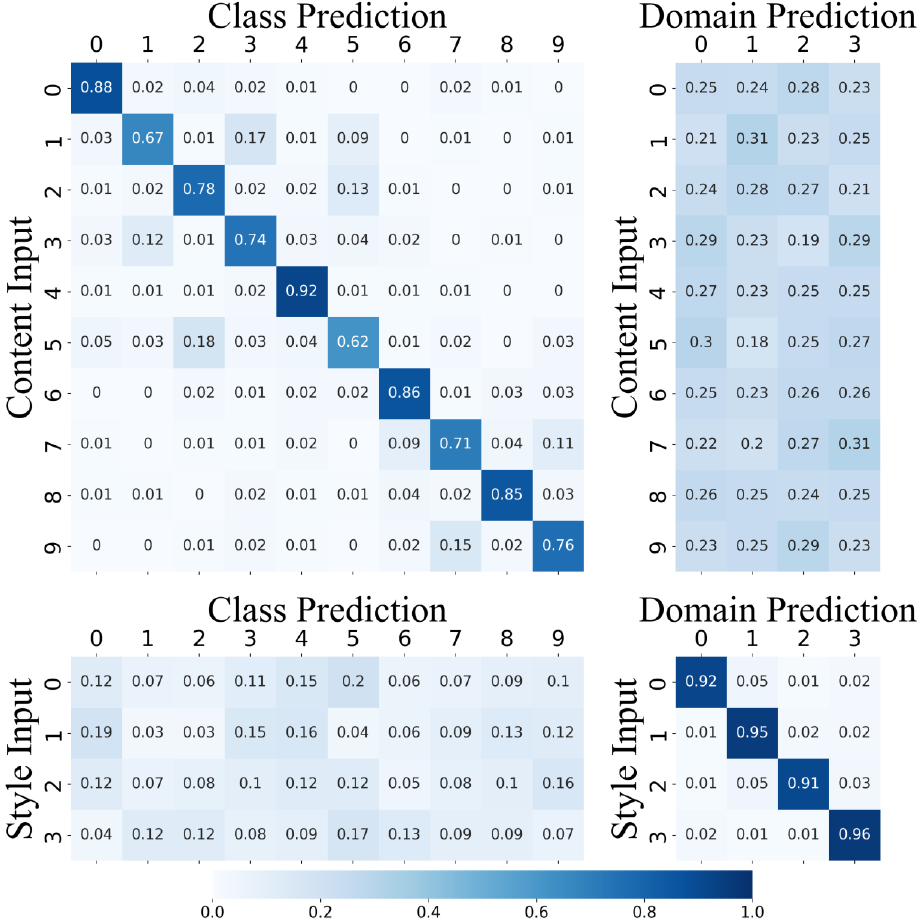}
	\caption{Illustration content ande style disentanglement on CIFAR10. The number in each cell denotes the prediction probability.}
	\label{HOODfig:disentanglement}
\end{figure}

\textbf{Content and Style Disentanglement:} 
To further testify that our disentanglement between content and style is effective, we select the latent variables from different content and style categories, and use the learned class and domain classifiers for cross-prediction. Specifically, there are four kinds of input-prediction types: content-class, content-domain, style-class, and style-domain. As we can see in Figure~\ref{HOODfig:disentanglement}, only the content features are meaningful for class prediction, and the same phenomenon goes for style input and domain prediction. However, neither of the style and content features can be identified by the class predictor and domain predictor, respectively. Therefore, we can reasonably conclude that our disentanglement between content and style is effectively achieved.

\begin{figure}[t]
	\centering
	\includegraphics[width=0.8\linewidth]{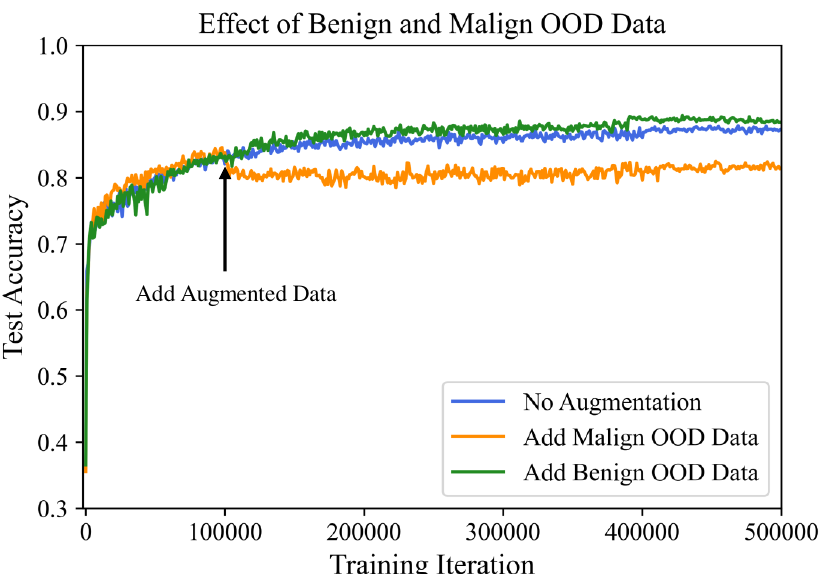}
	\caption{Effect of adding benign and malign OOD data into training.}
	\label{HOODfig:augmentation_effect}
\end{figure}

\textbf{Effect of Adding Benign and Malign OOD Data into Training:}
To give a illustrative comparison of adding benign OOD data and malign OOD data into training, we conduct experiments under the open-set SSL setting and separately augmenting benign OOD data and malign OOD data to compare their effects. Moreover, we conduct plain training as a baseline result which do not use either augmentations. The results are shown in Figure~\ref{HOODfig:augmentation_effect}. We can see that after adding augmented data, the effect of malign OOD data causes sudden performance degradation. On the contrary, benign OOD data can further improve the learning result compared to the plain training baseline. Which again shows that preserving content and augmenting style is beneficial, and eliminating content is harmful for generalization.

\begin{table}[t]
	\centering
	\caption{Averaged OOD scores on three applications.}
	\begin{tabular}{lcc}
		\toprule[1pt]
		\multirow{2}{*}{Application} & \multicolumn{2}{c}{OOD score} \\ \cline{2-3} 
		& Benign OOD data       & Malign OOD data      \\ 
		\midrule[0.6pt]
		OOD detection                & $0.16\pm0.3$     & $0.83\pm0.6$     \\
		Open-Set SSL                 & $0.08\pm0.5$     & $0.91\pm0.4$     \\
		Open-Set DA                  & $0.21\pm0.4$     & $0.88\pm0.3$     \\ 
		\toprule[1pt]
	\end{tabular}
	\label{HOODtab:oodscore}
\end{table}

\textbf{Analysis of OOD Score}
To show the effectiveness of identifying malign OOD data from benign OOD data, we test the performance of HOOD on three applications to observe the OOD scores of benign OOD data and malign OOD data and show the averaged OOD scores in Table~\ref{HOODtab:oodscore}. We can see that the OOD score produced by our one-vs-all classifier can clearly distinguish benign and malign OOD data during the test phase, which again validates the effectiveness of HOOD.

\begin{table}[t]
	\centering
	\caption{Execution efficiency comparisons on three applications.}
	\begin{tabular}{llllll}
		\toprule[1pt]
		\multicolumn{2}{c}{OOD Detection} & \multicolumn{2}{c}{Open-set SSL} & \multicolumn{2}{c}{Open-set DA} \\ 
		\midrule[0.6pt]
		\multicolumn{1}{l}{Method} & Time & \multicolumn{1}{l}{Method} & Time & \multicolumn{1}{l}{Method} & Time \\ \midrule[0.6pt]
		\multicolumn{1}{l}{Likelihood} & 6.2h & \multicolumn{1}{l}{DS3L} & 15.4h & \multicolumn{1}{l}{UAN} & 8.5h \\
		\multicolumn{1}{l}{OpenGAN} & 7.8h & \multicolumn{1}{l}{OpenMatch} & 10.5h & \multicolumn{1}{l}{STA} & 9.1h \\
		\multicolumn{1}{l}{HOOD} & 11.4h & \multicolumn{1}{l}{HOOD} & 13.7h & \multicolumn{1}{l}{HOOD} & 12.4h \\ 
		\toprule[1pt]
	\end{tabular}
	\label{HOODtab:execution}
\end{table}

\textbf{Execution Efficiency}
Additionally, to give a quantitative comparison on the execution efficiency of HOOD, here we provide the running time on 3090 GPU compared to some typical baseline methods. The results are shown in Table~\ref{HOODtab:execution}. Note that our method involves causal disentanglement as well as adversarial training, therefore, the training time is increased.

\section{Conclusion}
\label{HOOD:conclusion}
In this Chatper, we propose HOOD to effectively harness OOD examples. Specifically, we construct a SCM to disentangle content and style, which can be leveraged to identify benign and malign OOD data. Subsequently, by maximizing ELBO, we can successfully disentangle the content and style feature and break the spurious correlation between class and domain. As a result, HOOD can be more robust when facing distribution shifts and unseen OOD data. Furthermore, we augment the content and style through a novel intervention process to produce benign and malign OOD data, which can be leveraged to improve classification and OOD detection performance. Extensive experiments are conducted to empirically validate the effectiveness of HOOD on three typical OOD applications.

\chapter{Sharpness-Based Distribution Robust Optimization}
\label{cha:SharpDRO}
Robust generalization aims to tackle the most challenging data distributions, which are rare in the training set and contain severe noise, i.e., photon-limited corruptions. Common solutions, such as distributionally robust optimization (DRO), focus on the worst-case empirical risk to ensure low training error on the uncommon noisy distributions. However, due to the over-parameterized model being optimized on scarce worst-case data, DRO fails to produce a smooth loss landscape, thus struggling on generalizing well to the test set. Therefore, instead of focusing on the worst-case risk minimization, we propose SharpDRO by penalizing the sharpness of the worst-case distribution, which measures the loss changes around the neighborhood of learning parameters. Through worst-case sharpness minimization, the proposed method successfully produces a flat loss curve on the corrupted distributions, thus achieving robust generalization. Moreover, by considering whether the distribution annotation is available, we apply SharpDRO to two problem settings and design a worst-case selection process for robust generalization. Theoretically, we demonstrate that SharpDRO offers a strong convergence guarantee. Experimentally, we simulate photon-limited corruptions using the CIFAR10/100 and ImageNet30 datasets and demonstrate that SharpDRO exhibits a strong generalization ability against severe corruptions, outperforming well-known baseline methods by a significant margin.

\section{Introduction}
\label{SharpDRO:introduction}
Learning against corruptions has been a vital challenge in the practical deployment of computer vision models, as learning models are much more fragile to subtle noises than human perception systems~\cite{goodfellow2014explaining, hendrycks2016baseline, liu2021probabilistic}. During the training, the encountered corruptions are essentially perceived as a distribution shift, which would significantly hinder the prediction results~\cite{liang2017enhancing, long2015learning, tzeng2017adversarial, xia2019anchor, xia2020part,xia2023moderate}. Therefore, to mitigate the performance degradation, enhancing generalization to corrupted data distributions has drawn lots of attention~\cite{arjovsky2019invariant,sagawa2019distributionally}.

\begin{figure}
	\centering
	\includegraphics[width=0.7\linewidth]{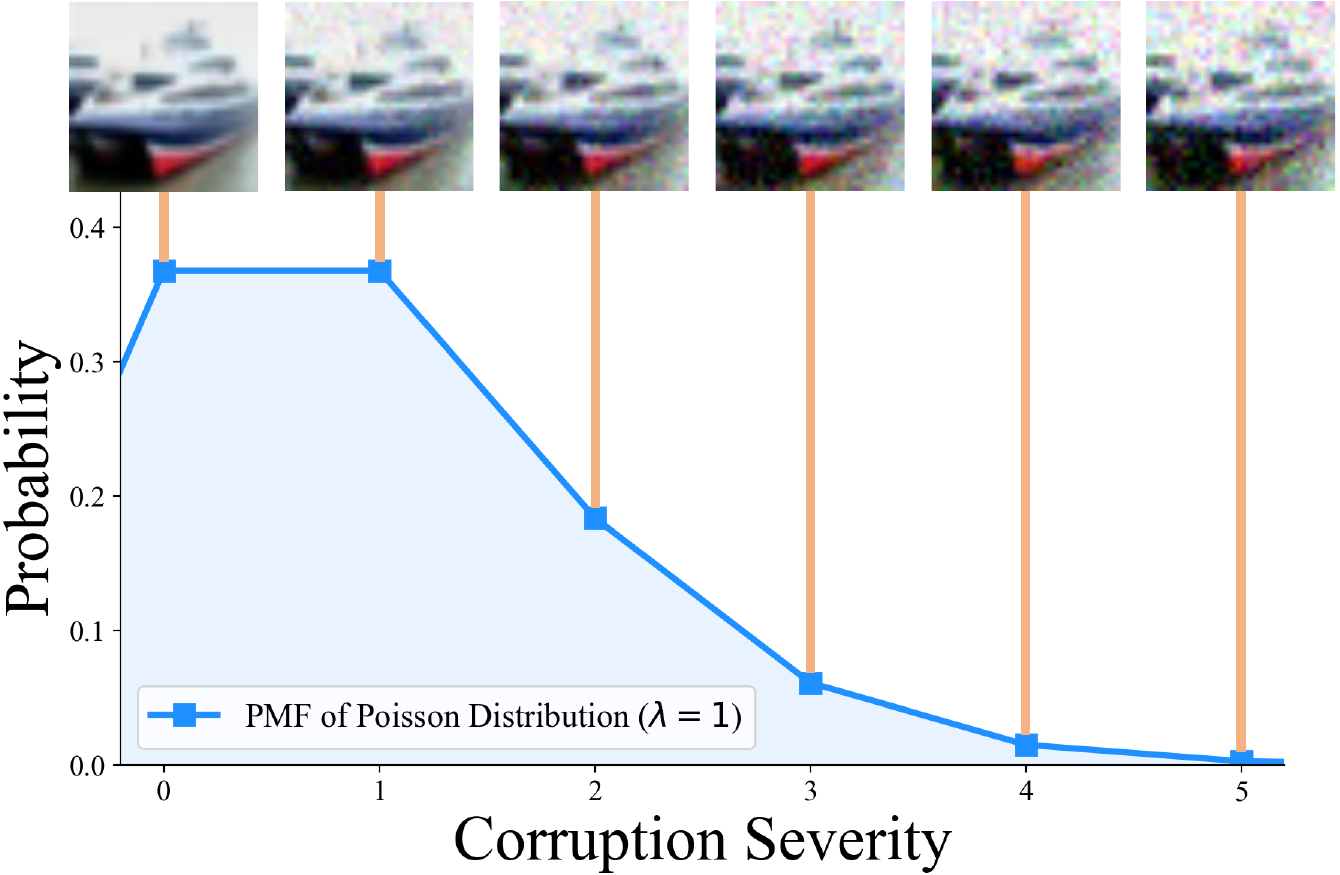}
	\caption{Illustration of photon-limited corruptions.}
	\label{SharpDROfig:poisson}
\end{figure}

In the real world, noise corruptions are commonly known as photon-limited imaging problems~\cite{ingle2021passive, li2021photon, luisier2010image, timmermann1999multiscale} which arise due to numerous small corruption photons arriving at an image sensor. Consequently, different numbers of captured photons would form different levels of corruption severity, further producing multiple data distributions and imposing varied impacts on learning models~\cite{hendrycks2019benchmarking}. Specifically, the encountered photon-limited corruption $\mathscr{E}$ is a composition of multiple noise photons $u$, which is triggered by some discrete factors with a certain probability during a time interval. For example, a photon $u$ can be triggered by each platform changing, redistribution, transmission, etc. More photons are captured, and severe corruption would be applied to the image. Therefore, the severity $s$ of the photon-limited corruptions $\mathscr{E}$ can be modeled by Poisson distribution, i.e., $s\sim P\left( s; \lambda \right) = \frac{{e^{-\lambda } \lambda ^s }}{{s!}}$, which is illustrated in Figure~\ref{SharpDROfig:poisson}. As a result, the real-world dataset is not completely composed of clean data, but contains corrupted data with various severities.

\begin{figure*}[t]
	\centering
	\includegraphics[width=\linewidth]{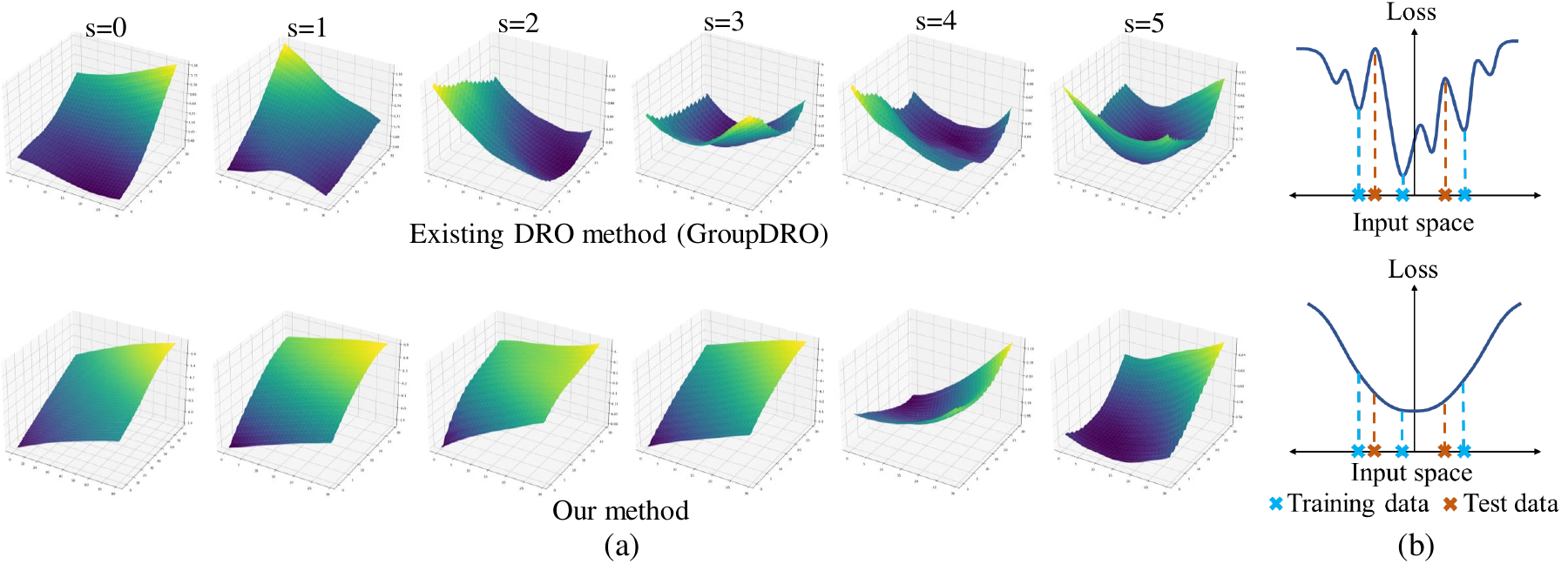}
	\caption{Illustration of our motivation. (a) Loss surface visualization of GroupDRO and the proposed SharpDRO. The columns from left to right stand for corrupted distributions with severity $s=0$ to $5$. (b) Illustration of why a sharp loss surface hinders generalization to test data.}
	\label{SharpDROfig:visualization}
\end{figure*}

Dealing with such a realistic problem by vanilla empirical risk minimization can achieve satisfactory average accuracy on the whole training set. However, due to the extremely limited number of severely corrupted data, the learning model would produce large training errors on the corrupted distributions, further hindering the robust performance under challenging real-world situations. A popular approach to achieve low error on the scarce corrupted data is distributionally robust optimization (DRO)~\cite{namkoong2016stochastic,sagawa2019distributionally,zhai2021doro, piratla2021focus, wang2022meta, wang2022improving}, which commonly optimizes the model parameter $\theta$ by optimizing:
\begin{equation}
	\min_{\theta \in \Theta} \sup_{Q \in \mathcal{Q}} \mathbb{E}_{(x, y)\sim Q}\left[\mathcal{L}(\theta; (x, y))\right],
	\label{SharpDROeq:general_dro}
\end{equation}
where $\mathcal{Q}$ denotes the uncertainty set that is utilized to estimate the possible test distribution. Intuitively, DRO assumes that $\mathcal{Q}$ consists of multiple sub-distributions, among which exists a worst-case distribution $Q$. By concentrating on the risk minimization of the worst-case distribution, DRO hopes to train a robust model that can deal with the potential distribution shift during the test phase. However, existing DRO methods usually leverage over-parameterized models to focus on a small portion of worst-case training data. Therefore, the worst-case data contaminated with severe corruption is highly possible to get stuck in sharp minima. As shown in the upper part of Figure~\ref{SharpDROfig:visualization} (a), a stronger corruption would cause the existing method to learn a sharper loss surface. Consequently, optimization via DRO fails to produce a flat loss landscape over the corrupted distributions, which leads to a large generalization gap between training and test set~\cite{keskar2017large,chaudhari2017entropy}.


To remedy this defect, in this Chapter, we propose the SharpDRO method to focus on learning a flat loss landscape of the worst-case data, which can largely mitigate the training-test generalization gap problem of DRO. Specifically, we adopt the largest loss difference formed by applying weight perturbation~\cite{foret2020sharpness,wu2020adversarial} to measure the sharpness of the loss function. Intuitively, a sharp loss landscape is sensitive to noise and cannot generalize well on the test set. On the contrary, a flat loss landscape produces consistent loss values and is robust against perturbations (Figure~\ref{SharpDROfig:visualization} (b)). By minimizing the sharpness, we can effectively enhance the generalization performance~\cite{keskar2017large, chaudhari2017entropy}. However, directly applying sharpness minimization on multiple distributions would yield poor results~\cite{cha2021swad}, as the computed sharpness could be influenced by the largest data distribution, and thus cannot generalize well to small corrupted data. Therefore, we only focus on worst-case sharpness minimization. In this way, as the lower part of Figure~\ref{SharpDROfig:visualization} (a) shows, SharpDRO successfully produces a flat loss surface, thus achieving robust generalization on the severely corrupted distributions.

In addition, identification of the worst-case distribution requires expensive annotations, which are not always practically feasible~\cite{liu2021just}. In this Chapter, we apply SharpDRO to solve two problem settings: (1) \textit{Distribution-aware robust generalization}, which assumes that distribution indices are accessible, and (2) \textit{Distribution-agnostic robust generalization}, where the distributions are no longer identifiable, making the worst-case data hard to find. Existing approaches, such as Just Train Twice (JTT), require two-stage training, which is rather inconvenient. To tackle this challenge, we propose a simple (Out-of-distribution) OOD detection~\cite{hendrycks2016baseline,huang2021universal, huang2022they,liang2017enhancing, liu2020energy, wang2022watermarking, wang2023out} process to detect the worst-case data, which can be further leveraged to enable worst-case sharpness minimization. Through constructing training sets according to the Poisson distributed noisy distribution using CIFAR10/100 and ImageNet30, we show that SharpDRO can achieve robust generalization results on both problem settings, surpassing well-known baseline methods by a large margin.

To sum up, our main contributions are threefold:
\begin{itemize}
	\item We proposed a sharpness-based DRO method that overcomes the poor worst-case generalization performance of distributionally robust optimization.
	
	\item We apply our SharpDRO to both distribution-aware and distribution-agnostic settings, which brings a practical capability to our method. Moreover, we propose an OOD detection approach to select worst-case data to enable robust generalization.
	
	\item Theoretically, we show that SharpDRO has a convergence rate of $\mathcal{O}(\frac{\kappa^2}{\sqrt{MT}})$. Empirically, we form a photon-limited corruption dataset that follows a Poisson distribution, and conduct extensive experiments to show a strong generalization ability of SharpDRO as well as its superiority to compared baseline methods.
\end{itemize}

In the following, we first briefly introduce the background and discuss the problem setting in section~\ref {SharpDRO:background}. Then, we specify our SharpDRO over two problem settings in Section~\ref{SharpDRO:method}. Moreover, we give a detailed optimization process and provide convergence analysis in Section~\ref{SharpDRO:optimization}. Further, we conduct extensive experiments to validate our SharpDRO in Section~\ref{SharpDRO:experiments}. At last, we conclude this Chapter in Section~\ref{SharpDRO:conclusion}.

\section{Robust Generalization Methods}
\label{SharpDRO:background}
Due to the practical significance of robust generalization, various approaches have been proposed to deal with distribution shift. Here, we briefly introduce three typical baseline methods, namely Invariant Risk Minimization, Risk Extrapolation, and GroupDRO.

\textbf{Invariant Risk Minimization (IRM)}~\cite{arjovsky2019invariant,chang2020invariant,creager2021environment} aims to extract the invariant feature across different distributions (also denoted as environments). Specifically, the learning model is separated into a feature extractor $G$ and a classifier $C$. IRM assumes an invariant model $C\circ G$ over various environments can be achieved if the classifier $C$ constantly stays optimal. Then, the learning objective is formulated as:
\begin{equation}
	\begin{aligned}
		&\min_{C^*\circ G} \big\{\mathcal{L}_{\text{IRM}}:=\sum_{e\in \mathcal{E}}\mathcal{L}^e(C^*\circ G)\big\} \\
		&\text{s. t.}\  C^*\in\arg\min_{G}\mathcal{L}^e(C\circ G), \text{for all}\  e \in \mathcal{E},
	\end{aligned}
	\label{SharpDROeq:irm}
\end{equation}
where $C^*$ stands for the optimal classifier, and $e$ denotes a specific environment from a given environmental set $\mathcal{E}$. By solving Equation~\eqref{SharpDROeq:irm}, the feature extractor $G$ can successfully learn invariant information without being influenced by the distribution shift between different environments.

\textbf{Risk Extrapolation (REx)}~\cite{krueger2021out} targets at generalization to out-of-distribution (OOD) environments. Inspired by the discovery that penalizing the loss variance across distributions helps achieve improved performance on OOD generalization, REx proposes to optimize via:
\begin{equation}
	\min_{\theta\in\Theta}\big\{ \mathcal{L}_{\text{REx}}:=\sum_{e\in \mathcal{E}}\mathcal{L}^e(\theta) + \beta Var(\mathcal{L}^e,..., \mathcal{L}^m)\big\},
	\label{SharpDROeq:rex}
\end{equation}
where $\beta$ controls the penalization level. Intuitively, REx seeks to achieve risk fairness among all $m$ training environments, so as to increase the similarity of different learning tasks. As a result, the training model can capture the invariant information that helps generalize to unseen distributions.

\textbf{GroupDRO}~\cite{sagawa2019distributionally, hashimoto2018fairness, piratla2021focus} deal with the situation when the correlation between class label $y$ and unknown attribute $a$ differs in the training and test set. Such a difference is called spurious correlation, which could seriously misguide the model prediction. As a solution, GroupDRO considers each combination of class and attribute as a group $g$. By conducting risk minimization through:
\begin{equation}
	\min_{\theta \in \Theta} \big\{\mathcal{L}_{\text{GroupDRO}}:=\max_{g} \mathbb{E}_{(x, y)\sim P_g}\left[\mathcal{L}(\theta; (x, y))\right]\big\}.
	\label{SharpDROeq:group_dro}
\end{equation}
The worst-case group from the distribution $P_g$, which commonly holds spurious correlation, is emphasized, thus breaking the spurious correlation.

\textbf{Discussion:} IRM and REx both focus on learning invariant knowledge across various environments. However, when the training set contains extremely imbalanced noisy distributions, as shown in Figure~\ref{SharpDROfig:poisson}, the invariant learning results would be greatly misled by the most dominating distribution. Thus, the extracted invariant feature would be questionable for generalization against distribution shift. Although emphasizing the risk minimization of worst-case data via GroupDRO can alleviate the imbalance problem, its generalization performance is still sub-optimal when facing novel test data. However, SharpDRO can not only focus on the uncommon corrupted data but also effectively improve the generalization performance on the test set by leveraging worst-case sharpness minimization.

Our investigated problem is closely related to OOD generalization which is a broad field that contains many popular research topics, such as Domain Generalization~\cite{carlucci2019domain, peng2019moment, qiao2020learning, shu2021open, mahajan2021domain, muandet2013domain, huang2023harnessing, zhang2022towards}, Causal Invariant Learning~\cite{arjovsky2019invariant, krueger2021out, li2018deep, yang2021causalvae, scholkopf2021toward, yue2021transporting}. Generally, existing works mainly study two types of research problems: (1) mitigating domain shift between the training and test datasets; and (2) breaking the spurious correlation between causal factors. However, as generalization against corruptions \textbf{does not introduce any domain shift or spurious correlation}, such a problem cannot be naively solved by domain generalization methods or causal representation learning techniques. Therefore, in this Chapter, we focus on complementing this rarely-explored field and propose SharpDRO to enforce robust generalization against corruption. In the next section, we elaborate on the methodology of SharpDRO.

\begin{figure*}
	\begin{minipage}[t]{0.32\textwidth}
		\centering
		\includegraphics[width=\linewidth]{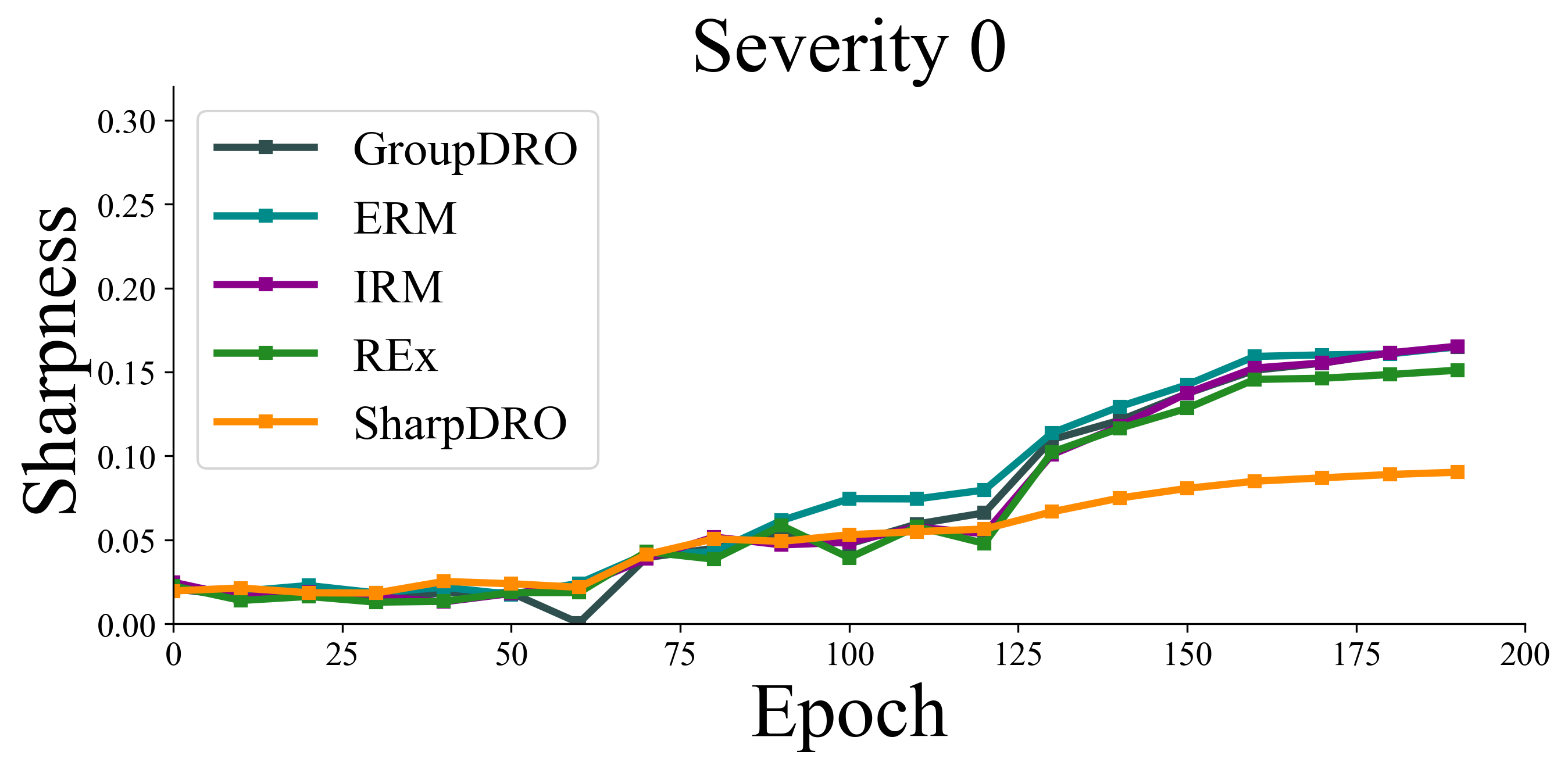}
	\end{minipage}
	\begin{minipage}[t]{0.32\textwidth}
		\centering
		\includegraphics[width=\linewidth]{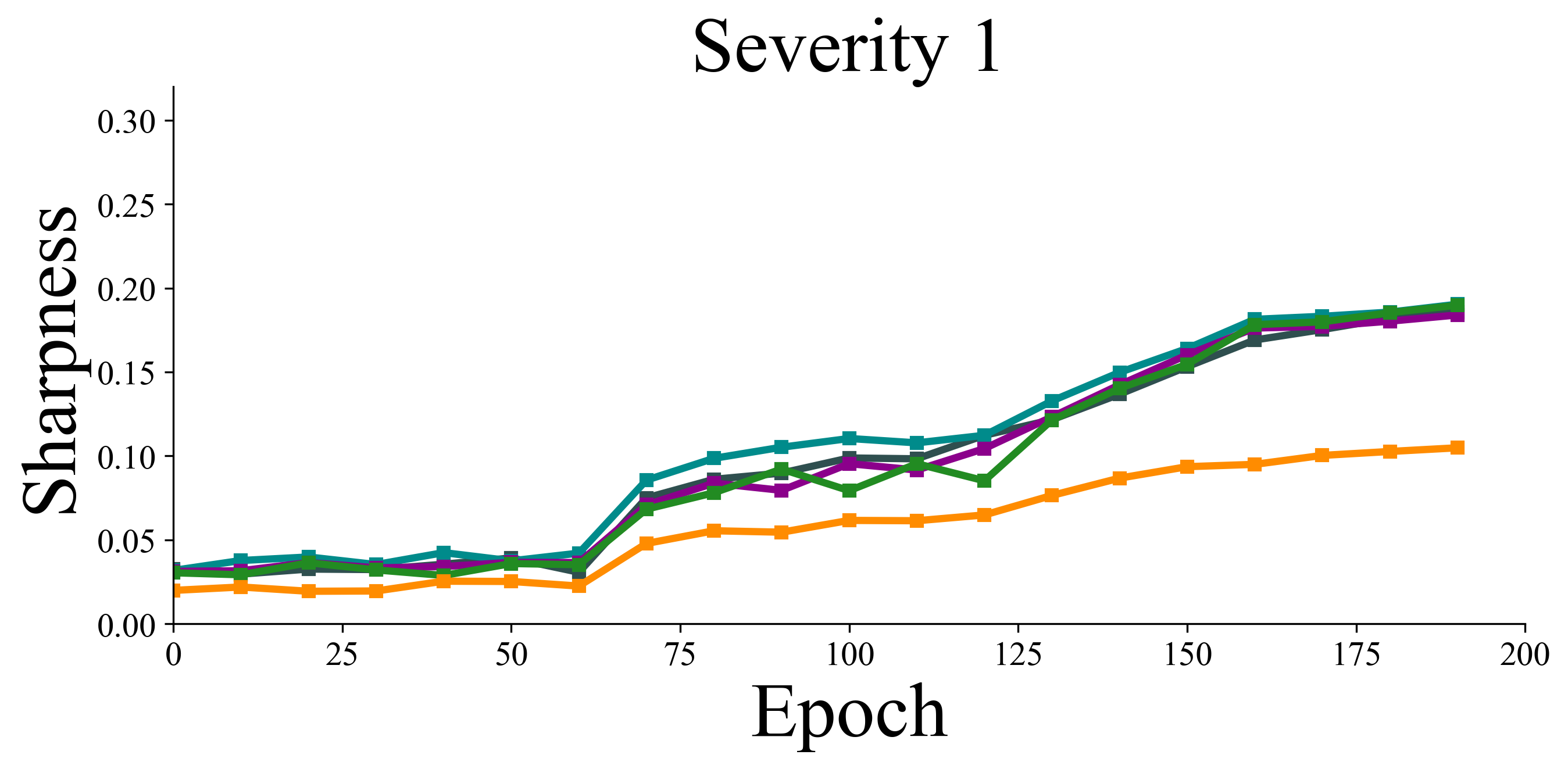}
	\end{minipage}
	\begin{minipage}[t]{0.32\textwidth}
		\centering
		\includegraphics[width=\linewidth]{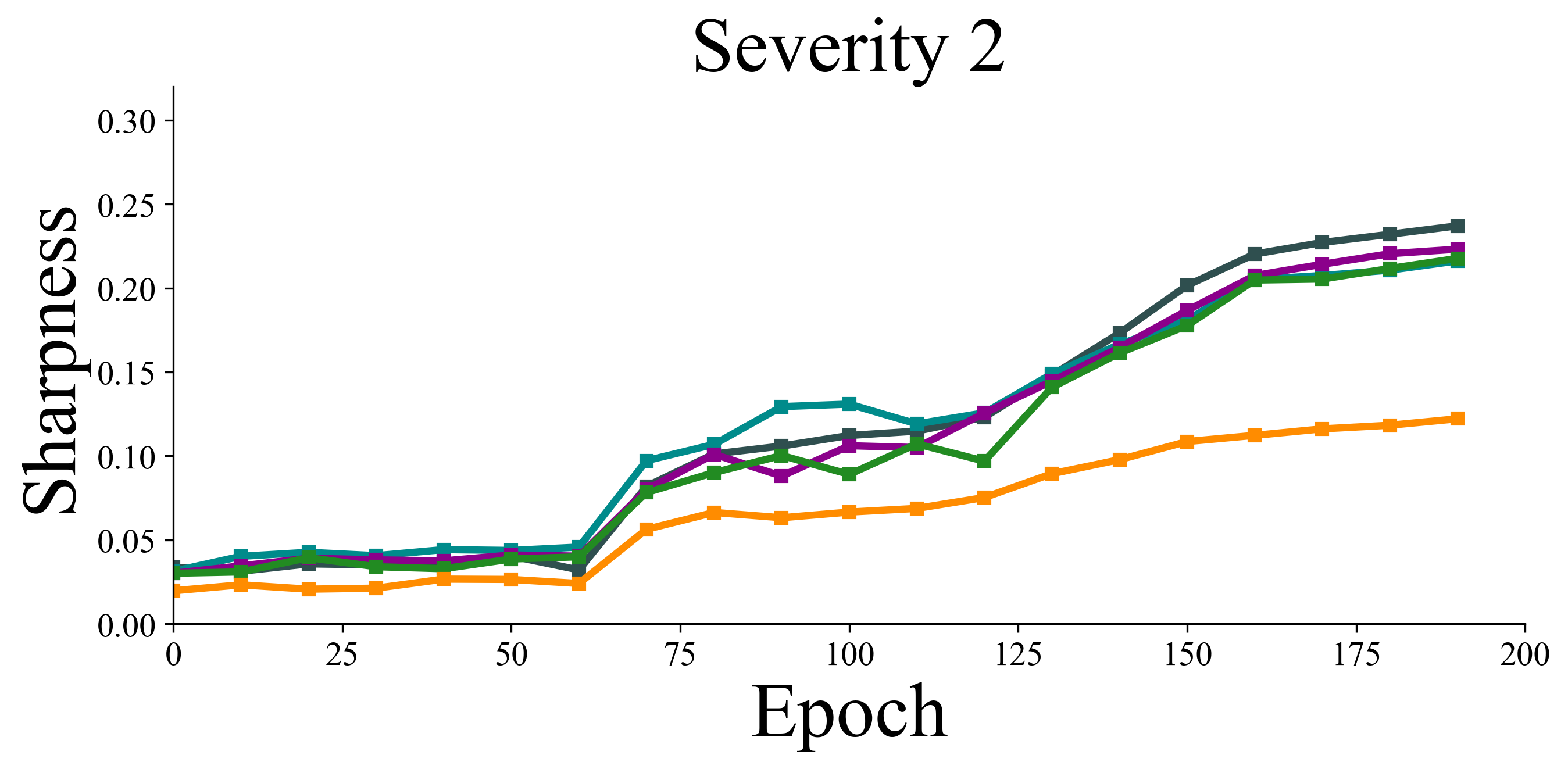}
	\end{minipage}\\
	\begin{minipage}[t]{0.32\textwidth}
		\centering
		\includegraphics[width=\linewidth]{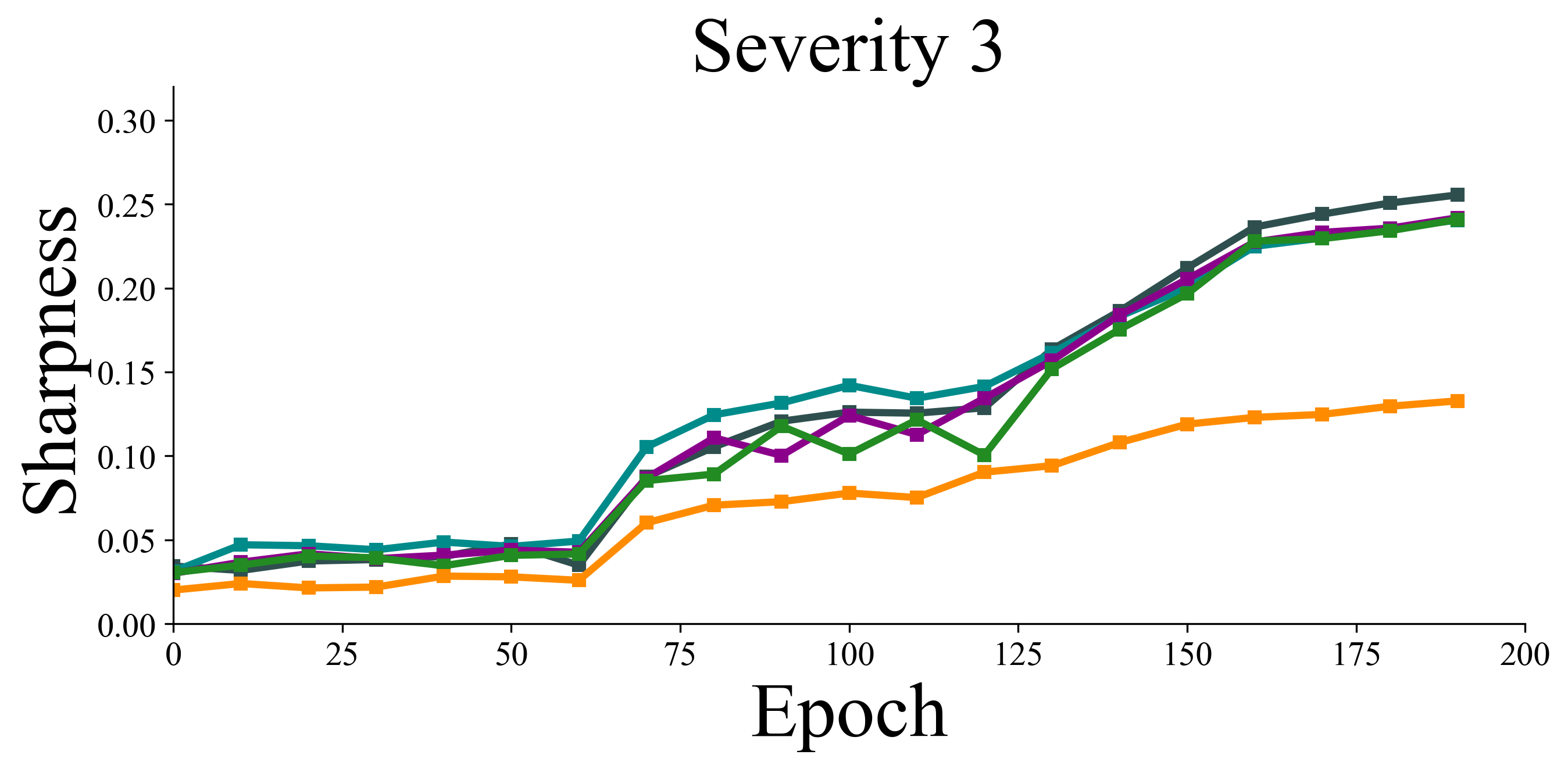}
	\end{minipage}
	\begin{minipage}[t]{0.32\textwidth}
		\centering
		\includegraphics[width=\linewidth]{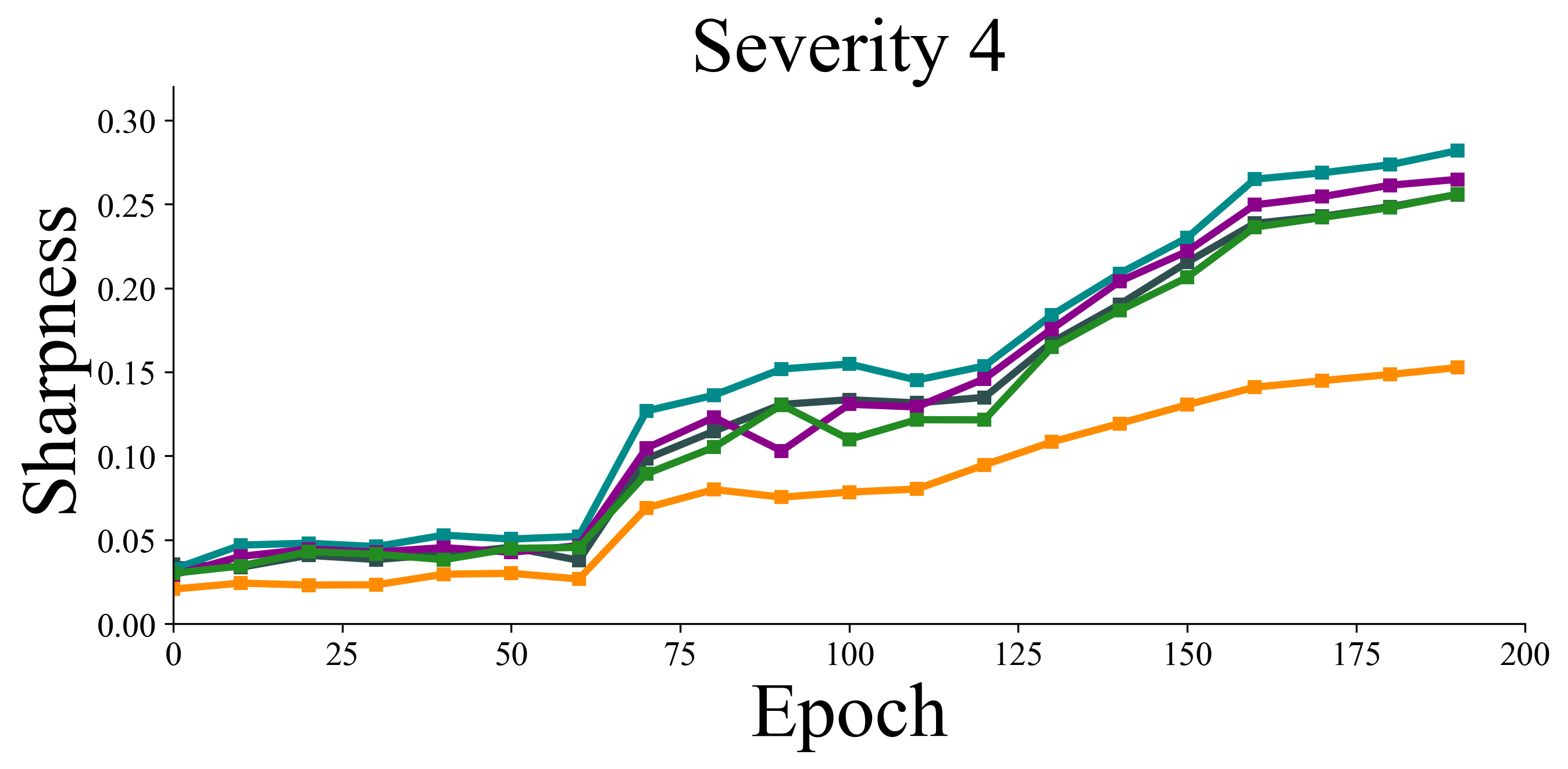}
	\end{minipage}
	\begin{minipage}[t]{0.32\textwidth}
		\centering
		\includegraphics[width=\linewidth]{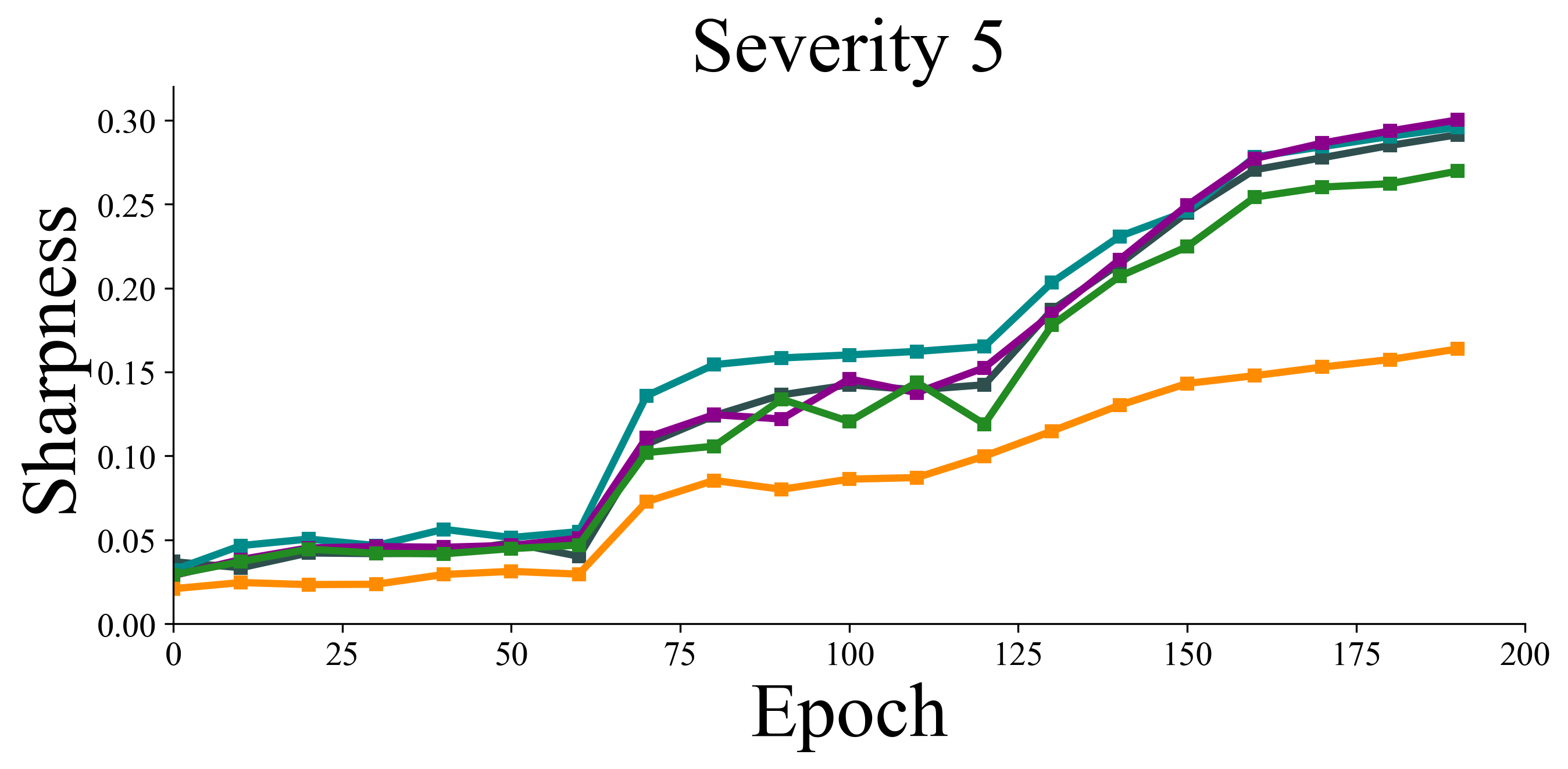}
	\end{minipage}
	\caption{\small Sharpness during networking training on clean ($s=0$) and corrupted distributions ($s=1$ to $5$).}
	\label{SharpDROfig:sharpness}
\end{figure*}

\section{Methodology}
\label{SharpDRO:method}

In robust generalization problems, we are given a training set $\mathcal{D}^{\text{train}}$ containing $n$ image examples, each example $x\in\mathcal{X}$ is given a class label $y\in\mathcal{Y}=\left\{1, 2, ..., c\right\}$. Moreover, the training set is corrupted by a certain type of noise whose severity $s$ follows a Poisson distribution $P(s; \lambda)$. Here we assume $\lambda=1$, which indicates that the mean number of noise photons $u$ that occurred during a time interval is $1$. Therefore, the distribution $P$ of the whole training set is composed of $S$ sub-distributions $P_s, s\in\left\{1, 2, ..., S\right\}$ with varied levels of corruption. Our goal is to learn a robust model $\theta \in \Theta$ that can achieve good generalization performance on challenging data distributions $P_s$ with large severity.

The general objective of SharpDRO is formulated as:
\begin{equation}
	\begin{aligned}
		\min_{\theta} \left\{\mathcal{L}_{\text{SharpDRO}}:= \mathbb{E}_{(x, y)\sim Q}\left[\mathcal{L}(\theta; (x, y))\right] + \mathbb{E}_{(x, y)\sim Q}\left[\mathcal{R}(\theta; (x, y))\right]\right\},
	\end{aligned}
	\label{SharpDROeq:SharpDRO}
\end{equation}
where the first term denotes the risk minimization using the loss function $\mathcal{L}$, meanwhile, a worst-case distribution $Q$ is selected based on the model prediction. The second term $\mathcal{R}$ indicates the sharpness minimization, which aims to maximally improve the generalization performance on the worst-case distribution $Q$. Specifically, as shown in Figure~\ref{SharpDROfig:sharpness}, the sharpness gradually increases as the corruption severity increases. Therefore, to accomplish robust generalization, we are motivated to emphasize the worst-case distribution. As a result, we can produce much smaller sharpness compared to other methods, especially on severely corrupted distributions.

In the following, we first introduce worst-case sharpness for robust generalization. Then, we demonstrate worst-case data selection on two scenarios. Finally, we provide a detailed optimization process and convergence analysis.

\subsection{Sharpness for Robust Generalization}

The main challenge of robust generalization is that the training distribution is extremely imbalanced, as shown in Figure~\ref{SharpDROfig:poisson}. The performance on the abundant clean data is quite satisfactory, but robustness regarding the corrupted distribution is highly limited, owing to the severe disturbance of corruption as well as the insufficiency of noisy data. To enhance the generalization performance, we leverage sharpness to fully exploit the worst-case data. Specifically, sharpness~\cite{foret2020sharpness, kim2022fisher, liu2022towards,wu2020adversarial,zheng2021regularizing} is measured by the largest loss change when model weight $\theta$ is perturbed with $\epsilon$, formally:
\begin{equation}
	\mathcal{R}:=\max_{\|\epsilon\|_2\le\rho}\{\mathcal{L}(\theta+\epsilon; (x, y)) - \mathcal{L}(\theta; (x, y))\},
	\label{SharpDROeq:sharpness}
\end{equation}
where $\rho$ is a scale parameter to control the perturbation magnitude. By supposing $\epsilon$ is small enough, we can have:
\begin{equation}
	\mathcal{L}(\theta+\epsilon) - \mathcal{L}(\theta)\approx\nabla\mathcal{L}(\theta)\epsilon.
\end{equation}
Further, we hope to obtain the largest loss change to find the optimal weight perturbation $\epsilon^*$, which is be computed as:
\begin{equation}
	\epsilon^*:=\arg\max_{\|\epsilon\|_2\le\rho}{\nabla\mathcal{L}(\theta)\epsilon}.
\end{equation}
By following dual norm problem, the optimal $\epsilon^*$ can be solved as $\rho\sign(\nabla\mathcal{L}(\theta))$~\cite{foret2020sharpness}, which is essentially the $\infty$-norm of the gradient $\nabla\mathcal{L}$ multiplied with a scale parameter $\rho$. Hence, common sharpness minimization aims to minimize:
\begin{equation}
	\mathbb{E}_{(x, y)\sim Q}\mathcal{R}:=\mathcal{L}(\theta+\rho\sign(\nabla\mathcal{L}(\theta; (x, y)))) - \mathcal{L}(\theta; (x, y)).
	\label{SharpDROeq:sharpness_minimization}
\end{equation}
The intuition is that the perturbation along the gradient norm direction increases the loss value significantly. When training on corrupted distributions, the scarce noisy data scatter sparsely in the high-dimensional space. As a consequence, the neighbor of each datum could not be sufficiently explored, thus producing a sharp loss curve. During testing, the unseen noisy data is likely to fall on an unexplored point with a large loss, further causing inaccurate model predictions.

Therefore, instead of directly applying sharpness minimization on the whole dataset, which leads to poor generalization performance~\cite{cha2021swad} (as demonstrated in Section~\ref{SharpDRO:ablation_study}), we focus on sharpness minimization over the worst-case distribution $Q$. By conducting the worst-case sharpness minimization, we can enhance the flatness of our classifier. Consequently, when predicting unknown data during the test phase, a flat loss landscape is more likely to produce a low loss than a sharp one; hence, our SharpDRO can generalize better than other DRO methods. However, the robust performance largely depends on the worst-case distribution $Q$, so next, we explain our worst-case data selection.

\subsection{Worst-Case Data Selection}
Generally, the worst-case data selection focuses on finding the most uncertain data distribution $Q$ from the uncertainty set $\mathcal{Q}$, which is an $f$-divergence ball from the training distribution $P$~\cite{ben2013robust, duchi2018learning, hu2018does}. Most works assume each distribution is distinguishable from the other. However, when the distribution index is not available, it would be very hard to select worst-case data. In this section, we investigate two situations: distribution-aware robust generalization and distribution-agnostic robust generalization.

\subsubsection{Distribution-Aware Robust Generalization}
When given annotations to denote different severities of corruptions, the image data $x$ is paired with class label $y$ and distribution index $s$. Then, the worst-case distribution $Q$ can be found by identifying the sub-distribution $P_s\in P$ that yields the largest loss. Hence, we can optimize through:
\begin{align}
	\min_{\theta} & \Big\{\max_{\omega_s; \atop Q = \{\omega_s P_s\}_{s=1}^S} \big\{\sum_{(x_i, y_i)\in Q}\left[\mathcal{L}(\theta,\omega_s; (x_i, y_i))\right]\big\} + \sum_{(x_i, y_i)\in Q}\left[\mathcal{R}(\theta,\omega_s; (x_i, y_i))\right]\Big\}, \label{SharpDROeq:distribution_aware}
\end{align}
where $\omega_s$ belongs to a $(S-1)$-dimensional probability simplex. The first term simply recovers the learning target of GroupDRO~\cite{sagawa2019distributionally,hu2018does} and helps find the worst-case distribution $Q$. Then, by emphasizing the selected $P_s$, the second sharpness minimization term can act as a sharpness regularizer. As a result, SharpDRO can learn a flatter loss surface on the worst-case data, thus generalizes better compared to GroupDRO, as discussed in Section~\ref{SharpDRO:experiments}.

\subsubsection{Distribution-Agnostic Robust Generalization}
Due to the annotations being extremely expensive in the real world, a practical challenge is how to learn a robust model without a distribution index. Unlike JTT~\cite{liu2021just}, which trains the model through two stages, we aim to solve this problem more efficiently by detecting the worst-case data during network training. As the corrupted data essentially lie out-of-distribution, we are motivated to conduct OOD detection~\cite{hendrycks2019deep, li2022out, liang2017enhancing, liu2020energy, wei2022mitigating} to find the worst-case data.

Particularly, we re-utilize the weight perturbation $\epsilon^*$ to compute an OOD score:
\begin{equation}
	\omega_i = \max f(\theta; (x_i)) - \max f(\theta+\epsilon^*; (x_i)),
	\label{SharpDROeq:ood_score}
\end{equation}
where $f(\cdot)$ stands for the $c$-dimensional label prediction in the label space, whose maximum value is considered as prediction confidence. Intuitively, as the model is much more robust to the clean distribution than the corrupted distribution, the prediction of clean data usually exhibits more stability than scarce noisy data when facing perturbations. Hence, if an example comes from a rarely explored distribution, its prediction certainty would deviate significantly from the original value, thus producing a large OOD score, as shown in Section~\ref{SharpDRO:ood_score}. Note that the major difference is that we target generalization on worst-case data, but OOD detection aims to exclude OOD data.

To this end, we can construct our worst-case dataset as:
\begin{equation}
    Q:=\left\{\sum_{i=1}^M \bar{\omega}_i\cdot(x_i, y_i): \bar{\omega}_i=\frac{\omega_i}{\frac{1}{M}\sum_{i=1}^M \omega_i}\right\},
\end{equation}
where normalization on $\omega_i$ is performed simultaneously. Then, the learning target of the distribution-agnostic setting becomes:
\begin{align}
		\min_{\theta}& \Big\{\max_{\bar{\omega}_i} \big\{\sum_{(x_i, y_i)\in Q}\left[\mathcal{L}(\theta,\bar{\omega}_i; (x, y))\right]\big\} +  \sum_{(x_i, y_i)\in Q}\left[\mathcal{R}(\theta,\bar{\omega}_i; (x, y))\right]\Big\}, \label{SharpDROeq:distribution_agnostic}
\end{align}
Therefore, the worst-case data can be selected by focusing on the examples with large OOD scores. In this way, our sharpDRO can be successfully deployed into the distribution-agnostic setting to ensure robust generalization, whose effectiveness is demonstrated by quantitative and qualitative results in Sections~\ref{SharpDRO:distribution_agnostic} and~\ref{SharpDRO:ood_score}. Next, we give details about implementing SharpDRO.

\begin{algorithm}[t]
    \caption{Optimization process of SharpDRO}
    \label{SharpDROalg:SharpDRO}
    \begin{algorithmic}[1]
        \STATE Training set $\mathcal{D}^{\text{train}}=\{x_i, y_i\}_{i=1}^M$ containing Poisson distributed noisy corruptions;
        \STATE Model parameter $\theta \gets \theta_0$;
        \STATE Weighting parameter $\omega \gets \omega_0$;
        \STATE Learning rate: $\eta_{\theta}$, $\eta_{\omega}$.
        
        \FOR{$t = 0,1, \ldots,T-1$}
            \IF{Distribution-aware}
                \STATE {\color{black!60} $\triangleright$ Loss maximization via optimizing $\omega_{t+1}$}
                \STATE $\omega_{t+1} := \arg\max_{\omega} \left\{\mathbb{E}_{(x, y)\sim \omega P_s} \left[\mathcal{L}(\theta_{t},\omega; (x, y))\right]\right\}$;
            \ELSIF{Distribution-agnostic}
                \STATE {\color{black!60} $\triangleright$ OOD detection for computing $\omega_{t+1}$}
                \STATE Update $\omega_{t+1}$ via Equation~\eqref{SharpDROeq:ood_score};
            \ENDIF
            
            \STATE {\color{black!60} $\triangleright$ Optimize variable $\theta$}
            \STATE $\theta_{t+1} = \arg\min_{\theta} \Big\{\mathbb{E}_{(x,y)\sim \omega_t P} [\mathcal{L}(\theta,\omega_t) + \mathcal{R}(\theta,\omega_t)]\Big\}$
        \ENDFOR
    \end{algorithmic}
\end{algorithm}

\subsection{Optimization for SharpDRO}
\label{SharpDRO:optimization}
In both distribution-aware and distribution-agnostic scenarios, the worst-case data distribution is identified using the distribution weighting parameter $\omega_s$ and OOD score $\omega$, respectively. Intuitively, their effect is similar: finding the worst data distribution that yields the maximum loss. Therefore, without loss of generality, we consider the maximization in Equations~\eqref{SharpDROeq:distribution_aware} and~\eqref{SharpDROeq:distribution_agnostic} as the optimization on the same weighting parameter $\omega$ and the samples $(x,y)$ can be considered i.i.d. from $\mathcal{Q}$ weighted by $\omega$ to compute the loss value $\mathcal{L}$. Moreover, the sharpness regularization can be reformulated in the same way as Equation~\eqref{SharpDROeq:sharpness} by including $\omega$: $\mathcal{R}(\theta,\omega;(x,y))=\max_{\|\epsilon\|_2\leq \rho}\{\mathcal{L}(\theta+\epsilon,\omega;(x,y))-\mathcal{L}(\theta,\omega;(x,y))\}$. Therefore, our general learning objective can be formulated as a bi-level optimization:
\begin{align}
& \min_{\theta} \mathbb{E}_{(x, y)\sim Q}\left[\mathcal{L}(\theta,\omega^{*}; (x, y))\right] + \mathcal{R}(\theta,\omega^{*}; (x, y))   \\
& \qquad \text{s. t.}\ \omega^{*} = \arg\max_{\omega} \mathbb{E}_{(x, y)\sim Q}\left[\mathcal{L}(\theta,\omega; (x, y))\right].
\end{align}
The optimization process is shown in Algorithm~\ref {SharpDROalg:SharpDRO}. Specifically, we first update the weighting parameter $\omega$ based on the empirical risk term $\mathcal{L}$ using stochastic gradient ascent. Then, by leveraging the updated $\omega$, we optimize the general objective, which contains both risk minimization of $\mathcal{L}$ and worst-case sharpness minimization of $\mathcal{R}$. We iterate these processes until convergence, hoping to minimize the risk on the target loss function $\mathcal{L}$ with the worst-case data distribution.

\textbf{Convergence Analysis:}
	First we give some brief notations:
    \begin{equation}
        \mathbb{L}(\theta,\omega):=\mathbb{E}_{(x,y)\sim Q}\mathcal{L}(\theta,\omega;(x,y)).
    \end{equation}
    The worst-case data distribution, which has the maximum loss, is denoted by $\omega^*(\theta):= \arg\max_{\omega}\mathbb{L}(\theta,\omega)$. We can obtain the convergence to a stationary point of
    \begin{equation}
        \mathbb{L}^*(\theta):=\max_{\omega}\mathbb{L}(\theta,\omega)=\mathbb{L}(\theta,\omega^*(\theta))
    \end{equation}
    by averaged gradient 
	$\frac{1}{T}\sum_{t=0}^{T-1}\mathbb{E}\|\nabla \mathbb{L}^*(\theta_t)\|^2$.

\begin{table*}[t]
    \scriptsize
    \centering
    \caption{Quantitative comparisons on distribution-aware robust generalization setting. Averaged accuracy ($\%$) with standard deviations is computed over three independent trials.}
    \setlength{\tabcolsep}{1.2mm}
    \renewcommand{\arraystretch}{1.1}
    \label{SharpDROtab:distribution_aware}
    \begin{tabular}{ccl cccccc}
        \toprule
        \multirow{2}{*}{Dataset} & \multirow{2}{*}{Type} & \multirow{2}{*}{Method} & \multicolumn{6}{c}{Corruption Severity} \\
        \cmidrule(lr){4-9}
        &  &  & 0 & 1 & 2 & 3 & 4 & 5 \\ 
        \midrule
        \multirow{10}{*}{\rotatebox{90}{CIFAR10}} & \multirow{5}{*}{\rotatebox{90}{Gaussian}} 
          & ERM      & $90.9\pm0.02$ & $89.2\pm0.02$ & $86.4\pm0.03$ & $85.9\pm0.01$ & $83.5\pm0.01$ & $78.8\pm0.01$ \\
        & & IRM      & $91.8\pm0.01$ & $90.3\pm0.01$ & $89.5\pm0.01$ & $86.7\pm0.02$ & $81.8\pm0.02$ & $80.0\pm0.02$ \\
        & & REx      & $91.3\pm0.03$ & $89.5\pm0.02$ & $88.1\pm0.02$ & $86.7\pm0.02$ & $83.3\pm0.01$ & $80.5\pm0.02$ \\
        & & GroupDRO & $90.2\pm0.03$ & $89.1\pm0.02$ & $88.4\pm0.04$ & $84.3\pm0.01$ & $83.0\pm0.02$ & $78.2\pm0.02$ \\
        & & SharpDRO & $\mathbf{93.1\pm0.05}$ & $\mathbf{92.2\pm0.03}$ & $\mathbf{91.4\pm0.03}$ & $\mathbf{89.2\pm0.04}$ & $\mathbf{87.1\pm0.03}$ & $\mathbf{84.5\pm0.03}$ \\ 
        \cmidrule(lr){2-9}
        & \multirow{5}{*}{\rotatebox{90}{Shot}} 
          & ERM      & $\mathbf{92.5\pm0.02}$ & $91.1\pm0.02$ & $89.9\pm0.01$ & $85.6\pm0.03$ & $85.7\pm0.01$ & $78.8\pm0.01$ \\
        & & IRM      & $90.4\pm0.01$ & $90.3\pm0.02$ & $89.4\pm0.02$ & $86.3\pm0.01$ & $84.3\pm0.02$ & $79.1\pm0.02$ \\
        & & REx      & $91.1\pm0.02$ & $90.6\pm0.02$ & $90.2\pm0.03$ & $86.8\pm0.02$ & $84.7\pm0.02$ & $80.5\pm0.01$ \\
        & & GroupDRO & $92.2\pm0.01$ & $91.4\pm0.01$ & $89.4\pm0.02$ & $84.0\pm0.01$ & $84.7\pm0.02$ & $78.3\pm0.01$ \\
        & & SharpDRO & $91.7\pm0.04$ & $\mathbf{93.3\pm0.04}$ & $\mathbf{92.5\pm0.05}$ & $\mathbf{91.0\pm0.02}$ & $\mathbf{88.2\pm0.03}$ & $\mathbf{83.9\pm0.04}$ \\
        \midrule
        \multirow{10}{*}{\rotatebox{90}{CIFAR100}} & \multirow{5}{*}{\rotatebox{90}{Gaussian}} 
          & ERM      & $68.2\pm0.01$ & $64.8\pm0.01$ & $60.6\pm0.01$ & $56.9\pm0.01$ & $53.9\pm0.01$ & $50.2\pm0.03$ \\
        & & IRM      & $64.7\pm0.02$ & $64.7\pm0.01$ & $62.2\pm0.01$ & $54.5\pm0.02$ & $53.4\pm0.03$ & $50.4\pm0.01$ \\
        & & REx      & $68.0\pm0.03$ & $65.1\pm0.03$ & $61.8\pm0.01$ & $56.8\pm0.01$ & $53.2\pm0.01$ & $51.5\pm0.01$ \\
        & & GroupDRO & $66.1\pm0.01$ & $61.7\pm0.02$ & $59.3\pm0.03$ & $53.6\pm0.01$ & $54.0\pm0.02$ & $50.6\pm0.02$ \\
        & & SharpDRO & $\mathbf{72.3\pm0.03}$ & $\mathbf{72.2\pm0.03}$ & $\mathbf{71.5\pm0.02}$ & $\mathbf{62.4\pm0.03}$ & $\mathbf{59.7\pm0.03}$ & $\mathbf{55.7\pm0.06}$ \\ 
        \cmidrule(lr){2-9}
        & \multirow{5}{*}{\rotatebox{90}{Shot}} 
          & ERM      & $67.6\pm0.03$ & $65.1\pm0.01$ & $62.9\pm0.01$ & $56.0\pm0.01$ & $55.1\pm0.01$ & $47.3\pm0.01$ \\
        & & IRM      & $67.5\pm0.02$ & $65.7\pm0.01$ & $62.7\pm0.01$ & $59.5\pm0.01$ & $55.8\pm0.01$ & $48.3\pm0.01$ \\
        & & REx      & $65.7\pm0.01$ & $63.8\pm0.02$ & $61.9\pm0.01$ & $59.3\pm0.03$ & $53.8\pm0.01$ & $48.1\pm0.01$ \\
        & & GroupDRO & $67.0\pm0.02$ & $65.8\pm0.01$ & $63.1\pm0.01$ & $58.9\pm0.01$ & $57.5\pm0.01$ & $49.3\pm0.01$ \\
        & & SharpDRO & $\mathbf{71.3\pm0.02}$ & $\mathbf{70.5\pm0.03}$ & $\mathbf{68.3\pm0.03}$ & $\mathbf{63.6\pm0.04}$ & $\mathbf{60.5\pm0.04}$ & $\mathbf{53.4\pm0.03}$ \\
        \midrule
        \multirow{10}{*}{\rotatebox{90}{ImageNet30}} & \multirow{5}{*}{\rotatebox{90}{Gaussian}} 
          & ERM      & $87.5\pm0.01$ & $84.6\pm0.01$ & $81.9\pm0.01$ & $76.5\pm0.01$ & $71.2\pm0.01$ & $65.3\pm0.01$ \\ 
        & & IRM      & $86.6\pm0.01$ & $84.4\pm0.03$ & $80.6\pm0.01$ & $75.2\pm0.01$ & $70.7\pm0.03$ & $64.8\pm0.01$ \\ 
        & & REx      & $86.3\pm0.01$ & $83.8\pm0.03$ & $81.1\pm0.02$ & $75.6\pm0.02$ & $71.5\pm0.01$ & $66.1\pm0.03$ \\ 
        & & GroupDRO & $85.1\pm0.02$ & $84.2\pm0.01$ & $81.2\pm0.03$ & $76.3\pm0.03$ & $72.0\pm0.02$ & $66.3\pm0.01$ \\ 
        & & SharpDRO & $\mathbf{89.6\pm0.03}$ & $\mathbf{88.3\pm0.02}$ & $\mathbf{85.5\pm0.03}$ & $\mathbf{82.8\pm0.04}$ & $\mathbf{77.5\pm0.03}$ & $\mathbf{70.2\pm0.05}$ \\ 
        \cmidrule(lr){2-9}
        & \multirow{5}{*}{\rotatebox{90}{Shot}} 
          & ERM      & $86.9\pm0.01$ & $84.8\pm0.01$ & $83.6\pm0.01$ & $79.7\pm0.01$ & $75.4\pm0.01$ & $64.6\pm0.01$ \\
        & & IRM      & $86.8\pm0.01$ & $85.1\pm0.03$ & $81.5\pm0.01$ & $73.5\pm0.02$ & $68.5\pm0.03$ & $62.5\pm0.03$ \\
        & & REx      & $83.8\pm0.01$ & $86.3\pm0.03$ & $82.5\pm0.02$ & $73.9\pm0.01$ & $70.6\pm0.03$ & $64.0\pm0.02$ \\
        & & GroupDRO & $86.7\pm0.01$ & $85.6\pm0.03$ & $84.5\pm0.01$ & $80.7\pm0.01$ & $76.2\pm0.04$ & $65.4\pm0.01$ \\
        & & SharpDRO & $\mathbf{88.3\pm0.02}$ & $\mathbf{88.1\pm0.03}$ & $\mathbf{86.4\pm0.02}$ & $\mathbf{85.3\pm0.04}$ & $\mathbf{78.2\pm0.04}$ & $\mathbf{68.2\pm0.03}$ \\
        \bottomrule
    \end{tabular}
\end{table*}

\begin{theorem}[\textbf{Informal}] Assuming the loss function $\mathbb{L}$ is $l$-Lipschitz smooth, satisfies $\mu$-Polyak-Łojasiewicz (PL) condition on the second variable $\omega$, and has unbiased estimation about the gradient as well as $\sigma^2$ bounded variance, we can get the convergence rate during $T$ iterations:
\begin{equation*}
\begin{aligned}
\!\frac{1}{T}\!\sum_{t=0}^{T-1}\mathbb{E}\|\nabla\mathbb{L}^*(\theta_t)\|^2
&    \!\leq\! 320\sqrt{\frac{3\kappa^4l(\mathbb{E}[\mathbb{L}^*(\theta_0)]\!-\!\min_{\theta}\mathbb{E}[\mathbb{L}^*(\theta)])\sigma^2}{11MT}}\\
&=\mathcal{O}\left(\frac{\kappa^2}{\sqrt{MT}}\right),
\end{aligned}
\end{equation*}
where the conditional number $\kappa=l/\mu$ and $M$ means the sample batch(here we can choose $M=1$)\footnote{The resulting bound here means our SharpDRO can converge to the $\epsilon$-stationary point in $\frac{1}{\epsilon^2}$ iterations.}.
\label{SharpDRO:theorem}
\end{theorem}

\section{Experiment}
\label{SharpDRO:experiments}
In experiments, we first give details about our experimental setup. Then, we conduct quantitative experiments to compare to proposed SharpDRO with the most popular baseline methods by considering both distribution-aware and distribution-agnostic settings, which shows the capability of SharpDRO to tackle the most challenging distributions. Finally, we conduct qualitative analyses to validate the effectiveness of SharpDRO in robust generalization.

\begin{table*}[t]
    \scriptsize
    \centering
    \caption{Quantitative comparisons on distribution-agnostic robust generalization setting. Averaged accuracy ($\%$) with standard deviations is computed over three independent trials.}
    \setlength{\tabcolsep}{1.2mm}
    \renewcommand{\arraystretch}{1.1} 
    \label{SharpDROtab:distribution_agnostic}
    \begin{tabular}{ccl cccccc}
        \toprule
        \multirow{2}{*}{Dataset} & \multirow{2}{*}{Type} & \multirow{2}{*}{Method} & \multicolumn{6}{c}{Corruption Severity} \\
        \cmidrule(lr){4-9} 
        &  &  & 0 & 1 & 2 & 3 & 4 & 5 \\ 
        \midrule
        \multirow{6}{*}{\rotatebox{90}{CIFAR10}} & \multirow{3}{*}{\rotatebox{90}{Gaussian}} 
        & JTT      & $89.9\pm0.02$ & $88.8\pm0.02$ & $86.5\pm0.02$ & $86.1\pm0.02$ & $83.4\pm0.03$ & $79.8\pm0.02$ \\
        & & EIIL     & $88.6\pm0.02$ & $87.5\pm0.03$ & $86.3\pm0.03$ & $85.4\pm0.02$ & $83.2\pm0.03$ & $78.8\pm0.01$ \\
        & & SharpDRO & $\mathbf{91.2\pm0.02}$ & $\mathbf{91.3\pm0.03}$ & $\mathbf{88.9\pm0.04}$ & $\mathbf{87.3\pm0.02}$ & $\mathbf{85.6\pm0.04}$ & $\mathbf{83.1\pm0.02}$ \\ 
        \cmidrule(lr){2-9} 
        & \multirow{3}{*}{\rotatebox{90}{Shot}} 
        & JTT      & $91.3\pm0.02$ & $90.5\pm0.03$ & $89.3\pm0.01$ & $86.5\pm0.02$ & $83.1\pm0.02$ & $79.8\pm0.02$ \\
        & & EIIL     & $90.3\pm0.03$ & $90.1\pm0.02$ & $88.3\pm0.01$ & $86.2\pm0.02$ & $82.3\pm0.03$ & $78.5\pm0.02$ \\
        & & SharpDRO & $\mathbf{91.9\pm0.02}$ & $\mathbf{91.1\pm0.02}$ & $\mathbf{90.2\pm0.04}$ & $\mathbf{88.6\pm0.04}$ & $\mathbf{86.5\pm0.05}$ & $\mathbf{83.3\pm0.04}$ \\ 
        \midrule
        \multirow{6}{*}{\rotatebox{90}{CIFAR100}} & \multirow{3}{*}{\rotatebox{90}{Gaussian}} 
        & JTT      & $68.0\pm0.02$ & $65.3\pm0.02$ & $61.3\pm0.01$ & $56.3\pm0.01$ & $54.2\pm0.03$ & $51.2\pm0.02$ \\
        & & EIIL     & $67.2\pm0.01$ & $66.2\pm0.02$ & $61.0\pm0.02$ & $55.8\pm0.02$ & $54.6\pm0.03$ & $52.1\pm0.02$ \\
        & & SharpDRO & $\mathbf{70.3\pm0.03}$ & $\mathbf{68.8\pm0.03}$ & $\mathbf{65.2\pm0.03}$ & $\mathbf{60.3\pm0.02}$ & $\mathbf{57.4\pm0.03}$ & $\mathbf{55.3\pm0.03}$ \\ 
        \cmidrule(lr){2-9}
        & \multirow{3}{*}{\rotatebox{90}{Shot}} 
        & JTT      & $66.3\pm0.02$ & $65.3\pm0.03$ & $63.4\pm0.02$ & $56.6\pm0.04$ & $55.5\pm0.04$ & $48.6\pm0.04$ \\
        & & EIIL     & $66.5\pm0.02$ & $65.3\pm0.03$ & $62.8\pm0.04$ & $57.5\pm0.02$ & $56.5\pm0.01$ & $49.5\pm0.01$ \\
        & & SharpDRO & $\mathbf{68.9\pm0.02}$ & $\mathbf{66.2\pm0.03}$ & $\mathbf{64.9\pm0.03}$ & $\mathbf{59.8\pm0.02}$ & $\mathbf{56.5\pm0.03}$ & $\mathbf{51.0\pm0.02}$ \\ 
        \midrule
        \multirow{6}{*}{\rotatebox{90}{ImageNet30}} & \multirow{3}{*}{\rotatebox{90}{Gaussian}} 
        & JTT      & $87.3\pm0.02$ & $84.5\pm0.02$ & $82.3\pm0.04$ & $75.6\pm0.01$ & $72.1\pm0.04$ & $66.5\pm0.02$ \\
        & & EIIL     & $\mathbf{88.2\pm0.02}$ & $85.2\pm0.03$ & $81.3\pm0.02$ & $74.5\pm0.02$ & $71.5\pm0.02$ & $65.0\pm0.04$ \\
        & & SharpDRO & $87.5\pm0.03$ & $\mathbf{86.6\pm0.03}$ & $\mathbf{85.3\pm0.03}$ & $\mathbf{79.3\pm0.04}$ & $\mathbf{75.3\pm0.02}$ & $\mathbf{70.0\pm0.02}$ \\ 
        \cmidrule(lr){2-9}
        & \multirow{3}{*}{\rotatebox{90}{Shot}} 
        & JTT      & $86.5\pm0.02$ & $85.4\pm0.03$ & $82.6\pm0.04$ & $79.6\pm0.04$ & $77.2\pm0.04$ & $65.0\pm0.01$ \\
        & & EIIL     & $85.5\pm0.01$ & $86.3\pm0.04$ & $81.6\pm0.02$ & $80.2\pm0.03$ & $75.3\pm0.02$ & $64.4\pm0.03$ \\
        & & SharpDRO & $\mathbf{87.1\pm0.02}$ & $\mathbf{87.1\pm0.03}$ & $\mathbf{84.8\pm0.04}$ & $\mathbf{83.0\pm0.02}$ & $\mathbf{76.5\pm0.03}$ & $\mathbf{69.2\pm0.04}$ \\ 
        \bottomrule
    \end{tabular}
\end{table*}

\subsection{Practical Implementation}
\label{SharpDRO:appendix_implementation}
Our SharpDRO requires two backward phases, so the time complexity is twice as much as plain training, for efficient sharpness computation, please refer to~\cite{du2022efficient, du2022sharpness, zhang2022ga, zhao2022penalizing, zhao2022ss}. In the first step, we record the label prediction $p$ of each data during inference and simultaneously compute the loss $\mathcal{L}$. Additionally, in the first backward pass, we store the computed gradient $\nabla\mathcal{L}(\theta)$. Further, by adding $\epsilon^*$, we use the perturbed model to compute the second label prediction $\hat{p}$, which is further leveraged to compute the sharpness regularization $\mathcal{R}$. Moreover, in the distribution-agnostic setting, the predictions $p$ and $\hat{p}$ from two forward steps are used to compute the OOD score $\omega_i$. Then, we add the recorded gradient $\nabla\mathcal{L}(\theta)$ back to the model parameter and conduct sharpness minimization over the selected worst-case data. In this way, our SharpDRO can be correctly performed.

\subsection{Experimental Setup}
For distribution-aware situation, we choose GroupDRO~\cite{sagawa2019distributionally}, IRM~\cite{arjovsky2019invariant}, REx~\cite{krueger2021out}, and ERM for comparisons. As for a distribution-agnostic situation, we pick JTT~\cite{liu2021just} and Environment Inference for Invariant Learning (EIIL)~\cite{creager2021environment} for baseline methods\footnote{Note that we do not include the sharpness minimization method SAM~\cite{foret2020sharpness} in this problem setting because its OOD generalization performance is worse than ERM. However, we conduct a detailed analysis between SharpDRO and SAM in Section~\ref{SharpDRO:qualitative_analysis}}. For each problem setting, we construct corruption using CIFAR10/100~\cite{krizhevsky2009learning} and ImageNet30~\cite{russakovsky2015imagenet} datasets. Specifically, we follow~\cite {hendrycks2019benchmarking} to perturb the image data with severity levels varying from $1$ to $5$ by using two types of corruption: ``Gaussian Noise'' and ``Shot Noise''. Moreover, the clean data are considered as having a corruption severity of $0$. For each corrupted distribution, we sample them with different probabilities by following the Poisson distribution $P(s; \lambda=1)$, i.e., for $s$ varies from $0$ to $5$, the sample probabilities are $\left\{0.367, 0.367, 0.184, 0.061, 0.015, 0.003\right\}$, respectively. Then, we test the robust performance on each data distribution. For hyper-parameter $\rho$, we follow~\cite{foret2020sharpness} by setting it to $0.05$ to control the magnitude of $\epsilon^*$. For each experiment, we conduct three independent trials and report the average test accuracy with standard deviations.

\begin{table}[t]
	\scriptsize
	\centering
	\caption{Quantitative comparisons on distribution-aware robust generalization setting on Snow corruption. Averaged accuracy ($\%$) with standard deviations are computed over three independent trails.}
    \setlength{\tabcolsep}{1.2mm}
    \renewcommand{\arraystretch}{1.1} 
	\label{SharpDROtab:appendix_distribution_aware}
	\begin{tabular}{lllcccccc}
		\toprule[1pt]
		\multirow{2}{*}{Dataset} & \multirow{2}{*}{Type} & \multirow{2}{*}{Method} & \multicolumn{6}{c}{Corruption Severity} \\
		&  &  & \multicolumn{1}{c}{0} & \multicolumn{1}{c}{1} & \multicolumn{1}{c}{2} & \multicolumn{1}{c}{3} & \multicolumn{1}{c}{4} & \multicolumn{1}{c}{5} \\ \midrule[0.6pt]
		\multirow{5}{*}{\rotatebox{90}{CIFAR10}} & \multirow{5}{*}{\rotatebox{90}{Snow}} & ERM & \multicolumn{1}{c}{$90.8\pm0.01$} & \multicolumn{1}{c}{$90.1\pm0.02$} & \multicolumn{1}{c}{$88.1\pm0.02$} & \multicolumn{1}{c}{$88.1\pm0.02$} & \multicolumn{1}{c}{$85.7\pm0.02$} & \multicolumn{1}{c}{$82.6\pm0.01$} \\
		&  & IRM & \multicolumn{1}{c}{$91.1\pm0.02$} & \multicolumn{1}{c}{$90.7\pm0.01$} & \multicolumn{1}{c}{$89.7\pm0.02$} & \multicolumn{1}{c}{$88.0\pm0.03$} & \multicolumn{1}{c}{$84.6\pm0.02$} & \multicolumn{1}{c}{$83.2\pm0.03$} \\
		&  & REx & \multicolumn{1}{c}{$91.8\pm0.02$} & \multicolumn{1}{c}{$\bm{91.9}\pm\bm{0.01}$} & \multicolumn{1}{c}{$88.4\pm0.01$} & \multicolumn{1}{c}{$88.3\pm0.01$} & \multicolumn{1}{c}{$88.6\pm0.01$} & \multicolumn{1}{c}{$83.0\pm0.02$} \\
		&  & GroupDRO & \multicolumn{1}{c}{$91.5\pm0.02$} & \multicolumn{1}{c}{$91.0\pm0.01$} & \multicolumn{1}{c}{$88.7\pm0.02$} & \multicolumn{1}{c}{$88.6\pm0.02$} & \multicolumn{1}{c}{$85.2\pm0.03$} & \multicolumn{1}{c}{$83.5\pm0.02$} \\
		&  & SharpDRO & \multicolumn{1}{c}{$\bm{93.1}\pm\bm{0.01}$} & \multicolumn{1}{c}{$91.8\pm0.01$} & \multicolumn{1}{c}{$\bm{90.5}\pm\bm{0.02}$} & \multicolumn{1}{c}{$\bm{90.8}\pm\bm{0.02}$} & \multicolumn{1}{c}{$\bm{87.9}\pm\bm{0.01}$} & \multicolumn{1}{c}{$\bm{84.3}\pm\bm{0.02}$} \\ \midrule[0.6pt]
		\multirow{5}{*}{\rotatebox{90}{CIFAR10}} & \multirow{5}{*}{\rotatebox{90}{Snow}} & ERM & $67.7\pm0.01$ & $68.1\pm0.01$ & $64.7\pm0.01$ & $63.1\pm0.01$ & $60.5\pm0.02$ & $57.3\pm0.01$ \\
		&  & IRM & $69.3\pm0.01$ & $67.5\pm0.02$ & $64.9\pm0.02$ & $61.0\pm0.01$ & $58.2\pm0.01$ & $55.1\pm0.01$ \\
		&  & REx & $66.4\pm0.01$ & $65.9\pm0.01$ & $62.4\pm0.01$ & $61.2\pm0.02$ & $57.5\pm0.03$ & $56.0\pm0.02$ \\
		&  & GroupDRO & $68.0\pm0.02$ & $68.2\pm0.01$ & $65.1\pm0.01$ & $60.9\pm0.03$ & $59.8\pm0.01$ & $58.1\pm0.02$ \\
		&  & SharpDRO & $\bm{71.5}\pm\bm{0.01}$ & $\bm{70.8}\pm\bm{0.03}$ & $\bm{67.5}\pm\bm{0.02}$ & $\bm{65.5}\pm\bm{0.01}$ & $\bm{62.3}\pm\bm{0.01}$ & $\bm{59.2}\pm\bm{0.03}$ \\ \midrule[0.6pt]
		\multirow{5}{*}{\rotatebox{90}{ImageNet30}} & \multirow{5}{*}{\rotatebox{90}{Snow}} & ERM & $86.7\pm0.03$ & $85.2\pm0.01$ & $83.4\pm0.01$ & $81.1\pm0.01$ & $75.3\pm0.01$ & $75.6\pm0.01$ \\
		&  & IRM & $85.6\pm0.01$ & $84.0\pm0.02$ & $82.1\pm0.03$ & $79.7\pm0.01$ & $75.0\pm0.01$ & $75.6\pm0.01$ \\
		&  & REx & $85.4\pm0.01$ & $84.6\pm0.02$ & $82.7\pm0.02$ & $80.5\pm0.03$ & $75.7\pm0.03$ & $75.9\pm0.03$ \\
		&  & GroupDRO & $86.7\pm0.01$ & $85.5\pm0.03$ & $83.4\pm0.01$ & $81.2\pm0.02$ & $76.3\pm0.01$ & $76.6\pm0.01$ \\
		&  & SharpDRO & $\bm{88.2}\pm\bm{0.02}$ & $\bm{88.2}\pm\bm{0.01}$ & $\bm{85.4}\pm\bm{0.02}$ & $\bm{81.9}\pm\bm{0.01}$ & $\bm{79.8}\pm\bm{0.03}$ & $\bm{79.5}\pm\bm{0.02}$ \\ \bottomrule[1pt]
	\end{tabular}
\end{table}

\begin{table}[t]
	\scriptsize
    \centering
	\caption{Quantitative comparisons on distribution-agnostic robust generalization setting on snow corruption. Averaged accuracy ($\%$) with standard deviations are computed over three independent trails.}
    \setlength{\tabcolsep}{1.2mm}
    \renewcommand{\arraystretch}{1.5} 
	\label{SharpDROtab:appendix_distribution_agnostic}
	\begin{tabular}{lllcccccc}
		\toprule[1pt]
		\multirow{2}{*}{Dataset} & \multirow{2}{*}{Type} & \multirow{2}{*}{Method} & \multicolumn{6}{c}{Corruption Severity} \\
		&  &  & \multicolumn{1}{c}{0} & \multicolumn{1}{c}{1} & \multicolumn{1}{c}{2} & \multicolumn{1}{c}{3} & \multicolumn{1}{c}{4} & \multicolumn{1}{c}{5} \\ \midrule[0.6pt]
		\multirow{3}{*}{\rotatebox{90}{CIFAR10}} & \multirow{3}{*}{\rotatebox{90}{Snow}} & JTT & \multicolumn{1}{c}{$88.6\pm0.02$} & \multicolumn{1}{c}{$87.8\pm0.03$} & \multicolumn{1}{c}{$86.5\pm0.02$} & \multicolumn{1}{c}{$87.2\pm0.02$} & \multicolumn{1}{c}{$84.2\pm0.02$} & \multicolumn{1}{c}{$83.2\pm0.03$} \\
		&  & EIIL & \multicolumn{1}{c}{$88.3\pm0.02$} & \multicolumn{1}{c}{$87.8\pm0.01$} & \multicolumn{1}{c}{$85.6\pm0.02$} & \multicolumn{1}{c}{$87.3\pm0.03$} & \multicolumn{1}{c}{$85.2\pm0.04$} & \multicolumn{1}{c}{$82.3\pm0.01$} \\
		&  & SharpDRO & $\bm{91.6}\pm\bm{0.01}$ & $\bm{91.1}\pm\bm{0.02}$ & $\bm{90.8}\pm\bm{0.01}$ & $\bm{89.7}\pm\bm{0.02}$ & $\bm{86.2}\pm\bm{0.01}$ & $\bm{83.8}\pm\bm{0.02}$ \\ \midrule[0.6pt]
		\multirow{3}{*}{\rotatebox{90}{CIFAR10}} & \multirow{3}{*}{\rotatebox{90}{Snow}} & JTT & $67.5\pm0.01$ & $68.1\pm0.02$ & $65.3\pm0.02$ & $64.3\pm0.02$ & $60.2\pm0.02$ & $57.8\pm0.02$ \\
		&  & EIIL & $68.2\pm0.03$ & $69.1\pm0.03$ & $65.2\pm0.02$ & $64.0\pm0.02$ & $61.0\pm0.04$ & $57.5\pm0.04$ \\
		&  & SharpDRO & $\bm{70.6}\pm\bm{0.02}$ & $\bm{69.9}\pm\bm{0.03}$ & $\bm{66.7}\pm\bm{0.03}$ & $\bm{64.4}\pm\bm{0.02}$ & $\bm{61.9}\pm\bm{0.03}$ & $\bm{60.7}\pm\bm{0.03}$ \\ \midrule[0.6pt]
		\multirow{3}{*}{\rotatebox{90}{ImageNet30}} & \multirow{3}{*}{\rotatebox{90}{Snow}} & JTT & $86.0\pm0.04$ & $85.8\pm0.02$ & $82.3\pm0.03$ & $80.4\pm0.02$ & $74.6\pm0.02$ & $73.5\pm0.02$ \\
		&  & EIIL & $87.5\pm0.01$ & $85.4\pm0.02$ & $83.5\pm0.04$ & $\bm{81.6}\pm\bm{0.01}$ & $76.3\pm0.01$ & $75.8\pm0.02$ \\
		&  & SharpDRO & $\bm{87.5}\pm\bm{0.03}$ & $\bm{86.7}\pm\bm{0.02}$ & $\bm{85.4}\pm\bm{0.02}$ & $81.5\pm0.03$ & $\bm{78.9}\pm\bm{0.02}$ & $\bm{78.5}\pm\bm{0.03}$ \\ \bottomrule[1pt]
	\end{tabular}
\end{table}

\subsection{Quantitative Comparisons}
In this part, we focus on three questions: (1) Can SharpDRO perform well in two situations of robust generalization? (2) Does SharpDRO generalize well on the most severely corrupted distributions? and (3) Is SharpDRO able to tackle different types of corruption? To answer these questions, we conduct experiments on both settings by testing on different corruption types and severity levels.

\textbf{Distribution-Aware Robust Generalization}
\label{SharpDRO:distribution_aware}
As shown in Table~\ref{SharpDROtab:distribution_aware}, we can see that SharpDRO surpasses other methods with larger performance gains as the corruption severity increases. Especially in the CIFAR10 dataset on ``Gaussian Noise'' corruption, the improvement margin between SharpDRO and the second-best method is $2.2\%$ with severity of $0$, which is further increased to about $\bm{5.7\%}$ with severity of $5$, which indicates the capability of SharpDRO on generalization against severe corruptions. Moreover, SharpDRO frequently outperforms other methods on all scenarios, which manifests the robustness of SharpDRO against various corruption types.

\textbf{Distribution-Agnostic Robust Generalization}
\label{SharpDRO:distribution_agnostic}
In Table~\ref{SharpDROtab:distribution_agnostic}, we can see a similar phenomenon as in Table~\ref{SharpDROtab:distribution_aware} that the more severe corruptions are applied, the larger performance gains SharpDRO achieves. Especially, in the ImageNet30 dataset corrupted by ``Shot Noise'', SharpDRO shows about $0.6\%$ performance gains upon the second-best method with severity $0$, which is further increased to almost $\bm{4.2\%}$ with severity $5$. Moreover, SharpDRO is general to all three corruption types, as it surpasses other methods in most cases. Therefore, SharpDRO can perfectly generalize even without the distribution annotations.

\section{Results on Additional Corruptions}
\label{SharDRO:appendix_exp_results}
In the main paper, we have provided the results using ``Gaussian Noise'' corruption and ``Shot Noise'' corruption. Here, we conduct additional experiments to show the effectiveness of SharpDRO under ``Snow'' corruption. The results on CIFAR10, CIFAR100, and ImageNet30 datasets in both distribution-aware and distribution-agnostic scenarios are shown in Tables~\ref{SharpDROtab:appendix_distribution_aware} and~\ref{SharpDROtab:appendix_distribution_agnostic}. We can see that SharpDRO still performs effectively and surpasses other methods with a large margin. Especially, on the ImageNet30 dataset in both problem settings, SharpDRO outperforms the second-best method by about $3\%$, which indicates the capability of SharpDRO on generalization against different corruptions.

\begin{figure*}[t]
	\centering
	\begin{minipage}[t]{0.195\textwidth}
		\centering
		\includegraphics[width=\linewidth]{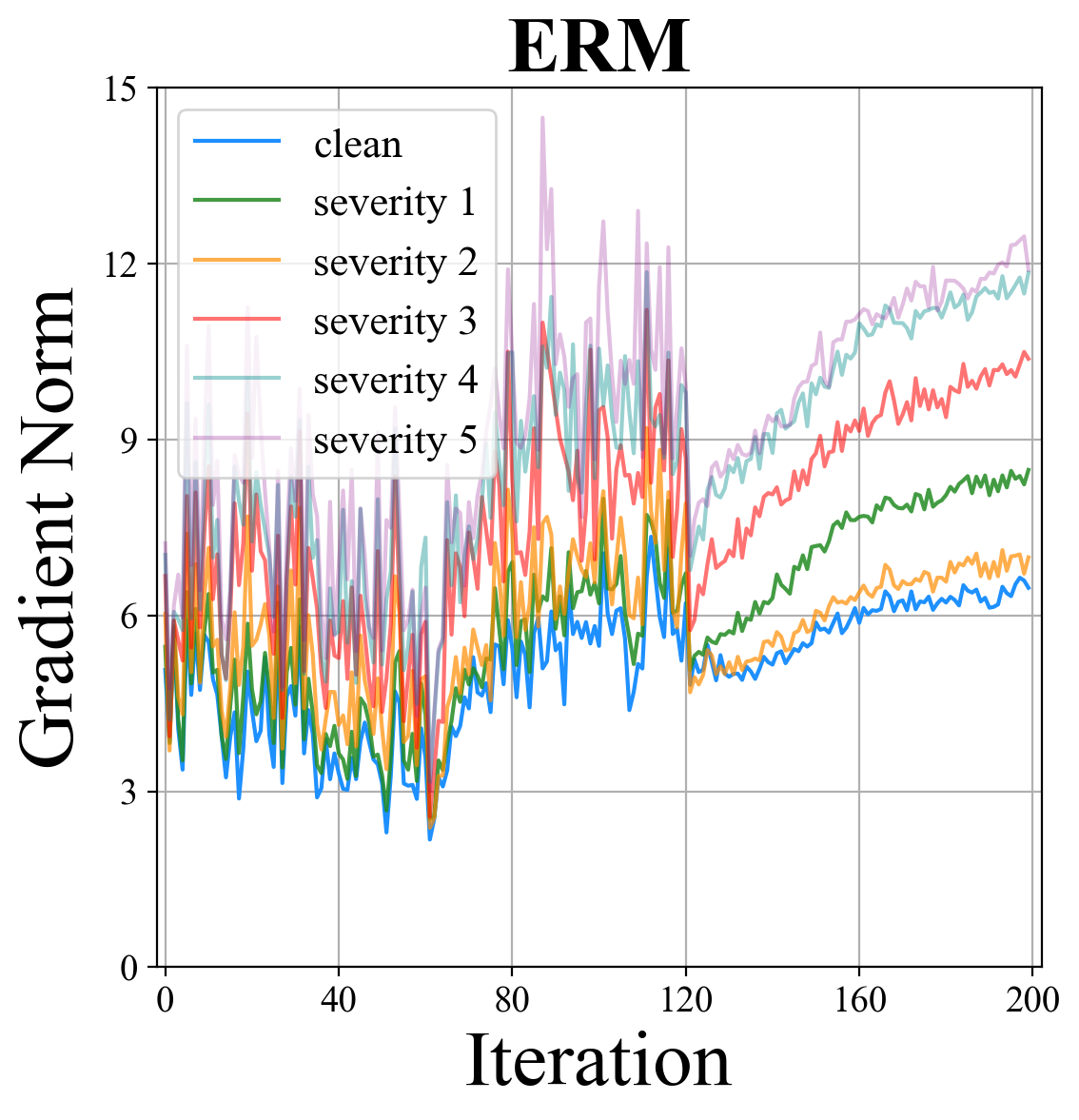}
	\end{minipage}
	\begin{minipage}[t]{0.195\textwidth}
		\centering
		\includegraphics[width=\linewidth]{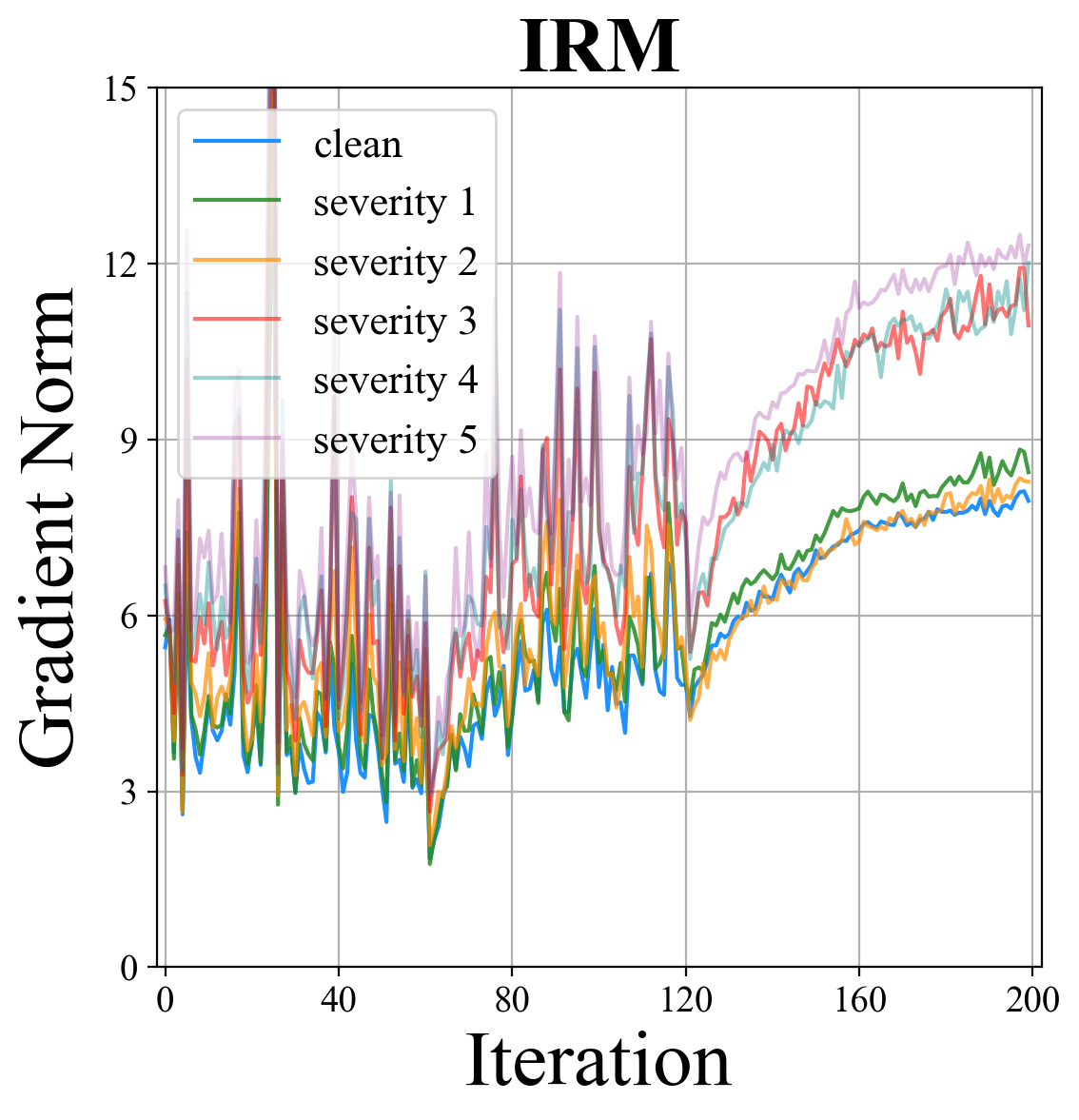}
	\end{minipage}
	\begin{minipage}[t]{0.195\textwidth}
		\centering
		\includegraphics[width=\linewidth]{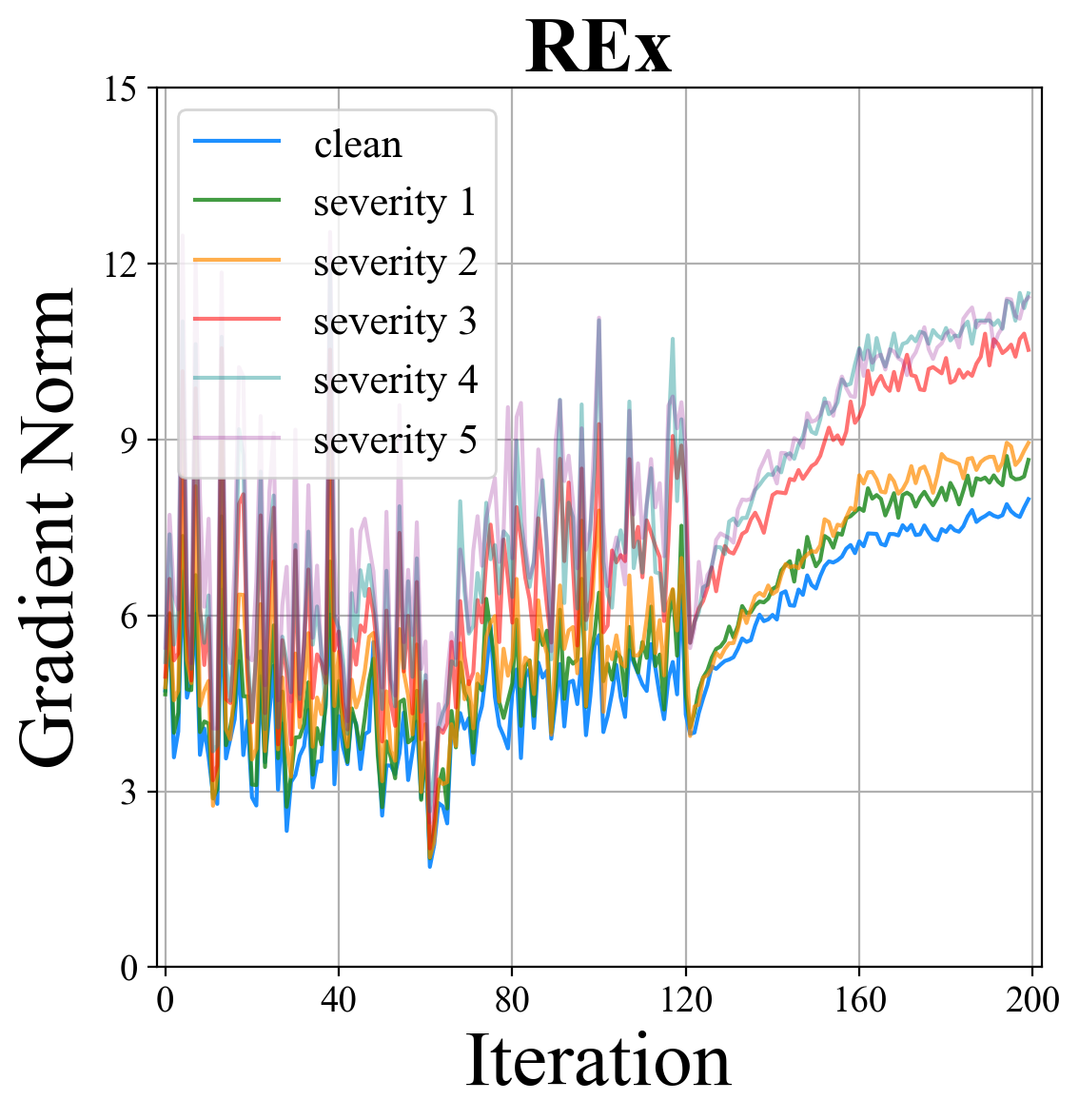}
	\end{minipage}
	\begin{minipage}[t]{0.195\textwidth}
		\centering
		\includegraphics[width=\linewidth]{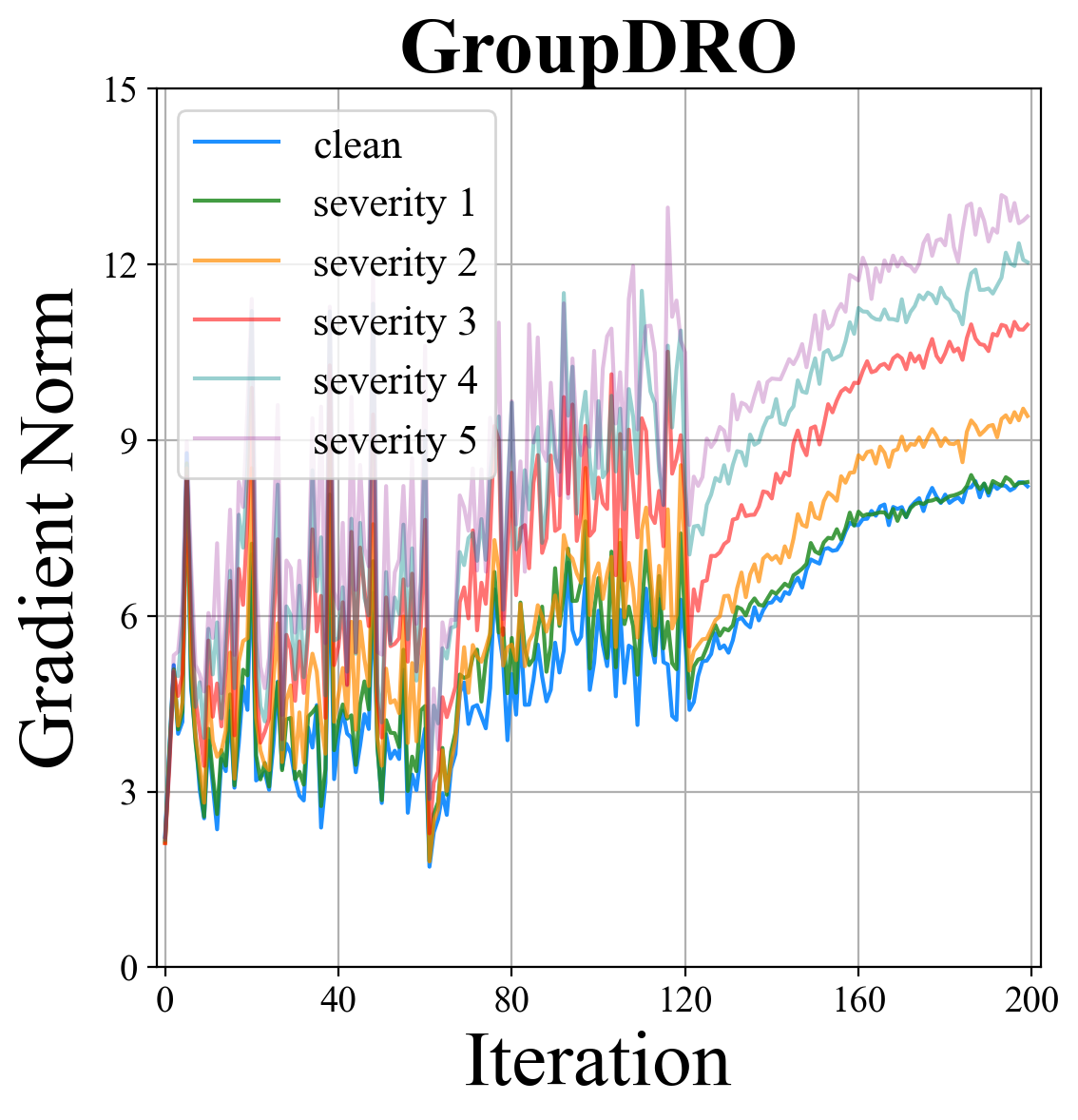}
	\end{minipage}
	\begin{minipage}[t]{0.195\textwidth}
		\centering
		\includegraphics[width=\linewidth]{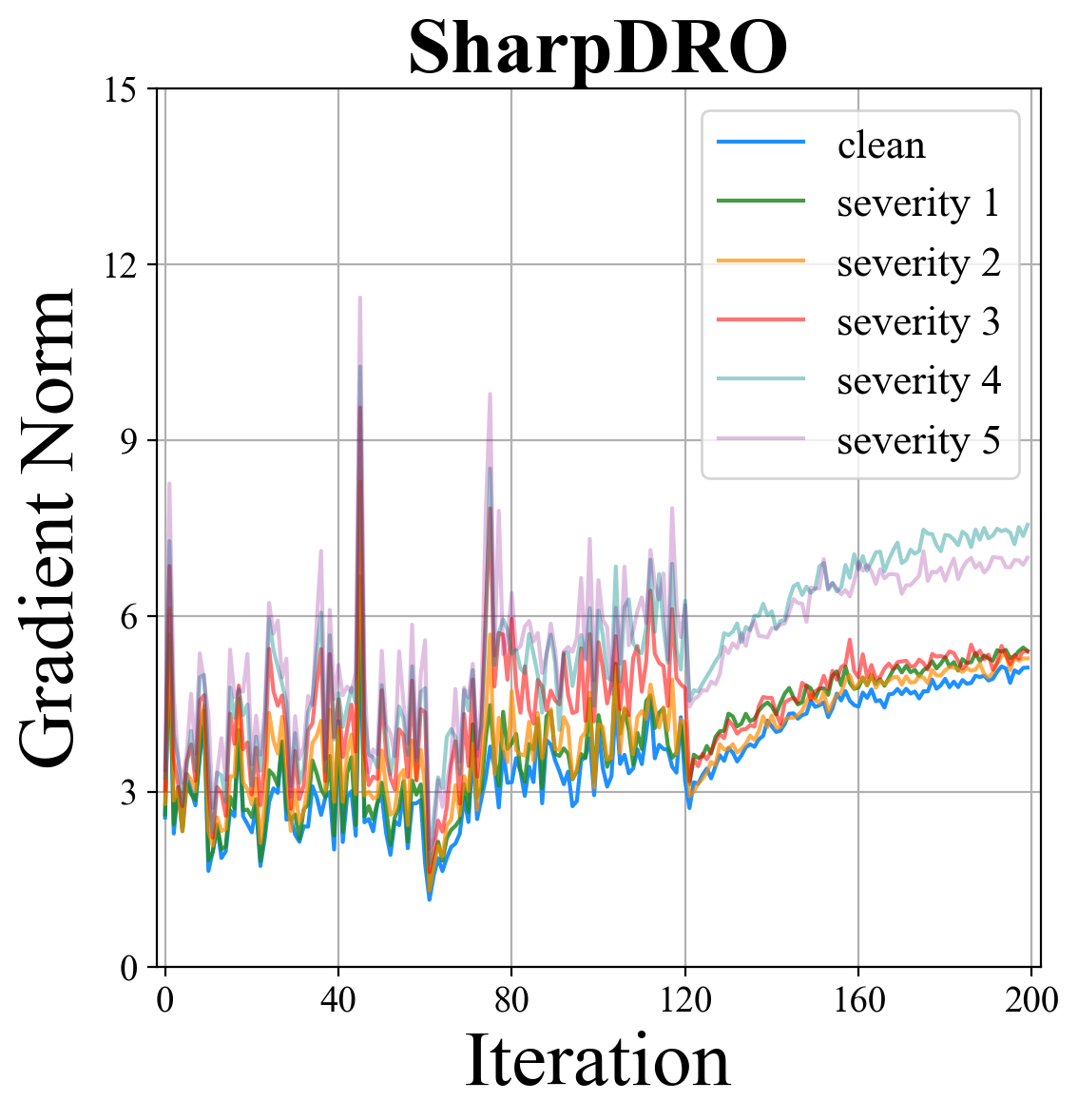}
	\end{minipage}
	\caption{\small Gradient norm comparisons between different methods over all corrupted distributions.}
	\label{SharpDROfig:gradnorm}
\end{figure*}

\subsection{Qualitative Analysis}
\label{SharpDRO:qualitative_analysis}
To investigate the effectiveness of SharpDRO, we first conduct an ablation study to show that the worst-case sharpness minimization is essential for achieving generalization with robustness. Then, we utilize gradient norm, an important criterion to present training stability, to validate that our method is stable for severely corrupted distributions. Then, we analyze the hyper-parameter $\rho$ and OOD score $\bar{w}$ to disclose the effectiveness of sharpness minimization and worst-case data selection. Finally, another second-order method, SAM~\cite{foret2020sharpness, zhong2022improving, mi2022make, sun2023adasam, sun2023fedspeed}, is investigated to discover the efficiency property of SharpDRO. All analyses are conducted using CIFAR10 with ``Gaussian Noise'' corruption.

\begin{table}[t]
	\small
	\caption{Ablation study. "w/o data selection" denotes training without worst-case data selection, which recovers SAM~\cite{foret2020sharpness}, and "w/o sharp min" indicates training without sharpness minimization, which is the same as GroupDRO~\cite{sagawa2019distributionally}.}
	\setlength{\tabcolsep}{3mm}
	\label{SharpDROtab:ablation_study}
	\begin{tabular}{lclllll}
		\toprule[1pt]
		\multirow{2}{*}{Method} & \multicolumn{6}{c}{Corruption Severity} \\
		& 0 & \multicolumn{1}{c}{1} & \multicolumn{1}{c}{2} & \multicolumn{1}{c}{3} & \multicolumn{1}{c}{4} & \multicolumn{1}{c}{5} \\ \midrule[0.6pt]
		w/o data selection (SAM) & 93.2 & \multicolumn{1}{c}{90.5} & \multicolumn{1}{c}{87.6} & \multicolumn{1}{c}{82.1} & \multicolumn{1}{c}{80.5} & \multicolumn{1}{c}{75.4} \\
		w/o sharp min (GroupDRO) & \multicolumn{1}{l}{90.2} & 89.1 & 88.4 & 84.3 & 83.0 & 78.2 \\
		\rowcolor{gray!25} SharpDRO & \multicolumn{1}{l}{93.1} &  92.2 & 91.4 & 89.2 & 87.1 & 84.5 \\ \bottomrule[1pt]
	\end{tabular}
\end{table}

\textbf{Ablation Study}
\label{SharpDRO:ablation_study}
By eliminating the worst-case data selection, we recover the original sharpness minimization method SAM~\cite{foret2020sharpness}. Then, we remove the sharpness minimization module, which is basically training via GroupDRO. The ablation results are shown in Table~\ref{SharpDROtab:ablation_study}. We can see that deploying SAM on the whole training dataset can achieve improved results on the clean dataset. However, the robust performance on corrupted distributions is even worse than GroupDRO. This could be because sharpness is easy to be dominated by principal distributions, which is misleading for generalization to small distributions. Thus, the sharpness of corrupted data would be sub-optimal. As for GroupDRO, it fails to produce a flat loss surface, hence it cannot generalize as well as the proposed SharpDRO.

\textbf{Distributional Stability}
To show our method can be stable even in the most challenging distributions, we show the gradient norm on a validation set including corruption severity from $0$ to $5$. As shown in Figure~\ref{SharpDROfig:gradnorm}, SharpDRO not only produces the smallest norm value but also can ensure almost equal gradient norm across all corruptions, which indicates that SharpDRO is the most distributionally stable method among all compared methods.

\begin{figure}[t]
	\centering
    \includegraphics[width=\linewidth]{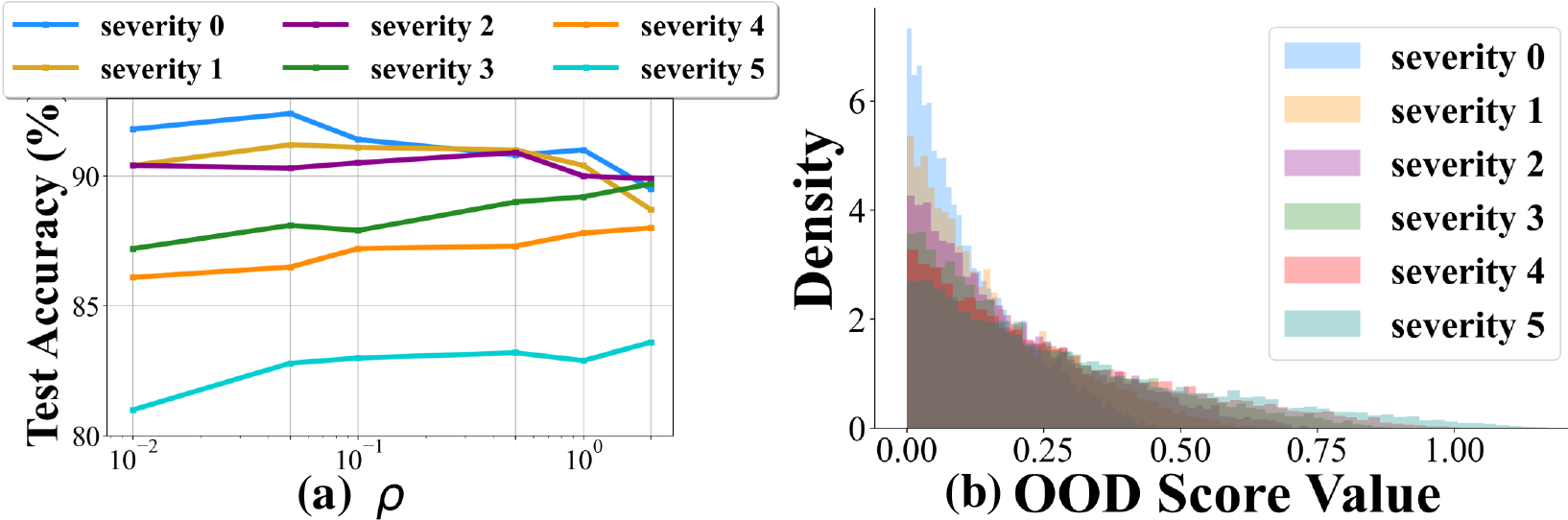}
\caption{\small (a) Sensitivity of $\rho$ whose value is set to $\left\{0.01, 0.05, 0.1, 0.5, 1, 2\right\}$. (b) Distribution of the normalized OOD score $\bar{\omega}$ on distribution $s=0$ to $5$.}
\label{SharpDROfig:sens_ood}
\end{figure}

\textbf{Parameter Analysis}
To understand how the scale parameter $\rho$ affects our generalization performance, we conduct a sensitivity analysis by changing this value and show the test results of different distributions. In Figure~\ref{SharpDROfig:sens_ood} (a), we find an interesting discovery that as $\rho$ increases, which indicates the perturbation magnitude $\epsilon^*$ enlarges, would enhance the generalization of severely corrupted data but degrade the performance of slightly corrupted data. This might be because the exploration of hard distributions needs to cover wa ide range of neighborhoods to ensure generalization. On the contrary, exploration too far on easy distributions can reach out-of-distribution, thus causing performance degradation. Therefore, for practitioners who aim to generalize on small and difficult datasets, we might be able to enhance performance by aggressively setting a large perturbation scale.

\textbf{OOD Score Analysis}
\label{SharpDRO:ood_score}
The OOD score helps to select worst-case data under the distribution-agnostic setting. To show its effectiveness in selecting the noisy data, we plot the value distribution of OOD scores from all corrupted distributions in epoch $30$ in Figure~\ref{SharpDROfig:sens_ood} (b). We can see the tendency that severe corruption has larger OOD scores. Therefore, our OOD score is a valid criterion to select worst-case data. Note that during the training process, the worst-case data would be gradually learned, thus the OOD score can become smaller, which explains why the value distribution of our score is not as separable as OOD detection does.

\begin{figure}[t]
	\centering
	\includegraphics[width=0.7\linewidth]{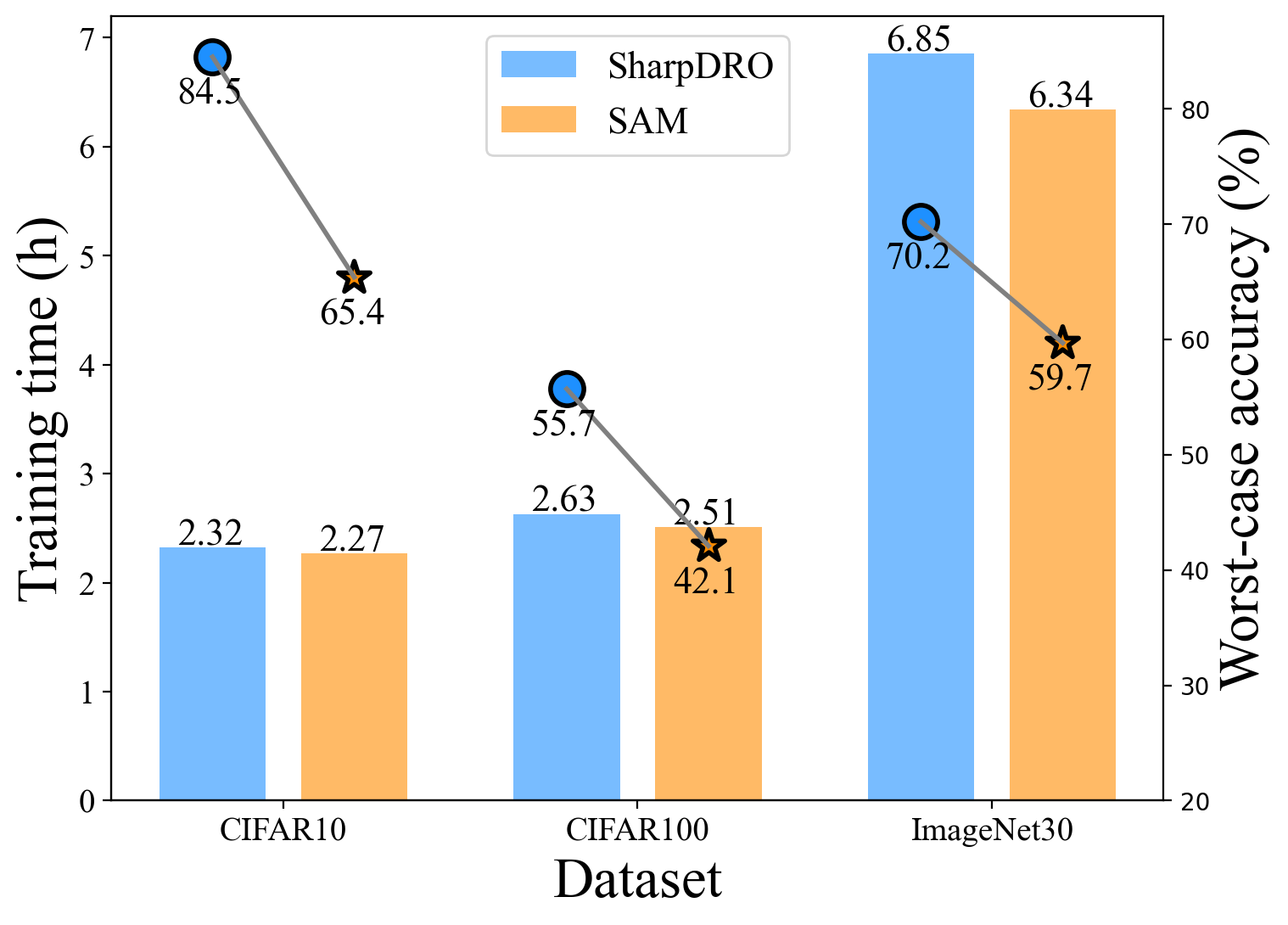}
	\caption{\small Efficiency comparison between SharpDRO and SAM.}
	\label{SharpDROfig:time}
\end{figure}

\textbf{Training Efficiency Analysis}
It is clear that the proposed SharpDRO method is a second-order optimization method. Hence, when compared to first-order methods such as GroupDRO and REx, computational cost is the price to pay for achieving improved generalization performance\footnote{Note that our method can be deployed with existing efficient sharpness-based methods~\cite{zhang2022ga, du2022efficient, du2022sharpness, zhao2022ss}.}. However, to further explore the advantage of SharpDRO compared to other second-order methods, here we use SAM~\cite{foret2020sharpness} as a competitor, and show their computational time as well as worst-case accuracy ($s=5$) in Figure~\ref {SharpDROfig:time}. We can see that on all three datasets, our SharpDRO requires nearly the same time to train, and significantly outperforms the worst-case performance of SAM, owing to our efficient worst-case data selection, which is vital for robust generalization against severe corruptions.

\section{Conclusion}
\label{SharpDRO:conclusion}
In this Chapter, we propose a SharpDRO approach to enhance the generalization performance of DRO methods. Specifically, we focus on minimizing the sharpness of worst-case data to learn flat loss surfaces. As a result, SharpDRO is more robust to severe corruptions compared to other methods. Moreover, we apply SharpDRO to distribution-aware and distribution-agnostic settings and propose an OOD detection process to select the worst-case data when the distribution index is not known. Extensive quantitative and qualitative experiments have been conducted to show that SharpDRO can deal with the most challenging corrupted distributions and achieve improved generalization results compared to well-known baseline methods.
\chapter{Exploring Variant Parameters for Invariant Learning}
\label{cha:EVIL}
Out-of-Distribution (OOD) Generalization aims to learn robust models that generalize well to various environments without fitting to distribution-specific features. Recent studies based on Lottery Ticket Hypothesis (LTH) address this problem by minimizing the learning target to find some of the parameters that are critical to the task. However, in OOD problems, such solutions are suboptimal as the learning task contains severe distribution noises, which can mislead the optimization process. Therefore, apart from finding the task-related parameters (i.e., invariant parameters), we propose Exploring Variant parameters for Invariant Learning (EVIL) which also leverages the distribution knowledge to find the parameters that are sensitive to distribution shift (i.e., variant parameters). Once the variant parameters are left out of invariant learning, a robust subnetwork that is resistant to distribution shift can be found. Additionally, the parameters that are relatively stable across distributions can be considered invariant ones to improve invariant learning. By fully exploring both variant and invariant parameters, our EVIL can effectively identify a robust subnetwork to improve OOD generalization. In extensive experiments on integrated testbed: DomainBed, EVIL can effectively and efficiently enhance many popular methods, such as ERM, IRM, SAM, etc.

\section{Introduction}
\label{EVIL:introduction}

\begin{figure}[t]
  \begin{center}
    \includegraphics[width=0.7\linewidth]{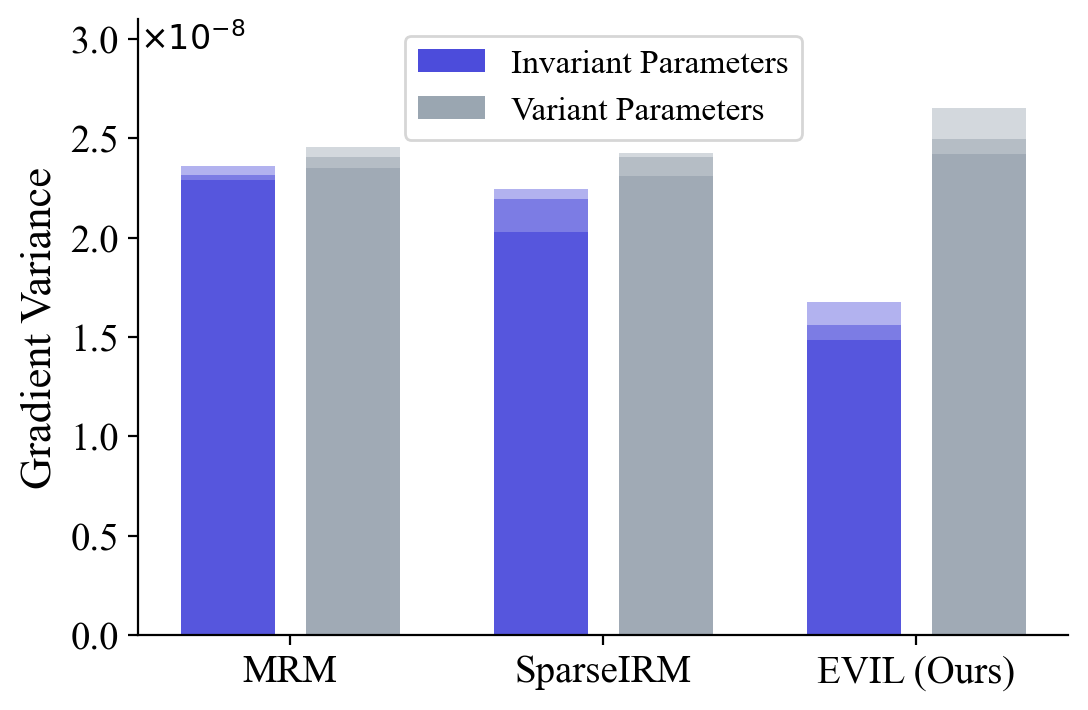}
  \end{center}
  \caption{Comparison of the gradient variance between the learned subnetwork, i.e., invariant parameters, and the pruned parameters, i.e., variant parameters. The gradient variance is computed through $\mathcal{V}=Mean(Var(\left[g_i\right]_{i=0}^d))$~\cite{rame2022fishr}, where $g_i$ denotes the $i$-th gradient among $d$ distributions, and $Var(\cdot)$ and $Mean(\cdot)$ denotes the mathematical variance and mean, respectively. The results are from three independent trials.}
\label{EVILfig:grad_var}
\end{figure}
The strong representation ability of deep neural networks~\cite{he2016deep,krizhevsky2009learning,lecun2015deep} has been one of the vital keys to the success of deep learning over the past decade. However, the realistic deployment of neural networks is often restricted to the IID assumption, where the training data and test data should be distributed independently and identically. When such an assumption is violated, a drastic degradation in learning performance is often observed, which seriously hinders the practical application of deep models. Therefore, Out-Of-Distribution (OOD) generalization~\cite{gulrajani2021search, li2017deeper, muandet2013domain} thrives as a promising direction that aims to enhance model robustness against unknown distribution shifts.

In order to achieve OOD generalization, one mainstream methodology is invariant learning~\cite{arjovsky2019invariant, creager2021environment, lu2021invariant, krueger2021out}, which enforces extracting invariant features to help make consistent predictions among various data distributions (or domains), meanwhile avoiding learning distribution-specific features that are irrelevant to label information. Recent advances based on Lottery Ticket Hypothesis (LTH)~\cite{frankle2019lottery, frankle2020linear, malach2020proving} show that sparse training optimized by learning task could select some critical parameters as a subnetwork, which are strongly responsible for invariant learning~\cite{morcos2019one, zhang2021can, zhou2022sparse}. However, in OOD problems, the sparsification guided by the learning task is problematic because the distribution noise could be erroneously incorporated into the optimization of sparse learning. As a consequence, existing methods fail to identify a robust subnetwork that is stable across different distributions. Particularly, we follow Rame et al.~\cite{rame2022fishr} by using gradient variance to indicate model sensitivity to distribution shift. Then, we compare the gradient variance between the subnetwork and the pruned parameters learned by different methods, as shown in Figure~\ref{EVILfig:grad_var}. We can see that the subnetwork learned by existing methods (MRM~\cite{zhang2021can} and SparseIRM~\cite{zhou2022sparse}) are almost as sensitive as the pruned parameters, which means that invariant information could not be fully captured.

To overcome this problem, we propose a novel sparse training framework by Exploring Variant parameters for Invariant Learning (EVIL). Specifically, by following common assumptions that input data can be decomposed into invariant features and spurious features~\cite{mitrovic2020representation, von2021self, zhang2021can, zhang2013domain}\footnote{Though some works investigate more complex situations where there are multiple factors causing the data generation process~\cite{huang2020causal, lu2021invariant, suter2019robustly}, our assumption is more common in OOD generalization.}, we can divide the network parameters into two types: \textit{invariant parameters} that are strongly related to invariant features, and \textit{variant parameters} that can mistakenly produce spurious features. Intuitively, the invariant parameters and variant parameters are mutually exclusive of each other, as they are either helpful or harmful to our learning task. In order to correctly identify an ideal subnetwork for OOD generalization, our EVIL method not only selects invariant parameters based on the learning task but also explores variant parameters via discriminating each distribution, i.e., classifying the data based on the distribution information. In this way, the connection between variant parameters and spurious features can be successfully established. By finding those variant parameters that are strongly activated by the distribution information, we can be sure that they should not be identified as invariant ones, which provides an alternative and effective way to improve invariant learning.

Furthermore, to dynamically improve our identification of invariant parameters during the course of network training, we propose to revisit some parameters that hardly vary when facing distribution shifts. Concretely, starting from an initialized partition of invariant parameters and variant ones, we select some variant parameters that show low response to the distribution information, as they might be critical for learning distribution-invariant features. Hence, such parameters are recollected as invariant ones to learn from label information. On the other hand, some invariant parameters that are insensitive to our learning task shall be rejected from sparse training, as they hardly contribute to invariant learning. Through this dynamic process, we are able to identify a robust subnetwork that is stable across different distributions. As shown in Figure~\ref{EVILfig:grad_var}, the invariant parameters learned by our EVIL method show much smaller gradient variance than the rest variant parameters, which manifests its effectiveness for capturing the invariant information. 

By applying our EVIL framework to many existing OOD generalization methods, we conduct extensive empirical comparison and analysis to show that EVIL brings promising improvement with little computation cost. Specifically, when combined with simple ERM, our method achieves $2.4\%$ gains on averaged performance from DomainBed. Furthermore, our EVIL framework can surpass existing sparse training methods for invariant learning by a large margin in various sparsity levels.

To sum up, our contributions are threefold:
\begin{itemize}
	\item[--] We propose a novel sparse training framework for OOD generalization, which can fully explore the variant parameters to capture invariant information.
	
	\item[--] An iterative strategy is designed to dynamically improve the identification of robust subnetworks.
	
	\item[--] The proposed EVIL framework can be deployed to many popular methods with great effectiveness and efficiency. Moreover, EVIL effectively surpasses existing sparse invariant learning methods.
\end{itemize}

\section{Related Work}
\textbf{Invariant learning for OOD generalization} seeks to enforce model predictive invariance when facing distribution shifts~\cite{arjovsky2019invariant, huang2021universal, huang2022they, bai2022rsa}. Invariant Risk Minimization (IRM)~\cite{arjovsky2019invariant} tries to find an optimal classifier for each data distribution such that the spurious information from each domain is left out. Then, Distributionally Robust Optimization (DRO)~\cite{hu2018does, huang2023robust, sagawa2019distributionally, wang2023doe} proposes to tackle the most challenging distribution to improve OOD generalization, which is shown to be effective by using strong regularization penalties. Moreover, Sharpness-Aware Minimization (SAM)~\cite{foret2020sharpness} hopes to learn a flat loss landscape via penalizing the sharpness measurement to improve generalization results~\cite{cha2021swad} and robustness to label noise~\cite{kang2023unleashing, xia2023combating, xia2021sample}. Further, Risk Extrapolation (REx)~\cite{krueger2021out} finds that only focusing on one of the known distributions might not help generalize to unknown distributions. Instead, REx shows it is beneficial to enforce comparable performance among all training distributions via penalizing risk variance. Additionally, other methods draw insights from causality~\cite{glymour2016causal, pearl2009causality, huang2023harnessing, wang2022exploring} to disentangle the invariant features from spurious ones~\cite{gong2016domain, lu2021invariant, mitrovic2020representation} so that prediction would not be affected by distribution shift~\cite{wang2023learning, wang2022watermark}.

Nonetheless, existing invariant learning methods suffer from two major drawbacks. Firstly, some of them a computationally expensive. For example, SAM requires second-order computation to manipulate gradient information, and causality-based methods often require training generative models, which is hard to deploy on large-scale datasets. Secondly, as found out by Gulrajani et al.~\cite{gulrajani2021search}, most methods have limited performances which are even worse than Empirical Risk Minimization (ERM)! However, our EVIL can not only avoid redundant optimization on the variant parameters but also fully capture the invariant feature to achieve superior generalization accuracy.

\textbf{Sparse training for OOD generalization} is first brought out by Morcos et al.~\cite{morcos2019one}, which aims to discover the generalization ability of sparse networks obtained via common initialization methods. Then, Modular Risk Minimization (MRM)~\cite{zhang2021can} shows that sparse training can possibly improve the OOD generalization performance compared to the original dense network. However, MRM is designed in a static way, which cannot be optimized along with network training, hindering the sparse learning results. To tackle this issue, Sparse Invariant Risk Minimization (SparseIRM)~\cite{zhou2022sparse} proposes to conduct the sparse training process and IRM simultaneously. As a result, its generalization performance is further improved compared to MRM.

Despite the improvement of existing sparse methods, they are still suboptimal as the sparse training could be affected by the noisy gradient from the learning task. Meanwhile, the pruned parameters are not properly leveraged, which would cause non-negligible information loss. Fortunately, EVIL can fully explore both variant and invariant parameters dynamically. Thus, it finds an ideal subnetwork minimally influenced by distribution shift.

\section{A Critical Analysis of Sparse Training with OOD Data}
\label{EVIL:theoretic_analysis}
OOD generalization aims to learn an invariant predictor by leveraging multiple distributions of training data such that the generalization performance on unseen test data distributions. Practically, we usually have multiple datasets correspondingly drawn from $m$ distributions (also termed environment), $\mathcal{E}=\{e_1, \ldots e_m\}$, where each distribution $e=\{(\mathbf{x}_i, y_i)\}_{i=0}^n$ contains $n$ examples $\mathbf{x}\in X\in \mathbb{R}$ with class label $y\in Y\in\mathbb{R}^c$. Therefore, for each example from distribution $e$, we can assign a distribution index $d\in\mathbb{R}^m$ and denote a data point as $(\mathbf{x}, y, d)$. Moreover, we have a test dataset sampled from unseen distributions $\mathcal{E}_{unseen}$ to evaluate the generalization performance of our invariant learning. Let $f_{\theta}: \mathbb{R}\rightarrow\mathbb{R}^D$ be a parameterized model with parameters $\theta\in\Theta$ which extracts feature $Z\in\mathbb{R}^M$. Our goal is to prune a sparse subnetwork from an overparameterized model so that variant features can be excluded from making the final prediction. Therefore, OOD generalization can be improved.

\textbf{Data Generation Process.} By following the same formulation and problem setting from Zhang et al.~\cite{zhang2021can}, we assume the input variable $X^e$ from environment $e$ is generated from latent variables $Z^e=(Z^e_{inv}, Z^e_{var})$. Intuitively, the input $X^e$ indicates the image pixels, while $Z^e_{inv}$ stands for the feature of the object-of-interest that stays invariant across different environments, and $Z^e_{var}$ denotes the spurious feature which is introduced by the change of environments. Then, the data is generated through $X^e = G(Z^e_{inv}, Z^e_{var})$ where $G(\cdot)$ denotes the data generating function. To obtain an OOD-robust model, we hope to extract learning representations $Z^e$ which can recover the invariant feature $Z^e_{inv}$, meanwhile excluding the variant feature $Z^e_{var}$. Such a process is modeled through $Z^e=f_{\theta}(X^e)$, where we hope $Z^e\approx \left[Z^e_{inv}, \mathbf{0}\right]$. Hence, based on the extracted feature, we can make predictions through a classification head $\hat{Y}^e=h(Z^e)$ and train our model by minimizing the error between the prediction $\hat{Y}^e$ and ground truth label $Y^e$.

\textbf{The Cause of Data Bias.} Based on the data generation process, here we explain why different distributions contain biases that hinder the generalization result. We consider a simple example where $Z^e_{inv}$ and $Z^e_{var}$ are multivariate variables with binary elements, i.e., $Z^e_{inv}\in \{-1,1\}^{M_{inv}}$ and $Z^e_{var}\in \{-1,1\}^{M_{var}}$, in which $M_{inv}$ and $M_{var}$ denotes the dimension of invariant feature and spurious feature, respectively. We have class label $Y^e\in\{-1, 1\}$ and distribution index $D\in\{-1, 1\}$. Since the invariant feature stays constant across environments, we assume each element of $Z^e_{inv}$ is equal to $Y^e$. On the other hand, we assume each element in $Z^e_{var}$ takes a value equal to $Y^e$ with probability $p^e$ and $-Y^e$ with probability $1-p^e$~\cite{zhang2021can}. When $p^e$ is large, the spurious feature would be closely correlated with the class label, hence being unlikely to introduce large data biases. Conversely, if $p^e$ is small, $Z^e_{var}$ can easily introduce noisy signals that might flip the prediction.

Additionally, we analyze the domain knowledge to provide an opposite perspective, which is overlooked by previous works~\cite{zhang2021can, zhou2022sparse}. Concretely, the change of distribution index $D$ is the cause of introducing a spurious feature, i.e., $D\rightarrow Z^e_{var}$, as described by many proposed causal structures~\cite{gong2016domain, liu2021learning, sun2021recovering}. Therefore, when given the distribution $D$, we can find a specific type of spurious feature. Hence, we assume each element of $Z^e_{var}$ is equal to $D$. On the other hand, we consider $Z^e_{inv}$ takes the value of $D$ with probability $q^e$ and $-D$ with probability $1-q^e$. It has been commonly assumed that the invariant feature and domain knowledge are independent~\cite{gong2016domain}, thus the probability $q^e$ could approximately be $0.5$.

\textbf{The Flaw of Common Sparse Training Strategy}
Existing studies on sparse invariant learning~\cite{zhang2021can, zhou2022sparse} have shown that when pruning an overparameterized model, the OOD generalization performance could be improved substantially. However, we find that the existing pruning strategy, which is based on ERM or objectives only related to labels, could be suboptimal. Specifically, we consider the same data setting described above, $Z^e_{inv}\in\{-1, 1\}^{M_{inv}}$ and $Z^e_{var}\in\{-1, 1\}^{M_{var}}$. The data generating function $G$ is simplified as an identity map~\cite{tsipras2018robustness, rosenfeld2020risks}, thus $X=(Z^e_{inv}, Z^e_{var})$. Suppose the classification model $f_{\theta}$ is a linear layer, we have a mask $\mathbf{m}$ randomly initialized with $0-1$ values to prune the parameter $\theta$, and its sparsity ratio is set to $R=\frac{M_{var}}{M}$. Particularly, we denote the selected invariant parameters as $\theta_{inv}=\mathbf{m}\circ\theta$ and the pruned variant parameters as $\theta_{var}=(\mathbf{1}-\mathbf{m})\circ\theta$ where $\circ$ is the element-wise production. To ease the calculation, let the parameter values follow a unit norm, i.e., $\theta=\mathbf{1}\frac{1}{\sqrt{M}}$\footnote{Note that our assumption is more general than that from Zhang et al.~\cite{zhang2021can}, in which only two extreme case are considered: an optimal sparse invariant network only extracts invariant feature and a network completely depending on spurious feature. Our assumption is practical since it is similar to an initial state where all the parameters are initialized with unit-norm.}.


\begin{proposition}
    Consider a biased dataset described above, where $Z^e_{\text{inv}} \in \{-1, 1\}^{M_{\text{inv}}}$ and $Z^e_{\text{var}} \in \{-1, 1\}^{M_{\text{var}}}$. Let mask $\mathbf{m}$ be randomly initialized to $\{0, 1\}$ values with sparsity ratio $R = \frac{M_{\text{var}}}{M}$, and assume $Z^e$ is a multivariate variable with independent elements. For a common sparse training strategy that aims to minimize empirical risk:
    \begin{equation}
        \text{Err}^e = \frac{1}{2}\mathbb{E}_{(X^e, Y^e)\sim e}\left[1 - Y^e\hat{Y}^e\right],
    \end{equation}
    we have:
    \begin{itemize}
        \item[--] The common strategy fails to find invariant parameters, i.e., $\mathbf{m}_{i \in \{0, \dots, M_{\text{inv}}\}}$ remains unupdated. When leveraging domain knowledge with regularization:
        \begin{equation}
            \text{Err}^d = \frac{1}{2}\mathbb{E}_{(X, Y)\sim \mathcal{E}}\left[1 - D\hat{D}\right],
        \end{equation}
        the invariant parameters can be effectively selected with probability at least $1 - \frac{q^e}{2}$;
        \item[--] On an unknown distribution, the performance of the common strategy is highly sensitive to $p^e$: $\text{Err}^e \le \mathcal{O}(e^{-(p^e)^4})$, while leveraging domain knowledge achieves a tighter error bound when $p^e$ is small: $\text{Err}^e \le \mathcal{O}(e^{-(p^e)^2})$.
    \end{itemize}
    \label{EVIL:theorem}
\end{proposition}
As we find out, the pruning strategy is not sufficient to find an ideal subnetwork that can exclude spurious features meanwhile extracting invariant features. This is because the invariant parameters do not produce any error. As a result, existing strategies based on connection sensitivity~\cite{lee2018snip}, weight value~\cite{evci2020rigging}, and Fisher information~\cite{sung2021training} could be suboptimal when dealing with OOD problems because the gradient information is not actually related to invariant parameters, but variant parameters. Based on this intuition, we proposed a simple yet effective strategy that leverages an additional domain knowledge regularization to explore the invariant parameters. Thanks to such a regularization, the invariant parameters can be selected because they generate gradients when calculating the distribution regularization, thus easy to find. Meanwhile, the variant parameters can still be excluded to avoid learning spurious features. Moreover, based on the error bounds, our method is insensitive to the spurious correlation $1-p^e$ compared to the common strategy. In a difficult scenario where $p^e$ is small, our method can still be robust to distribution shift.

\section{Methodology}
\begin{figure*}[t]
	\centering
	\includegraphics[width=0.9\linewidth]{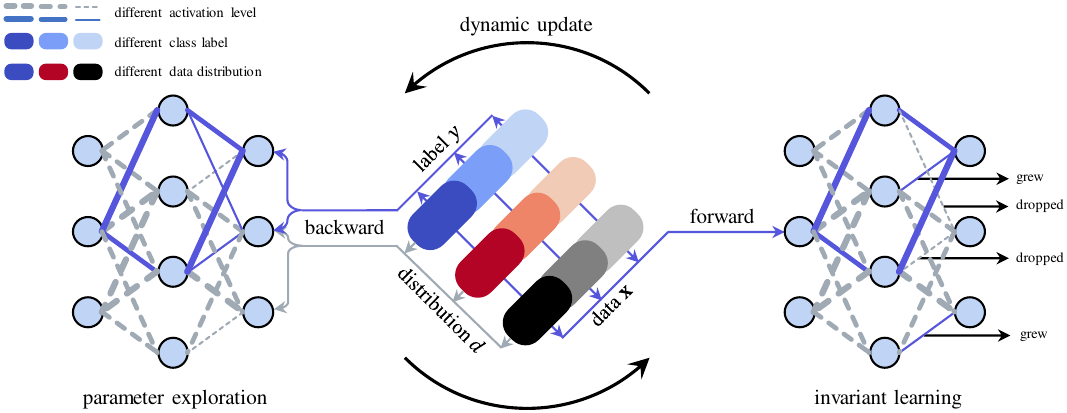}
	\caption{Learning flow of EVIL: The blocks in the middle are the training dataset where different levels of shades denote different classes, and different types of color indicate different data distributions. The blue arrows ({\color[rgb]{0.33725, 0.33725, 0.86667}$\boldsymbol{\rightarrow}$}) and gray arrows ({\color[rgb]{0.62745, 0.66667, 0.7098}$\boldsymbol{\rightarrow}$}) stand for the information flow related to label and distribution, respectively. Moreover, the blue solid lines ({\color[rgb]{0.33725, 0.33725, 0.86667} \rule[.5ex]{1em}{1.5pt}}) and gray dashed lines ({\color[rgb]{0.62745, 0.66667, 0.7098} \rule[.5ex]{0.2em}{1.5pt}}{\color[rgb]{1, 1, 1}\rule[.5ex]{0.2em}{1.5pt}}{\color[rgb]{0.62745, 0.66667, 0.7098}\rule[.5ex]{0.2em}{1.5pt}}{\color[rgb]{1, 1, 1}\rule[.5ex]{0.2em}{1.5pt}}{\color[rgb]{0.62745, 0.66667, 0.7098}\rule[.5ex]{0.2em}{1.5pt}}) that connect neurons are the selected invariant parameters and pruned variant ones, respectively.}
	\label{EVILfig:framework}
\end{figure*}

In this section, we introduce our EVIL framework as shown in Figure~\ref{EVILfig:framework}. In the learning flow of EVIL, there are two procedures: Parameter Exploration, in which we propose to not only study invariant parameters but also explore the variant ones; and Invariant Learning, where we train the identified subnetwork to optimize the invariant parameters. 

In the following content, we first introduce our EVIL framework, which contains the aforementioned two procedures. Then, we carefully demonstrate the realization of EVIL using an important optimization method: SAM~\cite{foret2020sharpness}, which largely improves OOD generalization.

\subsection{The Proposed EVIL Framework}
\label{EVIL:evil}
In order to get a good initialization, a few steps of pre-training are commonly conducted by minimizing a learning objective $\mathcal{L}(f_{\theta}(\mathbf{x}))$~\cite{chen2021lottery, dettmers2019sparse, evci2020rigging, morcos2019one}, where $f_{\theta}$ is a deep model with parameters $\theta\in\mathbb{R}^N$. To sparsify the deep model, a binary mask $\mathbf{m}$ is often applied through element-wise product $\mathbf{m}\circ\theta$. Such a mask $\mathbf{m}$ is either learned through optimization~\cite{csordas2020neural, dettmers2019sparse, louizoslearning, zhang2021can, zhou2022sparse}, or obtained based on certain criteria, such as connection sensitivity~\cite{lee2018snip}, weight value~\cite{evci2020rigging}, fisher information~\cite{sung2021training}, or even random initialization~\cite{frankle2019lottery, liu2022unreasonable}. By setting the sparsity ratio $R=1-\frac{\|\mathbf{m}\|_0}{|\theta|}$, we can decide how many parameters are rejected from sparse training. Then, we start from a pre-trained model with an initialized mask $\mathbf{m}$.

\textbf{Parameter Exploration.}\label{EVIL:parameter_exploration} In this step, we mainly have two optimization targets:
\begin{equation}
	\min_{f_{\theta_{inv}}\otimes h} \mathcal{L}_{inv}(h(f_{\theta_{inv}}(\mathbf{x})), y),
	\label{EVILeq:inv}
\end{equation}
\begin{equation}
	\min_{f_{\theta_{var}}\otimes g} \mathcal{L}_{var}(g(f_{\theta_{var}}(\mathbf{x})), d),
	\label{EVILeq:var}
\end{equation}
where $\theta_{inv}=\mathbf{m}\circ\theta$ and $\theta_{var}=(1 - \mathbf{m})\circ\theta$ denote the corresponding invariant parameters and variant ones divided by the mask $\mathbf{m}$, $h$ and $g$ are two fully-connected layers which map the extracted features into class label space $\mathbb{R}^c$ and distribution index space $\mathbb{R}^m$, respectively. Intuitively, the objective $\mathcal{L}_{inv}$ is the classification task, which tries to make predictions based on the label information, and $\mathcal{L}_{var}$ tries to discriminate each distribution based on the distribution information, which is spurious and unwanted. 

By minimizing $\mathcal{L}_{inv}$ in Equation~\eqref{EVILeq:inv}, the gradient magnitude $\nabla_{\theta_{inv}}\mathcal{L}_{inv}$~\cite{lee2018snip} can be used to find the most relevant parameters to our loss function (Note that other aforementioned sparse training criteria can be used). Similarly, by minimizing $\mathcal{L}_{var}$ in Equation~\eqref{EVILeq:var}, those parameters with large $\nabla_{\theta_{var}}\mathcal{L}_{var}$ are sensitive to the spurious information, which cannot help produce invariant features. Thus, we can sort the parameters based on the gradient magnitude to show how much they are activated by their corresponding objective.


Further, to dynamically improve our sparsification. We propose to update the mask $\mathbf{m}$ for every $\Delta T$ iterations by rejecting the least activated invariant parameters, meanwhile calling back the least activated variant parameters as invariant ones. Specifically, this process is conducted as:
\begin{equation}
	\begin{aligned}
		&\mathbf{m}\left[ArgTopK(-|\nabla_{\theta_{inv}}\mathcal{L}_{inv}|, \|\mathbf{m}\|_0S(t, \alpha, T))\right]=0,\\
		&\mathbf{m}\left[ArgTopK(-|\nabla_{\theta_{var}}\mathcal{L}_{var}|, \|\mathbf{m}\|_0S(t, \alpha, T))\right]=1,
		\label{EVILeq:update}	
	\end{aligned}	
\end{equation}
where $ArgTopK(v, k)$ returns the indices of top-$k$ elements regarding value $v$, $\mathbf{m}\left[\cdot\right]$ denotes indexing $\mathbf{m}$. Moreover, to decide how many parameters should be exchanged, we follow Dettmers \& Zettlemoyer~\cite{dettmers2019sparse} to use cosine annealing function $S(t, \alpha, T)=\frac{\alpha}{2}(1+cos(\frac{t\pi}{T}))$, where $t$ and $T$ are the current iteration and total iterations, respectively, and hyper-parameter $\alpha$ decides the largest value. Intuitively, such a cosine annealing function gradually changes from $\alpha<1$ to 0. Through Equation~\eqref{EVILeq:update}, the obtained new mask finds the parameters that are less affected by the distribution information and more related to our learning task than the previous one, further improving the invariant learning performance.

\begin{algorithm}[t]
    \caption{EVIL}
    \label{alg:EVIL}
    \begin{algorithmic}[1]
        \REQUIRE Multiple training sets $\mathcal{E}=\{e_1, \ldots e_m\}$; Learning model $f_{\theta}$; Cosine annealing function $S(t, \alpha, T)$, mask $\mathbf{m}$ initialized based on weight value, iteration number of pre-training $T_{\text{pre}}$.
        \FOR{$t = 0, 1, \ldots, T-1$}
            \STATE Optimize via Equation~\eqref{EVILeq:inv}; \COMMENT{\textit{\color{black!60} Invariant learning}}
            \IF{$t > T_{\text{pre}}$ \text{and} $t \% \Delta T == 0$}
                \STATE Obtain gradients of $\theta_{\text{inv}}$ and $\theta_{\text{var}}$ via Equation~\eqref{EVILeq:inv} and Equation~\eqref{EVILeq:var}, respectively; \COMMENT{\textit{\color{black!60} Parameter exploration}}
                \STATE Update the mask $\mathbf{m}$ via Equation~\eqref{EVILeq:update};
            \ENDIF
        \ENDFOR
    \end{algorithmic}
\end{algorithm}

\textbf{Invariant Learning.}\label{EVIL:invariant_learning}
After obtaining the updated mask $\mathbf{m}$, we then use the invariant parameters as a subnetwork to conduct invariant learning, which is generally formed as:
\begin{equation}
	\mathcal{L}_{inv}(h(f_{\theta_{inv}}(\mathbf{x})), y) = \mathcal{L}_{ce} + \lambda\mathcal{L}_{reg},
	\label{EVILeq:invariant_learning}
\end{equation}
where the first term is the empirical risk computed through cross-entropy loss, and the second term is the invariant learning regularization with penalty weight $\lambda$, which can be realized by many popular methods. For instance, to use IRM~\cite{arjovsky2019invariant}, the regularization term is
\begin{equation}
	\mathcal{L}_{reg}=\frac{1}{mn}\sum_{\mathbf{x}\in \mathcal{E}}\|\nabla_{h|h=\mathbf{1}}\mathcal{L}_{ce}(h(f_{\theta_{inv}}(\mathbf{x})), y)\|^2.
\end{equation}
To implement REx~\cite{krueger2021out}, we penalize the loss variance as
\begin{equation}
	\mathcal{L}_{reg}=\frac{1}{mn}\sum_{\mathbf{x}\in \mathcal{E}}Var(\{\mathcal{L}_{ce}(h(f_{\theta_{inv}}(\mathbf{x}|d)), y)\}_{d=1}^m).
\end{equation}
Moreover, we can focus on the worst-case distribution to realize DRO~\cite{hu2018does, sagawa2019distributionally}:
\begin{equation}
	\mathcal{L}_{inv}=\min_{\theta_{inv}} \max_{e\in \mathcal{E}}\frac{1}{n}\sum_{\mathbf{x}\in e}\mathcal{L}_{ce}(h(f_{\theta_{inv}}(\mathbf{x})), y).
\end{equation}
By combining with existing methods, their performance can be largely improved by EVIL, as shown in Section~\ref{EVIL:experiments}. The general process of EVIL is summarized in Algorithm~\ref{alg:EVIL}. Next, we describe one realization of EVIL by adopting the SAM optimizer~\cite{foret2020sharpness} to further improve the generalization performance.

\section{Experiment}
\label{EVIL:experiments}
In this section, we conduct extensive experiments to evaluate the performance of EVIL based on a well-known testbed for OOD generalization: DomainBed~\cite{gulrajani2021search}. Specifically, we first describe the experimental setup. Then, we improve the performance of well-known invariant learning methods by deploying EVIL, including ERM, IRM~\cite{arjovsky2019invariant}, REx~\cite{krueger2021out}, DRO~\cite{hu2018does, sagawa2019distributionally}, SAM~\cite{foret2020sharpness}, CORelation ALignment (CORAL)~\cite{sun2016deep}, SWAD~\cite{cha2021swad}, and MIRO~\cite{cha2022domain}. Further, we compare EVIL and its variant EVIL-SAM with other existing sparse invariant learning methods, including MRM~\cite{zhang2021can}, SparseIRM~\cite{zhou2022sparse}, and report the results under different sparsity levels ($20\%$, $40\%$, $60\%$, and $80\%$). Finally, we perform various analytical experiments to validate the effectiveness and efficiency of EVIL.

\subsection{Practical Implementation}
\setlength{\intextsep}{6pt}
\setlength{\columnsep}{10pt}
\begin{wrapfigure}{r}{5cm}
	\includegraphics[width=0.85\linewidth]{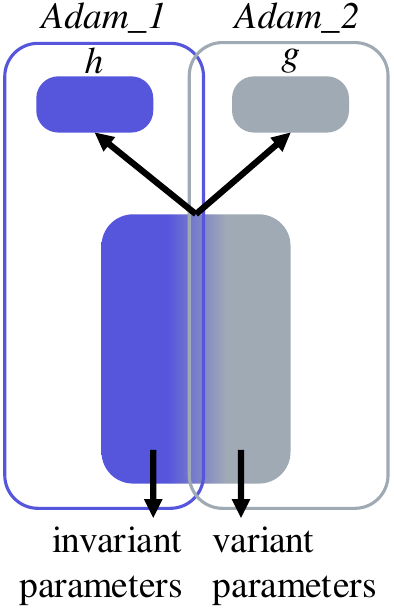}
	\caption{Illustration of our optimizers.}
	\label{EVILfig:optim}
\end{wrapfigure}

All our experiments are conducted on a single NVIDIA 3090 using PyTorch. To implement our EVIL framework, we first pre-train the models using ERM for 1,000 iterations. Then, a mask $\mathbf{m}$ is initialized based on the weight value. Specifically, by setting a sparsity ratio $R$, we can select parameters $R|\theta|$-largest weight values by setting their corresponding mask value as $1$. During parameter exploration, we first pass the gradient of invariant learning loss $\mathcal{L}_{inv}$, based on which we can sort the invariant parameters with their gradient magnitude from largest to smallest. Then, we reject the $S(t, \alpha, T)$-least invariant parameters by setting their corresponding mask to 0. Similarly, we use the gradient of $\mathcal{L}_{var}$ to sort the variant parameters and recollect top-$S(t, \alpha, T)$ parameters. During invariant learning, we can apply the mask to parameter values as well as their corresponding gradients to conduct sparse training.

To optimize our model, we use Adam~\cite{kingma2014adam} optimizer with an initial learning rate $1e-3$ without weight decay. Moreover, to avoid the conflict between optimizing invariant parameters and variant parameters, we adopt two Adam optimizers, denoted as $Adam_1$ and $Adam_2$, to correspondingly include the invariant and variant parameters. Moreover, $Adam_1$ would include the class prediction head $h$ and $Adam_2$ would include the distribution prediction head $g$, as illustrated in Figure~\ref{EVILfig:optim}. During the training, $Adam_1$ is mainly used to optimize the invariant parameters, but $Adam_2$ is just employed to optimize the variant ones.

For implementing baseline methods, we mainly follow~\cite{gulrajani2021search} to set the hyper-parameters. Specifically, for DRO, we set $\eta$ as $1e-2$ to update the group importance. For IRM, we set $\lambda=1e2$ to trade off the invariant regularizer. The similar $\lambda$ for penalization from REx is set to $1e1$. For CORAL and MMD, set the trade-off weight $\gamma$ as $1$. For implementing SagNet, we set the weight for adversarial loss as $0.1$. For SAM, we do not use Adaptive SAM~\cite{kwon2021asam} and set the perturbation magnitude $\rho$ as $0.05$.

\begin{table*}[t]
    \centering
    \scriptsize 
    \caption{Comparison between OOD generalization methods and our EVIL realization on some typical methods. Test accuracies on seven OOD generalization benchmarks from DomainBed. Best results and second best results are highlighted. $\dagger$ denotes results from \cite{cha2021swad}.}
    \setlength{\tabcolsep}{1.1mm}
    \renewcommand{\arraystretch}{1.2}
    \label{EVILtab:deployment}
    \begin{tabular}{l ccccccc c c}
        \toprule
        \textbf{Algorithm} & \textbf{CMNIST} & \textbf{RMNIST} & \textbf{PACS} & \textbf{VLCS} & \textbf{OfficeHome} & \textbf{TerraInc} & \textbf{DomainNet} & \textbf{Average} & \textbf{FLOPs} \\
        \midrule
        I-Mixup$^\dagger$ & 31.3 & 97.8 & 84.6 & 77.4 & 68.1 & 47.9 & 39.2 & 63.7 & -- \\ 
        MLDG$^\dagger$    & 36.9 & 98.0 & 84.9 & 77.2 & 66.8 & 47.8 & 41.2 & 64.6 & -- \\
        MMD$^\dagger$     & \textbf{42.6} & 98.1 & 84.7 & 77.5 & 66.4 & 42.2 & 23.4 & 62.1 & -- \\
        DANN$^\dagger$    & 29.0 & 89.1 & 83.7 & 78.6 & 65.9 & 46.7 & 38.3 & 61.6 & -- \\
        CDANN$^\dagger$   & 31.1 & 96.3 & 82.6 & 77.5 & 65.7 & 45.8 & 38.3 & 62.5 & -- \\
        MTL$^\dagger$     & 30.4 & 97.2 & 84.6 & 77.2 & 66.4 & 45.6 & 40.6 & 63.1 & -- \\
        SagNet$^\dagger$  & 34.2 & 96.4 & 86.3 & 77.8 & 68.1 & 48.6 & 40.3 & 64.5 & -- \\
        ARM$^\dagger$     & 32.6 & 98.1 & 85.1 & 77.6 & 64.8 & 45.5 & 35.5 & 62.5 & -- \\
        RSC$^\dagger$     & 35.2 & 96.3 & 85.2 & 77.1 & 65.5 & 46.6 & 38.9 & 63.5 & -- \\
        Mixstyle$^\dagger$& 38.5 & 97.2 & 85.2 & 77.9 & 60.4 & 44.0 & 34.0 & 62.4 & -- \\
        \midrule
        ERM$^\dagger$     & 34.2 & 98.0 & 83.3 & 76.8 & 67.3 & 46.2 & 40.8 & 63.8 & 1$\times$ \\ 
        EVIL              & 39.4 & 98.4 & 86.0 & 78.8 & 68.2 & 49.1 & 43.8 & 66.2 & 0.42$\times$ \\[-1.6ex]
        & \tiny($\pm1.1$) & \tiny($\pm0.1$) & \tiny($\pm0.1$) & \tiny($\pm0.2$) & \tiny($\pm0.2$) & \tiny($\pm0.2$) & \tiny($\pm0.3$) & {\color[rgb]{0.3, 0.3, 0.8}\tiny($\mathbf{\uparrow2.4}$)} & \\
        \cmidrule(lr){1-10}
        DRO$^\dagger$     & 32.2 & 97.9 & 84.4 & 76.7 & 66.0 & 43.2 & 33.3 & 61.9 & 1$\times$ \\
        EVIL-DRO          & 34.2 & 98.2 & 85.6 & 77.7 & 66.4 & 49.1 & 35.5 & 63.8 & 0.42$\times$ \\[-1.6ex]
        & \tiny($\pm1.7$) & \tiny($\pm0.1$) & \tiny($\pm0.2$) & \tiny($\pm0.2$) & \tiny($\pm0.1$) & \tiny($\pm0.2$) & \tiny($\pm0.2$) & {\color[rgb]{0.3, 0.3, 0.8}\tiny($\mathbf{\uparrow1.9}$)} & \\
        \cmidrule(lr){1-10}
        IRM$^\dagger$     & 36.3 & 97.7 & 83.5 & 78.6 & 64.3 & 47.6 & 33.9 & 63.1 & 1$\times$ \\
        EVIL-IRM          & 39.1 & 98.3 & 85.1 & 78.8 & 66.4 & 48.3 & 36.0 & 64.6 & 0.42$\times$ \\[-1.6ex]
        & \tiny($\pm2.2$) & \tiny($\pm0.2$) & \tiny($\pm0.1$) & \tiny($\pm0.1$) & \tiny($\pm0.1$) & \tiny($\pm0.3$) & \tiny($\pm0.3$) & {\color[rgb]{0.3, 0.3, 0.8}\tiny($\mathbf{\uparrow1.5}$)} & \\
        \cmidrule(lr){1-10}
        REx$^\dagger$     & 39.2 & 97.3 & 84.9 & 78.3 & 66.4 & 46.4 & 33.6 & 63.7 & 1$\times$ \\
        EVIL-REx          & 41.2 & \underline{98.7} & 86.0 & 79.1 & 68.0 & 48.4 & 34.5 & 65.1 & 0.42$\times$ \\[-1.6ex]
        & \tiny($\pm1.3$) & \tiny($\pm0.1$) & \tiny($\pm0.1$) & \tiny($\pm0.2$) & \tiny($\pm0.2$) & \tiny($\pm0.3$) & \tiny($\pm0.1$) & {\color[rgb]{0.3, 0.3, 0.8}\tiny($\mathbf{\uparrow1.4}$)} & \\
        \cmidrule(lr){1-10}
        CORAL$^\dagger$   & 29.9 & 98.1 & 86.2 & 78.8 & 68.7 & 47.7 & 41.5 & 64.4 & 1$\times$ \\ 
        EVIL-CORAL        & 34.5 & 98.6 & 86.9 & \underline{79.2} & \underline{69.0} & 49.2 & 42.6 & 65.7 & 0.43$\times$ \\[-1.6ex]
        & \tiny($\pm1.9$) & \tiny($\pm0.1$) & \tiny($\pm0.2$) & \tiny($\pm0.1$) & \tiny($\pm0.1$) & \tiny($\pm0.2$) & \tiny($\pm0.3$) & {\color[rgb]{0.3, 0.3, 0.8}\tiny($\mathbf{\uparrow1.3}$)} & \\
        \cmidrule(lr){1-10}
        SWAD              & 38.3 & 98.1 & 88.1 & 79.1 & 70.6 & 50.0 & 46.5 & 67.2 & 1$\times$ \\
        EVIL-SWAD         & 38.7 & 98.3 & \textbf{88.3} & 79.3 & \textbf{71.7} & \textbf{51.2} & \textbf{46.9} & \textbf{67.7} & 0.43$\times$ \\[-1.6ex]
        & \tiny($\pm2.3$) & \tiny($\pm0.3$) & \tiny($\pm0.1$) & \tiny($\pm0.1$) & \tiny($\pm0.2$) & \tiny($\pm0.3$) & \tiny($\pm0.2$) & {\color[rgb]{0.3, 0.3, 0.8}\tiny($\mathbf{\uparrow0.5}$)} & \\
        \cmidrule(lr){1-10}
        MIRO              & 39.4 & 97.5 & 85.4 & 79.0 & 70.5 & 50.4 & 44.3 & 66.6 & 1$\times$ \\
        EVIL-MIRO         & 40.2 & 98.6 & 85.8 & 79.4 & \underline{71.2} & \underline{50.9} & \underline{45.0} & 67.3 & 0.45$\times$ \\[-1.6ex]
        & \tiny($\pm2.3$) & \tiny($\pm0.3$) & \tiny($\pm0.1$) & \tiny($\pm0.1$) & \tiny($\pm0.2$) & \tiny($\pm0.3$) & \tiny($\pm0.2$) & {\color[rgb]{0.3, 0.3, 0.8}\tiny($\mathbf{\uparrow0.7}$)} & \\
        \cmidrule(lr){1-10}
        SAM               & 38.5 & 98.1 & 85.8 & 79.4 & 69.6 & 43.3 & 44.3 & 65.6 & 2$\times$ \\
        EVIL-SAM          & \underline{40.4} & \textbf{98.8} & \underline{87.8} & \textbf{80.1} & 70.3 & 50.5 & \underline{45.0} & \underline{67.5} & 0.89$\times$ \\[-1.6ex]
        & \tiny($\pm2.3$) & \tiny($\pm0.3$) & \tiny($\pm0.1$) & \tiny($\pm0.1$) & \tiny($\pm0.2$) & \tiny($\pm0.3$) & \tiny($\pm0.2$) & {\color[rgb]{0.3, 0.3, 0.8}\tiny($\mathbf{\uparrow1.9}$)} & \\ 
        \bottomrule
    \end{tabular}
\end{table*}

\subsection{Experimental Setup}
\textbf{Evaluation Protocol.} We follow the experimental setting of DomainBed~\cite{gulrajani2021search} to evaluation OOD generalization performance. Specifically, DomainBed contains seven benchmark datasets: CMNIST~\cite{arjovsky2019invariant} (60,000 images, 10 classes, and 3 domains), RMNIST~\cite{ghifary2015domain} (60,000 images, 10 classes, and 6 domains), PACS~\cite{li2017deeper} (9,991 images, 7 classes, 4 domains), VLCS~\cite{fang2013unbiased} (10,729 images, 5 classes, and 4 domains), OfficeHome~\cite{venkateswara2017deep} (15,588 images, 65 classes, and 4 domains), TerraIncognita~\cite{beery2018recognition} (24,788 images, 10 classes, and 4 domains), DomainNet~\cite{peng2019moment} (586,575 images, 345 classes, and 6 domains), WILDS~\cite{koh2021wilds} (a testbed contains various dataset with significant distribution shfit, here we use two typical datasets: iWildCam and FMoW), ImageNet~\cite{russakovsky2015imagenet} (contain 1000 classes, here we use ImageNet dataset for fine-tuning, and use many of its variant dataset for OOD evaluation, including: ImageNetV2~\cite{recht2019imagenet}, ImageNetR~\cite{hendrycks2021many}, ImageNetA~\cite{hendrycks2021natural}, ImageNetSketch~\cite{wang2019learning}, and ObjectNet~\cite{barbu2019objectnet}). For each benchmark dataset, we leave one domain out of the training dataset and use it as an OOD test dataset. Moreover, we use pre-trained ResNet-50~\cite{he2016deep} as our backbone model and train it for 5,000 iterations on all datasets except DomainNet, which requires 15,000 iterations to converge. The test accuracies generated by training models from the last step are provided. To avoid randomness, three independent trials are conducted.

\begin{table*}[t]
    \centering
    \scriptsize 
    \caption{Comparison between existing sparse invariant learning methods and EVIL varying sparsity levels. The test accuracies on seven OOD generalization benchmarks from DomainBed are provided. We highlight the \textbf{best results} and the \underline{second best results}.}
    \setlength{\tabcolsep}{1.1mm}
    \renewcommand{\arraystretch}{1.2} 
    \label{EVILtab:sparse_comparison}
    \begin{tabular}{cl cccccccc c}
        \toprule
        \textbf{R} & \textbf{Algorithm} & \textbf{CMNIST} & \textbf{RMNIST} & \textbf{PACS} & \textbf{VLCS} & \textbf{OfficeHome} & \textbf{TerraInc} & \textbf{DomainNet} & \textbf{Average} & \textbf{FLOPs} \\
        \midrule
        \multirow{6}{*}{20\%} 
        & MRM & 31.5 & 89.3 & 78.3 & 70.0 & 63.6 & 42.7 & 37.9 & 59.0 & 0.82$\times$ \\ 
        & SparseIRM & 31.5 & 92.2 & 80.8 & 71.2 & 63.7 & 43.0 & 39.6 & 60.3 & 0.81$\times$ \\
        & EVIL & 34.1 & 95.8 & 82.2 & 73.7 & 66.3 & 45.3 & 41.9 & 62.7 & 0.82$\times$ \\[-1.5ex]
        & & \tiny($\pm1.7$) & \tiny($\pm0.1$) & \tiny($\pm0.2$) & \tiny($\pm0.2$) & \tiny($\pm0.2$) & \tiny($\pm0.3$) & \tiny($\pm0.2$) & & \\
        & EVIL-SAM & 35.2 & 96.3 & 83.6 & 74.0 & 65.6 & 45.3 & 42.6 & 63.2 & 1.62$\times$ \\[-1.5ex]
        & & \tiny($\pm2.3$) & \tiny($\pm0.2$) & \tiny($\pm0.2$) & \tiny($\pm0.2$) & \tiny($\pm0.2$) & \tiny($\pm0.1$) & \tiny($\pm0.1$) & & \\ 
        
        \cmidrule(lr){1-11}
        
        \multirow{6}{*}{40\%} 
        & MRM & 36.2 & 95.8 & 81.9 & 73.5 & 63.1 & 45.6 & 40.4 & 62.3 & 0.62$\times$ \\ 
        & SparseIRM & 35.7 & 96.4 & 82.5 & 74.2 & 66.8 & 47.8 & 42.6 & 63.7 & 0.62$\times$ \\
        & EVIL & 38.9 & 97.3 & 84.7 & 75.3 & 66.4 & 47.1 & 44.0 & 64.8 & 0.62$\times$ \\[-1.5ex]
        & & \tiny($\pm1.6$) & \tiny($\pm0.2$) & \tiny($\pm0.1$) & \tiny($\pm0.2$) & \tiny($\pm0.1$) & \tiny($\pm0.1$) & \tiny($\pm0.1$) & & \\
        & EVIL-SAM & 38.8 & 97.9 & 84.8 & 77.4 & 66.9 & 48.1 & \textbf{45.2} & 65.6 & 1.33$\times$ \\[-1.5ex]
        & & \tiny($\pm2.4$) & \tiny($\pm0.3$) & \tiny($\pm0.3$) & \tiny($\pm0.2$) & \tiny($\pm0.1$) & \tiny($\pm0.2$) & \tiny($\pm0.2$) & & \\ 
        
        \cmidrule(lr){1-11}
        
        \multirow{6}{*}{60\%} 
        & MRM & 38.2 & 97.6 & 83.6 & 76.8 & 66.5 & 46.7 & 40.3 & 64.2 & 0.41$\times$ \\ 
        & SparseIRM & 37.9 & 97.9 & 84.9 & 77.3 & 65.1 & 48.8 & 42.0 & 64.8 & 0.42$\times$ \\
        & EVIL & \underline{39.4} & \underline{98.4} & \underline{86.0} & \underline{78.8} & \underline{68.2} & \underline{49.1} & 43.8 & \underline{66.3} & 0.42$\times$ \\[-1.5ex]
        & & \tiny($\pm1.4$) & \tiny($\pm0.2$) & \tiny($\pm0.1$) & \tiny($\pm0.2$) & \tiny($\pm0.2$) & \tiny($\pm0.2$) & \tiny($\pm0.3$) & & \\
        & EVIL-SAM & \textbf{40.4} & \textbf{98.8} & \textbf{87.8} & \textbf{80.1} & \textbf{70.3} & \textbf{50.5} & \underline{45.0} & \textbf{67.6} & 0.89$\times$ \\[-1.5ex]
        & & \tiny($\pm2.2$) & \tiny($\pm0.1$) & \tiny($\pm0.1$) & \tiny($\pm0.1$) & \tiny($\pm0.2$) & \tiny($\pm0.1$) & \tiny($\pm0.2$) & & \\ 
        
        \cmidrule(lr){1-11}
        
        \multirow{6}{*}{80\%} 
        & MRM & 37.7 & 96.3 & 80.3 & 72.0 & 61.2 & 42.7 & 35.4 & 60.8 & 0.21$\times$ \\ 
        & SparseIRM & 37.8 & 97.2 & 82.9 & 71.6 & 62.4 & 43.8 & 36.2 & 61.7 & 0.21$\times$ \\
        & EVIL & 38.5 & 98.1 & 84.7 & 74.1 & 64.3 & 46.4 & 40.1 & 63.7 & 0.21$\times$ \\[-1.5ex]
        & & \tiny($\pm1.3$) & \tiny($\pm0.2$) & \tiny($\pm0.0$) & \tiny($\pm0.2$) & \tiny($\pm0.2$) & \tiny($\pm0.1$) & \tiny($\pm0.0$) & & \\
        & EVIL-SAM & 38.9 & 98.3 & \textbf{87.8} & 76.8 & 65.7 & 47.6 & 42.7 & 65.4 & 0.57$\times$ \\[-1.5ex]
        & & \tiny($\pm1.2$) & \tiny($\pm0.3$) & \tiny($\pm0.2$) & \tiny($\pm0.2$) & \tiny($\pm0.2$) & \tiny($\pm0.1$) & \tiny($\pm0.3$) & & \\
        \bottomrule
    \end{tabular}
\end{table*}

\subsection{Improving Invariant Learning Using EVIL}
\label{EVIL:improve_invariant_learning}
In this section, we deploy our EVIL framework to some well-known invariant learning methods and compare them with some other typical baseline methods. To conduct a fair comparison, we only considered end-to-end training on one single model, so some other methods that conduct model ensembling or averaging~\cite{cha2021swad, rame2022diverse, izmailov2018averaging, bai2021me} are not considered. Moreover, we use floating point operations per second (FLOPs) as a criterion to denote the computational efficiency by denoting the FLOPs of ERM as 1$\times$ ($7.8e10$). Practically, we set the sparsity ratio $R=60\%$, hyper-parameter $\alpha=0.2$, and $\Delta T=300$ to implement EVIL. The results are shown in Table~\ref{EVILtab:deployment}. We can see that our EVIL can effectively improve the performance of all chosen backbone methods. Particularly, on ERM, DRO, and SAM, EVIL can increase their test accuracies for $2.4\%$, $1.9\%$, and $1.9\%$, respectively. Moreover, EVIL-SAM achieves the best OOD generalization performance among all compared methods. Especially on TerraIncognita dataset, EVIL-SAM can improve the original performance of SAM for $7.2\%$, which indicates the effectiveness of EVIL in improving the performance of invariant learning. Moreover, compared to the FLOPs of all baseline methods, our EVIL shows much less computational burden, which manifests the great efficiency of our method.

\subsection{Comparing EVIL to Sparse Invariant Learning}
\label{EVIL:comparing_evil_to_sparse}
Furthermore, to show that our method finds a more robust subnetwork for OOD generalization, we compare EVIL with two existing sparse invariant learning methods. Specifically, to validate the effectiveness of our method under different levels of sparsity, we vary to sparsity ratio to $20\%$, $40\%$, $60\%$, and $80\%$. The experimental results are shown in Table~\ref{EVILtab:sparse_comparison}. Generally, we can see that our EVIL and EVIL-SAM surpass both MRM and SparseIRM in almost all scenarios. Moreover, among all sparsity levels, both EVIL and the other two methods achieve the best results under sparsity $60\%$. Specifically, EVIL-SAM shows the best performance under sparsity $60\%$ on almost all datasets, and it surpasses the second-best opponent for $2.8\%$ on average accuracy. Besides, EVIL implementation with just ERM can also improve the second-best methods for $1.5\%$ on the averaged results. As for the computational efficiency, our EVIL is comparable to other sparse training methods, except EVIL-SAM, which requires an extra backward pass to compute the parameter perturbation. Therefore, by exploring the variant parameters, EVIL successfully achieves superior OOD generalization performance with comparable efficiency to the sparse invariant learning methods.

\subsection{Performance on Additional Invariant Learning Methods}
\begin{table}[t]
	\caption{Results on additional invariant learning methods.}
	\setlength{\tabcolsep}{6mm}
	\begin{tabular}{lcccc}
		\toprule[1pt]
		Method & MMD & SagNet & Mixstyle & ARM \\ 
		\midrule[0.6pt]
		w/o EVIL & 84.7 & 86.3 & 85.2 & 85.1 \\
		with EVIL & \textbf{85.3} & \textbf{87.1} & \textbf{86.5} & \textbf{86.6} \\
		& ($\pm0.1$) & ($\pm0.2$) & ($\pm0.2$) & ($\pm0.2$) \\
		\toprule[1pt]
	\end{tabular}
	\label{EVILtab:methods}
\end{table}
We have discussed several invariant learning methods in the main paper, here we conduct extra experiments on PACS dataset using additional invariant learning methods to show how EVIL affects their OOD generalization results. Moreover, we conduct experiments using different network architectures to show the effect of EVIL on various learning models.

Concretely, as we have provided results of IRM, REx, DRO, CORAL, and SAM in the main paper, here we implement EVIL using backbone methods including MMD, SagNet, Mixstyle, and ARM. The results on PACS dataset are shown in Table~\ref{EVILtab:methods}. We can see that our method can still improve the OOD generalization performance which is consistent with the observation in the main paper. Therefore, the proposed EVIL framework is generally effective among various invariant learning methods, which shows great 
deployment practicality.

\begin{table}[t]
	\centering
	\caption{Results on various model architectures. ResNet50 is pre-trained on ImageNet, and other models are trained from scratch.}
	\begin{tabular}{lcccccc}
		\toprule[1pt]
		{\small Arch.} & {\small ResNet50} & {\small WRN-20} & {\small WRN-32} & {\small WRN-44} & {\small  WRN-56} & {\small WRN-110} \\ 
		\midrule[0.6pt]
		ERM & 84.2 & 35.6 & 39.2 & 41.0 & 44.6 & 48.9 \\
		EVIL & \textbf{86.0} & \textbf{37.3} & \textbf{42.5} & \textbf{43.7} & \textbf{47.2} & \textbf{51.4} \\ \addlinespace[-0.1cm]
		& ($\pm0.1$) & ($\pm0.2$) & ($\pm0.3$) & ($\pm0.3$) &($\pm0.2$) & ($\pm0.3$) \\
		\toprule[1pt]
	\end{tabular}
	\label{EVILtab:arch}
\end{table}

\subsection{Performance on Additional ResNet Architectures}
Moreover, to evaluate the effectiveness of EVIL on different backbone models, we implement the Wide ResNet (WRN)~\cite{zagoruyko2016wide} with varied depths (20, 32, 44, 56, and 110) and train each model from scratch for 500,000 steps to ensure convergence. We also show the result of using ResNet50 pre-trained on ImageNet (Note that due to the pre-training, the performance on ResNet50 would be much better than training from scratch). The comparison between ERM and EVIL is shown in Table~\ref{EVILtab:arch}. Again, we can observe the superiority of EVIL over the baseline method ERM on all investigated architectures. Therefore, we can conclude that the performance improvement brought by EVIL is model-agnostic.

\subsection{Optimizing EVIL Using SAM}
\label{EVIL:ood_sam}
In this section, we first briefly describe the realization of EVIL optimized by SAM for OOD generalization (EVIL-SAM). Then, despite of orthogonality of flatness and OOD generalization as found before~\cite{cha2021swad, rame2022diverse}, we discuss some properties of SAM and demonstrate why combining EVIL and SAM can achieve great performance.

\textbf{Realization of EVIL-SAM.}
Generally, our EVIL can be optimized using SAM by minimizing the following objectives:
\begin{equation}
	\min_{\theta_{inv}}\max_{\|\bm{\epsilon}\circ\mathbf{m}\|_2\le \rho} \mathcal{L}(\theta_{inv}+\bm{\epsilon}\circ\mathbf{m}; x, y).
	\label{eq:sam_evil}
\end{equation}
Specifically, SAM seeks to compute an optimal parameter perturbation $\bm{\epsilon}^*=\arg\max_{\bm{\epsilon}}\mathcal{L}(\theta+\bm{\epsilon}; x, y)$ within $\rho$-radius neighbor that can maximally increase the loss value $\mathcal{L}$. By applying $\bm{\epsilon}^*$, the loss change $\mathcal{L}(\theta+\bm{\epsilon}^*; x, y)-\mathcal{L}(\theta; x, y)$ is denoted as \textit{sharpness} which indicates the flatness of the learned loss function. Intuitively, a flatter loss function often shows better generalization properties, as a slight shift imposed in the input space would not significantly change the loss value. Therefore, SAM has achieved promising in-distribution (ID) generalization performance~\cite{andriushchenko2022towards, kim2022fisher, kwon2021asam, liu2022towards, zhao2022penalizing}. To adopt SAM into EVIL, we just need to apply our mask $\mathbf{m}$ to the parameter perturbation $\bm{\epsilon}$ before computing the optimal $\bm{\epsilon}^*$. This process not only leaves out spurious information but also reduces the computational burden of SAM. As a result, SAM-EVIL can achieve low sharpness for invariant learning.

\begin{table}[t]
	\caption{Comparison of SAM~\cite{foret2020sharpness} and ERM under both ID and OOD situations on DomainBed.}
	\centering
	\begin{tabular}{clccccc|c}
		\toprule[1pt]
		& & PACS  & VLCS  & OfficeHome & TerraInc & DomainNet &  Average \\
		\midrule
		\multirow{2}{*}{{\scriptsize ID}} & {\scriptsize ERM}        & 96.6 & 84.7 & 78.9 & 91.3 & 81.4 & 86.5                      \\
		& {\scriptsize SAM} & 97.1 & 86.8 & 82.0 & 93.1 & 85.2 &  88.8                      \\
		\midrule
		\multirow{2}{*}{{\scriptsize OOD}} & {\scriptsize ERM}      & 85.5 & 77.5 & 66.5  & 46.1 & 40.9 &  63.3                      \\
		& {\scriptsize SAM} & 85.8 & 79.4 & 69.6 & 43.3 &  44.3 &  64.5                      \\
		\bottomrule[1pt]
	\end{tabular}
	\label{EVILtab:sam}
\end{table}

\textbf{Discussion.}
Although SAM has achieved promising ID results, its OOD performance is quite limited~\cite{cha2021swad, rame2022diverse} which is still unexplained. As shown in Table~\ref{EVILtab:sam}, in the ID scenario, SAM shows great effectiveness compared to ERM, but it merely achieves comparable results to ERM in the OOD setting, even worse in some scenarios. In our perspective, the limitation of SAM is caused by erroneously perturbing the variant parameters which encourages fitting to spurious features. Specifically, in OOD problems, the invariant features and spurious ones would activate $\theta_{inv}$ and $\theta_{var}$, respectively. Enforcing robustness (\textit{i.e.}, low sharpness) against perturbation on $\theta_{inv}$ can enhance extracting invariant features. However, by perturbing $\theta_{var}$, low sharpness $\mathcal{L}(\theta+\bm{\epsilon}^*; x, y)-\mathcal{L}(\theta; x, y)$ denotes encouraging the spurious features to bond with the label information. Therefore, SAM cannot extract invariant features as it is sensitive to spurious ones, thus damaging the OOD generalization results. Fortunately, our EVIL can perfectly solve this problem by filtering out the variant parameters which is strongly related to distribution noise. Thus SAM can be further leveraged to enhance the robustness of extracting invariant features. Its effectiveness and efficiency are demonstrated in Section~\ref{EVIL:experiments}.

\begin{table}
	\caption{Comparison of EVIL-SAM with other baseline methods on five datasets from DomainNet.}
	\begin{tabular}{lccccc|c}
		\toprule[1pt]
		{Method} & PACS  & VLCS  & OfficeHome & TerraInc & DomainNet &  Avg. \\
		\midrule
		SagNet-SAM  &  86.4  &  78.5  &  69.2  &  49.3  &  40.0  &  64.6 \\
		CORAL-SAM  &  86.6  &  79.0  &  69.3  &  47.9  &  42.1 & 65.0 \\
		MRM-SAM  &  83.9  &  77.1  &  67.0  &  47.4  &  40.6  & 63.2 \\
		SparseIRM-SAM  &  85.2  &  77.4  &  65.6  &  48.5  &  43.1 & 63.9 \\
		EVIL-SAM  &  \textbf{87.8}  &  \textbf{80.1}  & \textbf{ 70.3}  &  \textbf{50.5}  &  \textbf{45.0} & \textbf{66.7}\\
		\bottomrule[1pt]
	\end{tabular}
	\label{EVILtab:justify_sam}
\end{table}

\textbf{Compare with Other Methods using SAM optimization.}
To further validate our realization that combining EVIL with SAM indeed shows a positive effect, we compare EVIL-SAM to other algorithms as shown in Table~\ref{EVILtab:justify_sam}. We observe that EVIL-SAM achieves the best result among both dense and sparse methods with a significant margin, therefore we can justify our improvement on SAM as more effective than other methods.

\begin{table*}[t!]
	\centering
	\caption{Performance on ImageNet, iWildCam, and FMoW using CLIP ViT-B/16 as backbone.}
		\begin{tabular}{ccccccc}
			\toprule
			& \multicolumn{2}{c}{ImageNet} & \multicolumn{2}{c} {iWildCam}   & \multicolumn{2}{c}{FMoW}                  \\ \cmidrule(l){2-7} 
			\multirow{-2}{*}{Methods} & ID                  & \multicolumn{1}{c|}{OOD}                                         & ID                  & \multicolumn{1}{c|}{OOD}                 & ID                  & OOD                                          \\ \midrule
			Zeroshot          & 68.3 $\pm$0.0            & \multicolumn{1}{c|}{58.7 $\pm$0.0}           & 8.7 $\pm$0.0             & \multicolumn{1}{c|}{11.0 $\pm$0.0}    & 20.4 $\pm$0.0            & 18.7 $\pm$0.0           \\
			Finetuning       & 82.5 $\pm$0.1          & \multicolumn{1}{c|}{61.3 $\pm$0.1}    & 48.1 $\pm$0.5          & \multicolumn{1}{c|}{35.0 $\pm$0.5}                                  & 68.5 $\pm$0.1          & 39.2 $\pm$0.7          \\
			EVIL     & 81.8 $\pm$0.2          & \multicolumn{1}{c|}{\textbf{62.5} $\pm$0.6}     & 47.6 $\pm$0.8         & \multicolumn{1}{c|}{\textbf{37.4}$\pm$1.2}         & 68.2 $\pm$0.6          & \textbf{41.2} $\pm$1.3          \\ \midrule
		\end{tabular}
	\label{EVILtab:clip}
\end{table*}

\subsection{Performance on Large-Scale Architecture and Datasets}
In this section, we adopt pretrained CLIP ViT-B/16~\cite{radford2021learning} and conduct finetuning on training datasets from iWildCam, FMoW, and ImageNet, and further test the OOD generalization performance on the split OOD datasets. To extend our sparse training strategy into the CLIP model, we employ a linear layer on top of the ViT backbone and conduct the same pruning strategy by leveraging both class information and domain information. For all datasets, we set the finetuning epoch as 20 and keep the rest of the training parameters the same as described before. The results are shown in Table~\ref{EVILtab:clip}, we can see that although EVIL shows a slight performance drop on ID datasets, which is reasonable since we use fewer parameters than full finetuning, our method achieves the best OOD performance on all three datasets. Specifically, there are $1.2\%$, $2.4\%$, and $2.0\%$ performance gains on ImageNet, iWildCam, and FMoW datasets, respectively. Therefore, the effectiveness and superiority of the EVIL can be successfully extended to large-scale architectures and datasets.

\begin{figure*}[t]
	\centering
	\includegraphics[width=\linewidth]{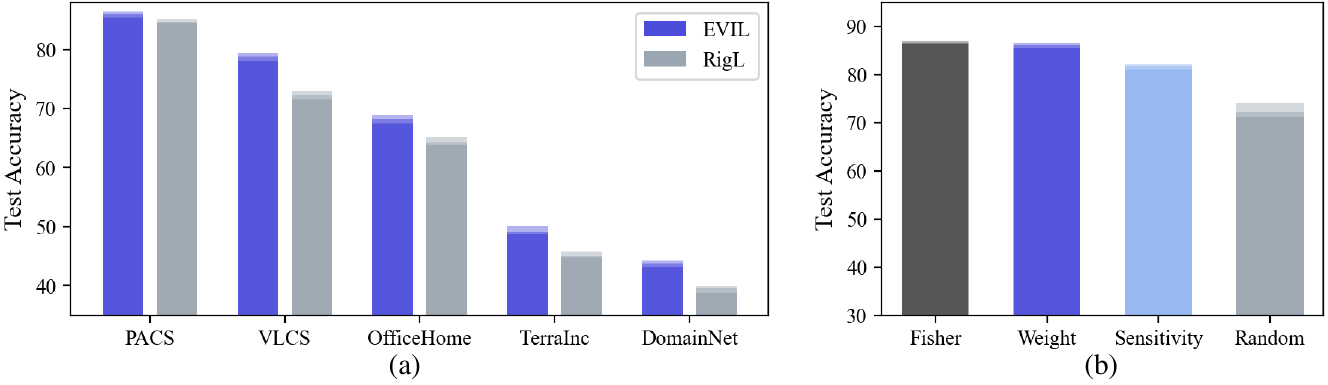}
	\caption{(a) Comparison of EVIL and RigL which leverages the label information to explore the variant parameters. (b) Comparison of different mask initialization strategies.}
	\label{EVILfig:ablation}
\end{figure*}

\subsection{Analytical Studies}
\label{EVIL:performance_analysis}
In this section, we conduct extensive empirical analyses to exploit why EVIL can achieve effective results. First, we conduct ablation studies to show the effect of leveraging distribution knowledge and the influence of choosing different mask initialization strategies. Then, we conduct parameter sensitivity analysis on the hyperparameter $\alpha$ and $\Delta T$. Further, we show a comparison of EVIL-SAM and SAM by visualizing their sharpness during training. Finally, we show the Hessian spectrum to explain why EVIL achieves effective generalization.

\textbf{Ablation Study.} To validate the effectiveness of exploring variant parameters using distribution knowledge, we change the optimization target $\mathcal{L}_{var}(g(f_{\theta_{var}}(\mathbf{x})), d)$ in Equation~\eqref{EVILeq:var} to $\mathcal{L}_{ce}(h(f_{\theta_{var}}(\mathbf{x})), y)$ which leverages the label information instead. As a result, the changed variant is actually Rigging the Lottery (RigL)~\cite{evci2020rigging}, which is an effective sparse training method. By comparing EVIL and RigL on DomainBed as shown in Figure~\ref{EVILfig:ablation} (a), we can see that EVIL surpasses RigL in all scenarios. Therefore, we can conclude that exploring variant parameters by using the distribution information is essential for OOD generalization. Moreover, to show the influence of choosing different mask initialization strategies as mentioned in Section~\ref{EVIL:evil}, we compare the weight value (as done in our method) with Fisher information~\cite{sung2021training}, connection sensitivity~\cite{lee2018snip}, and random initialization~\cite{frankle2019lottery} and show the result in Figure~\ref{EVILfig:ablation} (b). As we can see, the Fisher information and weight value are two better strategies than connection sensitivity and random initialization, which supports our choice of using the weight value.

\begin{figure}[t]
	\begin{minipage}{0.54\linewidth}
		\includegraphics[width=\textwidth]{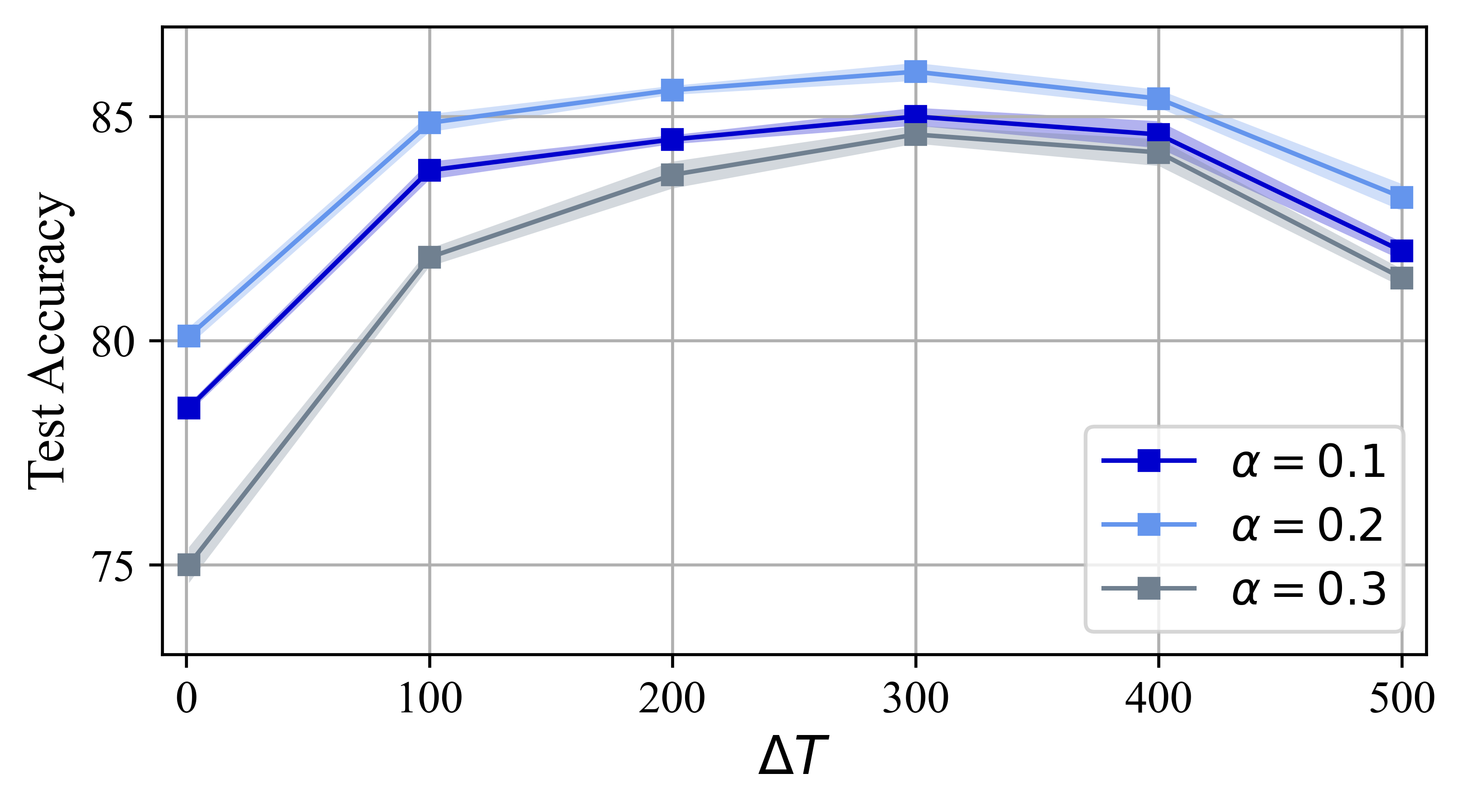}
		\caption{Parameter sensitivity analysis on $\alpha$ and $\Delta T$.}
		\label{EVILfig:parameter}
	\end{minipage}
	\begin{minipage}{0.44\linewidth}
		\includegraphics[width=\textwidth]{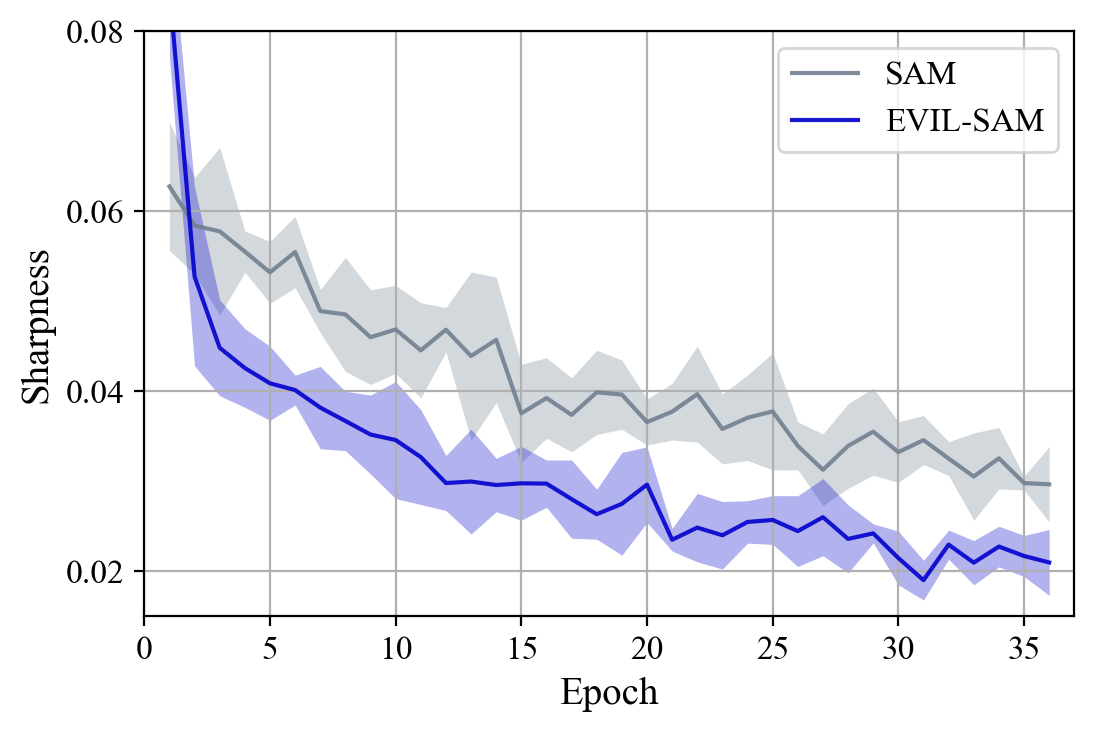}
		\caption{Sharpness comparison.}
		\label{EVILfig:sharpness}
	\end{minipage}
\end{figure}

\begin{wrapfigure}{r}{0.5\textwidth}
	\centering
	\includegraphics[width=\linewidth]{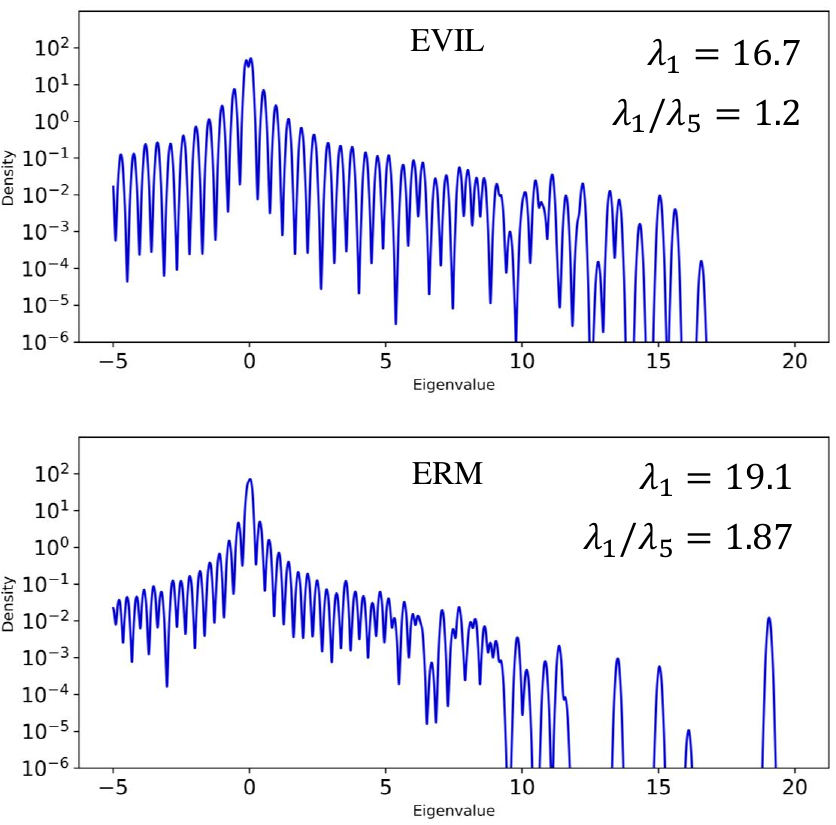}
	\caption{Hessian Spectrum of EVIL and ERM.}
	\label{EVILfig:hessian}
\end{wrapfigure}

\textbf{Parameter Sensitivity Analysis.}
To analyze the different choices of $\alpha$ and $\Delta T$, we set $\alpha$ to 0.1, 0.2, and 0.3 and vary $\Delta T$ to 1, 100, 200, 300, 400, and 500. As shown in Figure~\ref{EVILfig:parameter}, choosing $\alpha$ as 0.2, and $\Delta T$  as 300 is the best. Moreover, a lower $\alpha$ would hinder the dynamic update of the mask, and a higher $\alpha$ would cause incorporation of noises, thus both choices lead to a performance drop. On the other hand, $\Delta T$ controls the updating frequency. Both too small $\Delta T$ and too large $\Delta T$ would correspondingly cause insufficient update and redundant update, further leading to degradation.

\textbf{Sharpness Analysis.} To show how our EVIL affects SAM during OOD generalization, we visualize the sharpness obtained from the training process in Figure~\ref{EVILfig:sharpness}. As a result, our EVIL-SAM can produce smaller sharpness during training than SAM, which indicates that EVIL-SAM is more robust than SAM in OOD generalization problems.

\textbf{Hessian Spectrum.}
To analyze whether an algorithm can converge to a flat minima, the Hessian spectrum is commonly used as a criterion~\cite{ghorbani2019investigation}. Specifically, we follow Foret et al.~\cite{foret2020sharpness} to use the ratio of dominant eigenvalue to fifth largest eigenvalue, i.e., $\lambda_1 / \lambda_5$ as the criterion for comparing EVIL and ERM. Generally, a smaller $\lambda_1 / \lambda_5$ often means a flatter minima is found. Thus, we follow Ghorbani et al.~\cite{ghorbani2019investigation} by using the Lanczos algorithm to approximate the Hessian spectrum of ERM and EVIL in Figure~\ref{EVILfig:hessian}. As we can see, the $\lambda_1 / \lambda_5$ of EVIL is much smaller than that of ERM, which confirms that our method can converge to a flatter minima than ERM. Moreover, as the dominant eigenvalue $\lambda_1$ is also an important measurement, we can see that EVIL produces a smaller $\lambda_1$ than ERM as well, which again supports the effectiveness of EVIL. Therefore, it is reasonable that EVIL can achieve great generalization results.

\begin{figure}[t]
     \centering
     \begin{subfigure}[b]{0.48\textwidth}
         \centering
         \includegraphics[width=\textwidth]{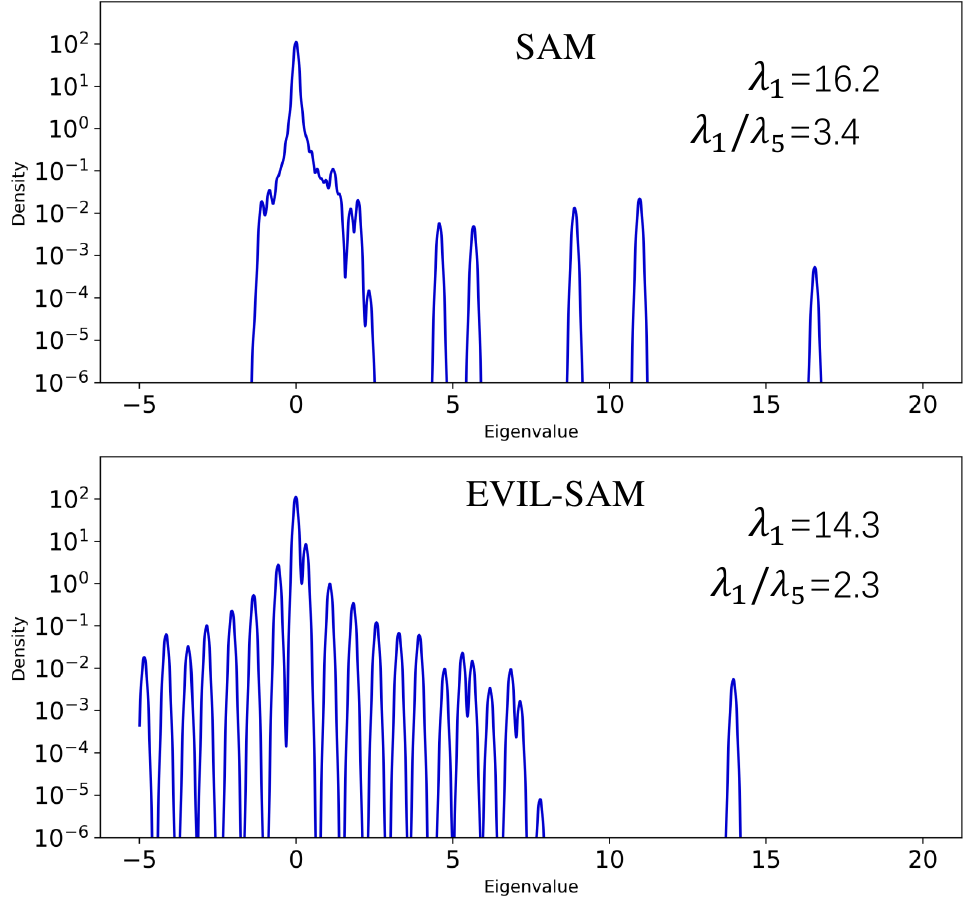}
         \caption{SAM vs. EVIL-SAM}
         \label{EVILfig:hessian_sam}
     \end{subfigure}
     \hfill 
     \begin{subfigure}[b]{0.48\textwidth}
         \centering
         \includegraphics[width=\textwidth]{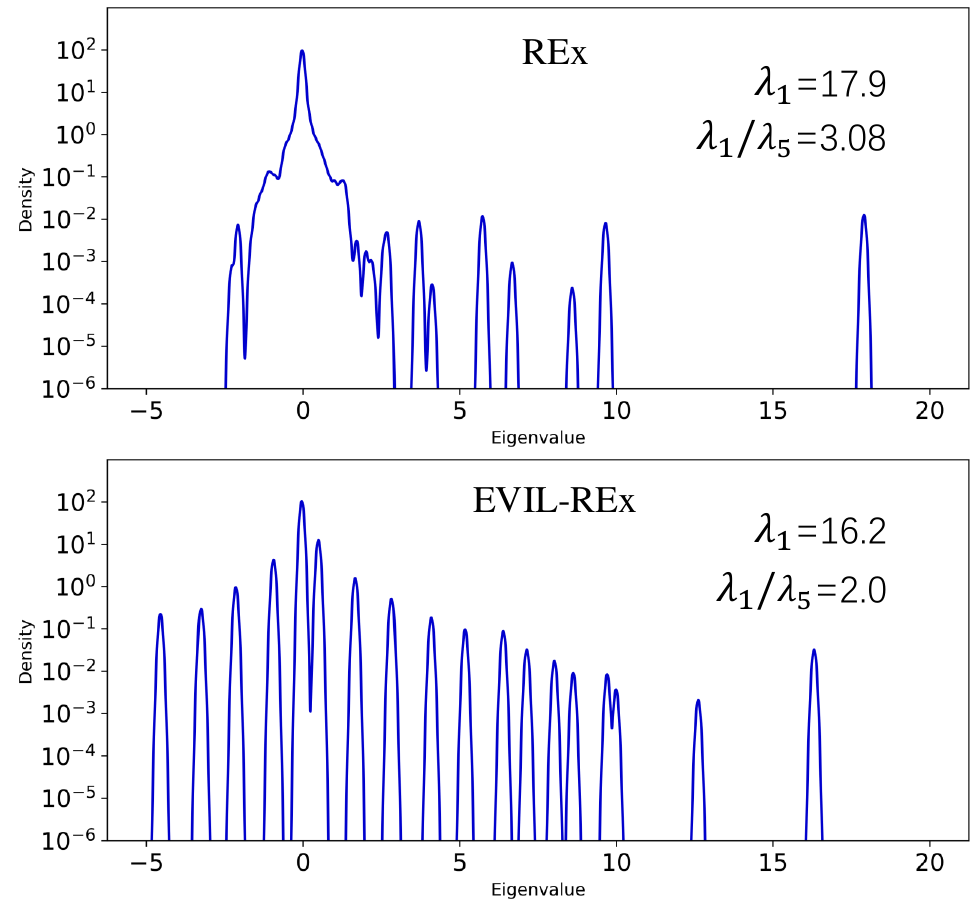}
         \caption{REx vs. EVIL-REx}
         \label{EVILfig:hessian_rex}
     \end{subfigure}
     
     \caption{Hessian spectrum comparison: EVIL realization significantly improves the sharpness across different methods.}
     \label{EVILfig:combined_hessian}
\end{figure}

\textbf{Additional Hessian Spectrum on SAM and REx.}
Since the proposed method shows effective generalization performance, as we have demonstrated in Section~\ref{EVIL:performance_analysis}, we further validate that the proposed EVIL framework can still help produce improved Hessian spectrum when compared to other methods such as SAM and REx. As shown in Figs.~\ref{EVILfig:hessian_sam} and ~\ref{EVILfig:hessian_rex}, We observe the same phenomenon as in the main paper: when combined with EVIL, the largest eigenvalue of both SAM and REx is smaller than its original ones, and the Hessian spectrum are more compact when using our EVIL framework. Therefore, we can again conclude that EVIL indeed helps produce flat minima.

\section{Conclusion}
In this Chapter, we aim to address the problem that existing sparse invariant learning methods fail to fully capture invariant information in OOD generalization problems, owing to the misleading influence of distribution shifts. Therefore, we propose EVIL by leveraging the distribution knowledge to explore the variant parameters. By finding the variant parameters that are highly sensitive to distribution shift, we can identify a robust subnetwork that effectively extracts invariant features. Moreover, we propose to improve our identification dynamically during network training. As a result, our EVIL framework can effectively and efficiently improve the OOD generalization performance of many invariant learning methods, meanwhile surpassing all compared sparse invariant learning methods. Exhaustive analyses are conducted to comprehensively validate the performance of EVIL.

\chapter{Machine Vision Therapy}
\label{cha:MVT}
Although pre-trained models such as Contrastive Language-Image Pre-Training (CLIP) show impressive generalization results, their robustness is still limited under Out-of-Distribution (OOD) scenarios. Instead of undesirably leveraging human annotation as commonly done, it is possible to leverage the visual understanding power of Multi-modal Large Language Models (MLLMs). However, MLLMs struggle with vision problems due to task incompatibility, thus hindering their effectiveness. In this Chapter, we propose to effectively leverage MLLMs via Machine Vision Therapy, which aims to rectify erroneous predictions of specific vision models. By supervising vision models using MLLM predictions, visual robustness can be boosted in a nearly unsupervised manner. Moreover, we propose a Denoising In-Context Learning (DICL) strategy to solve the incompatibility issue. Concretely, by examining the noise probability of each example through a transition matrix, we construct an instruction containing a correct exemplar and a probable erroneous one, which enables MLLMs to detect and rectify the incorrect predictions of vision models. Under mild assumptions, we theoretically show that our DICL method is guaranteed to find the ground truth. Through extensive experiments on various OOD datasets, our method demonstrates powerful capabilities for enhancing visual robustness under many OOD scenarios.

\section{Introduction}
\label{MVT:introduction}
Pre-trained vision models such as Vision Transformers (ViT)~\cite{dosovitskiy2020image, liu2021swin, wang2021pyramid} with Contrastive Language-Image Pretraining (CLIP)~\cite{chen2023clip2scene, radford2021learning, li2021supervision, li2022blip, wang2024clips, zheng2023enhancing} have been widely used thanks to their strong generalization performance meanwhile effectively avoiding training vision models from scratch. But when deployed to Out-of-Distribution (OOD) scenarios~\cite{dong2023benchmarking, kong2023robo3d, hendrycks2016baseline, huang2024winning, hong2024improving, li2024dynamic, peng2023sam, wang2019learning, zhu2023understanding}, their recognition performance could be seriously degraded~\cite{shu2023clipood}. Downstream fine-tuning has been a common practice to regain the generalizability~\cite{goyal2023finetune, wortsman2022robust, huang2025towards}, but it requires additional label acquisition through human labor, which is undesirable for large-scale applications. 

\begin{figure}
\includegraphics[width=\linewidth]{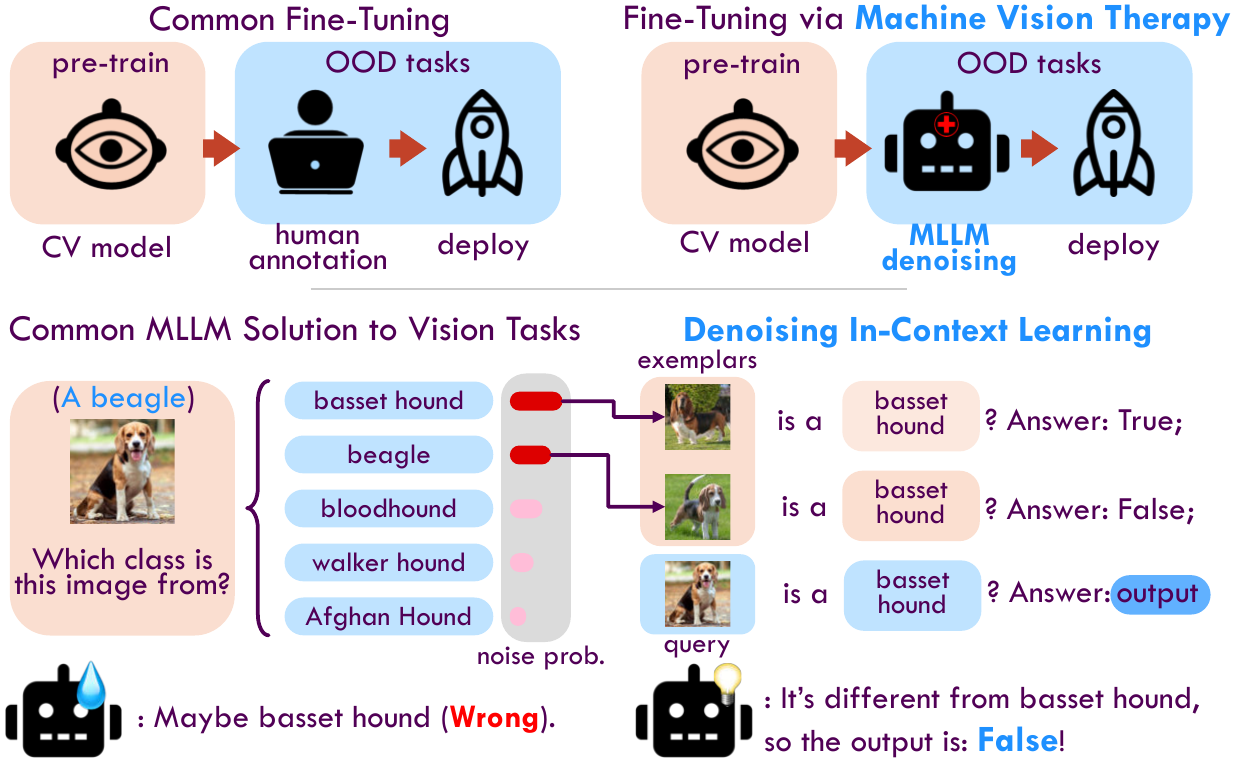}
\caption{Illustration of our methodology: Upper row: Comparison between common fine-tuning process and fine-tuning via Machine Vision Therapy. Our method potentially eliminates the necessity for human-annotation by leveraging the knowledge from MLLMs. Lower row: Comparison between previous MLLM solution to vision tasks and Denoising In-Context Learning strategy. Instead of considering all classes, our method make predictions by presenting a pair of positive and negative exemplars.}
\label{MVTfig:motivation}
\end{figure}

Fortunately, the thriving Multi-modal Large Language Models (MLLMs)~\cite{alayrac2022flamingo, awadalla2023openflamingo, chen2024towards, gong2023multimodal, li2023blip, li2023otter, liu2023llava, ye2023mplug, zhu2023minigpt}, which take advantage of the few-shot learning ability of Large Language Models (LLM)~\cite{brown2020language, chung2022scaling, floridi2020gpt, openai2023gpt4, scao2022bloom, touvron2023llama, touvron2023llama2, zheng2023judging, yang2025exploring}, have manifested powerful capabilities on understanding visual information with language interpretations, and excelled at recognizing novel objects in multimodal tasks such as image captioning, visual question answering, visual reasoning, etc. Considering the vulnerability of vision models under OOD situations, here we hope to refine vision models by leveraging the knowledge of MLLMs, as shown in the upper row of Figure~\ref{MVTfig:motivation}. However, due to the difficulty of aligning the text generation process with visual recognition tasks\footnote{Here, we mainly focus on classification task.}~\cite{alayrac2022flamingo, wang2022git}, MLLMs struggle with generating correct answers that match the ground-truth class names, thus underperforming the current dominant contrastive paradigms, even when employing them as own vision encoders~\cite{alayrac2022flamingo, awadalla2023openflamingo, huang2023language, wang2022git, zhai2023investigating}.

Focusing on enhancing the robustness of vision models, in this Chapter, we propose to effectively leverage MLLMs to conduct \textbf{Machine Vision Therapy (MVT)} which aims to diagnose and rectify the error predictions through a novel Denoising In-Context Learning (DICL) strategy. Then, we utilize the rectified supervision to guide the fine-tuning process in downstream OOD problems. Specifically, rather than giving a set of options to ask MLLMs for the exact answer~\cite{alayrac2022flamingo, huang2023language, zhai2023investigating}, we show that it is sufficient to query for the ground truth by using only two exemplars, i.e., (1) a correct one that demonstrates the exact match between a query class name with its image example and (2) an erroneous one that combines the same query class with an image from the most confusing category for the vision model. Since the erroneous predictions are essentially label noise, hence we draw inspiration from learning with noisy labels~\cite{han2018co, liu2015classification, lin2022we, lin2023cs, natarajan2013learning, wu2024mitigating, wu2023making, xia2020part, xia2020robust, yao2020dual, yao2021instance, yao2023better, yuan2024early}. Particularly, we can find the erroneous categories by estimating a transition matrix that captures the probability of one class being mistaken as another. By feeding the two exemplars, MLLMs can be instructed to leverage their few-shot learning power to distinguish the semantically similar images that are easily misclassified by vision models, as shown in the lower row of Figure~\ref{MVTfig:motivation}. To process such instructions, we leverage the multi-modal in-context learning ability of several existing MLLMs~\cite{chen2023lightweight, li2023mimicit, yasunaga2023retrieval, zhao2023mmicl} to realize our methodology. After the error predictions are diagnosed and rectified, vision models can be further fine-tuned to enhance their OOD robustness on downstream data distribution. Through a comprehensive empirical study on many challenging datasets and their OOD variants, such as ImageNet~\cite{deng2009imagenet}, WILDS~\cite{koh2021wilds}, and DomainBed~\cite{gulrajani2021search}, we carefully validate the effectiveness of MVT and demonstrate its superiority under various OOD scenarios on many well-known vision models.

To sum up, our contributions are threefold:
\begin{itemize}
	\item We design a novel Machine Vision Therapy paradigm to enhance computer vision models by effectively leveraging the knowledge of MLLMs without needing additional label information.
	
	\item We propose a Denoising In-Context Learning strategy to successfully align MLLMs with vision tasks.
	
	\item Through comprehensive quantitative and qualitative studies on many well-known datasets, we demonstrate that the proposed method can enhance:
    \textcolor[rgb]{0.05, 0.09, 0.58}{(1)} generalization on both ID and OOD data,
    \textcolor[rgb]{0.05, 0.09, 0.58}{(2)} robustness against domain shift,
    \textcolor[rgb]{0.05, 0.09, 0.58}{(3)} robustness against common corruptions,
    \textcolor[rgb]{0.05, 0.09, 0.58}{(4)} performance on recognizing fine-Grained attributes,
    \textcolor[rgb]{0.05, 0.09, 0.58}{(5)} robustness against spurious correlations,
    \textcolor[rgb]{0.05, 0.09, 0.58}{(6)} detection on prediction errors and OOD data.
\end{itemize}

\section{Related Work}
\label{MVT:related_work}

In this section, we provide a brief discussion of the OOD generalization problem and multimodal large-language models.
\subsection{OOD Generalization}
OOD data refers to those with different distributions from training data. OOD generalization aims at improving the performance of deep models to unseen test environments. Researchers attempted to tackle the problem from different perspectives, such as data augmentation, OOD detection, invariant causal mechanisms~\cite{huang2023harnessing, huang2023winning, zhou2021towards, zhou2022improving}, and so on. Data augmentation is effective in improving model generalization. Typical methods involve Cutout~\cite{devries2017improved}, which randomly occludes parts of an input image; CutMix~\cite{yun2019cutmix}, which replaces a part of the target image with a different image; Mixup~\cite{zhang2017mixup}, which produces a convex combination of two images; DeepAugment~\cite{hendrycks2021many}, which passes a clean image through an image-to-image network and introduces several perturbations during the forward pass. Some methods conduct OOD detection to separate OOD data. Typical methods include softmax confidence score~\cite{hendrycks2016baseline, huang2023robust}, which is a baseline for OOD detection; Outlier Exposure (OE)~\cite{hendrycks2018deep}, which uses unlabeled data as auxiliary OOD training data. Energy scores are shown to be better for distinguishing OOD samples from IID ones~\cite{liu2020energy}. Some work resort to causality to study the OOD generalization problem. Typical methods include MatchDG~\cite{mahajan2021domain}, which proposes matching-based algorithms when base objects are observed and approximate the objective when objects are not observed; INVRAT~\cite{chang2020invariant}, which leveraged some conditional independence relationships induced by the common causal mechanism assumption.

\subsection{Multimodal Large Language Models}
The field of vision-language models has witnessed significant advancements in recent years, driven by the growing synergy between computer vision and natural language processing. Notably, this synergy has led to the exceptional zero-shot performance~\cite{hou2024visual} of CLIP~\cite{radford2021learning}, a model that employs a two-tower contrastive pretraining approach to align image and text information. In the rapidly evolving landscape of LLMs, such as GPTs~\cite{brown2020language}, LLaMA~\cite{touvron2023llama}, and Vicuna~\cite{vicuna2023}, it has become evident that LLMs possess the capacity to process information from diverse domains. BLIP-2~\cite{li2023blip}, for instance, serves as a foundational model, aligning visual features and text features using a Querying Transformer (Q-former) and utilizing OPT~\cite{zhang2022opt} and FLAN~\cite{chung2022scaling} as language models. Building upon BLIP-2, Instruct-BLIP~\cite{dai2023instructblip} has enhanced instruction-following capabilities. To further bolster the instruction-following proficiency of multi-modal models, LLaVA~\cite{liu2023llava} and Mini-GPT4~\cite{zhu2023minigpt} have introduced meticulously constructed instruction sets, which have found widespread application in various multi-modal models. mPLUG-Owl~\cite{ye2023mplug} introduces a two-stage learning paradigm, first fine-tuning the visual encoder and then refining the language model with LoRA~\cite{hu2021lora}. This approach effectively fuses image and text features. Some models consider additional modalities, such as ImageBind~\cite{girdhar2023imagebind}, which simultaneously incorporates data from six modalities without the need for explicit supervision, and PandaGPT~\cite{su2023pandagpt} which enhances its instruction-following capabilities. Several multi-modal models prioritize the in-context learning abilities of LLMs. Flamingo~\cite{alayrac2022flamingo}, in one of the pioneering efforts, integrates a gated cross-attention module to align with the spaces of images and text. Otter~\cite{li2023otter} refines OpenFlamingo~\cite{awadalla2023openflamingo}, an open-source version of Flamingo, improves instruction-following abilities. Multi-Modal In-Context Learning (MMICL)~\cite{zhao2023mmicl} is a comprehensive vision-language model that incorporates Instruct-BLIP, enabling the analysis and comprehension of multiple images, as well as the execution of instructions. MLLMs possess the remarkable capacity to capture intricate details and engage in reasoning when presented with an image. Nevertheless, it remains uncertain about how to enhance visual perception by harnessing the knowledge embedded within LLMs.

\section{Methodology}
\label{MVT:method}

\begin{figure*}[t]
\centering
\includegraphics[width=0.93\linewidth]{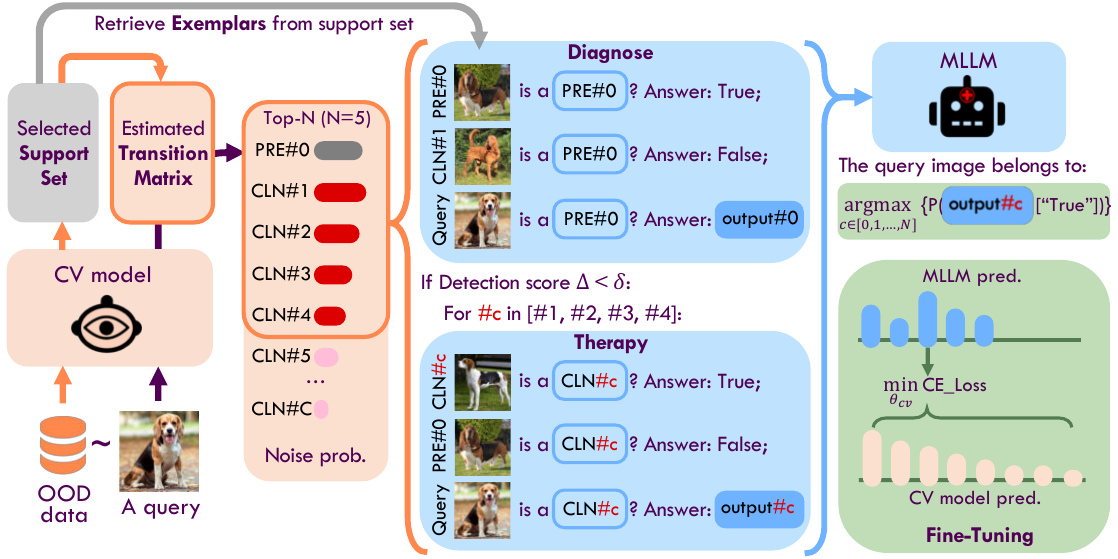}
\caption{Workflow of our Machine Vision Therapy: The orange part demonstrates the Transition Matrix Estimation, the blue part indicates the Denoising In-Context Learning process, and the green part illustrates the Fine-Tuning of vision models.}
\label{MVTfig:framework}
\end{figure*}

In this section, we carefully demonstrate the Machine Vision Therapy process which mainly contains three components, namely Transition Matrix Estimation, Denoising In-Context Learning, and Fine-Tuning of vision models. Next, we demonstrate problem setting and framework overview.

\subsection{Problem Formulation and Overview}
\label{MVT:formulation_overview}
Generalizing to Out-of-Distribution tasks has been a challenging topic in computer vision problems, where we normally have a vision model parameterized by $\theta_{cv}\in\Theta_{cv}$ pre-trained on massive labeled in-distribution (ID) data $\mathcal{D}^{id}=\{x^{id}_i, y^{id}_i\}_{i=0}^{m}\in\mathcal{X}\times\mathcal{Y}$, where $\mathcal{Y}=\mathbb{R}^C$. Here each ID example is sampled from a joint distribution, i.e., $(X^{id}, Y^{id})\sim p^{id}$, where $X^{id}$ and $Y^{id}$ stand for variables. After pretraining, we can assume the conditional distribution $P(Y^{id}|X^{id})$ can be perfectly captured by the inference function $\tilde{y}^{id}=f_{\theta_{cv}}(x^{id})$, where $\tilde{y}^{id}$ is the prediction. In OOD tasks, we are given a set of unlabeled examples $\mathcal{D}^{ood}=\{x^{ood}_i\}_{i=0}^{n}$ whose element $x^{ood}\in\mathcal{X}$ is drawn from an unknown data distribution $p^{ood}$. Due to the change of downstream task, some factors that affect the data generating process are shifted, causing a difference between $p^{ood}$ and $p^{id}$, further hindering the label prediction, i.e., $\tilde{y}^{ood}=f_{\theta_{cv}}(x^{ood})\not\sim P(Y^{ood}|X^{ood})$, where $Y^{ood}$ is the unknown ground truth. Fortunately, having been observed with extraordinary low-shot generalization capability, we leverage MLLM with parameters $\theta_{mllm}\in\Theta_{mllm}$ to enhance the OOD robustness of vision models.

Our framework is illustrated in Figure~\ref{MVTfig:framework} and our problem can be formulated as follows:
\begin{align}
    \min_{\theta_{cv}}\mathcal{L}(f_{\theta_{cv}}, z);\ z&=\left[\theta_{mllm}((X_c^{+}\!, Y_c^{+}); (X_c^{-}\!, Y_c^{+}); \!X_i)\right]_{c}^N\!; \nonumber \\
    Y_c&=T[c; \arg\!\max [f_{\theta_{cv}}(X_i)]],
\label{MVTeq:problem_formulation}
\end{align}
where $X_i^{+}$ and $X_i^{-}$ denotes the positive and negative exemplars, respectively, $X_i$ is the query image, and $T$ is the transition matrix. Intuitively, when a distribution shift occurs, the emerging prediction errors are essentially label noises that can be captured by estimating a transition matrix. Hence, by focusing on calibrating the examples with high noise probabilities, the visual robustness of downstream tasks can be improved effectively. In particular, we feed all OOD data into the vision model to obtain the noisy prediction distribution $P(\tilde{Y}^{ood}|X^{ood})$, based on which we can effectively estimate $T$ and provide exemplars to instruct MLLM\footnote{Although some manual annotation is required, we show in later experiments that our strategy has an acceptable labeling workload and demonstrates superior performance to vanilla fine-tuning on the support set. Furthermore, the support set is \textbf{ not used for parameter tuning} in our method, so our fine-tuning does not actually use any human annotation for training.}. Further, we conduct machine vision therapy to find the possible ground truth for $X_i$ based on the MLLM output $z$. Finally, $z$ is leveraged to minimize $\mathcal{L}$ to optimize $\theta_{cv}$. Next, we explain the details of each process.

\subsection{Transition Matrix Estimation}
\label{MVT:t_estimation}
The distribution shift from OOD data $x^{ood}$ leads to unreliable label prediction $\tilde{y}^{ood}$, which is highly unreliable due to instance-dependent feature noises~\cite{li2024instant, xia2020part} as shown in Section~\ref{MVT:ablation}. Hence, in order to capture the relationship between $\tilde{Y}^{ood}$ and $Y^{ood}$, we leverage a transition matrix $T\in [0, 1]^{C\times C}$~\cite{liu2015classification, natarajan2013learning, xia2019anchor} which satisfies $P(Y^{ood}|X^{ood})=T^\top P(\tilde{Y}^{ood}|X^{ood})$. However, estimating such a transition matrix is difficult without access to any noisy label supervision or strong assumption~\cite{liu2015classification, xia2019anchor}. Therefore, we propose a simple yet effective sample selection approach to construct a support set with clean labels. Specifically, we rank all OOD data within each class based on their prediction confidence, i.e., $\max_{c}\left[f_{\theta_{cv}}(x^{ood})\right]_c$, where $\left[\cdot\right]_c$ denotes the value of the $c$-th entry. From the sorted dataset $\{x_1^{ood, c}, x_2^{ood, c}, \cdots, x_{\frac{n}{C}}^{ood, c}\}_{c=1}^{C}$, we uniformly sample $\rho$ examples per class, where $\rho$ is the labeling budget, i.e., $\mathcal{D}^{supp}=\{\{x_{j\times\frac{n}{\rho C}}^{ood, c}\}_{j=1}^{\rho}\}_{c=1}^{C}$. In this way, we can effectively model the noisy posterior $P(\tilde{Y}^{ood}|X^{ood})$. Then, through an acceptable labeling process\footnote{We experimentally show that when there is a distribution shift between $\mathcal{D}^{supp}$ and $\mathcal{D}^{ood}$, the proposed method can still perform effectively. As a result, it is unnecessary to conduct the labeling process on each practical task. Instead, we can just use the existing support set to instruct most of OOD tasks.}, we can obtain the clean label posterior $P(Y^{ood}|X^{ood})$, thus effectively estimating the transition matrix $T$. Finally, the noise transition probability $T\left[:;\arg\max[f_{\theta_{cv}}]\right]$ of a query image can be obtained by indexing $T$ through its current prediction.

\subsection{Denoising In-Context Learning}
\label{MVT:dicl}
Thanks to the previously obtained noise probability list $T\left[:;\arg\max[f_{\theta_{cv}}]\right]$, we can further decide which one is the possible ground truth through DICL. In particular, we only consider the classes of the top-$N$ noise probability as potential candidates. If the label prediction denoted by ``PRE\#0'' is not in the candidates, we would fix it in the first place. Further, we conduct \textit{Diagnosing} which decides the fidelity of the current prediction, and \textit{Therapy} which finds the possible ground truth.

\textbf{Diagnosing.} Since the inference time of MLLMs is non-trivial, it is necessary to avoid redundant analysis on confident examples. Hence, to examine the fidelity of vision model predictions, our Diagnosing focuses on answering whether a query image belonging to class ``PRE\#0'' is ``True''. Specifically, we retrieve from $\mathcal{D}^{supp}$ to obtain one exemplar image belonging to ``PRE\#0'', and another exemplar image belonging to the class with the largest noise transition probability ``CLN\#1''\footnote{The performance of retrieve strategy is carefully studied in Section~\ref{MVT:performance_analysis}.}. Then, combined with the query image $X_q$, an in-context instruction is constructed:
\begin{tcolorbox}
Question: This image $<$IMG\_PRE\#0$>$ shows a photo of $<$PRE\#0$>$, True or False? Answer: True;\\
Question: This image $<$IMG\_CLN\#1$>$ shows a photo of $<$PRE\#0$>$, True or False? Answer: False;\\
Question: This image $<$IMG\_Query$>$ shows a photo of $<$PRE\#0$>$, True or False? Answer:
\end{tcolorbox}
The symbols $<$IMG\_PRE\#0$>$, $<$IMG\_CLN\#1$>$, and $<$IMG\_Query$>$ are replace tokens for the image features of exemplars from ``PRE\#0'' and ``CLN\#1'', and $X_q$, respectively. The first exemplar acts as the positive one to show MLLMs the true image from class ``PRE\#0'', and the second exemplar shows the negative one to show the highly probable false image from ``CLN\#1''. Then, based on the $X_q$ and ``PRE\#0'', MLLMs can effectively judge the correctness by outputting $z_0$:
\begin{equation}
	\!\!\!z_0\!=\!\theta_{mllm}((X_{PRE\#0},\! Y_{PRE\#0}); (X_{CLN\#1},\! Y_{PRE\#0});\! X_q)\!.\!\!
	\label{MVTeq:diagnose}
\end{equation}
To enable further quantitative analysis, we obtain the logits of ``True'' and ``False'' tokens from the MLLM output $z_0$ followed by a softmax function:
\begin{equation}
    z_0:=\text{softmax}(\left[z_0[\text{True}], z_0[\text{False}]\right]).
\end{equation}
Finally, we combine $z_0[\text{True}]$ and the prediction confidence of the vision model to obtain a detection score $\Delta$:
\begin{equation}
	\Delta = \frac{1}{2}(z_0[\text{True}]+\max_{c}\left[f_{\theta_{cv}}\right]_c(x^{ood})).
	\label{MVTeq:detection_score}
\end{equation}
If $\Delta$ is larger than a threshold $\delta$, we assume the current prediction ``PRE\#0'' is correct\footnote{Detailed analysis is shown in Section~\ref{MVT:performance_analysis}.}, otherwise, we conduct the next Therapy process.

\textbf{Therapy.} 
During therapy, we continue to use the instruction template above and traverse across the rest clean class candidates. Particularly, for each iteration $c$ in $N-1$ trials, we choose ``CLN\#c'' as the positive class and ``PRE\#0'' as the negative class, whose exemplars are correspondingly retrieved from $\mathcal{D}^{supp}$ to construct the prompt. Then, it is fed into MLLM to output whether the query image belongs to the class ``CLN\#c'', i.e., $z_c=\theta_{mllm}((X_{CLN\#c}, Y_{CLN\#c}); (X_{PRE\#0},\! Y_{CLN\#c});\! X_q)$,\!\! let $z_c := \text{softmax}(\left[z_c[\text{True}], z_c[\text{False}]\right])$. As a result, we can decide the final prediction through:
\begin{equation}
	y_{mllm} = \arg\max \left[z_c[\text{True}]\right]_{c=0}^N.
	\label{MVTeq:mllm_pred}
\end{equation}

As shown in Section~\ref{MVT:experiments}, the performance of MLLM prediction shows strong performance in many OOD scenarios. However, we still cannot directly employ MLLMs for inference, due to three main reasons: \textcolor[rgb]{0.05, 0.09, 0.58}{(1)} Non-negligible inference time: Since current MLLMs cannot handle large-batch data, it would be unimaginably slower (\textit{e.g.}, 1000$\times$) when using MLLMs rather than vision models; \textcolor[rgb]{0.05, 0.09, 0.58}{(2)} High requirements for computation: Inference through MLLM takes up huge memory of GPU. For MLLMs using large LLMs such as LLaMA-13B, it requires distributed inference on less advanced devices; \textcolor[rgb]{0.05, 0.09, 0.58}{(3)} Model privacy issue: Many MLLMs are highly sensitive with limited accessibility, therefore. Hence, we propose to fine-tune vision models based on the prediction of MLLMs.

\subsection{Fine-Tuning of Vision Models}
\label{MVT:fine-tuning}
After obtaining the MLLM prediction $y_{mllm}$, we propose to optimize vision models through the following objective:
\begin{equation}
min_{\theta_{cv}}\mathcal{L}_{ce}(f_{\theta_{cv}}, y_{mllm}),
\label{MVTeq:fine-tuning}
\end{equation}
where $\mathcal{L}_{ce}(\cdot)$ denotes the cross-entropy loss. Here we summarize our methodology in Algorithm~\ref{alg:MVT}. Further, we can directly deploy the fine-tuned vision models to OOD tasks whose effectiveness is demonstrated in Section~\ref{MVT:experiments}.

\begin{algorithm}[tb]
    \caption{Machine Vision Therapy.}
    \label{alg:MVT}
    \begin{algorithmic}[1]
        \REQUIRE Pre-trained vision model $\theta_{cv}$, MLLM $\theta_{mllm}$, OOD dataset $\mathcal{D}^{ood}$.
        \STATE Uniformly sample $\rho C$ examples from confidence-sorted $\mathcal{D}^{ood}$ to construct support set $\mathcal{D}^{supp}$;
        \STATE Estimate transition matrix $T$; \COMMENT{\textit{\color{black!60} Section~\ref{MVT:t_estimation}}}
        \FOR{$i = 0,1, \cdots, n$}
            \STATE Based on label prediction $\tilde{y}_i^{ood}$, obtain noisy transition probability $T\left[:;\arg\max[f_{\theta_{cv}}]\right]$;
            \STATE Conduct Diagnosing through Equation~\eqref{MVTeq:diagnose} and compute detection score $\Delta$ through Equation~\eqref{MVTeq:detection_score};
            \IF {$\Delta > \delta$}
                \STATE Accept current prediction;
            \ELSE 
                \STATE Conduct Therapy and obtain MLLM prediction through Equation~\eqref{MVTeq:mllm_pred}; \COMMENT{\textit{\color{black!60} Section~\ref{MVT:dicl}}}
                \STATE Based on the MLLM prediction, conduct fine-tuning through Equation~\eqref{MVTeq:fine-tuning}; \COMMENT{\textit{\color{black!60} Section~\ref{MVT:fine-tuning}}}
            \ENDIF
        \ENDFOR
    \end{algorithmic}
\end{algorithm}

\subsection{Theoretical Analysis}
We denote the MLLM is pretrained over a distribution $p$ defined by a latent concept $\phi\in\Phi$. During DICL, there are $n$ examples to form a prompt $S_n$ which are sampled from a prompt distribution $p_{prompt}$ defined by concept $\phi^*\in\Phi$. To justify the proposed DICL strategy, based on the theoretical framework proposed by Xie et al.~\cite{xie2021explanation}, we show that when MLLM achieving the most probable $z$ based on the given prompt $S_n$ and query image-text pair $x_q$-$y$ under a concept $\phi^*$, the corresponding $y$ is the same as the one found from $p_{prompt}$, which is $y_q$ that matches with $x_q$.

\begin{assumption}[Distribution consistency]
$\forall (x_q, y_q)\sim p_{prompt}, p(x_q, y_q)=p_{prompt}(x_q, y_q)$.
\end{assumption}

Moreover, the assumptions from Xie et al.~\cite{xie2021explanation} also hold, then we have:

\begin{theorem}
    Assume that the above assumptions hold, if for all $\phi\in\Phi$, $\phi\neq\phi^*$, the concept $\phi^*$ satisfies the distinguishability condition: $\sum_{j=1}^k KL_j(\phi^*\|\phi) > \epsilon_{start}^{\phi} + \epsilon_{delim}^{\phi}$, then as $n\rightarrow\infty$, the prediction according to the pretraining distribution is
    \begin{equation}
        \arg\max_y p(y\vert S_n, x_q, \phi^*) \rightarrow \arg\max_y p_{prompt}(y\vert x_q).
    \end{equation}
    Thus, the in-context predictor $f_n$ achieves the optimal $0-1$ risk: $\lim_{n\rightarrow\infty}\mathcal{L}_{0-1}(f_n)=\inf_f\mathcal{L}_{0-1}(f)$.
    \label{MVT:theorem_optimal_y}
\end{theorem}

\begin{lemma}
    Under the same condiction of Theorem~\ref{MVT:theorem_optimal_y}, the prediction $z$ according to the pretraining distribution is
    \begin{equation}
        \!\!\!\arg\!\max_z p(z\vert S_n,\! x_q,\! y_q,\! \phi^*) \!\!\rightarrow\! \arg\!\max_z p_{prompt}(z\vert x_q,\! y_q).\!\!\!\!
    \end{equation}
    \label{lemm:optimal_z_supp}
\end{lemma}

\begin{theorem}
    Assume that the above assumptions hold, as $n\rightarrow\infty$, when achieving the largest prediction probability of $z$ given prompt under concept $\phi^*$, the corresponding class description $y$ follows the same $y$ obtained from the prompt distribution:
    \begin{equation}
        \!\!\!\arg\!\max_y p(z\vert S_n,\! x_q,\! y,\! \phi^*)\! \rightarrow \!\arg\!\max_y p_{prompt}(z\vert x_q,\! y).\!\!\!
    \end{equation}
    \label{MVT:theorem}
\end{theorem}
We can see that if $n$ is large enough, the MLLM prediction $z$ achieves the largest value when $y_q$ is the exact match to $x_q$. As a result, we can justify that only when we feed the positive image-text pair to the MLLM, the prediction $z$ is the largest among all other combinations between $x_q$ and $y\in\mathcal{Y}, y\neq y_q$.

\begin{table*}[t]
\small
\centering
\caption{Classification accuracy (\%) of baseline CLIP models and our method on 5 ID datasets and 5 OOD datasets. The baseline methods includes ViT-L from CLIP~\cite{radford2021learning} and ViT-g from EVA~\cite{EVA}, VQA, and Vanilla FT.}
\setlength{\tabcolsep}{0.5mm}
\begin{tabular}{l|l|ccccc|ccccc}
\toprule
\multirow{2}{*}{Arch}&\multirow{2}{*}{Method}&\multicolumn{5}{c|}{ID}&\multicolumn{5}{c}{OOD}\\
\cline{3-12}
&& IN-Val & IN-V2 & CIFAR10 & CIFAR100 & MNIST & IN-A & IN-R & IN-SK & IN-V & iWildCam \\
\midrule\midrule
RN50 & \multirow{3}{*}{CLIP} & 59.7 & 52.6 & 71.5 & 41.9 & 58.5 & 23.9 & 60.7 & 35.4 & 31.1 & 8.2 \\
RN101 & & 61.7 & 56.2 & 80.8 & 48.8 & 51.6 & 30.2 & 66.7 & 40.9 & 35.4 & 12.3 \\
ViT-B & & 62.9 & 56.1 & 89.9 & 65.0 & 47.9 & 32.2 & 67.9 & 41.9 & 30.5 & 10.9 \\
\midrule
\multirow{5}{*}{ViT-L} & CLIP & 75.8 & 70.2& 95.6 & 78.2 & 76.4 & 69.3 & 86.6 & 59.4 & 51.8 & 13.4 \\
        & VQA & 64.9 & 59.9 & \underline{97.6} & \bf83.2 & 56.7 & 66.0 & 87.3 & 56.9 & 56.2 & 13.3 \\
        & Vanilla FT & \underline{76.1} & \bf70.8  & 96.1 & 80.3 & \underline{77.5} & 70.8 &  87.5 &  \underline{60.0} &  53.6 & \underline{15.2} \\
        & MVT & 75.2 & \bf70.8 & \bf97.9 & 78.9 & 53.0 & \underline{71.2} & \underline{88.1} & 59.0 & \underline{62.1} & \bf25.0 \\
        & +FT & \bf76.9 & \underline{70.5} & 96.7 & \underline{82.0} & \bf79.2 & \bf75.1 & \bf89.5 & \bf61.4 & \bf68.8 & - \\
\midrule
\multirow{5}{*}{ViT-g} & EVA & 78.8 & 71.2 & 98.3 & 88.8 & 62.2 & 71.9 & 91.4 & 67.7 & 64.9 & 21.9 \\
     & VQA & 64.3 & 59.6 & 97.9 & 84.5 & 55.7 & 64.6 & 87.4 & 58.2 & 59.2 & 19.7 \\
     & Vanilla FT & 78.9 & \underline{71.8} & \underline{98.7} & \underline{89.1} & 62.9 & 72.7 & \underline{91.6} & \underline{68.1} & 65.6 & \underline{22.4} \\
     & MVT & \bf79.1 & 71.6 & 98.1 & 89.0 & \underline{63.2} & \underline{73.2} & 91.4 & 67.9 & \underline{66.3} & \bf25.1 \\
     & +FT & \underline{79.0} & \bf72.2 & \bf98.9 & \bf91.2 & \bf65.7 & \bf75.5 & \bf92.8 & \bf68.6 & \bf70.6 & - \\
     \bottomrule
\end{tabular}
\label{MVTtab:1}
\end{table*}

\section{Experiment}
\label{MVT:experiments}
In this section, we first provide our experimental details. Then we conduct quantitative comparisons with the state-of-the-art vision models. Finally, we conduct ablation studies and analyses to qualitatively validate our method.

\subsection{Experimental Setup}
\textbf{Datasets.} In our experiments, we use well-known ID datasets including ImageNet-1K~\cite{deng2009imagenet} validation dataset, ImageNet-V2~\cite{recht2019imagenet}, CIFAR10~\cite{krizhevsky2009learning}, CIFAR100~\cite{krizhevsky2009learning} and MNIST~\cite{lecun1998gradient}. We also evaluate OOD generalization on datasets that are commonly considered OOD ones, ImageNet-A~\cite{hendrycks2021natural}, ImageNet-R~\cite{hendrycks2021many}, mageNet-Sketch~\cite{wang2019learning}, ImageNet-V~\cite{dong2022viewfool}, iWildCam~\cite{wilds2021}, and DomainBed~\cite{gulrajani2020search}.

\begin{table}[t]
\caption{Classification accuracy (\%) of baseline CLIP models and our method on 4 subsets of DomainBed datasets. The baseline methods include ViT-L from CLIP, ViT-g from EVA, VQA, and Vanilla FT.}
\label{MVTtab:2}
\centering
\setlength{\tabcolsep}{3.6mm}
\begin{tabular}{l|l|cccc|cccc}
\toprule
& Datasets & \multicolumn{4}{c|}{VLCS} & \multicolumn{4}{c}{PACS} \\
\cline{3-10}
& Method & 0 & 1 & 2 & 3 & 0 & 1 & 2 & 3 \\
\midrule
\multirow{4}{*}{\rotatebox{90}{ViT-L}} 
& CLIP       & 74.9 & 83.5 & 80.3 & 74.5 & 97.8 & 97.4 & 97.5 & \underline{99.4} \\
& Vanilla FT & 78.8 & 85.2 & 83.4 & 77.0 & \bf98.0 & 97.6 & 97.7 & 99.6 \\
& MVT        & \underline{83.8} & \underline{89.0} & \underline{87.2} & \underline{80.3} & 97.6 & \underline{97.5} & \bf98.0 & \underline{99.4} \\
& +FT        & \bf84.2 & \bf89.8 & \bf87.9 & \bf82.5 & \bf84.2 & \bf98.2 & \bf98.0 & \bf99.8 \\
\midrule
\multirow{4}{*}{\rotatebox{90}{ViT-g}} 
& EVA        & 72.5 & 80.0 & 79.8 & 72.8 & \underline{99.0} & \underline{98.8} & \underline{98.9} & \underline{99.8} \\
& Vanilla FT & 75.5 & 82.3 & 82.1 & 75.6 & 98.9 & 98.7 & \underline{98.9} & \underline{99.8} \\
& MVT        & \underline{81.2} & \underline{86.6} & \underline{86.1} & \underline{79.5} & 98.2 & 98.0 & 98.0 & 99.4 \\
& +FT        & \bf83.7 & \bf89.5 & \bf86.9 & \bf82.0 & \bf99.1 & \bf98.9 & \bf99.0 & \bf100.0 \\
\bottomrule
\end{tabular}

\setlength{\tabcolsep}{2.5mm}
\begin{tabular}{l|l|cccc|ccccc|c}
\toprule
& Datasets & \multicolumn{4}{c|}{OfficeHome} & \multicolumn{5}{c|}{DomainNet} & \multirow{2}{*}{Avg} \\
\cline{3-11}
& Method & 0 & 1 & 2 & 3 & 0 & 1 & 2 & 3 & 4 & \\
\midrule
\multirow{4}{*}{\rotatebox{90}{ViT-L}} 
& CLIP       & 87.7 & 92.7 & 85.7 & 85.6 & 61.1 & 62.1 & 60.2 & 78.4 & 51.1 & 80.6 \\
& Vanilla FT & \underline{87.9} & 93.1 & 87.1 & 86.9 & \underline{62.0} & \underline{62.5} & \underline{60.5} & 78.5 & 51.9 & 81.6 \\
& MVT        & 87.7 & \underline{93.4} & \underline{89.0} & \underline{88.5} & 61.3 & 62.1 & 60.4 & \underline{78.7} & \underline{53.4} & \underline{82.8} \\
& +FT        & \bf90.9 & \bf95.0 & \bf90.9 & \bf90.8 & \bf62.5 & \bf63.8 & \bf62.4 & \bf80.1 & \bf54.0 & \bf84.0 \\
\midrule
\multirow{4}{*}{\rotatebox{90}{ViT-g}} 
& EVA        & 90.5 & 94.2 & 88.6 & 88.7 & 61.4 & 64.7 & 61.2 & 81.6 & 54.9 & 81.6 \\
& Vanilla FT & \underline{90.6} & \underline{94.5} & 89.2 & 89.0 & 61.5 & 64.9 & 61.3 & 81.8 & 54.8 & 82.3 \\
& MVT        & 89.7 & 93.8 & \underline{89.7} & \underline{89.1} & \bf62.2 & \bf65.0 & \underline{61.6} & \bf82.3 & \underline{56.1} & \underline{83.3} \\
& +FT        & \bf91.6 & \bf95.1 & \bf90.7 & \bf90.6 & \underline{61.9} & \underline{64.8} & \bf63.2 & \underline{81.9} & \bf56.6 & \bf84.4 \\
\bottomrule
\end{tabular}
\end{table}

\textbf{Models and baselines.} For vision backbone, we employ CLIP models~\cite{radford2021learning} and utilize ViT-L/14 and ViT-g~\cite{zhai2022scaling} from EVA~\cite{EVA} as the vision model to be enhanced. For the MLLM backbone, we consider two existing works MMICL~\cite{zhao2023mmicl} and Otter~\cite{li2023otter} that possess multimodal ICL ability. Additionally, we conduct Visual Question Answering (VQA) to directly ask MLLMs the class of query images. Moreover, we conduct vanilla fine-tuning (Vanilla FT) using only $D^{supp}$ as a baseline. The performance of using MLLM prediction is denoted as MVT, and our fine-tuning result is denoted as FT.

\textbf{Settings.}
For model evaluation, we randomly select 5000 images independently from the ImageNet validation set (IN-Val), ImageNet-V2 (IN-V2), ImageNet-A (IN-A), ImageNet-R (IN-R), ImageNet-Sketch (IN-SK), ImageNet-V (IN-V) and 10000 images independently from CIFAR10, CIFAR100, MNIST to constitute the test samples. Additionally, we select $3$ images per category to construct a support set to provide in-context exemplars. We evaluate iWildCam from WILDS and VLCS, PACS, OfficeHome, and DomainNet from DomainBed. For the details of implementation, we choose the top-$6$ noisy classes to conduct MVT. Concretely, we set the threshold $\delta=0.6$ to diagnose incorrect predictions, then we retrieve exemplars from the support set based on the most similar logit prediction to query images. For each round of DICL, we repeat the process for $3$ times and average the model predictions. During fine-tuning, we optimize the vision models for $3$ epochs using Adam and SGD optimizers for ViT-L and ViT-g, respectively.

\begin{table*}[t]
\caption{Classification accuracy (\%) of baseline CLIP models and our method with MMICL~\cite{zhao2023mmicl} and Otter~\cite{li2023otter} as the VLMs on 5 ID datasets and 5 OOD datasets. We compare the performance of our method, and the fine-tuned models supervised by our method with the baseline models, i.e., ViT-L from CLIP~\cite{radford2021learning}. Fine-tuning with both MMICL and Otter improves the classification accuracy.}
\label{MVTtab:app_otter_imagenet}
\centering

\setlength{\tabcolsep}{3.8mm}
\begin{tabular}{l|l|ccccc}
\toprule
\multirow{2}{*}{MLLM} & \multirow{2}{*}{Method} & \multicolumn{5}{c}{ID} \\
\cline{3-7}
& & IN-Val & IN-V2 & CIFAR10 & CIFAR100 & MNIST \\
\midrule
None & CLIP & 75.8 & 70.2 & 95.6 & 78.2 & 76.4 \\
\midrule
\multirow{2}{*}{MMICL} 
& MVT & 75.2 & \bf70.8 & \bf97.9 & 78.9 & 53.0 \\
& +FT & \bf76.9 & \underline{70.5} & \underline{96.7} & \bf82.0 & \underline{79.2} \\
\midrule
\multirow{2}{*}{Otter} 
& MVT & 74.2 & 67.4 & 94.7 & 70.1 & 52.0 \\
& +FT & \underline{76.3} & 70.1 & 96.6 & \underline{81.8} & \bf81.3 \\
\bottomrule
\end{tabular}

\setlength{\tabcolsep}{2.6mm}
\begin{tabular}{l|l|ccccc|c}
\toprule
\multirow{2}{*}{MLLM} & \multirow{2}{*}{Method} & \multicolumn{5}{c|}{OOD} & \multirow{2}{*}{Avg (ID+OOD)} \\
\cline{3-7}
& & IN-A & IN-R & IN-SK & IN-V & iWildCam & \\
\midrule
None & CLIP & 69.3 & 86.6 & 59.4 & 51.8 & 13.4 & 61.7 \\
\midrule
\multirow{2}{*}{MMICL} 
& MVT & 71.2 & 88.1 & 59.0 & \underline{62.1} & \bf25.0 & \underline{64.3} \\
& +FT & \bf75.1 & \bf89.5 & \bf61.4 & \bf68.8 & - & \bf75.6 \\
\midrule
\multirow{2}{*}{Otter} 
& MVT & 64.1 & 85.2 & 59.5 & 51.9 & \underline{16.2} & 60.3 \\
& +FT & \underline{73.5} & \underline{88.7} & \underline{60.0} & 55.7 & - & 73.0 \\
\bottomrule
\end{tabular}
\end{table*}

\subsection{Quantitative Comparison}
First, we compare our MVT method with well-known vision models under both ID and OOD scenarios. As shown in Table~\ref{MVTtab:1}, we can see that our method with fine-tuning denoted as ``+FT'' achieves better performance in most settings. Specifically, on ``IN-V'', our method with fine-tuning can significantly surpass both CLIP and EVA for $17\%$ and $6\%$, respectively. Moreover, on ``IN-A'', our method achieves $4.3\%$ and $2.8\%$ performance improvement over the second-best method on both ViT-L and ViT-g backbone, respectively. We can also observe that even without fine-tuning, the prediction accuracy of MLLM denoted by ``MVT'' can still surpass all baselines on most scenarios, which denotes the strong performance enhancement of our MVT fine-tuning on vision models. Note that we did not provide fine-tuning on iWildCam because most of the predictions are incorrect. Though MVT can still achieve the best result, the vision encoders could be misled by erroneous decisions during the fine-tuning process.

\begin{table*}[t]
\caption{Classification accuracy (\%) of baseline CLIP models and our method with MMICL~\cite{zhao2023mmicl} and Otter~\cite{li2023otter} as the VLMs on 4 subsets of DomainBed datasets, including VLCS, PACS, OfficeHome, and DomainNet. We compare the performance of our method and the fine-tuned models supervised by our method with the baseline models, i.e., ViT-L from CLIP~\cite{radford2021learning}. Fine-tuning with both MMICL and Otter improves the classification accuracy.}
\label{MVTtab:app_otter_domainbed}
\centering

\setlength{\tabcolsep}{3.3mm}
\begin{tabular}{l|l|cccc|cccc}
\toprule
\multirow{2}{*}{MLLM} & Datasets & \multicolumn{4}{c|}{VLCS} & \multicolumn{4}{c}{PACS} \\
\cline{3-10}
& Method & 0 & 1 & 2 & 3 & 0 & 1 & 2 & 3 \\
\midrule
None & CLIP & 74.9 & 83.5 & 80.3 & 74.5 & \underline{97.8} & 97.4 & \underline{97.5} & \underline{99.4} \\
\midrule
\multirow{2}{*}{MMICL} 
& MVT & \underline{83.8} & \underline{89.0} & \underline{87.2} & \underline{80.3} & 97.6 & 97.5 & \bf98.0 & \underline{99.4} \\
& +FT & \bf84.2 & \bf89.8 & \bf87.9 & \bf82.5 & \bf98.0 & \bf98.2 & \bf98.0 & \bf99.8 \\
\midrule
\multirow{2}{*}{Otter} 
& MVT & 67.5 & 77.4 & 73.7 & 66.6 & 97.0 & 96.3 & 96.5 & 99.0 \\
& +FT & 76.8 & 87.7 & 82.3 & 77.4 & \bf98.0 & \underline{97.7} & \bf98.0 & \bf99.8 \\
\bottomrule
\end{tabular}

\setlength{\tabcolsep}{2.1mm}
\begin{tabular}{l|l|cccc|ccccc|c}
\toprule
\multirow{2}{*}{MLLM} & Datasets & \multicolumn{4}{c|}{OfficeHome} & \multicolumn{5}{c|}{DomainNet} & \multirow{2}{*}{Avg} \\
\cline{3-11}
& Method & 0 & 1 & 2 & 3 & 0 & 1 & 2 & 3 & 4 & \\
\midrule
None & CLIP & 87.7 & 92.7 & 85.7 & 85.6 & 61.1 & 62.1 & 60.2 & 78.4 & 51.1 & 80.6 \\
\midrule
\multirow{2}{*}{MMICL} 
& MVT & 87.7 & \underline{93.4} & \underline{89.0} & \underline{88.5} & 61.3 & 62.1 & 60.4 & \underline{78.7} & \underline{53.4} & \underline{82.8} \\
& +FT & \bf90.9 & \bf95.0 & \bf90.9 & \bf90.8 & \bf62.5 & \bf63.8 & \bf62.4 & \bf80.1 & \bf54.0 & \bf84.0 \\
\midrule
\multirow{2}{*}{Otter} 
& MVT & 85.6 & 89.9 & 83.6 & 83.3 & 56.5 & 58.6 & 56.3 & 74.1 & 46.5 & 77.0 \\
& +FT & \underline{88.7} & \underline{93.4} & 87.7 & 87.1 & \underline{62.0} & \underline{63.0} & \underline{61.3} & \underline{79.7} & \underline{53.4} & 82.0 \\
\bottomrule
\end{tabular}
\end{table*}

Furthermore, we consider domain shift by leveraging DomainBed datasets. Specifically, for each dataset, we leave one domain out as a test dataset and fine-tune on rest domains. By comparing two state-of-the-art vision backbones ViT-L and ViT-g, we show the performance comparison in Table~\ref{MVTtab:2}. As we can see, both MVT and MVT with fine-tuning can significantly surpass the baseline methods. For some scenarios such as the PACS dataset, our method can achieve nearly $100\%$ performance. Moreover, in several scenarios in the VLCS dataset, both our MVT and fine-tuning can achieve almost $10\%$ improvements. Additionally, we find that our method with fine-tuning largely surpasses vanilla fine-tuning baseline on both Tables~\ref{MVTtab:1} and~\ref{MVTtab:2}. Hence, we can conclude that our learning strategy can indeed provide effective supervisions which enhances vision robustness under distribution shift.

\begin{table*}[t]
\centering
\caption{Classification accuracy (\%) of baseline CLIP models and our method on 5 ID datasets and 5 OOD datasets. We compare the performance of our method, and the fine-tuned models supervised by our method with the baseline models, including ResNet-50 and ViT-B/32. The supervisor MLLM is MMICL~\cite{zhao2023mmicl}.}
\label{MVTtab:app_vision_model_imagenet}

\setlength{\tabcolsep}{4mm}
\begin{tabular}{l|l|ccccc}
\toprule
\multirow{2}{*}{Arch} & \multirow{2}{*}{Method} & \multicolumn{5}{c}{ID} \\
\cline{3-7}
& & IN-Val & IN-V2 & CIFAR10 & CIFAR100 & MNIST \\
\midrule
\multirow{3}{*}{RN50} 
& CLIP & 59.7 & 52.6 & 71.5 & 41.9 & \bf58.5 \\
& MVT & \bf76.2 & \bf70.8 & \bf80.2 & \bf49.7 & \underline{50.8} \\
& +FT & \underline{66.3} & \underline{65.7} & \underline{75.1} & \underline{46.9} & 47.3 \\
\midrule
\multirow{3}{*}{ViT-B} 
& CLIP & 62.9 & 56.1 & 89.9 & \bf65.0 & \underline{47.9} \\
& MVT & \bf77.5 & \bf71.0 & \bf92.5 & \underline{60.4} & \bf51.5 \\
& +FT & \underline{66.3} & \underline{66.0} & \underline{90.1} & 59.5 & 46.6 \\
\bottomrule
\end{tabular}

\setlength{\tabcolsep}{5mm}
\begin{tabular}{l|l|ccccc}
\toprule
\multirow{2}{*}{Arch} & \multirow{2}{*}{Method} & \multicolumn{5}{c}{OOD} \\
\cline{3-7}
& & IN-A & IN-R & IN-SK & IN-V & iWildCam \\
\midrule
\multirow{3}{*}{RN50} 
& CLIP & 23.9 & 60.7 & 35.4 & 31.1 & 8.2 \\
& MVT & \bf47.5 & \bf72.9 & \bf41.6 & \bf54.1 & \bf14.5 \\
& +FT & \underline{32.1} & \underline{64.4} & \underline{36.5} & \underline{38.2} & - \\
\midrule
\multirow{3}{*}{ViT-B} 
& CLIP & 32.2 & 67.9 & 41.9 & 30.5 & 10.9 \\
& MVT & \bf60.6 & \bf83.0 & \bf47.8 & \bf53.1 & \bf19.3 \\
& +FT & \underline{38.8} & \underline{68.7} & \underline{43.1} & \underline{37.6} & - \\
\bottomrule
\end{tabular}
\end{table*}

\begin{table*}[t]
\centering
\caption{Classification accuracy (\%) of baseline CLIP models and our method on 4 subsets of DomainBed datasets, including VLCS, PACS, OfficeHome, and DomainNet. We compare the performance of our method and the fine-tuned models supervised by our method with the baseline models, including ResNet-50 and ViT-B/32. The supervisor MLLM is MMICL~\cite{zhao2023mmicl}.}
\label{MVTtab:app_vision_model_domainbed}

\setlength{\tabcolsep}{3.6mm}
\begin{tabular}{l|l|cccc|cccc}
\toprule
\multirow{2}{*}{Arch} & \multirow{2}{*}{Method} & \multicolumn{4}{c|}{VLCS} & \multicolumn{4}{c}{PACS} \\
\cline{3-10}
& & 0 & 1 & 2 & 3 & 0 & 1 & 2 & 3 \\
\midrule
\multirow{3}{*}{RN50} 
& CLIP & 75.0 & 82.3 & 81.3 & 75.0 & 91.3 & 90.3 & 90.0 & 96.2 \\
& MVT  & \bf84.3 & \bf88.0 & \bf88.7 & \bf81.8 & \bf96.0 & \bf96.1 & \bf95.4 & \bf98.8 \\
& +FT  & \underline{83.7} & \underline{87.3} & \underline{88.1} & \underline{81.3} & \underline{95.6} & \underline{95.7} & \underline{95.1} & \underline{98.6} \\
\midrule
\multirow{3}{*}{ViT-B} 
& CLIP & 74.0 & 82.0 & 79.6 & 74.4 & \underline{93.6} & \underline{92.8} & \underline{93.0} & \underline{98.2} \\
& MVT  & \bf84.2 & \bf87.3 & \bf88.4 & \bf82.8 & \bf96.7 & \bf96.4 & \bf96.6 & \bf98.8 \\
& +FT  & \underline{76.0} & \underline{84.8} & \underline{81.3} & \underline{81.6} & 92.9 & 88.8 & 89.4 & 93.3 \\
\bottomrule
\end{tabular}

\setlength{\tabcolsep}{2.4mm}
\begin{tabular}{l|l|cccc|ccccc|c}
\toprule
\multirow{2}{*}{Arch} & \multirow{2}{*}{Method} & \multicolumn{4}{c|}{OfficeHome} & \multicolumn{5}{c|}{DomainNet} & \multirow{2}{*}{Avg} \\
\cline{3-11}
& & 0 & 1 & 2 & 3 & 0 & 1 & 2 & 3 & 4 & \\
\midrule
\multirow{3}{*}{RN50} 
& CLIP & 71.7 & 80.9 & 69.4 & 67.8 & \underline{47.2} & \underline{46.8} & \underline{44.9} & \underline{64.0} & \underline{32.9} & 71.0 \\
& MVT  & \bf77.2 & \bf85.3 & \bf77.5 & \bf75.4 & \bf46.1 & \bf46.3 & \bf43.4 & \bf61.7 & \bf33.2 & \bf75.0 \\
& +FT  & \underline{75.9} & \underline{85.0} & \underline{75.8} & \underline{74.7} & 45.3 & 45.6 & 43.0 & 60.4 & 32.6 & \underline{74.3} \\
\midrule
\multirow{3}{*}{ViT-B} 
& CLIP & 79.2 & 86.4 & 77.4 & 76.3 & \underline{49.7} & \underline{54.3} & \underline{51.0} & \underline{68.7} & \underline{40.7} & \underline{74.8} \\
& MVT  & \bf84.0 & \bf89.3 & \bf82.9 & \bf81.5 & \bf49.5 & \bf53.1 & \bf51.5 & \bf69.9 & \bf41.7 & \bf78.5 \\
& +FT  & \underline{81.1} & \underline{88.3} & \underline{80.8} & \underline{77.7} & 47.2 & 52.5 & 48.7 & 66.9 & 40.5 & \underline{74.8} \\
\bottomrule
\end{tabular}
\end{table*}

\subsection{Quantitative Comparison using Otter}
Similarly, here we conduct additional experiments on various ImageNet-based datasets and DomainBed datasets using ViT-L but a different MLLM backbone: Otter~\cite{li2023otter}. The results are shown in Tables~\ref{MVTtab:app_otter_imagenet} and~\ref{MVTtab:app_otter_domainbed}. We find that the performance of MVT is dependent on the MLLM backbone: when using Otter as the backbone model for MVT, the OOD performance would slightly degrade from the performance of MMICL, which could be due to the capability of MLLM to conduct ICL. However, the rectified predictions can still contain useful information to boost the performance of vision models. In several cases in ImageNet-Val, MNIST, and ImageNet-R, Otter with fine-tuning can still improve the visual robustness to the best or second-best results.

\begin{table*}[h]
    \small
    \caption{Classification accuracy (\%) of baseline CLIP models and our method with MMICL~\cite{zhao2023mmicl} as the VLM on 15 corruptions and 5 severities of ImageNet-C datasets. We compare the performance of our method and the fine-tuned models supervised by our method with the baseline models, i.e., ViT-L from CLIP~\cite{radford2021learning}. The fine-tuned models with our MVT method have the best performance.}
    \centering
    \setlength{\tabcolsep}{0.6mm}
    \renewcommand{\arraystretch}{1.2}
    \begin{tabular}{l|cccccc|cccccc|cccccc}
        \toprule
        \multirow{2}{*}{\shortstack{}} & \multicolumn{6}{c|}{Gaussian Noise} & \multicolumn{6}{c|}{Shot Noise} & \multicolumn{6}{c}{Impulse Noise} \\
        \cline{2-19}
        & 1 & 2 & 3 & 4 & 5 & avg & 1 & 2 & 3 & 4 & 5 & avg & 1 & 2 & 3 & 4 & 5 & avg \\
        \hline
        \hline
        CLIP & 69.8 & 66.7 & 59.7 & 46.9 & 30.6 & 54.7 & 70.5 & 64.9 & 57.7 & \underline{43.6} & 32.1 & 53.8 & 65.7 & 60.2 & 55.9 & 45.0 & 32.7 & 51.9 \\
        MVT & \underline{70.1} & \underline{67.5} & \underline{61.2} & \bf49.8 & \bf{33.6} & \underline{56.4} & \underline{70.8} & \underline{66.8} & \underline{59.2} & \bf46.1 & \bf35.6 & \underline{55.7} & \underline{66.3} & \underline{61.9} & \underline{58.4} & \underline{47.5} & \underline{35.6} & \underline{53.9} \\
        +FT & \bf71.0 & \bf67.9 & \bf61.3 & \underline{48.7} & \underline{33.5} & \bf56.5 & \bf72.0 & \bf67.1 & \bf60.1 & \bf46.1 & \underline{35.2} & \bf56.1 & \bf68.5 & \bf64.2 & \bf59.8 & \bf48.9 & \bf35.8 & \bf55.4 \\
        \cline{1-19}
        \cline{1-19}
        & \multicolumn{6}{c|}{Defocus Blur} & \multicolumn{6}{c|}{Glass Blur} & \multicolumn{6}{c}{Motion Blur} \\
        \cline{2-19}
        & 1 & 2 & 3 & 4 & 5 &avg& 1 & 2 & 3 & 4 & 5 &avg & 1 & 2 & 3 & 4 & 5 &avg\\
        \cline{1-19}
        \cline{1-19}
        CLIP & 66.1 & 62.4 & 53.0 & 43.4 & 35.0 & 52.0 & 65.5 & 59.3 & 40.5 & 33.8 & 25.4 & 44.9 & 70.9 & 66.8 & 59.9 & 49.5 & 41.8 & 57.8 \\
        MVT & \underline{67.1} & \underline{63.3} & \underline{55.8} & \bf47.6 & \bf38.8 & \underline{54.5} & \underline{67.1} & \underline{61.3} & \underline{42.8} & \underline{36.0} & \underline{29.4} & \underline{47.3} & \underline{71.9} & \underline{67.7} & \underline{60.9} & \underline{51.5} & \underline{43.2} & \underline{59.0} \\
        +FT & \bf68.8 & \bf64.1 & \bf56.3 & \underline{47.4} & \underline{38.4} & \bf55.0 & \bf68.9 & \bf64.8 & \bf45.2 & \bf37.6 & \bf30.2 & \bf49.3 & \bf72.8 & \bf69.1 & \bf62.1 & \bf52.7 & \bf45.3 & \bf60.4 \\
        \cline{1-19}
        \cline{1-19}
        & \multicolumn{6}{c|}{Zoom Blur} & \multicolumn{6}{c|}{Snow} & \multicolumn{6}{c}{Frost} \\
        \cline{2-19}
        & 1 & 2 & 3 & 4 & 5 &avg& 1 & 2 & 3 & 4 & 5 &avg & 1 & 2 & 3 & 4 & 5 &avg\\
        \cline{1-19}
        \cline{1-19}
        CLIP & 62.2 & 55.9 & 49.8 & 43.9 & 37.3 & 49.8 & 68.3 & 61.2 & 61.9 & 56.1 & 52.6 & 60.0 & 68.5 & 61.2 & 53.8 & 51.1 & 46.6 & 56.2 \\
        MVT & \underline{64.1} & \underline{57.3} & \underline{52.0} & \underline{45.7} & \underline{38.7} & \underline{51.6} & \underline{69.2} & \underline{61.5} & \underline{62.9} & \underline{57.1} & \underline{54.0} & \underline{60.9} & \underline{69.5} & \underline{61.5} & \underline{54.2} & \underline{52.7} & \underline{47.4} & \underline{57.1} \\
        +FT & \bf65.2 & \bf59.2 & \bf54.2 & \bf48.8 & \bf41.4 & \bf53.8 & \bf70.6 & \bf63.9 & \bf64.6 & \bf59.2 & \bf55.6 & \bf62.8 & \bf71.9 & \bf65.2 & \bf57.9 & \bf56.4 & \bf51.5 & \bf60.6 \\
        \cline{1-19}
        \cline{1-19}
        & \multicolumn{6}{c|}{Fog} & \multicolumn{6}{c|}{Brightness} & \multicolumn{6}{c}{Contrast} \\
        \cline{2-19}
        & 1 & 2 & 3 & 4 & 5 &avg& 1 & 2 & 3 & 4 & 5 &avg & 1 & 2 & 3 & 4 & 5 &avg \\
        \cline{1-19}
        \cline{1-19}
        CLIP & 69.8 & 67.9 & 65.0 & 61.3 & 52.0 & 63.2 & 74.3 & 74.0 & \underline{72.8} & 70.6 & 68.1 & 72.0 & 70.6 & 69.3 & 64.8 & 52.4 & 35.1 & 58.4 \\
        MVT & \underline{70.7} & \underline{69.2} & \underline{66.5} & \underline{62.6} & \underline{53.8} & \underline{64.6} & \underline{74.7} & \underline{74.1} & 72.6 & \underline{71.1} & \underline{68.8} & \underline{72.3} & \underline{70.9} & \underline{69.9} & \underline{65.1} & \underline{52.9} & \underline{36.9} & \underline{59.1} \\
        +FT & \bf72.5 & \bf71.3 & \bf69.5 & \bf67.1 & \bf60.3 & \bf68.1 & \bf76.0 & \bf75.1 & \bf74.3 & \bf73.1 & \bf71.1 & \bf73.9 & \bf73.5 & \bf73.5 & \bf70.2 & \bf59.2 & \bf42.7 & \bf63.8 \\
        \cline{1-19}
        \cline{1-19}
        & \multicolumn{6}{c|}{Elastic} & \multicolumn{6}{c|}{Pixelate} & \multicolumn{6}{c}{JPEG} \\
        \cline{2-19}
        & 1 & 2 & 3 & 4 & 5 &avg& 1 & 2 & 3 & 4 & 5 &avg& 1 & 2 & 3 & 4 & 5 & avg \\
        \cline{1-19}
        \cline{1-19}
        CLIP & 69.2 & 50.6 & 64.1 & 53.1 & 30.4 & 53.5 & 71.0 & 70.4 & 66.2 & 60.1 & 54.6 & 64.5 & 70.8 & 67.7 & 65.1 & 58.0 & 45.3 & 61.4 \\
        MVT & \underline{70.0} & \underline{51.1} & \underline{65.8} & \underline{55.2} & \bf32.7 & \underline{55.0} & \underline{71.7} & \underline{70.7} & \underline{66.5} & \underline{61.9} & \underline{57.3} & \underline{65.6} & \bf72.5 & \bf69.6 & \bf67.5 & \bf60.5 & \underline{47.7} & \bf63.6 \\
        +FT & \bf70.7 & \bf53.6 & \bf67.7 & \bf58.5 & \underline{32.2} & \bf56.5 & \bf72.8 & \bf71.8 & \bf69.1 & \bf62.9 & \bf57.7 & \bf66.9 & \underline{71.2} & \underline{68.9} & \underline{65.8} & \underline{60.0} & \bf48.8 & \underline{62.9} \\
        \bottomrule
    \end{tabular}
    \label{MVTtab:app_corruptions}
\end{table*}

\subsection{MVT on Additional Vision Models}
\label{MVT:rn_vitb}
Then, we conduct MVT using MMICL but using different vision backbone models such as ViT-B and ResNet-50 on ImageNet and DomainBed datasets. The results are shown in Tables~\ref{MVTtab:app_vision_model_imagenet} and~\ref{MVTtab:app_vision_model_domainbed}. We can see that the performance of MVT is quite strong compared to other vision models which shows over $10\%$ and $4\%$ improvements in ImageNet datasets and DomainBed datasets, respectively. Especially on ImageNet-V2,  ImageNet-A, ImageNet-R, and ImageNet-V, the performance improvement of MVT are encouragingly over $15\%$, $24\%$, $12\%$, and $23\%$, respectively. After fine-tuning, the performance can be improved in most cases, such as ResNet-50 is further improved by $13.1\%$ and $3.3\%$ correspondingly on ImageNet-V2 and DomainBed thanks to MMICL.

\subsection{Robustness against Visual Corruptions}
Further, we consider the visual robustness against corruptions by evaluating EVIL on a robustness benchmark: ImageNet-C~\cite{hendrycks2019benchmarking}. Specifically, there are 15 different types of corruption with different corruption severities varied from 1 to 5. Here we cover all scenarios to evaluate our method using MMICL as a backbone model and a baseline method CLIP ViT-L. The results are shown in Table~\ref{MVTtab:app_corruptions}. We can see that our method shows very strong performance in all scenarios. Compared to CLIP, using MVT can improve the performance by over $2\%$, and through fine-tuning, the performance is further boosted by over $4\%$. The encouraging results again demonstrate the effectiveness of our method.

\begin{figure}[t]
\centering
\includegraphics[width=\linewidth]{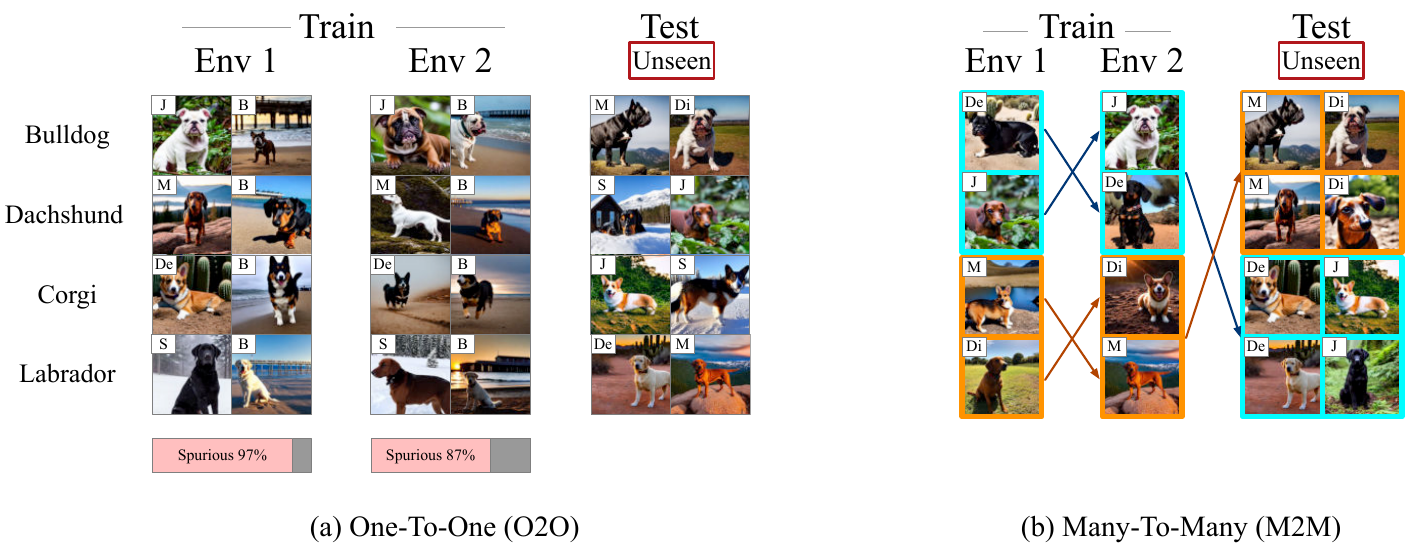}
\caption{Figures are from Lynch et al.~\cite{lynch2023spawrious}, the letters on each images denote a certain background. There are two spurious correlation types in the Spawrious dataset, namely O2O and M2M. In the O2O setting, each dog class is correlated to one certain background type and different distributions have different correlation probabilities as shown by the bar below the O2O figure. As for the M2M setting, multiple classes and backgrounds are correlated together and the correlation changes to different groups of classes and backgrounds during testing.}
\label{fig:spurious_example}
\end{figure}

\begin{table}[t]
	\caption{Performance comparison between MVT and CLIP on robustness against spurious correlation using Spawrious dataset.}
	\setlength{\tabcolsep}{0.6mm}
	\begin{tabular}{l|cccccc|c}
		\toprule
		Type  & O2O\_easy & O2O\_medium & O2O\_hard & M2M\_easy & M2M\_medium & M2M\_hard & Avg. \\ \hline
		ViT-L & 94.1      & 95.4        & 93.3      & 96.7      & 95.0        & 92.5      & 94.5 \\
		MVT   & \textbf{95.8}      & \textbf{96.3}        & \textbf{93.6 }     & \textbf{96.8 }     & \textbf{95.8}        & \textbf{92.9}      & \textbf{95.2} \\ \hline \hline
		ViT-g & 94.6      & 97.0        & 92.6      & 96.7      & 95.6        & 94.8      & 95.2 \\
		MVT   & \textbf{95.3 }     & \textbf{97.4}        &\textbf{ 92.8}      & \textbf{96.8}      & \textbf{96.6 }       & \textbf{95.4}      & \textbf{95.7} \\ \bottomrule
	\end{tabular}
	\label{MVTtab:spurious}
\end{table}

\subsection{Robustness against Spurious Correlation}
Moreover, we consider a common distribution shift scenario where the training dataset and test dataset have different foreground and background correlation, \textit{i.e.}, spurious correlation. Specifically, as shown in Figure~\ref{fig:spurious_example}, standing for the Spawrious dataset that we use, there are two different settings: One-To-One (O2O) correlation, where each class is correlated to one background type with a certain probability. The foreground objects in the training dataset and test dataset have different probabilities of being combined with a certain background. For the Many-To-Many (M2M) setting, the foregrounds and backgrounds are split into subgroups that contain multiple classes and background types. When different subgroups are correlated together between training and test datasets, the M2M spurious correlation is formed and brings more complexity. In the Spawrious dataset, there are three levels of hardness based on the correlation probability difference between the training and test datasets, namely easy, medium, and hard. Here, we consider all scenarios and show the results in Table~\ref{MVTtab:spurious}. We can see that the MVT method can outperform the ViT-L and ViT-g baseline methods in all scenarios, which leads to the conclusion that our method is robust to spurious correlations and can identify the class of interests despite the changing backgrounds.

\begin{figure}[h]
	\centering
	\includegraphics[width=\linewidth]{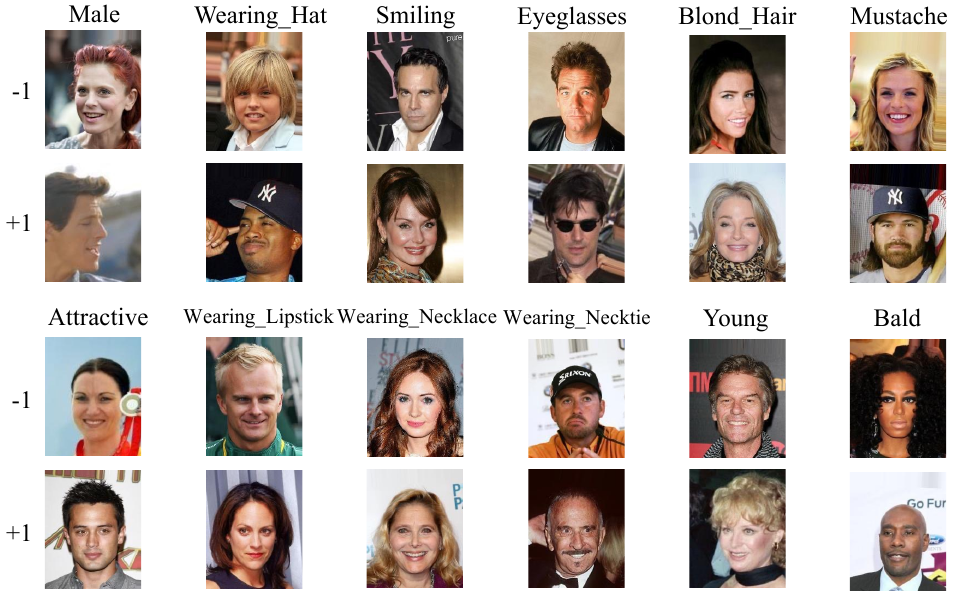}
	\caption{Examples of celebA photos with different attributes.}
	\label{fig:attribute}
\end{figure}

\begin{table}[h]
	\centering
	\caption{Class names for 12 chosen attributes.}
	\begin{tabular}{l|cc}
		\toprule
		Attribute         & \multicolumn{1}{c}{-1}                      & \multicolumn{1}{c}{+1}                  \\ \hline
		Male              & \multicolumn{1}{c}{a woman}                 & \multicolumn{1}{c}{a man}               \\
		Wear\_Hat         & \multicolumn{1}{c}{not wearing a hat}       & \multicolumn{1}{c}{wearing a hat}       \\
		Smiling           & \multicolumn{1}{c}{not smiling}             & \multicolumn{1}{c}{smiling}             \\
		Eyeglasses        & \multicolumn{1}{c}{not wearing eye glasses} & \multicolumn{1}{c}{wearing eye glasses} \\
		Blond\_Hair       & not having blond hair                       & having blind hair                       \\
		Mustache          & not having mustache                         & having mustache                         \\
		Attractive        & not attractive                              & attractive                              \\
		Wearing\_Lipstick & not wearing lipstick                        & wearing lipstick                        \\
		Wearing\_Necklace & not wearing necklace                        & wearing necklace                        \\
		Wearing\_Necktie  & not wearing necktie                         & wearing necktie                         \\
		Young             & not young                                   & young                                   \\
		Bald              & not bald                                    & bald                                    \\ \bottomrule
	\end{tabular}
	\label{MVTtab:attribute_classnames}
\end{table}

\subsection{Performance on Recognizing Fine-grained Attributes}

Additionally, here we further explore the capability of recognizing subtle attributes based on the CelebA dataset~\cite{liu2015faceattributes}. Particularly, we consider 12 face attributes, as shown in Figure~\ref{fig:attribute}. For each attribute, we testify whether a learning model could correctly identify the attribute in a given image. Here we compare our MVT method with CLIP ViT-L and ViT-g, and the performance of MVT produced by conducting therapy on ViT-L and ViT-g models.

Particularly, since CelebA is a binary classification task, here we design different prompts for vision models and our MLLM. For CLIP models, we use \code{The person in this image is <\#classname>} as text input, where \code{<\#classname>} of each attribute is shown in Table~\ref{MVTtab:attribute_classnames}. For our method, we still designed one positive prompt and one negative prompt for each ICL round. Specifically, for ``Male'' attribute, our instruction is as follows:
\begin{tcolorbox}
	Question: Is the person in this image \{replace\_roken\} a male? Answer: True;\\
	Question: Is the person in this image \{replace\_roken\} a female? Answer: False;\\
	Question: Is the person in this image \{replace\_roken\} a male? Answer:
\end{tcolorbox}
in which is first exemplar demonstrates an image of a male positively described as male, the second exemplar shows an image of a male negatively described as female, and finally, we ask whether the input image is a male and use the output of MLLM as the prediction.

\begin{table}[t]
	\caption{Performance comparison between MVT and CLIP on recognizing fine-grained attributes using CelebA dataset.}
	\setlength{\tabcolsep}{0.9mm}
	\begin{tabular}{l|cccccccccccc|c}
		\toprule
		\scriptsize{Attr.} & \scriptsize{Male} & \scriptsize{Hat} & \scriptsize{Smiling} & \scriptsize{Glasses} & \scriptsize{Blond} & \scriptsize{Mustache} & \scriptsize{Attract} & \scriptsize{Lipstick} & \scriptsize{Necklace} & \scriptsize{Necktie} & \scriptsize{Young} & \scriptsize{Bald} & \scriptsize{Avg.} \\ \hline
		
		ViT-L     & 63.0 & 60.8         & 64.5    & 75.8       & 36.2        & 29.0     & 42.0       & 30.8              & 38.0              & 37.5             & 66.6  & 86.3 & 52.5 \\
		MVT       & \textbf{74.0} & \textbf{67.0 }        & \textbf{65.4}    & \textbf{76.1}       & \textbf{53.0}        & \textbf{55.8}     & \textbf{42.4}       & \textbf{39.4}              & \textbf{38.6 }             & \textbf{53.9}             & \textbf{73.5}  & \textbf{88.1} & \textbf{60.6 }\\ \hline \hline
		ViT-g     & 98.5 & 75.5         & 70.4    & 83.8       & 46.0        & 66.6     & 58.2       & 72.5              & 43.5              & 28.4             & 54.1  & 91.3 & 65.7 \\
		MVT       & \textbf{98.9} & \textbf{77.2}         & \textbf{71.0}    & \textbf{84.1}       & \textbf{58.3}        & \textbf{74.9 }    & \textbf{59.0  }     & \textbf{73.2}              & \textbf{43.6}              & \textbf{41.2}             & \textbf{56.1}  & \textbf{91.8} & \textbf{69.1} \\ \bottomrule
	\end{tabular}
	\label{MVTtab:celeba}
\end{table}

The results on CelebA are shown in Table~\ref{MVTtab:celeba}, we observe that our method is quite effective in recognizing fine-grained attributes and its performance significantly surpasses ViT-L and ViT-g with a large margin. Especially in attributes such as  ``Blond\_Hair'', ``Mustache'', and ``Wearing\_Necktie'', the performance improvements are even over $10\%$ on both two CLIP models, and the final averaged results on all 12 attributes, the total improvements are $8.1\%$ and $3.4\%$ for ViT-L and ViT-g, respectively. Therefore, it is reasonable to conclude that our method can be effectively conducted on fine-grained attribute recognition and significantly outperforms several powerful vision models.

\subsection{Ablation Study}
\label{MVT:ablation}
In this part, we conduct ablation studies to analyze each module of MVT by using ViT-L backbone vision model.

\begin{figure}[t]
    \centering
    \includegraphics[width=\linewidth]{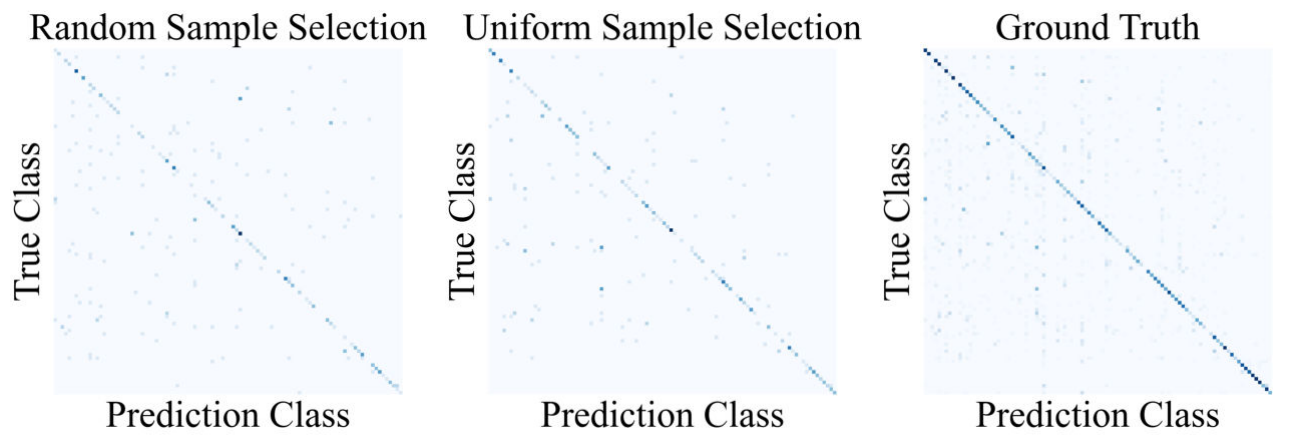}
\caption{Ablation study on transition matrix estimation by comparing our method with random sampling and ground truth.}
\label{MVTfig:estimate}
\end{figure}

\textbf{Ablation Study on Transition Matrix Estimation.}
To validate the performance of transition matrix estimation, we compare our confidence-based uniform sampling strategy to a random sampling baseline. The result on the ImageNet-V dataset is shown in Figure~\ref{MVTfig:estimate}. To quantitatively show the superiority of our method, we compute the $\ell_2$ norm of the difference between one estimation and ground truth which indicates the fidelity of the estimation. As a result, our estimation is much more accurate by achieving $3.83$ norm, compared to $4.46$ of random sampling.

\setlength{\intextsep}{1.4pt}
\setlength{\columnsep}{2pt}

\textbf{Ablation Study on Choosing Noisy Classes.}
Further, we justify the choice of using a transition matrix to obtain the noisy classes. As a comparison, we use the top-$6$ predictions as the therapy candidates and show the results in Table~\ref{MVTtab:4}. We can see on all datasets, our method can outperform the opponent with non-trivial improvements. Therefore, leveraging the transition matrix to find the potential noisy classes is more effective than using prediction.

\begin{table}[t]
\centering
\caption{Performance comparison between choosing noisy classes via transition matrix (MVT) and using Top-$N$ predictions.}
\label{MVTtab:4}
\setlength{\tabcolsep}{3mm}
\begin{tabular}{lcccccc}
\toprule
& IN-A & IN-SK & IN-Val & IN-R & IN-V2 & IN-V \\
\midrule
Top-$N$ Pred. & 60.3 & 58.4 & 74.2 & 85.3 & 67.7 & 58.3 \\
MVT & 65.5 & 59.0 & 75.1 & 86.0 & 70.7 & 61.6 \\
\bottomrule
\end{tabular}
\end{table}

\begin{figure}[t]
\centering
\includegraphics[width=\linewidth]{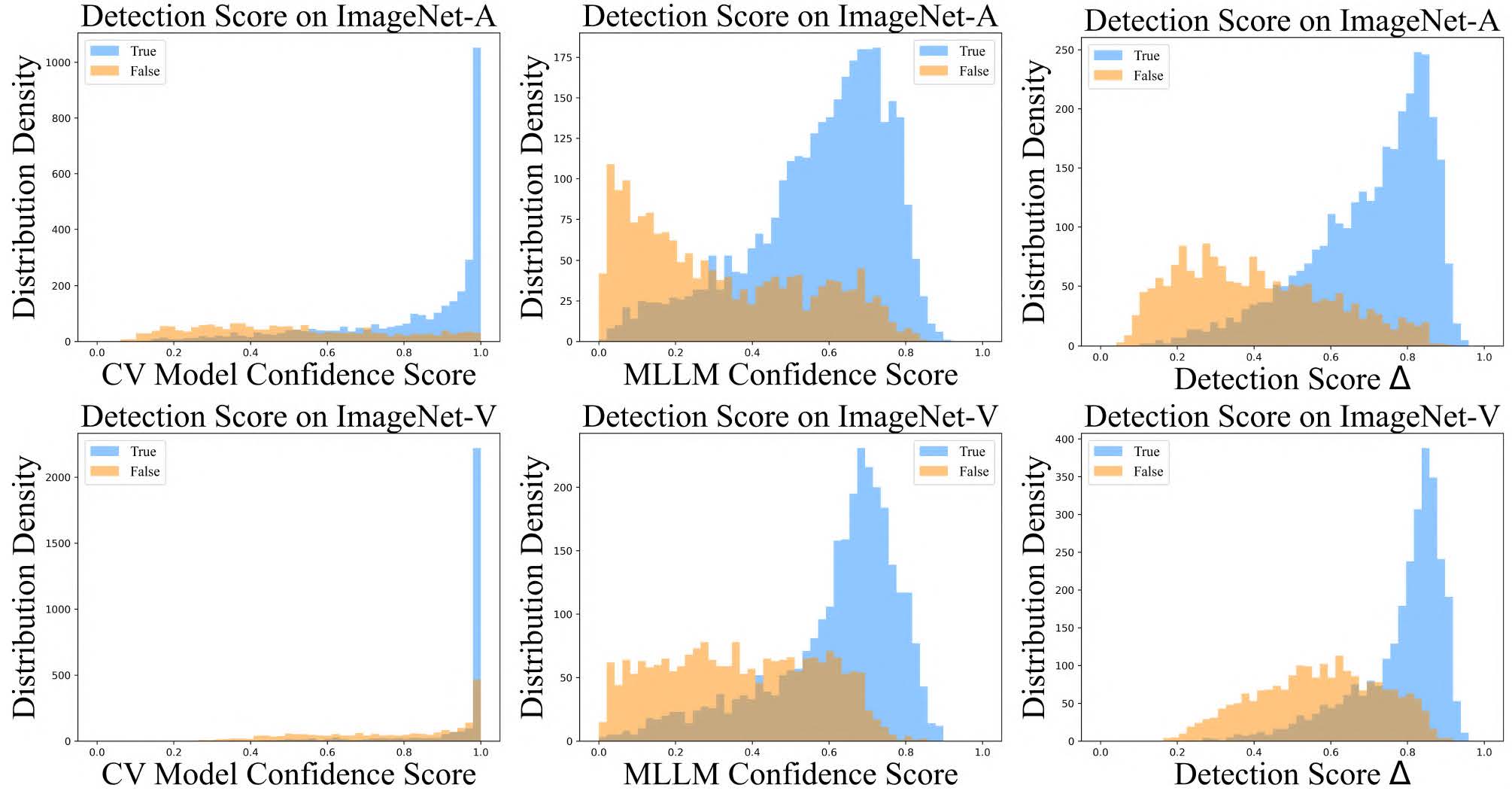}
\caption{Ablation study on detection score distribution.}
\label{MVTfig:detection_score}
\end{figure}

\textbf{Ablation Study on Detection Score.}
To analyze the proposed detection score on conducting diagnosing, we show the distribution of prediction confidence provided by the vision model, MLLM, and our detection score $\Delta$ in Figure~\ref{MVTfig:detection_score}. Based on the results, we can justify our design of $\Delta$: In the left column, we can see the confidence of correctly classified examples is very high, but the wrong ones show uniform distribution. Conversely, in the middle column, although MLLM poses slightly lower scores on correct ones, it significantly suppresses the confidence of wrong ones. As a result, we combine two scores to obtain $\Delta$, which can produce clearly separable distributions to benefit the diagnosing process. Unless specified, we set the threshold $\delta=0.6$ which works effectively in most scenarios.

\begin{table}[t]
\centering
\caption{Comparison of classification accuracy (\%) on 5 OOD datasets with Otter~\cite{li2023otter} and MMICL~\cite{zhao2023mmicl}. We compare the performance on CLIP ViT-L~\cite{radford2021learning} backbone.}
\setlength{\tabcolsep}{3mm}
\begin{tabular}{l|l|ccccc}
\toprule
MLLM & Method & IN-A & IN-R & IN-SK & IN-V & iWildCam \\
\midrule\midrule
None & CLIP & 69.3 & 86.6 & 59.4 & 51.8 & 13.4 \\

\midrule
\multirow{2}{*}{\shortstack{Otter}}
     & MVT & 64.1 & 85.2 & 59.5 & 51.9 & \bf16.2 \\
     & +FT & \bf73.5 & \bf88.7 & \bf60.0 & \bf55.7 & -  \\
\midrule
\multirow{2}{*}{\shortstack{MMICL}} 
     & MVT & 71.2 & 88.1 & 59.0 & 62.1 & \bf25.0 \\
     & +FT & \bf75.1 & \bf89.5 & \bf61.4 & \bf68.8 & - \\
\bottomrule
\end{tabular}
\label{MVTtab:mllm_model}
\end{table}

\textbf{Ablation Study on MLLM Backbone.}
To testify the effectiveness of MVT on different MLLM backbones, here we instantiate our method using Otter~\cite{li2023mimicit} and compare it to the previous realization on MIMIC~\cite{zhao2023mmicl}. The result is shown in Table~\ref{MVTtab:mllm_model}. We can see that both the implementation on MMICL and Otter show superior performance to the employed vision encoder backbone. Although the performance slightly differs between Otter and MMICL, which could be due to the model capacity and their training strategy, we can generally conclude that our MVT method is applicable to different MLLMs backbones with ICL and could further benefit from more sophisticated MLLMs in the future.

\subsection{Performance Analysis}
\label{MVT:performance_analysis}
Further, we conduct qualitative analysis to thoroughly validate the effectiveness of our MVT.

\begin{figure}[t]
    \centering
    \begin{minipage}{0.48\textwidth}
        \centering
        \includegraphics[width=\linewidth]{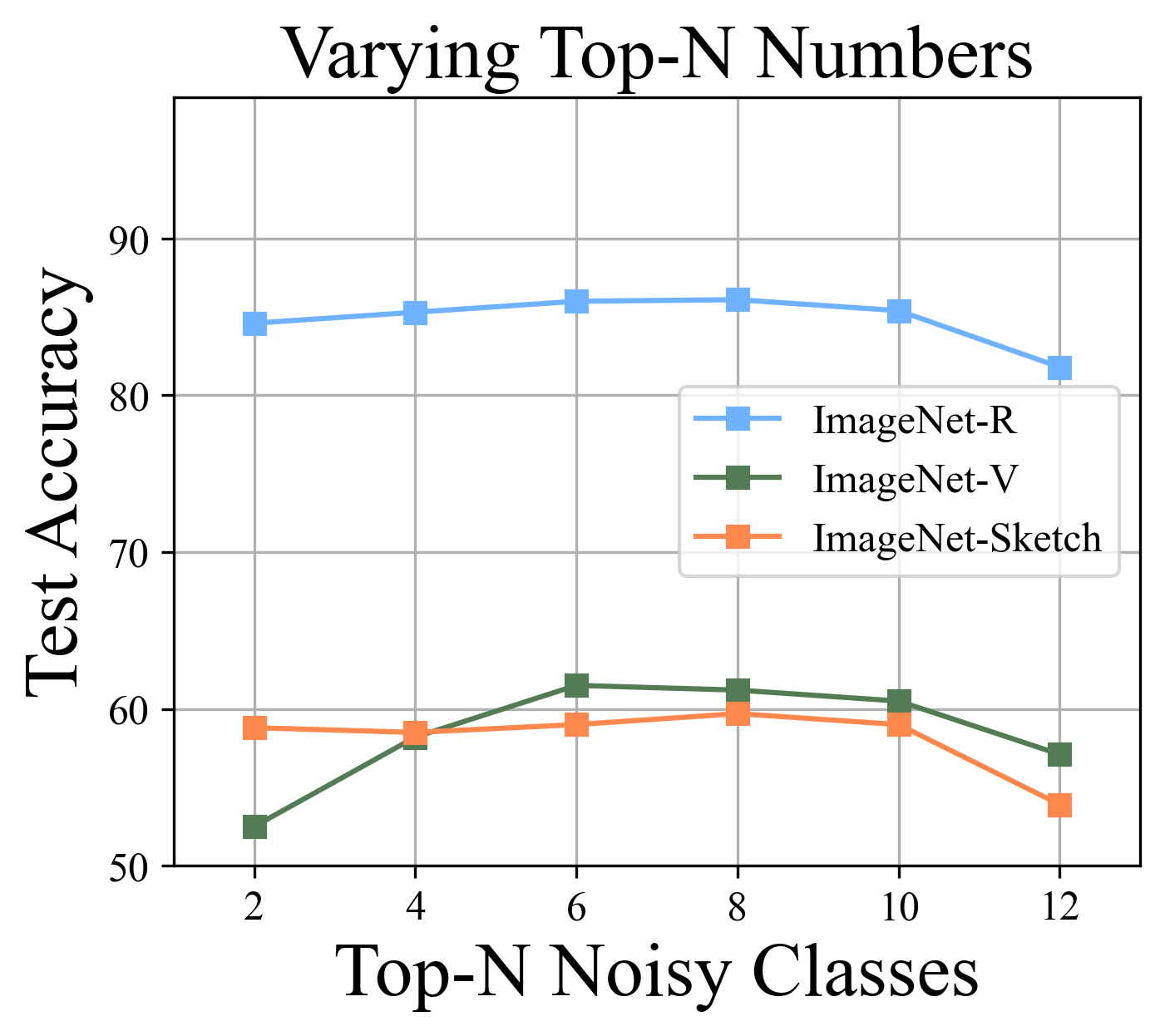}
    \end{minipage}
    \hfill 
    \begin{minipage}{0.48\textwidth}
        \centering
        \includegraphics[width=\linewidth]{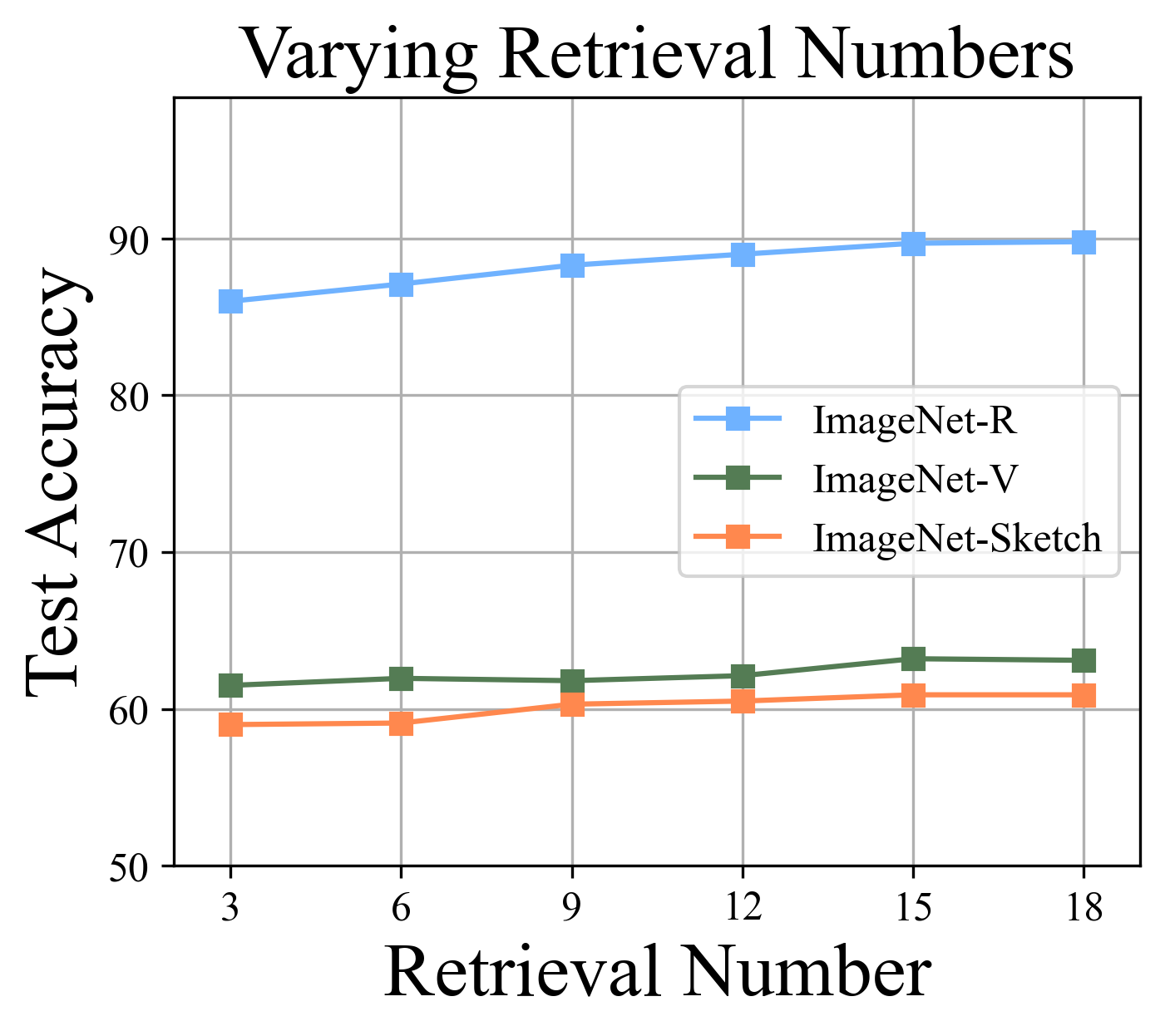}
    \end{minipage}
    
    \caption{Performance analysis: (left) varying the number of top-$N$ chosen noisy classes; (right) varying the number of retrieved exemplars.}
    \label{MVTfig:combined_analysis}
\end{figure}

\textbf{Choice of Top-$N$ Noisy Classes.}
To study how a varied number of chosen noisy classes could affect the performance of our method, we change the top-$N$ number from $2$ to $12$, and show the result on ImageNet-R, ImageNet-V, and ImageNet-Sketch datasets in Figure~\ref{MVTfig:combined_analysis} left. We find a common phenomenon that either too small or too large a number of $N$ could hurt the performance. This could be because that small $N$ would ignore too many potential ground-truth classes. In contrast, large $N$ includes too many choices that could interfere with the final prediction. Setting $N$ to $6$ could be an ideal choice for ImageNet-based datasets.

\textbf{Effect of Retrieval Numbers.}
In our experiments, we retrieve exemplars for $3$ times and average the predictions. To further investigate the effect of varied retrieval numbers, we change the number of retrievals from $3$ to $18$ and conduct experiments on the same OOD datasets as above. Specifically, we consider one positive and negative pair for a single DICL round as one retrieval. We repeat this process for $R$ times and ensemble the MLLM predictions through $\frac{1}{R}\sum_r^R\left[z_c[\text{True}]^r, z_c[\text{False}]^r\right]$. In this way, it is possible that MLLM predictions would be more accurate. The result is shown in Figure~\ref{MVTfig:combined_analysis} right. We observe that the performance steadily improves as the retrieval number increases, however, the performance gains vanish when the retrieval number becomes too large. Moreover, large retrieval numbers would multiply the computation cost. Therefore, it is suggested to set the number reasonably small.

\setlength{\intextsep}{8pt}
\setlength{\columnsep}{10pt}
\begin{wrapfigure}{r}{6cm}
    \centering
    \includegraphics[width=\linewidth]{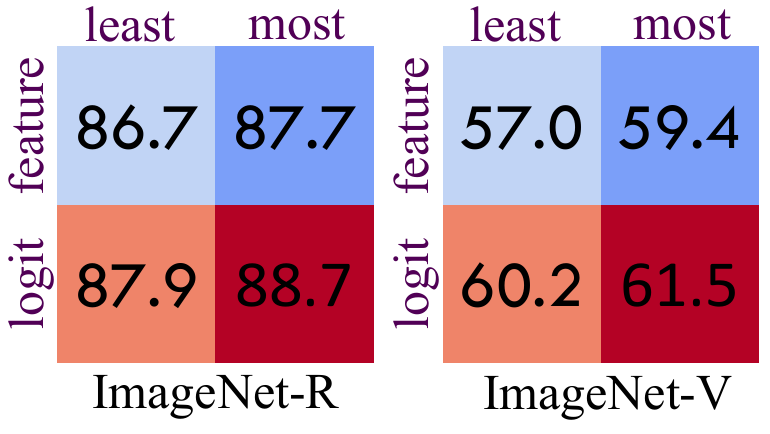}
    \caption{Performance analysis on different retrieval strategies.}
    \label{MVTfig:retrieve_strategy}
\end{wrapfigure}
\textbf{Performance of Different Retrieval Strategy.}
As shown by Alayrac et al.~\cite{alayrac2022flamingo}, Retrieval-based In-Context Example Selection (RICES) can significantly affect the ICL performance. Therefore, here we investigate its influence. Specifically, we propose two retrieval strategies, namely feature-based retrieval and logit-based retrieval. The former one is based on feature similarity and the latter one is based on the prediction logit. For each strategy, we conduct experiments on selecting the most similar examples and the least similar examples, which are denoted as ``most'' and ``least'', respectively. The results are shown in Figure~\ref{MVTfig:retrieve_strategy}. Apart from the intuitive finding that least-similar retrieval is inferior to selecting the most-similar one, we also observe that logit-based retrieval is more effective than feature-based one. We assume this is due to the image classification task is more related to logit value rather than feature similarity.

\begin{figure}[t]
    \centering
    \begin{minipage}{0.48\textwidth}
        \centering
        \includegraphics[width=\linewidth]{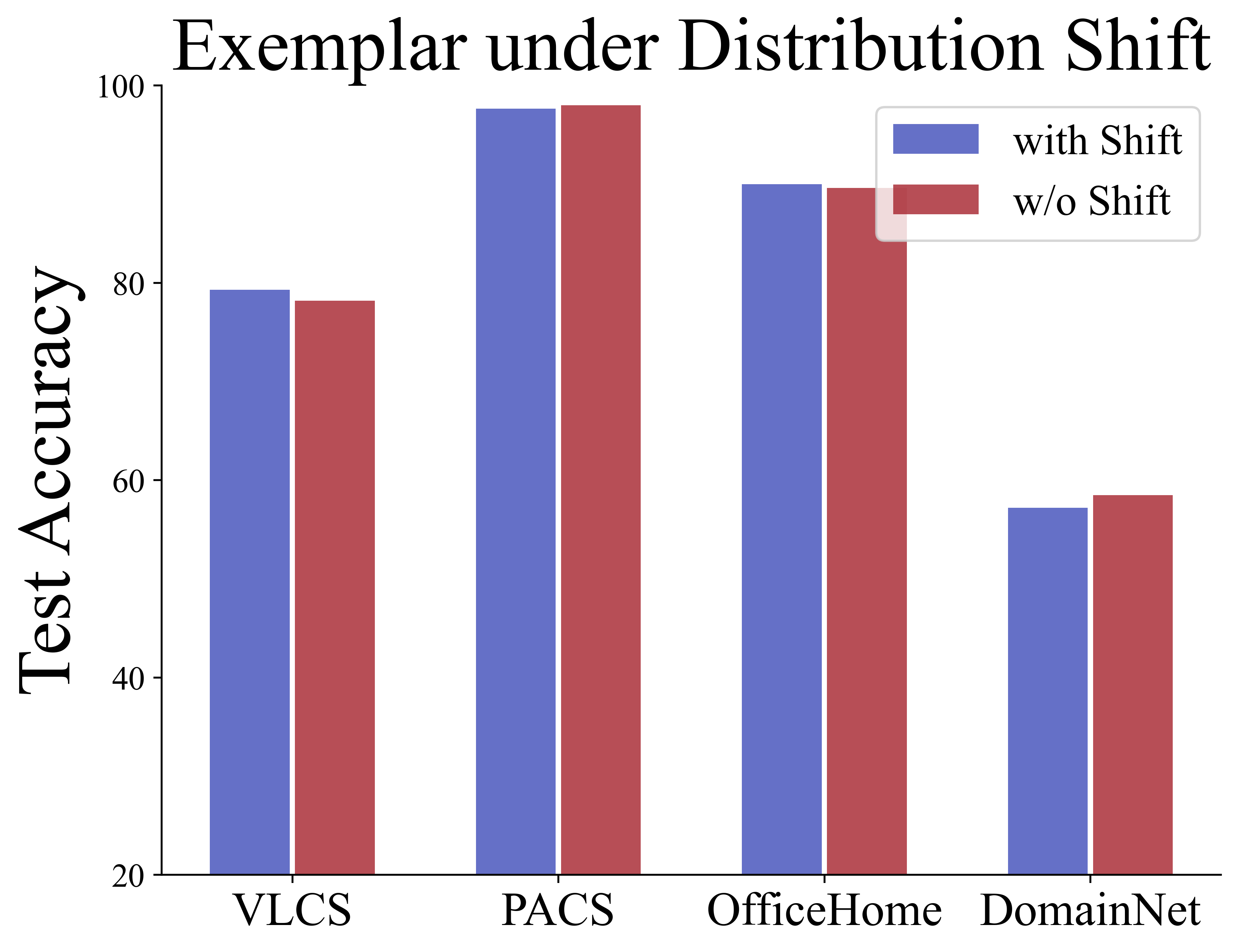}
    \end{minipage}
    \hfill 
    \begin{minipage}{0.48\textwidth}
        \centering
        \includegraphics[width=\linewidth]{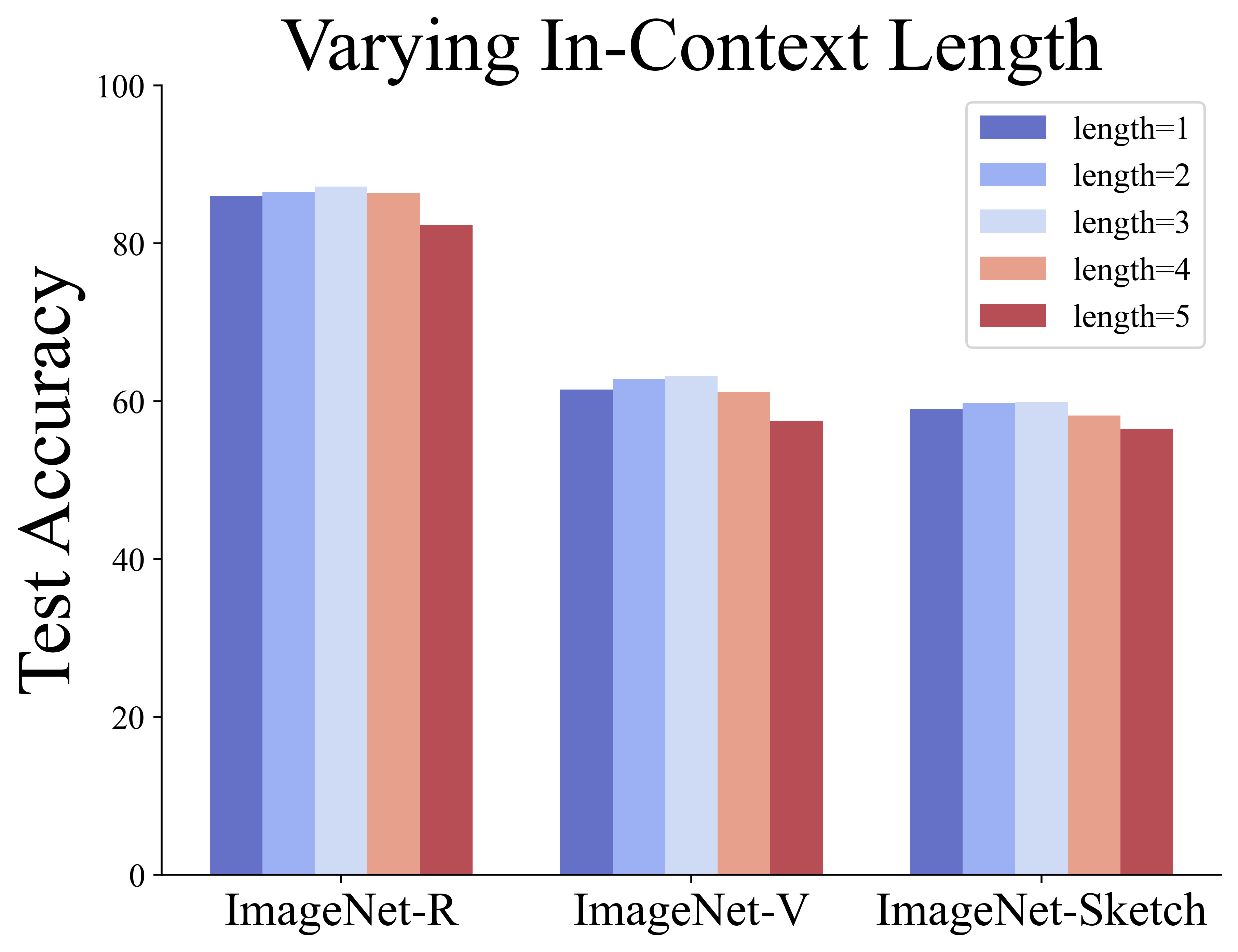}
    \end{minipage}
    \caption{Performance analysis of in-context exemplars: (left) under distribution shift; (right) varying the exemplar length.}
    \label{MVTfig:icl_analysis}
\end{figure}

\textbf{Effect of In-Context Exemplars with Distribution Shift.}
When the support set suffers from a distribution shift from the target OOD dataset, whether DICL can still perform robustly remains to be validated. Hence, we leave one domain out as our support set and leverage the rest domains as our target OOD dataset. In comparison, we choose a small hold-out data split as the support set which shares the same distribution as the OOD dataset. The results are shown in Figure~\ref{MVTfig:icl_analysis} left. Surprisingly, we find that the performance is not influenced by the distribution shift, which demonstrates that our MVT can still be effective when exemplars are retrieved from different distributions.

\textbf{Effect of Varying In-Context Length.}
Further, we analyze the effect of increasing exemplar length during inference. Particularly, we consider one positive and negative exemplar pair as length $1$. Here we vary the length from $1$ to $5$ and show the results in Figure~\ref{MVTfig:icl_analysis} right. We observe slight improvement when the length gradually increases which is consistent with the theoretical findings~\cite{xie2021explanation}. However, when the length is longer than $4$ the performance drops and the predictions of MLLM become unstable which could be other than ``True'' or ``False''. This might be due to the limited capacity of MLLMs on handling a certain amount of information, which is worth conducting studies on sophisticated MLLMs in the future.

\begin{figure}[t]
\centering
\includegraphics[width=\linewidth]{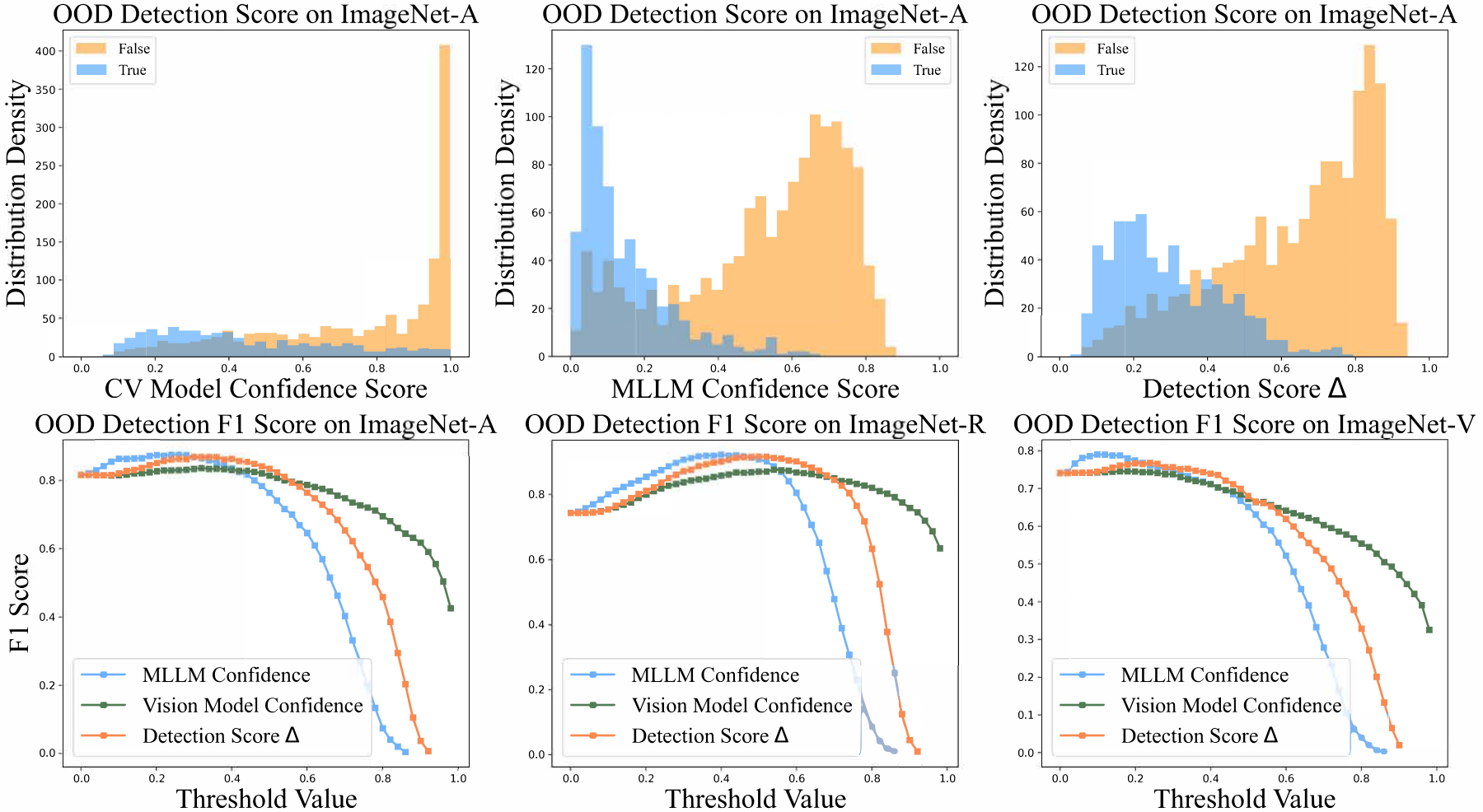}
\caption{OOD detection analysis. Upper: Detection score distribution on ImageNet-A; Lower: F1 scores of vision model confidence, MLLM diagnosing confidence, and our $\Delta$ score in ImageNet-A, ImageNet-R, and ImageNet-V.}
\label{MVTfig:ood_detection}
\end{figure}

\textbf{Performance of OOD Detection.}
At last, we consider a more challenging scenario where data from open classes could exist in the target dataset. Here we simulate this situation by choosing $60\%$ of the classes as closed classes and the rest are open classes. To detect such open-class data, i.e., OOD detection~\cite{hendrycks2016baseline, wang2024learning}\footnote{Note that OOD detection here is different from the previous setting: here we focus on detecting open-class data, and previous one focuses on detecting prediction errors.}, we use the vision model prediction confidence as a baseline and compare it with the MLLM diagnosing confidence as well as the detection score $\Delta$ in Equation~\eqref{MVTeq:detection_score}. The result is shown in Figure~\ref{MVTfig:ood_detection}. In the upper row, we observe the similar clearly distinguishable distributions using our score $\Delta$ as in Figure~\ref{MVTfig:detection_score}. In the lower row, we show the F1 score of each detection criterion under a threshold varied from $0$ to $1$ on three datasets. When a criterion produces confidence larger than the threshold, it would predict as close-class data, other as open-class ones. Based on the result, we find that MLLM achieves better detection performance when the threshold is small, but vision model confidence is relatively better when the threshold is large, i.e., MLLM can effectively detect open classes while vision models are better at recognizing close classes. However, an effective detection should have a reasonable threshold value that is neither too large nor too small and meanwhile has a high F1 score. Hence, by combining them together, our detection score $\Delta$ can achieve the best F1 score when the threshold is around the middle range.

\section{Conclusion}
In this Chapter, we propose a novel paradigm of fine-tuning vision models via leveraging MLLMs to improve visual robustness on downstream OOD tasks. Specifically, we effectively estimate a transition matrix to help find the most probable noisy classes. By using a positive exemplar and a negative exemplar retrieved based on the noisy classes, we can conduct DICL to rectify incorrect vision model predictions through two stages dubbed diagnosing and therapy. Thanks to the rectified predictions, the robustness of vision models can be further improved through fine-tuning. We conduct detailed theoretical analysis and extensive quantitative and qualitative experiments to justify the proposed method. Our framework can significantly reduce the cost of training vision models and provide insights into many visual recognition problems, such as OOD detection, OOD generalization, weakly-supervised learning, etc.

\chapter{Out-of-Modal Generalization}
\label{cha:OOM}
The world is understood from various modalities, such as appearance, sound, and language. Since each modality only partially represents objects in a certain meaning, leveraging additional ones is beneficial in both theory and practice. However, exploiting novel modalities normally requires cross-modal pairs corresponding to the same instance, which is extremely resource-consuming and sometimes even impossible, making knowledge exploration of novel modalities largely restricted. To seek practical multi-modal learning, here we study Out-of-Modal (OOM) Generalization as an initial attempt to generalize to an unknown modality without given instance-level modal correspondence. Specifically, we consider Semi-Supervised and Unsupervised scenarios of OOM Generalization, where the first has scarce correspondences and the second has none, and propose Connect\&Explore (COX) to solve these problems. COX first connects OOM data and known In-Modal (IM) data through a variational information bottleneck framework to extract shared information. Then, COX leverages the shared knowledge to create emergent correspondences, which is theoretically justified from an information-theoretic perspective. As a result, the label information on OOM data emerges along with the correspondences, which helps explore the OOM data with unknown knowledge, thus benefiting generalization results. We carefully evaluate the proposed COX method under various OOM generalization scenarios, verifying its effectiveness and extensibility.

\begin{figure}[t]
\centering
\includegraphics[width=0.5\textwidth]{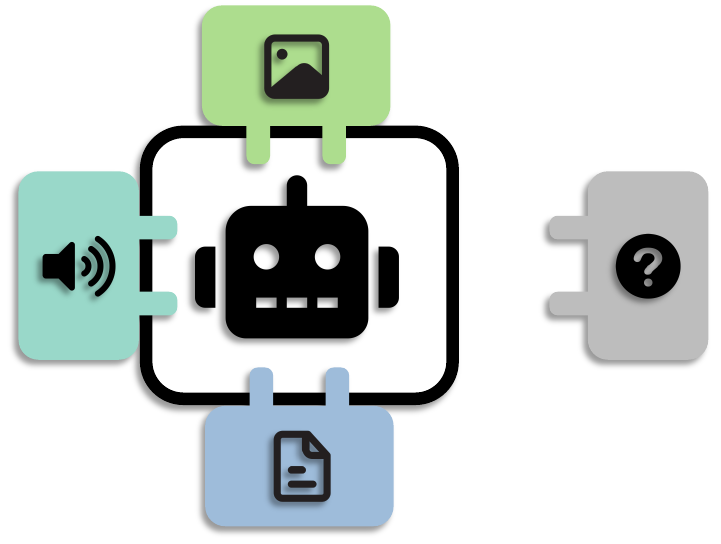}
\caption{AI is enhanced as more modalities are incorporated, so how can AI learns from novel modalities based on the ones it already know?}
\label{OOMfig:motivation}
\end{figure}

\section{Introduction}
\label{OOM:introduction}
To understand the world, we use various data \textit{modalities}, such as image \cite{he2016deep, he2017mask, ren2015faster} and text \cite{devlin2018bert, vaswani2017attention}. Each modality describes objects through a certain physical perspective, thus contributing to understanding objects. Therefore, \emph{multi-modal learning}~(MML) \cite{alayrac2022flamingo, huang2024machine, ngiam2011multimodal, radford2021learning, socher2013zero, wang2024noisegpt, liu2025towards} which learns from multiple modality data has been a core research topic in AI. Thanks to the utilization of various modalities, the learning performance has shown benefits on various tasks compared to uni-modal learning \cite{huang2021makes, lu2024theory, radford2021learning, sun2020tcgm}, such as cross-modal retrieval \cite{yasunaga2023retrieval, zhang2021cross, zhen2019deep}, human-computer interaction \cite{pantic2003toward, rahman2022explainable}, and robotics \cite{jiang2023vima, yu2023fusing}.

However, existing states of the art are not satisfactory, and emerging modalities need to be leveraged effectively just like the relatively new data modalities of geomagnetic fields \cite{hashimoto1926origin}, sound waves \cite{harley2003bottlenose}, and electromagnetic waves \cite{weinstein1988electromagnetic}. Therefore, emerging technologies have constantly leveraged new sensors to enhance their performance. For example, Embodied AIs \cite{savva2019habitat} already possess abilities like 3D vision and language, but they are still exploring novel skills, such as tactile and bio-sensing. Since it is hard to leverage such uncommon and inexperienced skills in practice, adapting the knowledge from common modalities to understand the novel ones could be beneficial, as shown in Figure \ref{OOMfig:motivation}. In practice, most existing MML investigations \cite{radford2021learning, girdhar2023imagebind, wang2024freebind, zhu2023languagebind} require \textit{instance-level modal correspondence}, i.e., multi-modal data are paired with the same instance, which is often hard to satisfy in real-world scenarios when facing novel modalities \cite{liang2023multimodal, liang2021cross, sun2020tcgm, xia2024achieving}. For a robotic example, some modalities are common and easy to acquire, \textit{e.g.}, vision and language. However, others like tactility need special sensors to resample from the same objects seen or spoken. Unfortunately, the resample could no longer be accessible in practice. As a result, the new modalities usually have incomplete or even no correspondence, which could seriously block the knowledge interaction across modalities and hinder the benefits brought by MML. Hence, a question naturally occurs: \textit{Do we really need instance-level modal correspondence to explore novel modalities?}

This Chapter studies a practical yet unexplored problem named \textit{Out-of-Modal (OOM) Generalization}. Particularly, given several modalities, \textit{i.e.,} In-Modal (IM) data, the goal is to generalize to an unknown modality without or sometimes only with scarce correspondence. Such a setting implies the real-world utilization of novel modalities: Even though human perception is limited to certain modalities, \textit{e.g.}, touch, sight, sound, and smell, we can still understand unperceivable ones such as magnetism by utilizing inherently-possessed senses, \textit{e.g.}, feel the force when pulling two magnets together; or see the magnetic field by observing the alignment of iron filings around a magnet. 

Based on this insight, we utilize IM perceptors that contain prior knowledge to encode known IM data, which can be implemented using existing MML models \cite{radford2021learning, girdhar2023imagebind, zhu2023languagebind, wang2024freebind}, and an OOM learner which learns novel modalities without any prior knowledge. By analyzing the interactions between latent features, we show theoretically and empirically that the OOM learner can be trained to gradually discover the OOM knowledge, as shown in Figure \ref{OOMfig:abstraction}. First, we consider \emph{semi-supervised OOM generalization} where few correspondences are given. Based on the correspondence, we can capture the prior probability distribution and learn mappings that connect OOM data and IM data. Through an information-theoretic perspective, we propose \emph{Connect\&Explore} (COX), which encourages the agreement on mappings across modalities, further sharing the cross-modal knowledge and exploring OOM knowledge. Then, we extend COX to an \emph{unsupervised OOM generalization} scenario where there is no instance-level correspondence at all. To tackle such a challenge, we enhance the OOM-IM connections by maximizing cross-modal interaction. First, we select data pairs from cross-modal mappings according to feature similarity. By assuming that the data pairs closing to OOM mappings can be considered as correspondence, we can create emerging correspondence and solve the unsupervised case via the semi-supervised solution. To validate the proposed COX, we carefully design experiments using various multi-modal datasets to validate its effectiveness. Moreover, we provide extensive analyses to understand the OOM problem and inspire future research. To sum up, our contributions are threefold:
\begin{itemize}
    \item We discover a novel and practical problem named OOM Generalization, which aims to explore a novel modality using the knowledge from known modalities.
    \item We consider two typical situations: Semi-Supervised OOM generalization and Unsupervised OOM generalization, and propose a Connect\&Explore framework to tackle both problems from an information-theoretic perspective.
    \item We conduct extensive experiments to tackle the OOM generalization on various datasets and provide intuitive insights to help inspire future research.
\end{itemize}

\section{Related Work}
\label{OOM:related_work}

\textbf{Modality Generalization}~\cite{liu2024towards} generally focuses on leveraging the knowledge from some modalities and generalizing to another one. Existing studies are conducted in different settings and with various tasks. Cross-modal fine-tuning mimics transfer learning by adapting the distribution of IM data to OOM data using the same model. \cite{shen2023cross} proposed to conduct distribution alignment to achieve this goal which requires both pre-trained knowledge and labeled target modality data. Based on a similar problem setting, \cite{cai2024enhancing} designed a gradual modality generation scheme that selects the top-$k$ active feature patches from target modalities, and replaces them with source modalities patches. Such a progressive strategy can align target modal data to ensure generalization. Cross-Modal Generalization uses separate encoders and focus on generalizing to a different modality data from the same instance. \cite{liang2021cross} used meta-learning to align OOM data to IM space and generalize to OOM tasks dynamically. \cite{xia2024achieving} studied a different setting where IM and OOM data are both known during training. Then, a unified representation space is learned to help with the downstream generalization of OOM data. Some other studies consider generalization when all modalities are available, \cite{ma2019unpaired} studied cross-modal generalization without paired data, \cite{wang2023visual} applied the information bottleneck to CLIP training, \cite{fang2024dynamic} conducted multi-modal fusion under limited clinical data, and \cite{dong2023simmmdg} considered domain generalization with fully-paired multi-modal data. A recent study MML without Labeled Multi-Modal Data \cite{liang2023multimodal} proposed a different setting where both IM and OOM data have labels, but they are not paired. Instead, additional unlabeled paired multi-modal data is given for learning the interaction between modalities. Moreover, \cite{xue2022modality} understood the interactions and applied it to knowledge distillation. Except for cross-modal fine-tuning which follows transfer learning, existing MML works mostly require instance-level correspondence. This work proposes OOM Generalization, where there is no correspondence and the OOM knowledge is barely provided. The comparison of related works is shown in Table \ref{OOMtab:setting_comparison}.

\begin{table}[t]
\centering
\setlength{\tabcolsep}{1.2mm}
\scriptsize
\caption{A comparison of different MML problems and their corresponding settings.}
\begin{tabular}{l|l|ccc}
\toprule[1.2pt]
\textbf{Problem}                            & \textbf{References}           & \textbf{IM Knowledge}       & \textbf{OOM Knowledge}      & \textbf{Correspondence}    \\ \midrule
Cross-Modal Fine-Tuning                     & \cite{shen2023cross, cai2024enhancing}  & pre-trained \& labeled     & labeled  & \XSolidBrush  \\ \midrule
\multirow{2}{*}{Cross-Modal Generalization} & \cite{liang2021cross}          & pre-trained \& labeled      & pre-trained                & \CheckmarkBold\\
 & \cite{xia2024achieving}  & pre-trained \& labeled      & pre-trained \& labeled     & \CheckmarkBold\\ \midrule
MML w/o labeled MM Data            & \cite{liang2023multimodal}     & partially labeled      & partially labels & \CheckmarkBold\\ \midrule
\multirow{2}{*}{OOM Generalization}         & Section \ref{OOM:semi}           & pre-trained \& labeled      & scarcely labeled           & A few \\
& Section \ref{OOM:unsup}   & pre-trained \& labeled      & \XSolidBrush  & \XSolidBrush  \\ 
\bottomrule[1.2pt]
\end{tabular}
\label{OOMtab:setting_comparison}
\end{table}

\textbf{Modality Binding} aims to learn a joint embedding space across different modalities. Contrastive Language-Image Pre-training CLIP \cite{radford2021learning} is the first work that aligns image with language data. Then, ImageBind \cite{girdhar2023imagebind} proposed to use vision modalities to bind various modalities into the same representation space. Further, LanguageBind \cite{zhu2023languagebind} proposed using language as an alternative solution, which binds various modalities similarly. Recently, FreeBind \cite{wang2024freebind} extended the existing unified space into an additional expert space. Specifically, two types of binding were considered, namely space displacement bond and space combination bind. Since modality binding often requires a large amount of data with correspondence, the selected modalities are often quite common. Therefore, the OOM generalization problem can take advantage of the development of modality binding by leveraging the encoders as our IM perceptors to learn novel modalities.

\section{OOM Generalization}
\label{OOM:methodology}

In this section, we first formalize the OOM generalization setting. Then, we demonstrate the proposed method. Further, we consider a Semi-Supervised case where a few correspondences are available and an Unsupervised scenario where there is no correspondence, showing that the proposed method can successfully tackle both settings and effectively leverage unpaired OOM data.

\subsection{Problem Setting}
In OOM generalization, we are given a set of known modalities $\{\mathcal{M}^{\mathrm{I}}_1,\ldots, \mathcal{M}^{\mathrm{I}}_K\}$ where $\mathcal{M}^{\mathrm{I}}_{k\in\{1,\ldots, K\}}=\{(x_{k, i}^{\mathrm{I}}, y_{k,i}^{\mathrm{I}})_{i=1}^N\in\mathcal{X}\times\mathcal{Y}\}$ is composed of $N$ number of labeled IM examples with its subscript $i$ denoting the correspondence across different modalities. Moreover, we have an unknown modality $\mathcal{M}^{\mathrm{O}}=\{(x_j^{\mathrm{O}})_{j=1}^M\}$ containing $M$ unlabeled OOM examples. In some cases, it is possible to obtain few correspondences with IM data, then our OOM data could be $\mathcal{M}^{\mathrm{O}}=\{(x_i^{\mathrm{O}}, y_i^{\mathrm{O}})\}_{i=1}^L\cup\{(x_j^{\mathrm{O}})\}_{j=L+1}^M$, where $L\ll M$ and the subscript $i$ traces the corresponding IM data instance and label.

\begin{figure}[t]
\centering
\includegraphics[width=0.8\textwidth]{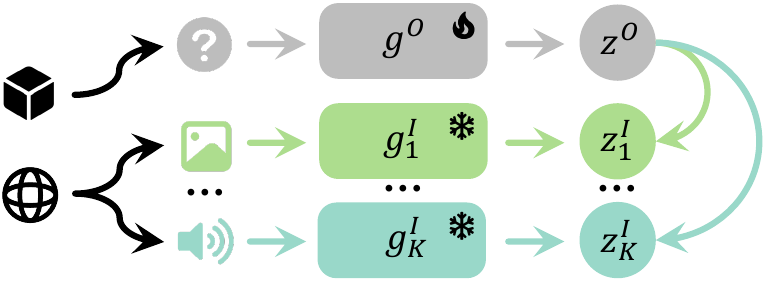}
\caption{Learning framework of our OOM generalization.}
\label{OOMfig:abstraction}
\end{figure}
To tackle OOM generalization, we propose a learning framework as shown in Figure \ref{OOMfig:abstraction}. Particularly, we use a set of IM perceptors $\{g^{\mathrm{I}}_1, \ldots, g^{\mathrm{I}}_K\}$ to perceive IM data, which can be realized by many existing modality-binding models, such as ImageBind \cite{girdhar2023imagebind} and LanguageBind \cite{zhu2023languagebind}. Then, the features of IM data are obtained via $z^{\mathrm{I}}_k = g^{\mathrm{I}}_k(x^{\mathrm{I}}_k)$. Moreover, we use an OOM learner $g^{\mathrm{O}}$ to learn features $z^{\mathrm{O}}$ from OOM data through $z^{\mathrm{O}}=g^{\mathrm{O}}(x^{\mathrm{O}})$. Our goal is to effectively generalize to OOM data by exploring the relationships between the OOM feature $z^{\mathrm{O}}$ and IM features $\{z^{\mathrm{I}}_k\}_{k=1}^K$. Note that we only focus on the generalization performance of OOM data, the improvement of learning IM data is not the goal of this Chapter. Therefore, we freeze the parameters of all IM perceptors and only train the OOM learner during experiments. On top of the above models, we further define classifiers $h^{\mathrm{O}}(x^{\mathrm{O}}):=h^{\mathrm{O}}(x^{\mathrm{O}}; g^{\mathrm{O}})$ and $h_k(x_k^{\mathrm{I}}):=h_k(x_k^{\mathrm{I}}; g_k^{\mathrm{I}})$ that make predictions.

\subsection{Methodology: Connect\&Explore (COX)}
\label{OOM:cox}
Here we elucidate the proposed method based on the interactive relationship between modalities \cite{liang2023multimodal, williams2010nonnegative}. Specifically, the total information of two modalities under a certain task is decomposed into (1) \textit{commonality}\footnote{It is originally termed ``redundancy'' which is negative. However, such property is quite positive for tackling our problem, and hence we rename it ``commonality''.} which indicates common attributes across modalities, (2) \textit{uniqueness} that is only presented in each modality, and (3) \textit{synergy} denoting the emerging information when modalities are presented together. Note that we do not consider (3) in this Chapter as our goal is generalizing to OOM data.

To generalize to an unknown modality based on common ones, we aim to extract the commonality that can help partially comprehend OOM data based on IM data. Then, we model the posterior distribution of OOM data by selecting anchor points with minimum uniqueness. To this end, the OOM generalization can be successfully established. The proposed COX method comprises two steps: (1) learning connections by mapping IM data to OOM data to extract commonality, and (2) exploring high uniqueness OOM data by matching their posterior to high-commonality OOM data.

\textbf{Connection through Commonality} aims to capture common knowledge across modalities using generative models \cite{lu2024theory}. Here we follow the variational information bottleneck (VIB) framework \cite{alemi2016deep} to achieve this goal. We assume that given IM data $X^{\mathrm{I}}$ and OOM data $X^{\mathrm{O}}$, the latent variable $V$ extracted from $X^{\mathrm{I}}$\footnote{Note that the latent variable $V$ here is different from the feature representation $z^{\mathrm{I}}$ and $z^{\mathrm{O}}$.}, and label $Y$, the joint distribution is factorized as
\begin{equation}
    p(X^{\mathrm{I}}, X^{\mathrm{O}}, V, Y) = p(V, Y|X^{\mathrm{O}}, X^{\mathrm{I}})p(X^{\mathrm{O}}|X^{\mathrm{I}})P(X^{\mathrm{I}}),
\end{equation}
where we assume $p(V, Y|X^{\mathrm{O}}, X^{\mathrm{I}})=p(V|X^{\mathrm{I}})p(Y|X^{\mathrm{I}})$, corresponding to the Markov chains $V\leftrightarrow X^{\mathrm{I}}\leftrightarrow X^{\mathrm{O}}$ and $X^{\mathrm{I}}\leftrightarrow Y \not\leftrightarrow X^{\mathrm{O}}$. Such an assumption means that $V$ is not related to $X^{\mathrm{O}}$ \cite{alemi2016deep} and the given label $Y$ is not directly connected to $X^{\mathrm{O}}$ under our OOM setting. Intuitively, given an IM datum, i.e., dog image, it is sufficient to infer the label ``dog'', and the same for inferring from an unknown OOM datum, i.e., dog bark. Thus, in common multi-modal settings, the label prediction using IM information dog image is not further conditioned on OOM knowledge dog bark, because here the OOM knowledge is redundant when IM data is given.

Our goal is to extract valuable knowledge from IM data to leverage OOM data by maximizing the information commonality \cite{liang2023multimodal, williams2010nonnegative}:
\begin{equation}
    \max I(X^{\mathrm{O}}; X^{\mathrm{I}}; Y)=I(X^{\mathrm{O}}; X^{\mathrm{I}}) - I(X^{\mathrm{O}}; X^{\mathrm{I}}|Y),
\end{equation}
where $I(X^{\mathrm{O}}; X^{\mathrm{I}}; Y)$ denotes the mutual information between $X^{\mathrm{O}}$ and $X^{\mathrm{I}}$ regarding the task $Y$, i.e., the label; and $I(X^{\mathrm{O}}; X^{\mathrm{I}}|Y)$ indicates the conditional mutual information irrelevant to $Y$. We start with the first term:
\begin{align}
    \!\!\!I(X^{\mathrm{O}}; X^{\mathrm{I}})&\!=\!\int dx^{\mathrm{O}}\!dx^{\mathrm{I}} p(x^{\mathrm{O}}\!, x^{\mathrm{I}})\log\frac{p(x^{\mathrm{O}}\!, x^{\mathrm{I}})}{p(x^{\mathrm{O}})p(x^{\mathrm{I}})}\!=\!\int dx^{\mathrm{O}}\!dx^{\mathrm{I}} p(x^{\mathrm{O}}\!, x^{\mathrm{I}})\log\frac{p(x^{\mathrm{O}}|x^{\mathrm{I}})}{p(x^{\mathrm{O}})},
\end{align}
where $p(x^{\mathrm{O}}|x^{\mathrm{I}})=\int dv p(x^{\mathrm{O}}, v|x^{\mathrm{I}})=\int dv p(x^{\mathrm{O}}|v)p(v|x^{\mathrm{I}})$ can be approximated via a decoder $q(x^{\mathrm{O}}|v)$. Since the Kullback Leibler (KL) divergence is always non-negative, we have $\text{KL}[p(X^{\mathrm{O}}|V)\parallel q(X^{\mathrm{O}}|V)]\ge 0\Rightarrow \int dx^{\mathrm{O}} p(x^{\mathrm{O}}|v)\log p(x^{\mathrm{O}}|v)\ge \int dx^{\mathrm{O}} p(x^{\mathrm{O}}|v)\log q(x^{\mathrm{O}}|v)$, and leveraging Jensen's inequity, we can have
\begin{align}
    \!\!\!I(X^{\mathrm{O}}; X^{\mathrm{I}})&\!\ge\!\int dx^{\mathrm{O}}\!dx^{\mathrm{I}} p(x^{\mathrm{O}}\!, x^{\mathrm{I}})\log\frac{\int dv q(x^{\mathrm{O}}|v)p(v|x^{\mathrm{I}})}{p(x^{\mathrm{O}})}\\
    &\!\ge\!\int dx^{\mathrm{O}}\!dx^{\mathrm{I}}\!dv p(x^{\mathrm{O}}\!, x^{\mathrm{I}})\log q(x^{\mathrm{O}}|v)p(v|x^{\mathrm{I}}) + H(X^{\mathrm{O}}),
\end{align}
where the last term is independent of our optimization process. Further, we rewrite:
\begin{equation}
    p(x^{\mathrm{O}}\!, x^{\mathrm{I}})=\int dv p(x^{\mathrm{O}}\!, x^{\mathrm{I}}, v)=\int dv p(x^{\mathrm{I}})p(x^{\mathrm{O}}|x^{\mathrm{I}})p(v|x^{\mathrm{I}}).
\end{equation}
Then, we have the following lower bound:
\begin{align}
    \!\!\!I(X^{\mathrm{O}}; X^{\mathrm{I}})&\!\ge\!\int dx^{\mathrm{O}}\!dx^{\mathrm{I}}\!dv p(x^{\mathrm{I}})p(x^{\mathrm{O}}|x^{\mathrm{I}})p(v|x^{\mathrm{I}})\log q(x^{\mathrm{O}}|v)p(v|x^{\mathrm{I}}),
    \label{OOMeq:first_lowerbound}
\end{align}
which is realized by sampling from the joint data distribution, the latent variable from our encoder $p(v|x^{\mathrm{I}})$, and the tractable variational approximation $q(x^{\mathrm{O}}|v)$.

Similarly, we can upper-bound the second term $I(X^{\mathrm{O}}; X^{\mathrm{I}}|Y)$:
\begin{align}
    \!\!\!\!I(X^{\mathrm{O}}\!; X^{\mathrm{I}}|Y)\!&\le\!\int\!\! dx^{\mathrm{O}}\!dx^{\mathrm{I}}\!dy p(x^{\mathrm{O}}, x^{\mathrm{I}}, y)\log p(y|x^{\mathrm{I}})p(x^{\mathrm{O}}|x^{\mathrm{I}})p(x^{\mathrm{I}})\! -\! \log h^{\mathrm{O}}(y|x^{\mathrm{O}}),
    \label{OOMeq:second_upperbound}
\end{align}
where $h^{\mathrm{O}}(y|x^{\mathrm{O}})$ is our classifier model for predicting OOM data. To this end, we can lower-bound our objective by combining Equations.~\eqref{OOMeq:first_lowerbound} and \eqref{OOMeq:second_upperbound}:
\begin{equation}
\begin{split}
    &I(X^{\mathrm{O}}; X^{\mathrm{I}}; Y)\ge\int\! dx^{\mathrm{O}}\!dx^{\mathrm{I}}\!dv p(x^{\mathrm{I}})p(x^{\mathrm{O}}|x^{\mathrm{I}})p(v|x^{\mathrm{I}})\log q(x^{\mathrm{O}}|v)p(v|x^{\mathrm{I}}) \\ 
    &-\int\! dx^{\mathrm{O}}\!dx^{\mathrm{I}}\!dy p(x^{\mathrm{O}}, x^{\mathrm{I}}, y)\log p(y|x^{\mathrm{I}})p(x^{\mathrm{O}}|x^{\mathrm{I}})p(x^{\mathrm{I}})\! +\! \log h^{\mathrm{O}}(y|x^{\mathrm{O}})=\mathcal{L}_\mathrm{con}.
    \label{OOMeq:final_lowerbound} 
\end{split}  
\end{equation}
The above lower bound contains two parts: (1) OOM data reconstruction where we reconstruct $X^{\mathrm{O}}$ using the latent $V$ and (2) OOM data label prediction where we model the label distribution $Y$. In practice, we can approximate $p(x^{\mathrm{O}}, x^{\mathrm{I}}, y)$ using empirical samples from IM and OOM data. Moreover, we use encoder $p(v|x^{\mathrm{I}})$ without any prior assumptions because we can leverage the feature distribution from the pre-trained IM perceptors. Additionally, a classifier $h(y|x^{\mathrm{O}})$ is optimized to categorize OOM data based on given labels. Empirically, we can minimize
\begin{equation}
    \mathcal{L}_{\text{con}}:=\frac{1}{M}\sum_{i=1}^M \|x_i^{\mathrm{O}}-q(x_i^{\mathrm{O}}|v_i)p(v_i|x_i^{\mathrm{I}})\|_2^2-\log h^{\mathrm{O}}(y_i|x_i^{\mathrm{O}}),
\end{equation}
where we use the reconstruction error $\|\cdot\|_2^2$\footnote{Though training generative models in input space is computationally inefficient, we propose to connect modalities in the feature space in experiments. Therefore, the raw data $x$ is replaced by latent feature $z$.} to realize the log-likelihood $q(x^{\mathrm{O}}|v)p(v|x^{\mathrm{I}})$, as similarly done by Kingma and Welling~\cite{kingma2013auto}. After building the connections, we can ensure the task-relevant information shared across modalities is learned, which helps partially understand OOM data regarding its commonality. However, note that the second term in Equation~\eqref{OOMeq:final_lowerbound} is not fully leveraged which contains $p(y|x^{\mathrm{I}})$ modeled by the IM perceptors. Take a step further, we can obtain $-\int\! dx^{\mathrm{O}}\!dx^{\mathrm{I}}\!dy p(x^{\mathrm{O}}, x^{\mathrm{I}}, y)\log\frac{p(y|x^{\mathrm{I}})p(x^{\mathrm{O}}|x^{\mathrm{I}})p(x^{\mathrm{I}})}{h^{\mathrm{O}}(y|x^{\mathrm{O}})}$. Since $p(x^{\mathrm{O}}|x^{\mathrm{I}})p(x^{\mathrm{I}})$ is fixed in label prediction, we can derive $-\text{KL}(p(y|x^{\mathrm{I}})\parallel h^{\mathrm{O}}(y|x^{\mathrm{O}}))$ which implies that the label information related IM data can be harnessed to explore commonality. Next, we demonstrate how the commonality helps OOM generalization, and provide a solution to explore uniqueness.

\begin{wrapfigure}{r}{0.55\textwidth}
\centering
\includegraphics[width=0.54\textwidth]{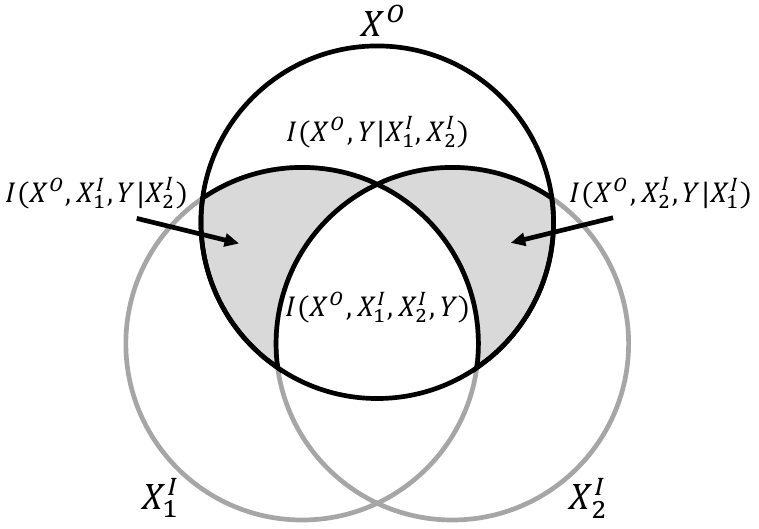}
\caption{Decomposition of $I(X^{\mathrm{O}}, Y)$.}
\label{OOMfig:MI_decomposition}
\end{wrapfigure}

\textbf{Exploration of Uniqueness} can be achieved via selecting and exploring the OOM data with high uniqueness. To identify these data, we can leverage the agreement and disagreement achieved by the optimal classifiers from various IM data. Our final goal is to optimize via
\begin{align}
    \min_{h^{\mathrm{O}}}\text{KL}(h^{\mathrm{O}}(y|x_d^{\mathrm{O}})\parallel h^{\mathrm{O}}(y|x_a^{\mathrm{O}})), \text{where } x_d^{\mathrm{O}}\in\mathcal{D}, x_a^{\mathrm{O}}\in\mathcal{A},
\end{align}
in which $h_1^*$ and $h_2^*$ denote the optimal classifiers found in two IM data $x_1^{\mathrm{I}}$ and $x_2^{\mathrm{I}}$, respectively, and $x_d^{\mathrm{O}}$ and $x_a^{\mathrm{O}}$ are selected from OOM data with modality disagreement $\mathcal{D}:=\{x^{\mathrm{O}}:h_1^*(x^{\mathrm{O}})\ne h_2^*(x^{\mathrm{O}})\}$ and agreement $\mathcal{A}:=\{x^{\mathrm{O}}:h_1^*(x^{\mathrm{O}})= h_2^*(x^{\mathrm{O}})\}$, respectively. Here we use two in-modalities for simplicity, but the conclusion can be extended to multiple modalities. Moreover, the data with agreement is considered anchor points that guide the exploration of those with disagreement. This objective aims to match the posterior of OOM data with uniqueness $h^{\mathrm{O}}(y|x_d^{\mathrm{O}})$ to the one of anchor points $h^{\mathrm{O}}(y|x_a^{\mathrm{O}})$. To justify this, we first define modality disagreement:
\begin{definition}[Modality disagreement]
Given $X_1, X_2$ and target $Y$, as well as their corresponding optimal classifiers $h_1^*$ and $h_2^*$, their modality disagreement is defined as $\alpha(h_1^*, h_2^*)=\mathbb{E}_{p(x_1, x_2)}[d(h_1^*, h_2^*)]$ where $d:\mathcal{Y}\times\mathcal{Y}\rightarrow\mathbb{R^+}$ is a distance function in the label space scoring the disagreement between $h_1^*$ and $h_2^*$.
\end{definition}
\begin{theorem}
    Given two Bayes' optimal classifiers $h_1^*$ and $h_2^*$ from two in-modalities, under relaxed triangle inequality,  inverse Lipschitz condition, and classifier optimality assumptions \cite{sridharan2008information}, the modalities disagreement is upper-bounded by
    \begin{equation}
        \alpha(h_1^*, h_2^*)\le I(X^{\mathrm{O}}, X_2^{\mathrm{I}}, Y|X_1^{\mathrm{I}})+I(X^{\mathrm{O}}, X_1^{\mathrm{I}}, Y|X_2^{\mathrm{I}})+2I(X^{\mathrm{O}}, Y|X_1^{\mathrm{I}}, X_2^{\mathrm{I}}).
    \end{equation}
    \label{OOM:theorem}
\end{theorem}
Finally, based on the decomposition of the task-related mutual information $X^{\mathrm{O}}$:
\begin{equation}
    I(X^{\mathrm{O}}, Y)=I(X^{\mathrm{O}}, X_2^{\mathrm{I}}, Y|X_1^{\mathrm{I}})+I(X^{\mathrm{O}}, X_1^{\mathrm{I}}, Y|X_2^{\mathrm{I}})+I(X^{\mathrm{O}}, Y|X_1^{\mathrm{I}}, X_2^{\mathrm{I}})+I(X^{\mathrm{O}}, X_1^{\mathrm{I}}, X_2^{\mathrm{I}}, Y),
\end{equation}
as shown in Figure \ref{OOMfig:MI_decomposition}, we can achieve
\begin{equation}
    \alpha(h_1^*, h_2^*)\le I(X^{\mathrm{O}}, Y)-I(X^{\mathrm{O}}, X_1^{\mathrm{I}}, X_2^{\mathrm{I}}, Y)+I(X^{\mathrm{O}}, Y|X_1^{\mathrm{I}}, X_2^{\mathrm{I}}),
    \label{OOMeq:final_disagreement}
\end{equation}

where the first term denotes the overall information, the second term indicates the commonality shared between all modalities, and the third term stands for the uniqueness only preserved in OOM data. Intuitively, when we try to increase the modality disagreement, the commonality is decreased and OOM uniqueness is increased, which successfully justifies our learning objective: In order to explore the uniqueness of OOM data, we can explore the ones with high modality disagreement; conversely, the OOM data with high commonality and low uniqueness is found where agreement is achieved among $h_1^*$ and $h_2^*$. Therefore, we select such data as anchor points that provide informative guidance to help explore uniqueness.

Next, we consider two realistic scenarios of OOM generalization and demonstrate how the proposed COX method can tackle them.

\begin{figure}[t]
\centering
\includegraphics[width=0.8\textwidth]{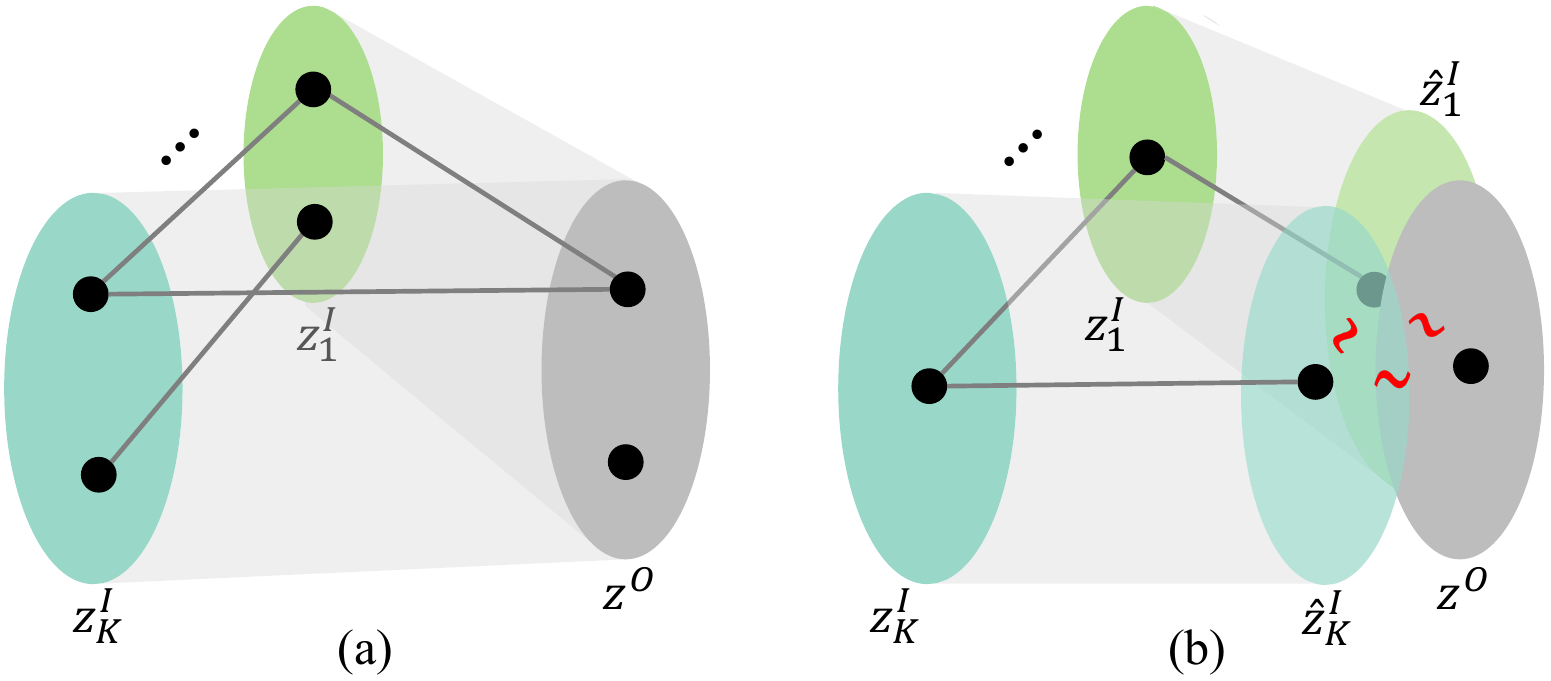}
\caption{Two scenarios: (a) Semi-Supervised OOM Generalization and (b) Unsupervised OOM Generlaizaiton.}
\label{OOMfig:setting}
\end{figure}

\subsection{Semi-Supervised OOM Generalization}
\label{OOM:semi}
We start with a semi-supervised case where a few correspondences are available in OOM data, as shown in Figure \ref{OOMfig:setting} (a). Based on the VIB framework proposed in Section \ref{OOM:cox}, we first leverage the OOM data $\{(x_i^{\mathrm{O}}, y_i^{\mathrm{O}})\}_{i=1}^L$ corresponding to IM data $\{(x_{k,i}^{\mathrm{I}}, y_{k,i}^{\mathrm{O}})\}_{i=1}^L, \forall k\in\{1,\ldots,K\}$ to build $K$ connections using additional generative models that can be trained via a point-to-point mapping. As a result, the mappings on the OOM feature space can successfully match the OOM feature distribution, which allows us to directly apply IM data posteriors to select and explore the uniqueness of OOM data. Hence, we formulate our objective as
\begin{equation}
 \min_{h^{\mathrm{O}}} \mathcal{L}_\mathrm{ssl}:=\frac{1}{L}\sum_{i=1}^L\text{CE}(h^{\mathrm{O}}(x_i^{\mathrm{O}}), y_i^{\mathrm{O}})+\frac{1}{L\!+\!|\mathcal{D}|}\!\sum_{x_{d,j}\in\mathcal{D}}\sum_{x_{i=1}^L}\text{KL}(h^{\mathrm{O}}(x_{d,j}^{\mathrm{O}})\!\parallel\! h^{\mathrm{O}}(x_{i}^{\mathrm{O}}); h_1^*, h_2^*),
\end{equation}
where the first term exploits labeled OOM data with correspondence and the second term explores OOM data $\mathcal{D}$ with modality disagreement by minimizing its KL divergence from the label posterior. Through the above objective, we can maximally exploit the uniqueness of OOM data to achieve effective OOM generalization.

\subsection{Unsupervised OOM Generalization}
\label{OOM:unsup}
As for the unsupervised case, we propose two-phase training: (1) we first conduct a warm-up training to initialize the OOM feature space and the connection, and (2) then, we enhance the connection by creating emergent correspondence and further exploring OOM data.

Specifically, we select anchor points from OOM data by directly applying modality agreement among all Bayes' optimal classifiers from IM data via
\begin{equation}
\!\!\!\!\!\mathcal{A}_{\text{sorted}}\!=\!\text{SORT}_T(\mathcal{A}, \frac{1}{K}\sum_{k=1}^K\max h_k^*(x^{\mathrm{O}})), \text{where }\mathcal{A}\!=\!\{\forall x^{\mathrm{O}}\!\in\!\mathcal{M}^{\mathrm{O}}\!\!:h_1^*(x^{\mathrm{O}})\!=\!\cdots\!=\!h_K^*(x^{\mathrm{O}})\},
\end{equation}
where the $\text{SORT}_T(\cdot,\cdot)$ is a sort function, which ranks each element $x^{\mathrm{O}}$ in $\mathcal{A}$ based on the value of $\frac{1}{K}\sum_{k=1}^K\max h_k^*(x^{\mathrm{O}})$ from large to small. Here, we select anchor points with the top-$T$ largest likelihood averaged over all $K$ IM classifiers. Then, we warm up the OOM learner via minimizing cross-entropy loss $\min \frac{1}{T}\sum_{x^{\mathrm{O}}\in\mathcal{A}_{\text{sorted}}}\text{CE}(h^{\mathrm{O}}(x^{\mathrm{O}}), \arg\max h_k^*(x^{\mathrm{O}}))$. Additionally, we also warm up the connection by leveraging class-wise information. Specifically, we compute the cluster centroids for each modality via $\frac{1}{|\mathcal{C}_y|}\sum_{x_i^{\mathrm{O}}\in\mathcal{C}_y:=\{x^{\mathrm{O}}:h^{\mathrm{O}}(x^{\mathrm{O}})=y,y\in\mathcal{Y}\}}z^{\mathrm{O}}_i$ and pair them to each IM centroid correspondingly. To this end, we can build up initial connections by following the VIB framework.

After the warm-up, we aim to further enhance both our connection and OOM exploration by creating emergent correspondence, as shown in Figure \ref{OOMfig:setting} (b). To tackle this, we map all IM data into the OOM feature space. If an OOM feature is close to all mappings $v_{k,i}, \forall k=\{1,\ldots,K\}$, then they can form a strong correspondence. Further, we select such OOM data as anchor points, which is further labeled the same as the corresponding IM data. Formally, we optimize OOM learners via
\begin{equation}
 \min_{h^{\mathrm{O}}} \mathcal{L}_\mathrm{uns}\!:=\!\frac{1}{|\mathcal{A}|}\!\!\sum_{(x_a^{\mathrm{O}}, y)\in\mathcal{A}}\!\!\!\!\!\text{CE}(h^{\mathrm{O}}(x_a^{\mathrm{O}}), y)\!+\!\frac{1}{|\mathcal{A}|\!\!+\!\!|\mathcal{D}|}\!\!\sum_{x_{d}^{\mathrm{O}}\in\mathcal{D}}\sum_{x_{a}^{\mathrm{O}}\in\mathcal{A}}\!\!\text{KL}(h^{\mathrm{O}}(x_{d}^{\mathrm{O}})\!\parallel\! h^{\mathrm{O}}(x_{a}^{\mathrm{O}}); h_1^*, h_2^*),
\end{equation}
where $\mathcal{A}$ denotes the updated anchor points which are realized by sorting the Euclidean distance: $\mathcal{A}:=\text{SORT}_S(\{(x_j^{\mathrm{O}}, y^{\mathrm{I}}_i)\}_{j=1}^M, -\min_{i\in\{1,\ldots,N\}}\frac{1}{K}\sum_{k=1}^K\|z_j^{\mathrm{O}}-v_{k,i}\|)$, where the first term computes the cross-entropy loss from the anchor points, and the second term calculates the KL divergence between the OOM data with modality disagreement and the anchor points.

After these two steps, we can effectively tackle the unsupervised OOM generalization. In practice, we connect modalities and select anchor points in the feature space, and hence our application to both two scenarios can be efficient. In the next section, we carefully conduct extensive experiments to justify the effectiveness and extendibility of the proposed COX method under various settings.

\section{Experiment}
\label{OOM:experiments}

In our experiments, we first elucidate the experimental details. Then, we provide performance comparisons to various baseline methods on different datasets. Finally, we conduct empirical analyses to provide an intuitive understanding of the proposed method.

\subsection{Implementation Details}
\textbf{Datasets.} We consider datasets with at least three modalities: (1) TVL dataset \cite{fu2024a} contains tactile sensing, RGB image, and class name which can be transformed into language; (2) LLVIP \cite{jia2021llvip} dataset has infrared thermal data, RGB image, and annotations for pedestrian detection. We follow \cite{zhu2023languagebind} to crop the pedestrian and background which stand for two classes. Further, we use the OpenAI template \cite{radford2021learning} to create language description; (3) NYU-D dataset \cite{silberman2012indoor} contains RGB image, depth data, and class name that can be transformed into language description as well; (4) VGGS dataset \cite{chen2020vggsound} includes video data, corresponding sound, and the language description; (5) MSR-VTT \cite{xu2016msr} includes videos and text description, we break down the videos into video frames and the audio data; (6) MOSEI dataset \cite{zadeh2018multimodal} contains videos from 7 classes of emotions, we extract audio data from the videos and use the emotion type to create language descriptions.

\textbf{Models.} We employ two types of IM perceptors, namely ImageBind \cite{girdhar2023imagebind} and LanguageBind \cite{zhu2023languagebind} which correspondingly contain 6 and 5 encoders to process different modalities. We select one modality for each experiment as OOM and then choose the rest as IM. For IM data, we use the existing encoders to extract their features. As for OOM data, we conduct preprocessing to ensure its compatibility. Then, we initialize an OOM learner from scratch using ViT-T/16 to learn from the OOM data using the guidance from IM perceptors. Note that for the TVL dataset, there are no existing encoders to process tactile modality. Therefore, when the tactile modality is chosen as IM data, we fine-tune the encoder using contrastive learning on the training set. For ImageBind, the tactile encoder is aligned with the image encoder, and for LanguageBind, it is aligned with the language encoder, which is the same as the original training process. For training the connection between modalities, we employ multi-layer perceptrons to realize the variational information bottleneck framework. Moreover, to obtain the optimal classifier from each in-modality, we utilize the extracted features and train a linear layer as classification heads.

\begin{table}[t]
    \centering
    \caption{Classification performance comparison of different methods across multiple datasets with different OOM modalities.}
    \begin{adjustbox}{max width=\textwidth}
        \begin{tabular}{ccc|ccc|ccc|ccc|ccc}
            \toprule[1.2pt]
            \multicolumn{1}{c}{\multirow{2}{*}{\textbf{Setting}}} & \multirow{2}{*}{\textbf{IM Perceptor}} & \multirow{2}{*}{\textbf{Method}} & \multicolumn{3}{c}{\textbf{TVL}} & \multicolumn{3}{c}{\textbf{LLVIP}} & \multicolumn{3}{c}{\textbf{NYU-D}} & \multicolumn{3}{c}{\textbf{VGGS}} \\
            \cmidrule(lr){4-15}
            \multicolumn{1}{c}{} & & & \textbf{RGB} & \textbf{Lan} & \textbf{Tac} & \textbf{RGB} & \textbf{Lan} & \textbf{The} & \textbf{RGB} & \textbf{Dep} & \textbf{Lan} & \textbf{Aud} & \textbf{Vid} & \textbf{Lan} \\
            \midrule
            \multirow{10}{*}{\rotatebox[origin=c]{90}{\textbf{Semi-Supervised\quad}}} & \multirow{5}{*}{ImageBind} & Random & 0.4 & 0.3 & 0.2 & 48.2 & 47.3 & 51.0 & 10.2 & 11.3 & 10.2 & 0.3 & 0.3 & 0.3 \\ &  & ERM & 23.1 & 19.5 & 22.7 & 54.6 & 53.1 & 54.1 & 45.2 & 44.5 & 38.1 & 9.3 & 10.2 & 8.4 \\
             &  & EntMin & 24.0 & 21.8 & 23.6 & 56.7 & 57.0 & 55.4 & 48.0 & 46.3 & 39.3 & 10.5 & 13.3 & 8.9 \\
             &  & COX & \textbf{31.2} & \textbf{25.3} & \textbf{26.5} & \textbf{59.2} & \textbf{58.3} & \textbf{58.3} & \textbf{52.3} & \textbf{50.7} & \textbf{44.2} & \textbf{16.8} & \textbf{18.4} & \textbf{11.7} \\ \cmidrule(lr){3-15} 
             &  & aligned & {\color[HTML]{9B9B9B} 79.5} & {\color[HTML]{9B9B9B} 29.8} & {\color[HTML]{9B9B9B} 35.8} & {\color[HTML]{9B9B9B} 65.4} & {\color[HTML]{9B9B9B} 61.8} & {\color[HTML]{9B9B9B} 63.4} & {\color[HTML]{9B9B9B} 61.8} & {\color[HTML]{9B9B9B} 54.0} & {\color[HTML]{9B9B9B} 52.7} & {\color[HTML]{9B9B9B} 27.8} & {\color[HTML]{9B9B9B} 29.3} & {\color[HTML]{9B9B9B} 19.1} \\ \cmidrule(lr){2-15} 
             & \multirow{5}{*}{LanguageBind} & Random & 0.4 & 0.3 & 0.2 & 48.2 & 47.3 & 51.0 & 10.2 & 11.3 & 10.2 & 0.3 & 0.3 & 0.3 \\
             &  & ERM & 23.6 & 20.1 & 22.6 & 56.5 & 54.9 & 58.3 & 44.8 & 44.5 & 39.9 & 9.8 & 13.7 & 9.9 \\
             &  & EntMin & 25.7 & 23.1 & 25.1 & 59.8 & 60.0 & 62.2 & 49.4 & 47.3 & 42.7 & 11.9 & 14.5 & 12.8 \\
             &  & COX & \textbf{33.5} & \textbf{26.3} & \textbf{27.3} & \textbf{61.2} & \textbf{62.3} & \textbf{66.4} & \textbf{58.8} & \textbf{53.5} & \textbf{48.4} & \textbf{18.3} & \textbf{22.1} & \textbf{13.4} \\ \cmidrule(lr){3-15} 
             &  & aligned & {\color[HTML]{9B9B9B} 81.6} & {\color[HTML]{9B9B9B} 31.2} & {\color[HTML]{9B9B9B} 38.3} & {\color[HTML]{9B9B9B} 74.1} & {\color[HTML]{9B9B9B} 73.2} & {\color[HTML]{9B9B9B} 87.2} & {\color[HTML]{9B9B9B} 68.6} & {\color[HTML]{9B9B9B} 65.1} & {\color[HTML]{9B9B9B} 57.7} & {\color[HTML]{9B9B9B} 38.6} & {\color[HTML]{9B9B9B} 32.5} & {\color[HTML]{9B9B9B} 20.9} \\ \midrule
            \multirow{8}{*}{\rotatebox[origin=c]{90}{\textbf{Unsupervised\quad}}} & \multirow{4}{*}{ImageBind} & Random & 0.4 & 0.3 & 0.2 & 48.2 & 47.3 & 51.0 & 10.2 & 11.3 & 10.2 & 0.3 & 0.3 & 0.3 \\
             &  & SSL & 6.3 & 4.3 & 5.1 & 52.3 & 56.1 & 52.4 & 14.6 & 13.6 & 18.9 & 2.5 & 6.9 & 3.8 \\
             &  & COX & \textbf{18.9} & \textbf{15.4} & \textbf{17.1} & \textbf{54.8} & \textbf{57.2} & \textbf{53.8} & \textbf{21.7} & \textbf{22.0} & \textbf{19.5} & \textbf{9.3} & \textbf{10.2} & \textbf{10.5} \\ \cmidrule(lr){3-15} 
             &  & aligned & {\color[HTML]{9B9B9B} 79.5} & {\color[HTML]{9B9B9B} 29.8} & {\color[HTML]{9B9B9B} 35.8} & {\color[HTML]{9B9B9B} 65.4} & {\color[HTML]{9B9B9B} 61.8} & {\color[HTML]{9B9B9B} 63.4} & {\color[HTML]{9B9B9B} 61.8} & {\color[HTML]{9B9B9B} 54.0} & {\color[HTML]{9B9B9B} 52.7} & {\color[HTML]{9B9B9B} 27.8} & {\color[HTML]{9B9B9B} 29.3} & {\color[HTML]{9B9B9B} 19.1} \\ \cmidrule(lr){2-15} 
             & \multirow{4}{*}{LanguageBind} & Random & 0.4 & 0.3 & 0.2 & 48.2 & 47.3 & 51.0 & 10.2 & 11.3 & 10.2 & 0.3 & 0.3 & 0.3 \\
             &  & SSL & 6.8 & 6.5 & 5.1 & 54.6 & 57.8 & 53.8 & 16.9 & 18.1 & 16.3 & 7.2 & 5.6 & 4.8 \\
             &  & COX & \textbf{19.3} & \textbf{19.2} & \textbf{18.6} & \textbf{55.0} & \textbf{56.4} & \textbf{55.7} & \textbf{24.5} & \textbf{23.1} & \textbf{20.4} & \textbf{10.0} & \textbf{11.6} & \textbf{10.4} \\ \cmidrule(lr){3-15} 
             &  & aligned & {\color[HTML]{9B9B9B} 81.6} & {\color[HTML]{9B9B9B} 31.2} & {\color[HTML]{9B9B9B} 38.3} & {\color[HTML]{9B9B9B} 74.1} & {\color[HTML]{9B9B9B} 73.2} & {\color[HTML]{9B9B9B} 87.2} & {\color[HTML]{9B9B9B} 68.6} & {\color[HTML]{9B9B9B} 65.1} & {\color[HTML]{9B9B9B} 57.7} & {\color[HTML]{9B9B9B} 38.6} & {\color[HTML]{9B9B9B} 32.5} & {\color[HTML]{9B9B9B} 20.9} \\
            \bottomrule[1.2pt]
        \end{tabular}
    \end{adjustbox}
    \label{OOMtab:classification}
\end{table}

\textbf{Setup.} We consider two scenarios of OOM generalization: For the semi-supervised case, we sample $10\%$ of the training data as labeled data with each class having a balanced number of labels. For the unsupervised case, we have no labels at all. For selecting the number of anchor points, we choose the same number of examples for the warm-up and training phases, which is $10\%$ of the total training set. To train the OOM learner, we use the Adam optimizer with an initial learning rate of $1e-3$ with weight decay $1e-5$, and train for $50$ epochs.

\textbf{Baseline methods.} Since there is no existing baseline method to compare with under our setting, we implement four methods for comparison, namely: Random where the model is randomly initialized, ERM where only labeled data is used to minimize the empirical risk, EntMin \cite{grandvalet2004semi} which minimize the entropy of unlabeled data meanwhile conduct ERM, SSL which conducts self-supervised learning using Gaussian noise perturbation on the input, and MoCo \cite{he2020momentum} which updates model parameters with ensembling and meanwhile conducts contrastive learning. Note that we use MoCo for comparison for the retrieval task in Table \ref{OOMtab:retrieval} because it is not for classification, and it is combined with EntMin in the semi-supervised case. Moreover, we use a pre-trained encoder as an upper-limit baseline ``aligned''. Next, we carefully compare the performance of our COX to these baseline methods.

\subsection{Performance Comparison}
For performance comparisons, we conduct classification and cross-modal retrieval to validate the proposed COX. There are seven modalities are considered, namely RGB image, language, tactile, thermal, depth, audio, and video which are simplified as RGB, Lan, Tac, The, Dep, Aud, and Vid, respectively. For each column, we choose one modality as OOM data, the rest modalities are selected IM data. For the retrieval task, we report the recall rate in both top 1 (R@1) and top 5 (R@5). The results are shown in Tables \ref{OOMtab:classification} and \ref{OOMtab:retrieval}. We can see that the proposed COX clearly shows the best performance in both scenarios. Specifically, COX can achieve more than $5\%$ performance improvement for most of the OOM setting, which justifies that leveraging the knowledge from IM perceptors can indeed help OOM generalization compared to using OOM data alone. Moreover, even though the performance is relatively limited compared to the fully pre-trained baseline under the unsupervised case, considering it is an extremely challenging setting, we can still largely improve the performance for over $10\%$ compared to the Random baseline, which demonstrates that the unsupervised OOM generalization is indeed learnable further leads to a novel research direction for improving the generalization performance. Additionally, note that the performance of COX is affected by the quality of IM perceptors, as using LanguageBind shows relatively higher performance compared to using ImageBind. Thus, it would be potentially helpful to leverage sophisticated IM perceptors to benefit the generalization performance.

\begin{table}[t]
\centering
\caption{Cross-modal retrieval performance comparison of different methods across multiple datasets with different OOM modalities.}
\begin{adjustbox}{max width=\textwidth}
\begin{tabular}{ccc|cc|cc|cc|cc|cc|cc}
\toprule[1.2pt]
\multirow{3}{*}{\textbf{Setting}} & \multirow{3}{*}{\textbf{IM Perceptor}} & \multirow{3}{*}{\textbf{Method}} & \multicolumn{6}{c|}{\textbf{MSR-VTT}} & \multicolumn{6}{c}{\textbf{MOSEI}} \\ \cmidrule(lr){4-15} 
 &  &  & \multicolumn{2}{c|}{\textbf{Aud}} & \multicolumn{2}{c|}{\textbf{Lan}} & \multicolumn{2}{c|}{\textbf{Vid}} & \multicolumn{2}{c|}{\textbf{Aud}} & \multicolumn{2}{c|}{\textbf{Lan}} & \multicolumn{2}{c}{\textbf{Vid}} \\ \cmidrule(lr){4-15} 
 &  &  & \textbf{R@1} & \textbf{R@5} & \textbf{R@1} & \textbf{R@5} & \textbf{R@1} & \textbf{R@5} & \textbf{R@1} & \textbf{R@5} & \textbf{R@1} & \textbf{R@5} & \textbf{R@1} & \textbf{R@5} \\ \midrule
&  & Random & 5.4 & \multicolumn{1}{c|}{25.1} & 5.0 & \multicolumn{1}{c|}{25.4} & 5.4 & 24.2 & 14.3 & \multicolumn{1}{c|}{42.5} & 14.4 & \multicolumn{1}{c|}{42.8} & 14.1 & 42.1 \\
 &  & ERM & 15.6 & \multicolumn{1}{c|}{30.3} & 16.1 & \multicolumn{1}{c|}{35.2} & 18.5 & 38.2 & 28.0 & \multicolumn{1}{c|}{45.3} & 29.3 & \multicolumn{1}{c|}{47.1} & 33.4 & 48.2 \\
 &  & EntMin & 18.5 & \multicolumn{1}{c|}{32.4} & 19.2 & \multicolumn{1}{c|}{38.5} & 21.0 & 39.4 & 29.6 & \multicolumn{1}{c|}{46.7} & 32.0 & \multicolumn{1}{c|}{48.7} & 35.4 & 50.5 \\
 &  & MoCo & 20.5 & \multicolumn{1}{c|}{33.9} & 21.1 & \multicolumn{1}{c|}{38.9} & 23.4 & 43.5 & 30.1 & \multicolumn{1}{c|}{47.3} & 32.7& \multicolumn{1}{c|}{50.1} & 36.2 & 51.0 \\
 &  & COX & \textbf{23.3} & \multicolumn{1}{c|}{\textbf{35.8}} & \textbf{23.4} & \multicolumn{1}{c|}{\textbf{39.1}} & \textbf{26.5} & \textbf{48.8} & \textbf{32.4} & \multicolumn{1}{c|}{\textbf{48.0}} & \textbf{33.8} & \multicolumn{1}{c|}{\textbf{50.4}} & \textbf{38.8} & \textbf{53.7} \\ \cmidrule(lr){3-15} 
 & \multirow{-6}{*}{ImageBind} & Aligned & {\color[HTML]{9B9B9B} 35.5} & \multicolumn{1}{c|}{{\color[HTML]{9B9B9B} 51.5}} & {\color[HTML]{9B9B9B} 32.3} & \multicolumn{1}{c|}{{\color[HTML]{9B9B9B} 52.4}} & {\color[HTML]{9B9B9B} 36.8} & {\color[HTML]{9B9B9B} 61.8} & {\color[HTML]{9B9B9B} 42.9} & \multicolumn{1}{c|}{{\color[HTML]{9B9B9B} 66.4}} & {\color[HTML]{9B9B9B} 48.2} & \multicolumn{1}{c|}{{\color[HTML]{9B9B9B} 69.4}} & {\color[HTML]{9B9B9B} 50.5} & {\color[HTML]{9B9B9B} 71.6} \\ \cmidrule(lr){2-15} 
 &  & Random & 5.2 & \multicolumn{1}{c|}{24.3} & 5.4 & \multicolumn{1}{c|}{25.1} & 5.0 & 25.6 & 13.5 & \multicolumn{1}{c|}{43.1} & 14.2 & \multicolumn{1}{c|}{42.7} & 14.6 & 41.9 \\
 &  & ERM & 16.3 & \multicolumn{1}{c|}{31.1} & 16.5 & \multicolumn{1}{c|}{36.2} & 18.7 & 37.9 & 27.3 & \multicolumn{1}{c|}{45.5} & 28.4 & \multicolumn{1}{c|}{47.6} & 33.4 & 49.3 \\
 &  & EntMin & 19.6 & \multicolumn{1}{c|}{33.4} & 19.8 & \multicolumn{1}{c|}{38.6} & 22.4 & 37.9 & 30.2 & \multicolumn{1}{c|}{45.5} & 33.5 & \multicolumn{1}{c|}{49.0} & 36.0 & 49.7 \\
 &  & MoCo & 21.1 & \multicolumn{1}{c|}{34.8} & 20.9 & \multicolumn{1}{c|}{39.2} & 24.5 & 38.6 & 31.1 & \multicolumn{1}{c|}{46.7} & 34.5 & \multicolumn{1}{c|}{\textbf{50.5}} & 37.0 & 51.7 \\
 &  & COX & \textbf{25.2} & \multicolumn{1}{c|}{\textbf{36.0}} & \textbf{24.1} & \multicolumn{1}{c|}{\textbf{40.0}} & \textbf{28.7} & \textbf{49.5} & \textbf{34.6} & \multicolumn{1}{c|}{\textbf{49.8}} & \textbf{34.6} & \multicolumn{1}{c|}{50.2} & \textbf{39.2} & \textbf{55.4} \\ \cmidrule(lr){3-15} 
\multirow{-12}{*}{\rotatebox[origin=c]{90}{\textbf{\quad Semi-Supervised}}} & \multirow{-6}{*}{LanguageBind} & Aligned & {\color[HTML]{9B9B9B} 42.0} & \multicolumn{1}{c|}{{\color[HTML]{9B9B9B} 53.6}} & {\color[HTML]{9B9B9B} 38.8} & \multicolumn{1}{c|}{{\color[HTML]{9B9B9B} 58.6}} & {\color[HTML]{9B9B9B} 44.8} & {\color[HTML]{9B9B9B} 70.0} & {\color[HTML]{9B9B9B} 44.6} & \multicolumn{1}{c|}{{\color[HTML]{9B9B9B} 68.9}} & {\color[HTML]{9B9B9B} 49.5} & \multicolumn{1}{c|}{{\color[HTML]{9B9B9B} 67.4}} & {\color[HTML]{9B9B9B} 51.1} & {\color[HTML]{9B9B9B} 68.3} \\ \midrule
 &  & Random & 5.4 & \multicolumn{1}{c|}{25.1} & 5.0 & \multicolumn{1}{c|}{25.4} & 5.4 & 24.2 & 14.3 & \multicolumn{1}{c|}{42.5} & 14.4 & \multicolumn{1}{c|}{42.8} & 14.1 & 42.1 \\
 &  & SSL & 8.9 & \multicolumn{1}{c|}{28.4} & 9.3 & \multicolumn{1}{c|}{28.1} & 10.1 & 29.5 & 17.4 & \multicolumn{1}{c|}{48.8} & 16.2 & \multicolumn{1}{c|}{45.2} & 16.0 & 45.0 \\
 &  & MoCo & 9.2 & \multicolumn{1}{c|}{28.9} & 9.5 & \multicolumn{1}{c|}{28.4} & 10.6 & 30.0 & 17.8 & \multicolumn{1}{c|}{50.3} & 16.6 & \multicolumn{1}{c|}{45.8} & 17.1 & 44.4 \\
 &  & COX & \textbf{13.5} & \multicolumn{1}{c|}{\textbf{30.4}} & \textbf{16.5} & \multicolumn{1}{c|}{\textbf{32.4}} & \textbf{15.2} & \textbf{34.8} & \textbf{20.8} & \multicolumn{1}{c|}{\textbf{53.7}} & \textbf{18.7} & \multicolumn{1}{c|}{\textbf{46.7}} & \textbf{18.2} & \textbf{48.9} \\ \cmidrule(lr){3-15} 
 & \multirow{-5}{*}{ImageBind} & Aligned & {\color[HTML]{9B9B9B} 35.5} & \multicolumn{1}{c|}{{\color[HTML]{9B9B9B} 51.5}} & {\color[HTML]{9B9B9B} 32.3} & \multicolumn{1}{c|}{{\color[HTML]{9B9B9B} 52.4}} & {\color[HTML]{9B9B9B} 36.8} & {\color[HTML]{9B9B9B} 61.8} & {\color[HTML]{9B9B9B} 42.9} & \multicolumn{1}{c|}{{\color[HTML]{9B9B9B} 66.4}} & {\color[HTML]{9B9B9B} 48.2} & \multicolumn{1}{c|}{{\color[HTML]{9B9B9B} 69.4}} & {\color[HTML]{9B9B9B} 50.5} & {\color[HTML]{9B9B9B} 71.6} \\ \cmidrule(lr){2-15} 
 &  & Random & 5.2 & \multicolumn{1}{c|}{24.3} & 5.4 & \multicolumn{1}{c|}{25.1} & 5.0 & 25.6 & 13.5 & \multicolumn{1}{c|}{43.1} & 14.2 & \multicolumn{1}{c|}{42.7} & 14.6 & 41.9 \\
 &  & SSL & 9.2 & \multicolumn{1}{c|}{28.9} & 11.0 & \multicolumn{1}{c|}{28.8} & 10.3 & 28.7 & 18.0 & \multicolumn{1}{c|}{48.9} & 18.4 & \multicolumn{1}{c|}{45.0} & 17.8 & 45.6 \\
 &  & MoCo & 9.6 & \multicolumn{1}{c|}{29.4} & 11.1 & \multicolumn{1}{c|}{28.5} & 11.0 & 29.3 & 18.8 & \multicolumn{1}{c|}{50.7} & 18.5 & \multicolumn{1}{c|}{45.2} & 18.0 & 45.5 \\
 &  & COX & \textbf{14.8} & \multicolumn{1}{c|}{\textbf{31.1}} & \textbf{18.4} & \multicolumn{1}{c|}{\textbf{34.4}} & \textbf{15.4} & \textbf{35.0} & \textbf{23.1} & \multicolumn{1}{c|}{\textbf{52.8}} & \textbf{19.4} & \multicolumn{1}{c|}{\textbf{47.2}} & \textbf{20.4} & \textbf{49.9} \\ \cmidrule(lr){3-15} 
\multirow{-10}{*}{\rotatebox[origin=c]{90}{\textbf{\quad Unsupervised}}} & \multirow{-5}{*}{LanguageBind} & Aligned & {\color[HTML]{9B9B9B} 42.0} & \multicolumn{1}{c|}{{\color[HTML]{9B9B9B} 53.6}} & {\color[HTML]{9B9B9B} 38.8} & \multicolumn{1}{c|}{{\color[HTML]{9B9B9B} 58.6}} & {\color[HTML]{9B9B9B} 44.8} & {\color[HTML]{9B9B9B} 70.0} & {\color[HTML]{9B9B9B} 44.6} & \multicolumn{1}{c|}{{\color[HTML]{9B9B9B} 68.9}} & {\color[HTML]{9B9B9B} 49.5} & \multicolumn{1}{c|}{{\color[HTML]{9B9B9B} 67.4}} & {\color[HTML]{9B9B9B} 51.1} & {\color[HTML]{9B9B9B} 68.3} \\ \bottomrule[1.2pt]
\end{tabular}
\end{adjustbox}
\label{OOMtab:retrieval}
\end{table}

\subsection{Empirical Analysis}
To provide an intuitive justification for the proposed method, here we conduct empirical analyses using the MSR-VTT dataset on various OOM scenarios and modalities.

\begin{figure}[h]
\centering
\includegraphics[width=\textwidth]{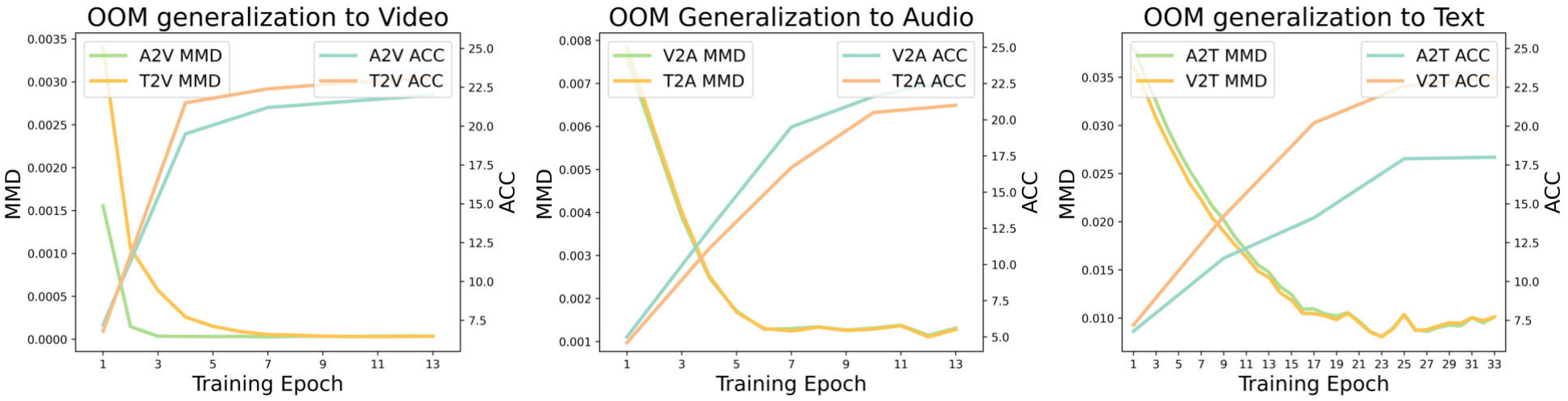}
\caption{Connection effect on maximum mean discrepancy and accuracy across modalities.}
\label{OOMfig:mmd_acc}
\end{figure}

\textbf{Connection mitigates modality gap.} To understand the performance of our VIB-based connection learning, here we show its effect on generalization out-of-modal. Specifically, during connection training, we compute the maximum mean discrepancy (MMD) between the mapping of each IM data and the OOM data. Meanwhile, we evenly select 6 points during the training and extract the IM mappings which are used to learn a classification head as the optimal classifier. Based on our theoretical result, we apply the classifiers to OOM data and compute their accuracies, as shown in Figure \ref{OOMfig:mmd_acc}. We can see that as training goes on, the MMD between each IM mapping and OOM data is decreasing and the corresponding accuracy is increasing, which shows that: (1) our connection can indeed close the modality gap between their features and (2) as the mappings of IM data getting close to OOM data, the optimal classifier shows better classification results on OOM data, which benefits the knowledge transfer from known modalities to unknown ones.

\begin{figure}[t]
\centering
\includegraphics[width=0.8\textwidth]{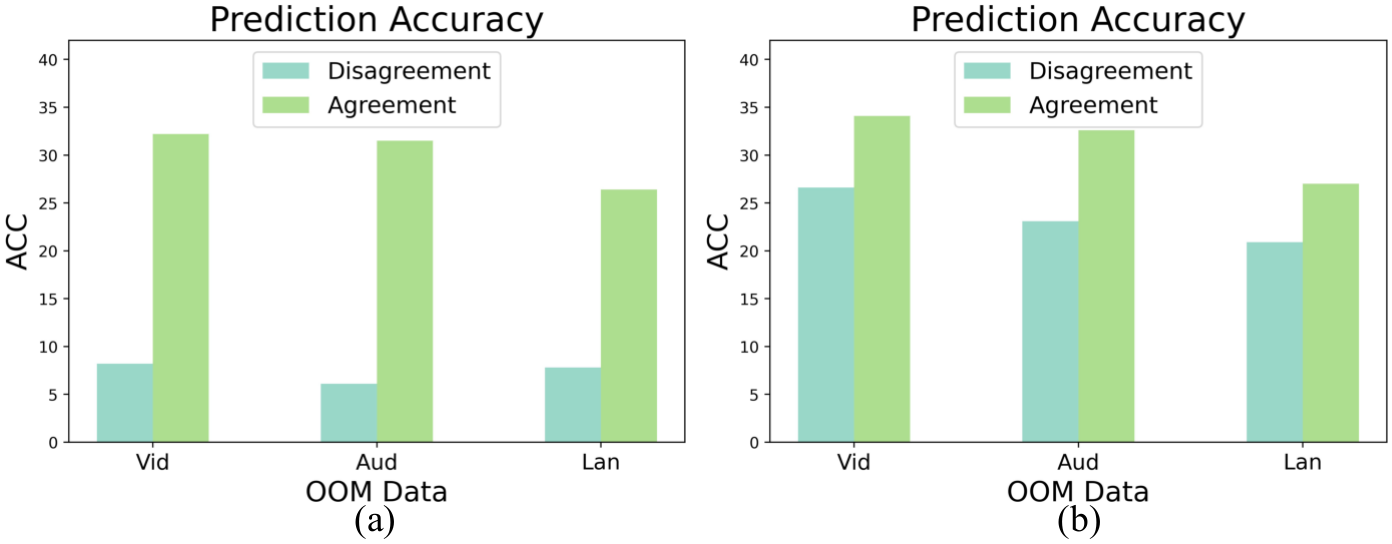}
\caption{Prediction accuracy of OOM data with modality disagreement and modalities agreement, respectively. (a) Before exploration. (b) After exploration.}
\label{OOMfig:disagreement}
\end{figure}

\textbf{Modality disagreement identifies uncertainty.} 
To understand the effect of modality disagreement, we analyze the semi-supervised scenario by training the OOM learner to use only labeled data for 10 epochs. Then, we leverage the modality disagreement criteria to separate OOM data into those with disagreement and agreement and show their prediction accuracies in Figure \ref{OOMfig:disagreement} (a). We can see that the accuracy for OOM data with disagreement is significantly lower than those with agreement, meaning that the prediction uncertainty, i.e., data with low accuracy, is effectively identified by the modality disagreement.

\textbf{Modality agreement alleviates uncertainty.} 
Further, we conduct training by following the procedure proposed in Section \ref{OOM:semi} and again show the accuracies of OOM data with disagreement and agreement in Figure \ref{OOMfig:disagreement} (b). We can see that the performance gap between the two types of data is largely mitigated, which justifies the methodology of exploring OOM data using the guidance of modality agreement. As a result, we can achieve almost comparable performance on both types of data, benefiting the overall generalization.

\begin{table}[t]
\caption{Ablation study on various settings.}
\setlength{\tabcolsep}{8mm}
\begin{tabular}{cc|ccc}
\toprule[1.2pt]
\multirow{2}{*}{Setting} & \multirow{2}{*}{Ablation} & \multicolumn{3}{c}{MSR-VTT R@1} \\ \cmidrule(lr){3-5} 
 &  & \multicolumn{1}{c|}{Aud} & \multicolumn{1}{c|}{Lan} & Vid \\ \midrule
\multirow{3}{*}{Semi} & w/o connection & 8.7 & 7.9 & 10.3 \\
 & w/o exploration & 16.4 & 16.5 & 18.8 \\
 & COX & \textbf{25.2} & \textbf{24.1} & \textbf{40.0} \\ \midrule
\multirow{2}{*}{Unsup.} & w/o warm-up & 7.4 & 11.5 & 10.5 \\
 & COX & \textbf{14.8} & \textbf{18.4} & \textbf{15.4} \\ \bottomrule[1.2pt]
\end{tabular}
\label{OOMtab:ablation}
\end{table}

\textbf{Ablation study.}
Additionally, we conduct an ablation study to justify the effect of our methodology. Specifically, we consider three ablations: (1) ``w/o connection'' where we remove the connection and directly apply the modality disagreement criteria on the original features of IM data and OOM data, (2) ``w/o exploration'' where we only leverage the OOM data with agreement for training, (3) For unsupervised scenario, we consider ``w/o warm-up'' where we do not conduct the warm-up phase and directly training the model. The results in Table \ref{OOMtab:ablation} show that all modules are essential for achieving effective OOM generalization. Specifically, the connection is vital for the knowledge transduction of IM data to OOM data, without which the generalization performance is largely degraded. The conclusion is consistent with the connection analysis where directly applying optimal classifiers across modalities leads to poor accuracy. Moreover, removing exploration also hinders the performance because the uniqueness of OOM data is largely ignored. Additionally, we find that the warm-up phase is essential for the unsupervised case. As initialized models have no classification capability, we need pre-training to form basic feature clusters that are consistent with IM data, further enabling effective OOM generalization.

\textbf{Discussion on computational efficiency.} Note that we conduct the feature connection mostly on the feature space, the computational cost of training VIB framework work is quite acceptable. The main cost is training the OOM learner which is the basic training with cross-entropy loss optimization and can be implemented on a single NVIDIA 3090/4090 GPU.

\section{Conclusion and Limitation}
In this Chapter, we study a novel and promising research direction dubbed Out-of-Modal (OOM) Generalization which aims to leverage knowledge from existing modalities to generalize to an unknown modality without instance-level correspondence. We consider two scenarios where there are a few correspondences and there is no correspondence, i.e., semi-supervised and unsupervised cases, respectively. To tackle these problems, we propose a Connect\&Explore (COX) method which first learns connections across modalities to extract common knowledge and then explores the unique knowledge of OOM data based on modality disagreement. Extensive experiments are conducted to justify the proposed method and intuitive insights are provided to inspire future studies. However, our research is limited to several aspects which we hope to address in the future. First, although challenging as it is, the performance is relatively limited compared to fully-aligned models, which requires more investigations to enhance generalization. Second, our OOM generalization is mostly conducted within the modalities from the same dataset. In the future, we hope to discover scenarios where the OOM data is from a different dataset with a large modality gap.

\chapter{Proofs and Theoretical Analyses}
\label{cha:theorem}
In this Chapter, we provide proofs and theoretical analyses, including the convergence analysis and theorem for sharpness-based worst-case optimization in Chapter~\ref{cha:SharpDRO}, analysis and proposition for the pruning strategy in Chapter~\ref{cha:EVIL}, theoretical proof for achieving optimal prediction via In-Context Learning in Chapter~\ref{cha:MVT}, and theoretical justification for our modality disagreement in Chapter~\ref{cha:OOM}.

\section{Proof for SharpDRO}
This section provides the proof of Theorem~\ref{SharpDRO:theorem}. We first give some notations before we start our proof for the convergence.
\begin{enumerate}
    \item We denote the loss expectation as $\mathbb{L}(\theta,\omega):=\mathbb{E}_{(x,y)\sim Q}\mathcal{L}(\theta,\omega;(x,y))$, and so as the SAM function that $\mathbb{R}(\theta,\omega)=\mathbb{E}_{(x,y)\sim Q}R(\theta,\omega;(x,y))$. So our objective can be turned into: $\min_{\theta}\{\max_{\omega}\mathbb{L}(\theta,\omega)\}+\mathbb{R}(\theta,\omega)$. And recalling our SharpDRO algorithm, we restate the meaning of the parameters: the model is parameterized by $\theta$ and $\omega$ means the weighted sampling.
    \item $\kappa$ is the condition number that $\kappa=\frac{l}{\mu}$, where $l$ is the Lipschitz-smoothness in Assumption \ref{as:smooth} and $\mu$ means the PL condition in Assumption \ref{as:pl}.
    \item We define $\mathbb{L}^*(\theta)=\max_{\omega}\mathbb{L}(\theta,\omega)$ and $\omega^*(\theta)=arg\max_{\omega}\mathbb{L}(\theta,\omega)$.
\end{enumerate}

\subsection{Update Rule}
Before our theoretical analyses, we need to make the update rule for each variable explicit. We have to pay attention to the fact that our algorithm is stochastic that we can not directly get the real value of the gradient $\nabla\mathbb{L}(\theta,\omega)$, rather we estimate it by batches of samples $g_{\theta}(\theta,\omega)=\frac{1}{M}\sum_{i=1}^M\frac{\partial \mathcal{L}}{\partial \theta}(\theta,\omega;(x_i,y_i))$ and $g_{\omega}(\theta,\omega)=\frac{1}{M}\sum_{i=1}^M\frac{\partial \mathcal{L}}{\partial \omega}(\theta,\omega;(x_i,y_i))$, who hold some properties we will introduce in Assumption \ref{as:bv}. So the optimization iteration is executed as follows in reality:
\begin{equation}
    \begin{aligned}
    &\theta_{t+1}=\theta_t-\eta_{\theta}g_{\theta}(\theta_t+\rho g_{\theta}(\theta_t,\omega_t),\omega_t);\\
    &\omega_{t+1}=\omega_t+\eta_{\omega}\nabla_{\omega}g_{\omega}(\theta_t,\omega_t).
    \end{aligned}
\end{equation}

We further give a notation for brief that $\theta_{t+1/2}\triangleq\theta_t+\rho g_{\theta}(\theta_t,\omega_t)$, so the update for $\theta$ can be simplified as: $\theta_{t+1}=\theta_t-\eta_{\theta}g_{\theta}(\theta_{t+1/2},\omega_t)$.

\subsection{Assumptions}
We also have to make some necessary assumptions on our problem setting for this convergence proof:
\begin{assumption}[Bounded variance]\label{as:bv}
The unbiased estimation about the gradient of the loss function also has bounded variance that:
\begin{equation*}
    \begin{aligned}
    \mathbb{E}_{(x,y)\sim Q}[\frac{\partial\mathcal{L}}{\partial\theta}(\theta,\omega;(x,y))]=\nabla_{\theta}\mathbb{L}(\theta,\omega),&\quad \mathbb{E}_{(x,y)\sim Q}\|\frac{\partial \mathcal{L}}{\partial \theta}(\theta,\omega;(x,y))-\nabla_{\theta}\mathbb{L}(\theta,\omega)\|^2\leq\sigma^2;\\
    \mathbb{E}_{(x,y)\sim Q}[\frac{\partial\mathcal{L}}{\partial\omega}(\theta,\omega;(x,y))]=\nabla_{\omega}\mathbb{L}(\theta,\omega),&\quad \mathbb{E}_{(x,y)\sim Q}\|\frac{\partial \mathcal{L}}{\partial \omega}(\theta,\omega;(x,y))-\nabla_{\omega}\mathbb{L}(\theta,\omega)\|^2\leq\sigma^2.
    \end{aligned}
\end{equation*}
\end{assumption}
\begin{remark}
Since $g_{\theta}$ and $g_{\omega}$ are the averaged samples that: $g_{\theta}=\frac{1}{M}\sum_{i=1}^M \frac{\partial \mathcal{L}}{\partial \theta }(\theta,\omega;(x_i,y_i))$ and $g_{\omega}=\frac{1}{M}\sum_{i=1}^M\frac{\partial\mathcal{L}}{\partial\omega}(\theta,\omega;(x_i,y_i))$ respectively, they also have the unbiased property and have bounded variance:
\begin{equation*}
    \begin{aligned}
    \mathbb{E}_{(x,y)\sim Q}[g_{\theta}(\theta,\omega;(x,y))]=\nabla_{\theta}\mathbb{L}(\theta,\omega),&\quad \mathbb{E}_{(x,y)\sim Q}\|g_{\theta}(\theta,\omega;(x,y))-\nabla_{\theta}\mathbb{L}(\theta,\omega)\|^2\leq\frac{\sigma^2}{M};\\
    \mathbb{E}_{(x,y)\sim Q}[g_{\omega}(\theta,\omega;(x,y))]=\nabla_{\omega}\mathbb{L}(\theta,\omega),&\quad \mathbb{E}_{(x,y)\sim Q}\|g_{\omega}(\theta,\omega;(x,y))-\nabla_{\omega}\mathbb{L}(\theta,\omega)\|^2\leq\frac{\sigma^2}{M}.
    \end{aligned}
\end{equation*}
\end{remark}

\begin{assumption}[Lipschitz smooth]\label{as:smooth}
$\mathcal{L}(\theta,\omega;(x,y))$ is differential and $l$-Lipschitz smooth for every given sample $(x,y)$:
\begin{equation*}
    \begin{aligned}
    \|\nabla_{\theta}\mathcal{L}(\theta_1,\omega;(x,y))-\nabla_{\theta}\mathcal{L}(\theta_2,\omega;(x,y))\|&\leq l\|\theta_1-\theta_2\|,\quad\forall \omega, (x,y);\\
    \|\nabla_{\omega}\mathcal{L}(\theta,\omega_1;(x,y))-\nabla_{\omega}\mathcal{L}(\theta,\omega_2;(x,y))\|&\leq l\|\omega_1-\omega_2\|,\quad\forall \theta, (x,y).
    \end{aligned}
\end{equation*}
\end{assumption}
\begin{remark}
So the expectation function $\mathbb{L}$ also have the Lipschitz smooth property that:
\begin{equation*}
    \begin{aligned}
    \|\nabla_{\theta}\mathbb{L}(\theta_1,\omega)\!\!-\!\!\nabla_{\theta}\mathbb{L}(\theta_2,\omega)\|&\leq\mathbb{E}\|\nabla_{\theta}\mathcal{L}(\theta_1,\omega;(x,y))\!\!-\!\!\nabla_{\theta}\mathcal{L}(\theta_2,\omega;(x,y))\|\leq l\|\theta_1-\theta_2\|,\forall \omega;\\
    \|\nabla_{\omega}\mathbb{L}(\theta,\omega_1)\!\!-\!\!\nabla_{\omega}\mathbb{L}(\theta,\omega_2)\|&\leq \mathbb{E}\|\nabla_{\omega}\mathcal{L}(\theta,\omega_1;(x,y))\!\!-\!\!\nabla_{\omega}\mathcal{L}(\theta,\omega_2;(x,y))\|\leq l\|\omega_1-\omega_2\|,\forall \theta.
    \end{aligned}
\end{equation*}
\end{remark}

\begin{assumption}[PL condition]\label{as:pl}
The loss function $\mathbb{L}(\theta,\cdot)$ satisfies PL condition on every given $\theta$, i.e., there exists $\mu>0$ such that $\|\nabla_{\omega}\mathbb{L}(\theta,\omega)\|^2\geq2\mu[\max_{\omega}\mathbb{L}(\theta,\omega)-\mathbb{L}(\theta,\omega)], \forall \theta,\omega$.
\end{assumption}

\subsection{Useful Lemmas}
In this part, we will prove some necessary lemmas for us to prove the convergence bound. And we will give the definition of the stationary point of our problem at the beginning.

\begin{definition}[Stationary measure]
$\theta$ is defined as the $\epsilon$-stationary point of our problem if $\mathbb{E}\|\nabla\mathbb{L}^*(\theta)\|\leq\epsilon$ for any $\epsilon\geq0$.
\end{definition}
\begin{remark}
For minmax problem, there are usually two ways to measure the stationary point. One is measured two-side when $\mathbb{E}\|\nabla_{\theta}\mathbb{L}(\theta,\omega)\|\leq\epsilon$ and $\mathbb{E}\|\nabla_{\omega}\mathbb{L}(\theta,\omega)\|\leq\epsilon$, we claim $(\theta,\omega)$ is the $(\epsilon,\epsilon)$-stationary point. It has been proved in \cite{yang2022faster} that these two measures can be translated into each other when $\mathbb{L}^*$ is smooth which will be shown in Lemma \ref{le:L*smooth}. But what we compute is the model parameter $\theta$ using the algorithm SharpDRO. So we choose the measure by $\mathbb{E}\|\mathbb{L}^*(\theta)\|$ here.
\end{remark}

\begin{lemma}~\cite{nouiehed2019solving}\label{le:L*smooth}
Under Assumption \ref{as:smooth} and \ref{as:pl}, $\mathbb{L}^*(\theta)$ is $(l+\frac{l^2}{2\mu})$-Lipschitz smooth with the gradient:
$$\nabla_{\theta}\mathbb{L}^*(\theta,\omega)=\nabla_{\theta}\mathbb{L}(\theta,\omega^*(\theta)).$$
\end{lemma}

\begin{lemma}~\cite{nouiehed2019solving}\label{le:w^*smooth}
Under Assumption \ref{as:smooth} and \ref{as:pl}, $\omega^*(\cdot)$ is smooth about its variable:
$$\|\omega^*(\theta_1)-\omega^*(\theta_2)\|\leq\frac{l}{2\mu}\|\theta_1-\theta_2\|, \quad\forall \theta_1,\theta_2.$$
\end{lemma}

\begin{lemma}\label{proofeq:gradientbound}
We give an estimation that:
\begin{equation}
    \mathbb{E}\|g_{\theta}(\theta_{t+1/2},\omega_t)\|^2\leq(4\rho^2l^2+2\rho l+2)\mathbb{E}\|\nabla_{\theta}\mathbb{L}(\theta_t,\omega_t)\|^2+(5\rho^2l^2+2)\frac{\sigma^2}{M}.
\end{equation}
\end{lemma}
\begin{proof}
\begin{align}
    \mathbb{E}\|g_{\theta}(\theta_{t+1/2},\omega_t)\|^2=&-\mathbb{E}\|\nabla_{\theta}\mathbb{L}(\theta_t,\omega_t)\|^2+\mathbb{E}\|g_{\theta}(\theta_{t+1/2},\omega_t)-\nabla_{\theta}\mathbb{L}(\theta_t,\omega_t)\|^2\\
    &\quad+2\mathbb{E}\langle g_{\theta}(\theta_{t+1/2},\omega_t),\nabla_{\theta}\mathbb{L}(\theta_t,\omega_t)\rangle.
\end{align}

For the cross-product term, we divide it as follows:
\begin{equation}
    \begin{aligned}
    &\quad\mathbb{E}\langle g_{\theta}(\theta_{t+1/2},\omega_t),\nabla_{\theta}\mathbb{L}(\theta_t,\omega_t)\rangle\\
    &=\mathbb{E}\langle g_{\theta}(\theta_{t+1/2},\omega_t)-g_{\theta}(\theta_t+\rho\nabla_{\theta}\mathbb{L}(\theta_t,\omega_t),\omega_t),\nabla_{\theta}\mathbb{L}(\theta_t,\omega_t)\rangle\\
    &\quad+\mathbb{E}\langle g_{\theta}(\theta_t+\rho\nabla_{\theta}\mathbb{L}(\theta_t,\omega_t),\omega_t),\nabla_{\theta}\mathbb{L}(\theta_t,\omega_t)\rangle\\
    &=\mathbb{E}\langle \nabla_{\theta}\mathbb{L}(\theta_{t+1/2},\omega_t)-\nabla_{\theta}\mathbb{L}(\theta_t+\rho\nabla_{\theta}\mathbb{L}(\theta_t,\omega_t),\omega_t),\nabla_{\theta}\mathbb{L}(\theta_t,\omega_t)\rangle\\
    &\quad+\mathbb{E}\langle\nabla_{\theta}\mathbb{L}(\theta_t+\rho\nabla_{\theta}\mathbb{L}(\theta_t,\omega_t),\omega_t),\nabla_{\theta}\mathbb{L}(\theta_t,\omega_t)\rangle\overset{(i)}{\leq}\frac{1}{2}\mathbb{E}\|\nabla_{\theta}\mathbb{L}(\theta_{t+1/2},\omega_t)\\
    &\quad-\nabla_{\theta}\mathbb{L}(\theta_t+\rho\nabla_{\theta}\mathbb{L}(\theta_t,\omega_t),\omega_t)\|^2+\frac{1}{2}\mathbb{E}\|\nabla_{\theta}\mathbb{L}(\theta_t,\omega_t)\|^2+\mathbb{E}\|\nabla_{\theta}\mathbb{L}(\theta_t,\omega_t)\|^2\\
    &\quad+\mathbb{E}\langle \nabla_{\theta}\mathbb{L}(\theta_t+\rho\nabla_{\theta}\mathbb{L}(\theta_t,\omega_t),\omega_t)-\nabla_{\theta}\mathbb{L}(\theta_t,\omega_t),\nabla_{\theta}\mathbb{L}(\theta_t,\omega_t)\rangle\\
    &\overset{(ii)}{\leq}\frac{\rho^2l^2}{2}\mathbb{E}\|g_{\theta}(\theta_t,\omega_t)-\nabla_{\theta}\mathbb{L}(\theta_t,\omega_t)\|^2+\frac{3}{2}\mathbb{E}\|\nabla_{\theta}\mathbb{L}(\theta_t,\omega_t)\|^2 +\rho l\mathbb{E}\|\nabla_{\theta}\mathbb{L}(\theta_t,\omega_t)\|^2\\
    &\overset{(iii)}{\leq}(\rho l+\frac{3}{2})\mathbb{E}\|\nabla_{\theta}\mathbb{L}(\theta_t,\omega_t)\|^2+\frac{\rho^2l^2\sigma^2}{2M},
    \end{aligned}
\end{equation}
where the inequality $(i)$ is due to the Cauchy-Schwarz inequality; the inequality $(ii)$ is because of the Lipschitz-smoothness of $\mathbb{L}$ that:
\begin{equation}
    \mathbb{E}\|\nabla_{\theta}\mathbb{L}(\theta_{t+1/2},\omega_t)-\nabla_{\theta}\mathbb{L}(\theta_t+\rho\nabla_{\theta}\mathbb{L}(\theta_t,\omega_t),\omega_t)\|^2\leq l^2\mathbb{E}\|\theta_{t+1/2}-\theta_t-\rho\nabla_{\theta}\mathbb{L}(\theta_t,\omega_t)\|^2,
\end{equation}
and the property of Lipschitz-smoothness that:
\begin{align}
    &\langle \nabla_{\theta}\mathbb{L}(\theta_t+\rho\nabla_{\theta}\mathbb{L}(\theta_t,\omega_t),\omega_t)-\nabla_{\theta}\mathbb{L}(\theta_t,\omega_t),\nabla_{\theta}\mathbb{L}(\theta_t,\omega_t)\rangle=\\
    &\frac{1}{\rho}\langle \nabla_{\theta}\mathbb{L}(\theta_t+\rho\nabla_{\theta}\mathbb{L}(\theta_t,\omega_t),\omega_t)-\nabla_{\theta}\mathbb{L}(\theta_t,\omega_t),\rho\nabla_{\theta}\mathbb{L}(\theta_t,\omega_t)\rangle\leq\frac{l}{\rho}\|\rho\nabla_{\theta}\mathbb{L}(\theta_t,\omega_t)\|^2,
\end{align}
and the inequality $(iii)$ makes use of the Assumption \ref{as:bv}.

As for the second term, we have:
\begin{equation}
    \begin{aligned}
    &\quad\mathbb{E}\|g_{\theta}(\theta_{t+1/2},\omega_t)-\nabla_{\theta}\mathbb{L}(\theta_t,\omega_t)\|^2\\
    &\leq2\mathbb{E}\|g_{\theta}(\theta_{t+1/2},\omega_t)-\nabla_{\theta}\mathbb{L}(\theta_{t+1/2},\omega_t)\|^2+2\mathbb{E}\|\nabla_{\theta}\mathbb{L}(\theta_{t+1/2},\omega_t)-\nabla_{\theta}\mathbb{L}(\theta_t,\omega_t)\|^2\\
    &\quad\leq\frac{2\sigma^2}{M}+2l^2\mathbb{E}\|\theta_{t+1/2}-\theta_t\|^2\\
    &=\frac{2\sigma^2}{M}+2\rho^2l^2\mathbb{E}\|g_{\theta}(\theta_t,\omega_t)\|^2\\
    &\quad\leq2\frac{\sigma^2}{M}(2\rho^2l^2+1)+4\rho^2l^2\mathbb{E}\|\nabla_{\theta}\mathbb{L}(\theta_t,\omega_t)\|^2,
    \end{aligned}
\end{equation}
where the last inequality comes from the fact that:
\begin{equation}
    \mathbb{E}\|g_{\theta}(\theta_t,\omega_t)\|^2\leq2\mathbb{E}\|g_{\theta}(\theta_t,\omega_t)-\nabla_{\theta}\mathbb{L}(\theta_t,\omega_t)\|^2+2\mathbb{E}\|\nabla_{\theta}\mathbb{L}(\theta_t,\omega_t)\|^2.
\end{equation}

By combining the above inequalities, we can get:
\begin{equation}
    \mathbb{E}\|g_{\theta}(\theta_{t+1/2},\omega_t)\|^2\leq(4\rho^2l^2+2\rho l+2)\mathbb{E}\|\nabla_{\theta}\mathbb{L}(\theta_t,\omega_t)\|^2+(5\rho^2l^2+2)\frac{\sigma^2}{M}.
\end{equation}
\end{proof}

\begin{lemma}\label{le:L^*gradient}
For the descending relationship of the function $\mathbb{L}^*$, we have:
\begin{equation*}
    \begin{aligned}
    \mathbb{E}[\mathbb{L}^*(\theta_{t+1})]
    &\leq\mathbb{E}[\mathbb{L}^*(\theta_t)]-\frac{\eta_{\theta}}{2}(1-5\rho l-2L\eta_{\theta}(4\rho^2l^2+2\rho l+2))\mathbb{E}\|\nabla\mathbb{L}^*(\theta_t)\|^2\\
    &\quad+[\frac{\eta_{\theta}}{2}(1+\frac{1}{2}\rho l)+L\eta_{\theta}^2(4\rho^2 l^2+2\rho l+2)]\mathbb{E}\|\nabla\mathbb{L}^*(\theta_t)-\nabla_{\theta}\mathbb{L}(\theta_t,\omega_t)\|^2\\
    &\quad+(5\rho^2l^2+2)\frac{L\eta_{\theta}^2\sigma^2}{2M},
    \end{aligned}
\end{equation*}
where we use the brief notation that $L=l+\frac{l\kappa}{2}$.
\end{lemma}
\begin{proof}
Since $\mathbb{L}^*(\theta)$ is $(l+\frac{l\kappa}{2})$-Lipschitz smooth according to Lemma \ref{le:L*smooth}, we have:
\begin{equation}\label{proofeq:phi_smooth}
    \begin{aligned}
    \mathbb{L}^*(\theta_{t+1})&\leq \mathbb{L}^*(\theta_t)+\langle\nabla\mathbb{L}^*(\theta_t),\theta_{t+1}-\theta_t\rangle+\frac{1}{2}(l+\frac{l\kappa}{2})\|\theta_{t+1}-\theta_t\|^2\\
    &=\mathbb{L}^*(\theta_t)-\eta_{\theta}\langle\nabla\mathbb{L}^*(\theta_t),g_{\theta}(\theta_{t+1/2},\omega_t)\rangle+\frac{1}{2}(l+\frac{l\kappa}{2})\eta_{\theta}^2\|g_{\theta}(\theta_{t+1/2},\omega_t)\|^2.
    \end{aligned}
\end{equation}

Taking expectation conditioned on $(\theta_t,\omega_t)$ and we get:
\begin{align}
    &\mathbb{E}[\mathbb{L}^*(\theta_{t+1})|\theta_t,\omega_t]\leq\\
    &\mathbb{L}^*(\theta_t)-\eta_{\theta}\langle\nabla\mathbb{L}^*(\theta_t),\nabla_{\theta}\mathbb{L}(\theta_{t+1/2},\omega_t)\rangle+\frac{1}{2}(l+\frac{l\kappa}{2})\eta_{\theta}^2\mathbb{E}[\|g_{\theta}(\theta_{t+1/2},\omega_t)\|^2|\theta_t,\omega_t].
\end{align} 

We again take expectation on both side on above inequality so we have:
\begin{equation}\label{proofeq:stophi_smooth}
    \mathbb{E}[\mathbb{L}^*(\theta_{t+1})]\leq\mathbb{E}[\mathbb{L}^*(\theta_t)]-\eta_{\theta}\mathbb{E}\langle\nabla\mathbb{L}^*(\theta_t),\nabla_{\theta}\mathbb{L}(\theta_{t+1/2},\omega_t)\rangle+\frac{1}{2}(l+\frac{l\kappa}{2})\eta_{\theta}^2\mathbb{E}\|g_{\theta}(\theta_{t+1/2},\omega_t)\|^2.
\end{equation}

For the second term, we decompose it as follows:
\begin{equation}\label{proofeq:phi_second}
    \begin{aligned}
    &\quad\; \, \mathbb{E}\langle\nabla\mathbb{L}^*(\theta_t),\nabla_{\theta}\mathbb{L}(\theta_{t+1/2},\omega_t)\rangle\\
    &=\mathbb{E}\langle\nabla\mathbb{L}^*(\theta_t),\nabla_{\theta}\mathbb{L}(\theta_t,\omega_t)+\nabla_{\theta}\mathbb{L}(\theta_{t+1/2},\omega_t)-\nabla_{\theta}\mathbb{L}(\theta_t,\omega_t)\rangle\\
    &\geq \mathbb{E}\langle\nabla\mathbb{L}^*(\theta_t),\nabla_{\theta}\mathbb{L}(\theta_t,\omega_t)\rangle-\mathbb{E}\|\nabla\mathbb{L}^*(\theta_t)\|\|\nabla_{\theta}\mathbb{L}(\theta_{t+1/2},\omega_t)-\nabla_{\theta}\mathbb{L}(\theta_t,\omega_t)\|\\
    &\geq \mathbb{E}\langle\nabla\mathbb{L}^*(\theta_t),\nabla_{\theta}\mathbb{L}(\theta_t,\omega_t)\rangle-\rho l\mathbb{E}\|\nabla\mathbb{L}^*(\theta_t)\|\|g_{\theta}(\theta_t,\omega_y)\|\\
    &\geq\mathbb{E}\langle\nabla\mathbb{L}^*(\theta_t),\nabla\mathbb{L}^*(\theta_t)+\nabla_{\theta}\mathbb{L}(\theta_t,\omega_t)-\nabla\mathbb{L}^*(\theta_t)\rangle\\
    &\quad-\rho l\mathbb{E}\|\nabla\mathbb{L}^*(\theta_t)\|(\|\nabla_{\theta}\mathbb{L}(\theta_t,\omega_t)\|+\|g_{\theta}(\theta_t,\omega_t)-\nabla_{\theta}\mathbb{L}(\theta_t,\omega_t)\|)\\
    &\geq\mathbb{E}\|\nabla\mathbb{L}^*(\theta_t)\|^2-\frac{1}{2}\mathbb{E}\|\nabla\mathbb{L}^*(\theta_t)\|^2-\frac{1}{2}\mathbb{E}\|\nabla_{\theta}\mathbb{L}(\theta_t,\omega_t)-\nabla\mathbb{L}^*(\theta_t)\|^2\\
    &\quad-\rho l\mathbb{E}\|\nabla\mathbb{L}^*(\theta_t)\|\|\nabla_{\theta}\mathbb{L}(\theta_t,\omega_t)\|\\
    &\quad -\frac{1}{2}\rho l\mathbb{E}\|\nabla\mathbb{L}^*(\theta_t)\|^2-\frac{1}{2}\rho l\mathbb{E}\|g_{\theta}(\theta_t,\omega_t)-\nabla_{\theta}\mathbb{L}(\theta_t,\omega_t)\|^2\\
    &\geq\frac{1-\rho l}{2}\mathbb{E}\|\nabla\mathbb{L}^*(\theta_t)\|^2-\frac{1}{2}\mathbb{E}\|\nabla_{\theta}\mathbb{L}(\theta_t,\omega_t)-\nabla\mathbb{L}^*(\theta_t)\|^2\\
    &\quad-\rho l\mathbb{E}\|\nabla\mathbb{L}^*(\theta_t)\|\|\nabla_{\theta}\mathbb{L}(\theta_t,\omega_t)\|-\frac{\rho l\sigma^2}{2M}.
    \end{aligned}
\end{equation}

We continue estimating the last term in above inequality \ref{proofeq:phi_second}
\begin{equation}\label{proofeq:phi_second_last}
    \begin{aligned}
    &\quad\;\,\mathbb{E}\|\nabla\mathbb{L}^*(\theta_t)\|\|\nabla_{\theta}\mathbb{L}(\theta_t,\omega_t)\|\\
    &=\mathbb{E}\|\nabla\mathbb{L}^*(\theta_t)\|\|\nabla_{\theta}\mathbb{L}(\theta_t,\omega_t)-\nabla\mathbb{L}^*(\theta_t)+\nabla\mathbb{L}^*(\theta_t)\|\\
    &\leq\mathbb{E}\|\nabla\mathbb{L}^*(\theta_t)\|^2+\mathbb{E}\|\nabla\mathbb{L}^*(\theta_t)\|\|\nabla_{\theta}\mathbb{L}(\theta_t,\omega_t)-\nabla\mathbb{L}^*(\theta_t)\|\\
    &\overset{(i)}{\leq}\mathbb{E}\|\nabla\mathbb{L}^*(\theta_t)\|^2+\mathbb{E}\|\nabla\mathbb{L}^*(\theta_t)\|^2+\frac{1}{4}\mathbb{E}\|\nabla_\theta\mathbb{L}(\theta_t,\omega_t)-\nabla\mathbb{L}^*(\theta_t)\|^2,
    \end{aligned}
\end{equation}
where the last inequality $(i)$ is due to Young's inequality.

By combining inequality \ref{proofeq:stophi_smooth} with \ref{proofeq:phi_second_last}, we can get:
\begin{equation}\label{proofeq:phi_second_final}
    \begin{aligned}
    &\quad\; \, \mathbb{E}\langle\nabla\mathbb{L}^*(\theta_t),\nabla_{\theta}\mathbb{L}(\theta_{t+1/2},\omega_t)\rangle\\
    &\geq \frac{1-\rho l}{2}\mathbb{E}\|\nabla\mathbb{L}^*(\theta_t)\|^2-\frac{1}{2}\mathbb{E}\|\nabla_{\theta}\mathbb{L}(\theta_t,\omega_t)-\nabla\mathbb{L}^*(\theta_t)\|^2\\
    &\quad-2\rho l\mathbb{E}\|\nabla\mathbb{L}^*(\theta_t)\|^2-\frac{\rho l}{4}\mathbb{E}\|\nabla_{\theta}\mathbb{L}(\theta_t,\omega_t)-\nabla\mathbb{L}^*(\theta_t)\|^2-\frac{\rho l\sigma^2}{2M}\\
    &=\frac{1}{2}(1-5\rho l)\mathbb{E}\|\nabla\mathbb{L}^*(\theta_t)\|^2-\frac{1}{2}(1+\frac{1}{2}\rho l)\mathbb{E}\|\nabla\mathbb{L}^*(\theta_t)-\nabla_{\theta}\mathbb{L}(\theta_t,\omega_t)\|^2-\frac{\rho l\sigma^2}{2M}.
    \end{aligned}
\end{equation}

Finally, we combine inequality \ref{proofeq:stophi_smooth} with Lemma \ref{proofeq:gradientbound} and inequality \ref{proofeq:phi_second_final}:
\begin{equation}
    \begin{aligned}
    &\quad\;\,\mathbb{E}[\mathbb{L}^*(\theta_{t+1})]\\
    &\leq\mathbb{E}[\mathbb{L}^*(\theta_t)]-\frac{\eta_{\theta}}{2}(1-5\rho l)\mathbb{E}\|\nabla\mathbb{L}^*(\theta_t)\|^2+\frac{\eta_{\theta}}{2}(1+\frac{1}{2}\rho l)\mathbb{E}\|\nabla\mathbb{L}^*(\theta_t)-\nabla_{\theta}\mathbb{L}(\theta_t,\omega_t)\|^2\\
    &\quad+\frac{1}{2}(l+\frac{l\kappa}{2})\eta_{\theta}^2((4\rho^2l^2+2\rho l+2)\mathbb{E}\|\nabla_{\theta}\mathbb{L}(\theta_t,\omega_t)\|^2+(5\rho^2l^2+2)\frac{\sigma^2}{M})\\
    &\overset{(i)}{\leq}\mathbb{E}[\mathbb{L}^*(\theta_t)]-\frac{\eta_{\theta}}{2}(1-5\rho l-\eta_{\theta}(2l+l\kappa)(4\rho^2l^2+2\rho l+2))\mathbb{E}\|\nabla\mathbb{L}^*(\theta_t)\|^2\\
    &\quad+[\frac{\eta_{\theta}}{2}(1+\frac{1}{2}\rho l)+\eta_{\theta}^2(l+\frac{l\kappa}{2})(4\rho^2 l^2+2\rho l+2)]\mathbb{E}\|\nabla\mathbb{L}^*(\theta_t)-\nabla_{\theta}\mathbb{L}(\theta_t,\omega_t)\|^2\\
    &\quad+\frac{1}{2}(l+\frac{l\kappa}{2})(5\rho^2l^2+2)\frac{\eta_{\theta}^2\sigma^2}{M},
    \end{aligned}
\end{equation}
where the last inequality $(i)$ uses the Cauchy-Schwarz inequality that $\|\nabla_{\theta}\mathbb{L}(\theta_t,\omega_t)\|^2\leq2\|\nabla\mathbb{L}^*(\theta_t)\|^2+2\|\nabla_{\theta}\mathbb{L}(\theta_t,\omega_t)-\nabla\mathbb{L}^*(\theta_t)\|^2$.
\end{proof}

\subsection{Theorem}
\vspace{-10mm}
\begin{theorem}\label{le:pl:phi}
Under Assumption \ref{as:bv}, \ref{as:smooth}, \ref{as:pl}, and the learning rate satisfy that:
\begin{equation}
    \eta_{\theta}\leq\min\{\frac{1}{128\kappa^2l},\sqrt{\frac{M(\mathbb{E}[\mathbb{L}^*(\theta_0)]-\min_{\theta}\mathbb{E}[\mathbb{L}^*(\theta)])}{132T\kappa^4l\sigma^2}}\},
\end{equation}
$\eta_{\omega}\leq64\kappa^2\eta_{\theta}$ and $\rho\leq\frac{\eta_{\theta}}{2l}$, we have the convergence bound for our problem:
\begin{equation}
    \frac{1}{T}\sum_{t=0}^{T-1}\mathbb{E}\|\nabla\mathbb{L}^*(\theta_t)\|^2\leq320\sqrt{\frac{3\kappa^4l(\mathbb{E}[\mathbb{L}^*(\theta_0)]-\min_{\theta}\mathbb{E}[\mathbb{L}^*(\theta)])\sigma^2}{11MT}}=\mathcal{O}(\frac{\kappa^2}{\sqrt{MT}}).
\end{equation}
\end{theorem}
\begin{proof}
First recall the descending relationship of the function $\mathbb{L}^*$ in Lemma \ref{le:L^*gradient}:
\begin{equation}
    \begin{aligned}
    &\mathbb{E}[\mathbb{L}^*(\theta_{t+1})]\leq\mathbb{E}[\mathbb{L}^*(\theta_t)]-\frac{\eta_{\theta}}{2}(1-5\rho l-2L\eta_{\theta}(4\rho^2l^2+2\rho l+2))\mathbb{E}\|\nabla\mathbb{L}^*(\theta_t)\|^2\\
    &\quad+[\frac{\eta_{\theta}}{2}(1+\frac{1}{2}\rho l)+L\eta_{\theta}^2(4\rho^2 l^2+2\rho l+2)]\mathbb{E}\|\nabla\mathbb{L}^*(\theta_t)-\nabla_{\theta}\mathbb{L}(\theta_t,\omega_t)\|^2\\
    &\quad+(5\rho^2l^2+2)\frac{L\eta_{\theta}^2\sigma^2}{2M}.
    \end{aligned}
\end{equation}

Then, using the smoothness of the variables $\theta$ and $\omega$ respectively, we can get:
\begin{equation*}
    \begin{aligned}
    \mathbb{L}(\theta_{t+1},\omega_t)&\geq \mathbb{L}(\theta_t,\omega_t)+\langle\nabla_{\theta}\mathbb{L}(\theta_t,\omega_t),\theta_{t+1}-\theta_t\rangle-\frac{l}{2}\|\theta_{t+1}-\theta_t\|^2;    \\
    \mathbb{L}(\theta_{t+1},\omega_{t+1})&\geq \mathbb{L}(\theta_{t+1},\omega_t)+\langle\nabla_{\omega}\mathbb{L}(\theta_{t+1},\omega_t),\omega_{t+1}-\omega_t\rangle-\frac{l}{2}\|\omega_{t+1}-\omega_t\|^2.
    \end{aligned}
\end{equation*}

Taking expectation, we can get:
\begin{equation}
    \begin{aligned}
    \mathbb{E}[\mathbb{L}(\theta_{t+1},\omega_t)]&\geq\mathbb{E}[\mathbb{L}(\theta_t,\omega_t)]-\eta_{\theta}\mathbb{E}\langle\nabla_{\theta}\mathbb{L}(\theta_t,\omega_t),\nabla_{\theta}\mathbb{L}(\theta_{t+1/2},\omega_t)\rangle\\
    &\quad-\frac{l\eta_{\theta}^2}{2}\mathbb{E}\|g_{\theta}(\theta_{t+1/2},\omega_t)\|^2\\
    &\geq\mathbb{E}[\mathbb{L}(\theta_t,\omega_t)]-\eta_{\theta}\mathbb{E}\|\nabla_{\theta}\mathbb{L}(\theta_t,\omega_t)\|^2-\frac{\eta_{\theta}}{2}\mathbb{E}\|\nabla_{\theta}\mathbb{L}(\theta_t,\omega_t)\|^2\\&\quad-\frac{\eta_{\theta}}{2}\mathbb{E}\|\nabla_{\theta}\mathbb{L}(\theta_{t+1/2},\omega_t)-\nabla_{\theta}\mathbb{L}(\theta_t,\omega_t)\|^2-\frac{l\eta_{\theta}^2}{2}\mathbb{E}\|g_{\theta}(\theta_{t+1/2},\omega_t)\|^2\\
    &\geq \mathbb{E}[\mathbb{L}(\theta_t,\omega_t)]-\frac{3\eta_{\theta}}{2}\mathbb{E}\|\nabla_{\theta}\mathbb{L}(\theta_t,\omega_t)\|^2-\frac{l^2\rho^2\eta_{\theta}}{2}\mathbb{E}\|g_{\theta}(\theta_t,\omega_t)\|^2\\
    &\quad-\frac{l\eta_{\theta}^2}{2}\mathbb{E}\|g_{\theta}(\theta_{t+1/2},\omega_t)\|^2\\
    &\geq\mathbb{E}[\mathbb{L}(\theta_t,\omega_t)]-(\frac{3\eta_{\theta}}{2}+\frac{l^2\rho^2\eta_{\theta}}{2}+l\eta_{\theta}^2(2\rho^2l^2+\rho l+1))\mathbb{E}\|\nabla_{\theta}\mathbb{L}(\theta_t,\omega_t)\|^2\\&\quad-(\frac{l^2\rho^2\eta_{\theta}}{2}+\frac{l\eta_{\theta}^2}{2}(5\rho^2l^2+2))\frac{\sigma^2}{M};
    \notag
    \end{aligned}
\end{equation}
\begin{equation}
    \begin{aligned}
    \mathbb{E}[\mathbb{L}(\theta_{t+1},\omega_{t+1})]&\geq\mathbb{E}[\mathbb{L}(\theta_{t+1},\omega_t)]+\eta_{\omega}\mathbb{E}\langle\nabla_{\omega}\mathbb{L}(\theta_{t+1},\omega_t),\nabla_{\omega}\mathbb{L}(\theta_t,\omega_t)\rangle-\frac{l\eta_{\omega}^2}{2}\mathbb{E}\|g_{\omega}(\theta_t,\omega_t)\|^2\\
    &\geq\mathbb{E}[\mathbb{L}(\theta_{t+1},\omega_t)]+\frac{\eta_{\omega}}{2}\mathbb{E}\|\nabla_{\omega}\mathbb{L}(\theta_t,\omega_t)\|^2-\frac{\eta_{\omega}}{2}\mathbb{E}\|\nabla_{\omega}\mathbb{L}(\theta_{t+1},\omega_t)\\
    &\quad-\nabla_{\omega}\mathbb{L}(\theta_t,\omega_t)\|^2-\frac{l\eta_{\omega}^2}{2}\mathbb{E}\|g_{\omega}(\theta_t,\omega_t)\|^2\\
    &\geq\mathbb{E}[\mathbb{L}(\theta_{t+1},\omega_t)]+\frac{\eta_{\omega}}{2}\mathbb{E}\|\nabla_{\omega}\mathbb{L}(\theta_t,\omega_t)\|^2-\frac{l\eta_{\theta}^2\eta_{\omega}}{2}\mathbb{E}\|g_{\theta}(\theta_{t+1/2},\omega_t)\|^2\\
    &\quad-\frac{l\eta_{\omega}^2}{2}\mathbb{E}\|g_{\omega}(\theta_t,\omega_t)\|^2\\
    &\geq\mathbb{E}[\mathbb{L}(\theta_{t+1},\omega_t)]+(\frac{\eta_{\omega}}{2}-\frac{l\eta_{\omega}^2}{2})\mathbb{E}\|\nabla_{\omega}\mathbb{L}(\theta_t,\omega_t)\|^2\\
    &\quad-(l\eta_{\theta}^2\eta_{\omega}(2\rho^2l^2+\rho l+1))\mathbb{E}\|\nabla_{\theta}\mathbb{L}(\theta_t,\omega_t)\|^2-(\frac{l\eta_{\omega}^2}{2}+\frac{l\eta_{\theta}^2\eta_{\omega}}{2}(5\rho^2l^2+2))\frac{\sigma^2}{M}.
    \end{aligned}
\end{equation}

Then we construct a potential function in the same way as~\cite{yang2022faster}:
$$V_t=V(\theta_t,\omega_t)=\mathbb{L}^*(\theta_t)+\alpha[\mathbb{L}^*(\theta_t)-\mathbb{L}(\theta_t,\omega_t)],$$
where $\alpha>0$ is a preset parameter. Then we come to evaluate the descending relationship of the potential function $V_t$.

Combining the above inequalities, we get the potential function descending relationship:
\begin{align}
    &\quad \mathbb{E}[V_{t+1}]-\mathbb{E}[V_t]  \notag\\
    &=(1+\alpha)(\mathbb{E}[\mathbb{L}^*(\theta_{t+1})]-\mathbb{E}[\mathbb{L}^*(\theta_t)])-\alpha(\mathbb{E}[\mathbb{L}(\theta_{t+1},\omega_{t+1})]-\mathbb{E}[\mathbb{L}(\theta_t,\omega_t)])  \notag\\
    &\leq(1+\alpha)\{-\frac{\eta_{\theta}}{2}(1-5\rho l-2L\eta_{\theta}(4\rho^2l^2+2\rho l+2))\mathbb{E}\|\nabla\mathbb{L}^*(\theta_t)\|^2  \notag\\
    &\quad+[\frac{\eta_{\theta}}{2}(1+\frac{1}{2}\rho l)+L\eta_{\theta}^2(4\rho^2 l^2+2\rho l+2)]\mathbb{E}\|\nabla\mathbb{L}^*(\theta_t)-\nabla_{\theta}\mathbb{L}(\theta_t,\omega_t)\|^2\\
    &\quad+(5\rho^2l^2+2)\frac{L\eta_{\theta}^2\sigma^2}{2M}\}  \notag-\alpha\{-(\frac{3\eta_{\theta}}{2}+\frac{l^2\rho^2\eta_{\theta}}{2}+l\eta_{\theta}^2(2\rho^2l^2+\rho l+1))\mathbb{E}\|\nabla_{\theta}\mathbb{L}(\theta_t,\omega_t)\|^2\\
    &\quad-(\frac{l^2\rho^2\eta_{\theta}}{2}+\frac{l\eta_{\theta}^2}{2}(5\rho^2l^2+2))\frac{\sigma^2}{M}+(\frac{\eta_{\omega}}{2}-\frac{l\eta_{\omega}^2}{2})\mathbb{E}\|\nabla_{\omega}\mathbb{L}(\theta_t,\omega_t)\|^2\notag\\
    &\quad-(l\eta_{\theta}^2\eta_{\omega}(2\rho^2l^2+\rho l+1))\mathbb{E}\|\nabla_{\theta}\mathbb{L}(\theta_t,\omega_t)\|^2-(\frac{l\eta_{\omega}^2}{2}+\frac{l\eta_{\theta}^2\eta_{\omega}}{2}(5\rho^2l^2+2))\frac{\sigma^2}{M}\}  \tag{\stepcounter{equation}\theequation}\\
    &=-\frac{\eta_{\theta}}{2}(1+\alpha)(1-5\rho l-2L\eta_{\theta}(4\rho^2l^2+2\rho l+2))\mathbb{E}\|\nabla\mathbb{L}^*(\theta_t)\|^2  \notag\\
    &\quad+(1+\alpha)(\frac{\eta_{\theta}}{2}(1+\frac{1}{2}\rho l)+L\eta_{\theta}^2(4\rho^2 l^2+2\rho l+2))\mathbb{E}\|\nabla\mathbb{L}^*(\theta_t)-\nabla_{\theta}\mathbb{L}(\theta_t,\omega_t)\|^2  \notag\\
    &\quad+\alpha[(\frac{3\eta_{\theta}}{2}+\frac{l^2\rho^2\eta_{\theta}}{2}+l\eta_{\theta}^2(2\rho^2l^2+\rho l+1))+l\eta_{\theta}^2\eta_{\omega}(2\rho^2l^2+\rho l+1)]\mathbb{E}\|\nabla_{\theta}\mathbb{L}(\theta_t,\omega_t)\|^2  \notag\\
    &\quad-\alpha(\frac{\eta_{\omega}}{2}-\frac{l\eta_{\omega}^2}{2})\mathbb{E}\|\nabla_{\omega}\mathbb{L}(\theta_t,\omega_t)\|^2  \notag\\
    &\quad+[(1+\alpha)(5\rho^2l^2+2)\frac{L\eta_{\theta}^2}{2}+\alpha(\frac{l^2\rho^2\eta_{\theta}}{2}+\frac{l\eta_{\theta}^2}{2}(5\rho^2l^2+2))\notag\\
    &\quad+\alpha(\frac{l\eta_{\omega}^2}{2}+\frac{l\eta_{\theta}^2\eta_{\omega}}{2}(5\rho^2l^2+2))]\frac{\sigma^2}{M}\\
    &\leq-\{\frac{\eta_{\theta}}{2}(1+\alpha)(1-5\rho l-2L\eta_{\theta}(4\rho^2l^2+2\rho l+2))\notag\\
    &\quad-2\alpha[(\frac{3\eta_{\theta}}{2}+\frac{l^2\rho^2\eta_{\theta}}{2}+l\eta_{\theta}^2(2\rho^2l^2+\rho l+1))+l\eta_{\theta}^2\eta_{\omega}(2\rho^2l^2+\rho l+1)]\}\notag\\&\quad\mathbb{E}\|\nabla\mathbb{L}^*(\theta_t)\|^2+\{(1+\alpha)(\frac{\eta_{\theta}}{2}(1+\frac{1}{2}\rho l)+L\eta_{\theta}^2(4\rho^2 l^2+2\rho l+2))\notag\\
    &\quad+2\alpha[(\frac{3\eta_{\theta}}{2}+\frac{l^2\rho^2\eta_{\theta}}{2}+l\eta_{\theta}^2(2\rho^2l^2+\rho l+1))+l\eta_{\theta}^2\eta_{\omega}(2\rho^2l^2+\rho l+1)]\}  \notag\\
    &\quad\mathbb{E}\|\nabla\mathbb{L}^*(\theta_t)-\nabla_{\theta}\mathbb{L}(\theta_t,\omega_t)\|^2-\alpha(\frac{\eta_{\omega}}{2}-\frac{l\eta_{\omega}^2}{2})\mathbb{E}\|\nabla_{\omega}\mathbb{L}(\theta_t,\omega_t)\|^2+[(1+\alpha)(5\rho^2l^2+2)\frac{L\eta_{\theta}^2}{2}  \notag\\
    &\quad+\alpha(\frac{l^2\rho^2\eta_{\theta}}{2}+\frac{l\eta_{\theta}^2}{2}(5\rho^2l^2+2))+\alpha(\frac{l\eta_{\omega}^2}{2}+\frac{l\eta_{\theta}^2\eta_{\omega}}{2}(5\rho^2l^2+2))]\frac{\sigma^2}{M}.
\end{align}

Since we have the following property according to Lemma \ref{le:L*smooth} and the PL condition \ref{as:pl}:
$$\|\nabla\mathbb{L}^*(\theta_t)-\nabla_{\theta}f(\theta_t,\omega_t)\|\leq l\|\omega^*(\theta_t)-\omega_t\|\leq\kappa\|\nabla_{\omega}f(\theta_t,\omega_t)\|.$$

So we can further the above inequality as follows:
\begin{equation}
    \begin{aligned}
    &\quad\mathbb{E}[V_{t+1}]-\mathbb{E}[V_t]\\
    &\leq-\{\frac{\eta_{\theta}}{2}(1+\alpha)(1-5\rho l-2L\eta_{\theta}(4\rho^2l^2+2\rho l+2))\\
    &\quad-2\alpha[(\frac{3\eta_{\theta}}{2}+\frac{l^2\rho^2\eta_{\theta}}{2}+l\eta_{\theta}^2(2\rho^2l^2+\rho l+1))+l\eta_{\theta}^2\eta_{\omega}(2\rho^2l^2+\rho l+1)]\}\mathbb{E}\|\nabla\mathbb{L}^*(\theta_t)\|^2\\
    &\quad-\{\alpha(\frac{\eta_{\omega}}{2}-\frac{l\eta_{\omega}^2}{2})-\kappa^2[(1+\alpha)(\frac{\eta_{\theta}}{2}(1+\frac{1}{2}\rho l)+L\eta_{\theta}^2(4\rho^2 l^2+2\rho l+2))\\&\quad+2\alpha[(\frac{3\eta_{\theta}}{2}+\frac{l^2\rho^2\eta_{\theta}}{2}+l\eta_{\theta}^2(2\rho^2l^2+\rho l+1))+l\eta_{\theta}^2\eta_{\omega}(2\rho^2l^2+\rho l+1)]]\}\mathbb{E}\|\nabla_{\omega}\mathbb{L}(\theta_t,\omega_t)\|^2\\
    &\quad+[(1+\alpha)(5\rho^2l^2+2)\frac{L\eta_{\theta}^2}{2}\notag\\
    &\quad+\alpha(\frac{l^2\rho^2\eta_{\theta}}{2}+\frac{l\eta_{\theta}^2}{2}(5\rho^2l^2+2))+\alpha(\frac{l\eta_{\omega}^2}{2}+\frac{l\eta_{\theta}^2\eta_{\omega}}{2}(5\rho^2l^2+2))]\frac{\sigma^2}{M}.
    \end{aligned}
\end{equation}

Then we require the parameters satisfy: $\alpha=\frac{1}{16}$, $\rho l \leq\frac{1}{16}$, $\eta_{\theta}(2\rho l+1)^2\kappa l\leq\frac{1}{64}$, $\kappa^2\eta_{\theta}l\leq\frac{1}{128}$, $\rho\leq\frac{\eta_{\theta}}{2l}$ and $\eta_{\omega}\leq64\kappa^2\eta_{\theta}$.

So the inequality can be further simplified as:
\begin{equation}
    \begin{aligned}
    &\quad\mathbb{E}[V_{t+1}]-\mathbb{E}[V_t]\\
    &\leq-\frac{11}{80}\eta_{\theta}\mathbb{E}\|\nabla\mathbb{L}^*(\theta_t)\|^2-\frac{41}{32}\eta_{\theta}\kappa^2\mathbb{E}\|\nabla_{\omega}f(\theta_t,\omega_t)\|^2+129\kappa^4l\eta_{\theta}^2\frac{\sigma^2}{M}.
    \end{aligned}
\end{equation}

Telescoping the above inequality we can get:
\begin{equation}\label{proofeq:bound_1}
    \begin{aligned}
    \frac{1}{T}\sum_{t=0}^{T-1}\mathbb{E}\|\nabla\mathbb{L}^*(\theta_t)\|^2\leq\frac{80}{11\eta_{\theta}T}(\mathbb{E}[V_0]-\mathbb{E}[V_T])+960\kappa^4l\eta_{\theta}\frac{\sigma^2}{M}.
    \end{aligned}
\end{equation}

Further, we can evaluate the first term that:
\begin{equation*}
    \begin{aligned}
    \mathbb{E}[V_0]-\mathbb{E}[V_T]&\leq\mathbb{E}[V_0]-\min_{\theta,\omega}\mathbb{E}[V(\theta,\omega)]\\
    &\leq \mathbb{E}[\mathbb{L}^*(\theta_0)]-\min_{\theta}\mathbb{E}[\mathbb{L}^*(\theta)]+\frac{1}{16}(\mathbb{E}[\mathbb{L}^*(\theta_0)]-\mathbb{E}[\mathbb{L}(\theta_0,\omega_0)])\\
    &=\mathbb{E}[\mathbb{L}^*(\theta_0)]-\min_{\theta}\mathbb{E}[\mathbb{L}^*(\theta)]+\frac{1}{16}\Delta_0,
    \end{aligned}
\end{equation*}
where we denote the initial error as: $\Delta_0=\mathbb{E}[\mathbb{L}^*(\theta_0)]-\mathbb{E}[\mathbb{L}(\theta_0,\omega_0)].$

Therefore, the inequality \ref{proofeq:bound_1} can be further evaluated as:
\begin{equation}\label{proofeq:bound_2}
    \begin{aligned}
    \frac{1}{T}\sum_{t=0}^{T-1}\mathbb{E}\|\nabla\mathbb{L}^*(\theta_t)\|^2\leq\frac{80}{11\eta_{\theta}T}(\mathbb{E}[\mathbb{L}^*(\theta_0)]-\min_{\theta}\mathbb{E}[\mathbb{L}^*(\theta)])+\frac{5}{11\eta_{\theta}T}\Delta_0+960\kappa^4l\eta_{\theta}\frac{\sigma^2}{M},
    \end{aligned}
\end{equation}
when we select $\eta_{\theta}=\sqrt{\frac{M(\mathbb{E}[\mathbb{L}^*(\theta_0)]-\min_{\theta}\mathbb{E}[\mathbb{L}^*(\theta)])}{132T\kappa^4l\sigma^2}}$, and samples can be minibatch, the convergence can be bounded by:
\begin{equation}
    \frac{1}{T}\sum_{t=0}^{T-1}\mathbb{E}\|\nabla\mathbb{L}^*(\theta_t)\|^2\leq320\sqrt{\frac{3\kappa^4l(\mathbb{E}[\mathbb{L}^*(\theta_0)]-\min_{\theta}\mathbb{E}[\mathbb{L}^*(\theta)])\sigma^2}{11MT}}=\mathcal{O}(\frac{\kappa^2}{\sqrt{MT}}).
\end{equation}

\end{proof}

\section{Proof for EVIL}
This section provides the proof of Proposition~\ref{EVIL:theorem}. First, by applying the selected invariant parameters, we can have the label prediction $\hat{Y}^e=\sgn(\theta_{inv}^\top Z^e)=\sgn((\mathbf{m}\circ\theta)^\top Z^e)$ where $\sgn(\cdot)$ returns the sign of input value. Further, we have:
\vspace*{-5mm}
\begin{align}
	(\mathbf{m}\circ\theta)^\top Z^e &=\frac{1}{\sqrt{M}}\mathbf{m}^\top Z^e\\
	\notag
	&= \sqrt{M}\frac{1}{M}\left[\mathbf{m}_{i\in\left[0, M_{inv}\right]}, \mathbf{m}_{i\in\left[M_{inv}, M\right]}\right]^\top \left[Z^e_{inv}, Z^e_{var}\right]\\
	&= \sqrt{M}\left[\frac{1}{M_{inv}}\mathbf{m}_{i\in\left[0, M_{inv}\right]}^\top Z^e_{inv}+\frac{1}{M_{var}}\mathbf{m}_{i\in\left[M_{inv}, M\right]}^\top Z^e_{var}\right].\label{proofeq:appendix_matrix_multiply}
\end{align}
Then the error produced by the current sparse network for a given environment is:
\begin{align}
	\Err^e&=\frac{1}{2}\mathbb{E}_{(X^e, Y^e)\sim e}\left[1-Y^e\hat{Y}^e\right]=\frac{1}{2}\left[1-\mathbb{E}^e\left[Y^e\hat{Y}^e\right]\right].
\end{align}
Here we simplify the expectation on samples from distribution $e$ as $\mathbb{E}^e$.
\begin{align}
	\notag
	\mathbb{E}^e\left[Y^e\hat{Y}^e\right]&=\mathbb{E}^e\left[\sgn(\frac{1}{M}\mathbf{m}^\top Z^e)Y^e\right]\\
	&=\sum_{y\in\left\{-1, 1\right\}}\mathbb{P}\left[Y^e=y\right]\mathbb{E}^e\left[\sgn(\frac{1}{M}\mathbf{m}^\top Z^e)\mid Y^e=y\right]y\label{proofeq:appendix_expected_prediction}.
\end{align}
\begin{align}
	\notag
	\mathbb{E}^e\!\!\left[\sgn(\frac{1}{M}\mathbf{m}^\top Z^e)\mid Y^e=1\right]&\!\!=\!\!\mathbb{P}\left[\frac{1}{M}\mathbf{m}^\top Z^e>0\mid Y^e=1\right]\!\!-\!\!\mathbb{P}\left[\frac{1}{M}\mathbf{m}^\top Z^e\le0\mid Y^e=1\right]\\
	&=2\mathbb{P}\left[\frac{1}{M}\mathbf{m}^\top Z^e>0\mid Y^e=1\right]-1\label{proofeq:appendix_expected_prediction_one}.
\end{align}
Similar to Zhang et al.~\cite{zhang2021can}, we observe that:
\begin{equation}
    \mathbb{P}\left[\frac{1}{M}\mathbf{m}^\top Z^e\le0\mid Y^e=1\right]=\mathbb{P}\left[\frac{1}{M}\mathbf{m}^\top Z^e>0\mid Y^e=-1\right],
\end{equation}
and plug Equation~\eqref{proofeq:appendix_expected_prediction_one} into Equation~\eqref{proofeq:appendix_expected_prediction} to get:
\begin{align}
	\Err^e=\mathbb{P}\left[\frac{1}{M}\mathbf{m}^\top Z^e\le0\mid Y^e=1\right]\label{proofeq:appendix_expected_error}.
\end{align}
Similar to Equation~\eqref{proofeq:appendix_matrix_multiply}, we further decompose Equation~\eqref{proofeq:appendix_expected_error}:
\begin{align}
	\notag
	\Err^e&=\mathbb{P}\left[\frac{1}{M_{inv}}\mathbf{m}_{i\in\left[0, M_{inv}\right]}^\top Z^e_{inv}+\frac{1}{M_{var}}\mathbf{m}_{i\in\left[M_{inv}, M\right]}^\top Z^e_{var}\le0\mid Y^e=1\right]\\
	&\le\mathbb{P}\left[\frac{1}{M_{inv}}\mathbf{m}_{i\in\left[0, M_{inv}\right]}^\top Z^e_{inv}\le0\mid Y^e=1\right]+\mathbb{P}\left[\frac{1}{M_{var}}\mathbf{m}_{i\in\left[M_{inv}, M\right]}^\top Z^e_{var}\le0\mid Y^e=1\right].
\end{align}
Since all elements in $Z^e_{inv}$ are equal to $Y^e$, the first term equals $0$, thus the equality also holds. Then, we have:
\begin{align}
	\Err^e&=\mathbb{P}\left[\frac{1}{M_{var}}\mathbf{m}_{i\in\left[M_{inv}, M\right]}^\top Z^e_{var}\le0\mid Y^e=1\right].
\end{align}
Here we assume each element of $Z^e$ and $\mathbf{m}$ are independent with each other, we can have $\Err^e=\mathbb{P}\left[\mathbf{m}_i Z^e_{var, i}\le0\mid Y^e=1\right]$. It is obvious that there is only one situation when the error occurs, i.e., $\mathbf{m}_i=1$ and $Z^e_{var, i}\le0$. Only in this scenario, the sparse training strategy would update the mask to value $0$. Therefore, $\mathbb{P}\left[\mathbf{m}_i=0\right]=1-p^e$. For other cases where $\mathbb{P}\left[Z^e_{var,i}>0\right]=p^e$, the value of each $\mathbf{m}_i$ is randomly initialized and stays intact since there is no error, hence, $\mathbb{P}\left[\mathbf{m}_i=0\right]=\mathbb{P}\left[\mathbf{m}_i=1\right]=\frac{p^e}{2}$. So, for $i\in\left[M_{inv},M\right]$, $\mathbb{P}\left[\mathbf{m}_i=0\right]=1-\frac{p^e}{2}$ and $\mathbb{P}\left[\mathbf{m}_i=1\right]=\frac{p^e}{2}$.

To further bound the error, we denote $\overline{\left[\mathbf{m}Z^e\right]}_{var}=\frac{1}{M_{var}}\mathbf{m}_{i\in\left[M_{inv}, M\right]}^\top Z^e_{var}$, and have:
\begin{align}
	\notag
	\Err^e&=\mathbb{P}\left[\overline{\left[\mathbf{m}Z^e\right]}_{var}\le0\mid Y^e=1\right]\\
	\notag
	&=\mathbb{P}\left[\overline{\left[\mathbf{m}Z^e\right]}_{var}-\mathbb{E}\left[\overline{\left[\mathbf{m}Z^e\right]}_{var}\right]\le-\mathbb{E}\left[\overline{\left[\mathbf{m}Z^e\right]}_{var}\right]\mid Y^e=1\right]\\
	&\le\mathbb{P}\left[\lvert\overline{\left[\mathbf{m}Z^e\right]}_{var}-\mathbb{E}\left[\overline{\left[\mathbf{m}Z^e\right]}_{var}\right]\rvert\ge\mathbb{E}\left[\overline{\left[\mathbf{m}Z^e\right]}_{var}\right]\mid Y^e=1\right].
\end{align}
\begin{align}
	\notag
	\mathbb{E}\left[\overline{\left[\mathbf{m}Z^e\right]}_{var}\right]&=\mathbb{E}\left[\sgn(\frac{1}{M_{var}}\mathbf{m}_{i\in\left[M_{inv}, M\right]}^\top Z^e_{var})\mid Y^e=1\right]\\
	\notag
	&=2\mathbb{P}\left[\frac{1}{M_{var}}\mathbf{m}_{i\in\left[M_{inv}, M\right]}^\top Z^e_{var}>0\mid Y^e=1\right]-1\\
	&=(p^e)^2-1.
\end{align}
Therefore, $\Err^e\le2e^{-2((p^e)^2-1)^2M_{var}}=\mathcal{O}(e^{-(p^e)^4})$. In contrast to the idealized bound that achieves $0$ error in Zhang et al.~\cite{zhang2021can}, when $\theta$ is initialized with the unit norm and given a small $p^e$, the error could be considerably large. This is because the error produced by the variant parameters is largely decided by the probability $p^e$, which could further affect the pruning process. Based on such an intuition, we propose to enhance the pruning strategy by adding an additional regularization that leverages domain knowledge.

Specifically, the regularization considers the errors from distinguishing different distributions using variant parameters:
\begin{align}
	\notag
	&\Err^d=\frac{1}{2}\mathbb{E}_{(X, Y)\sim \mathcal{E}}\left[1-D\hat{D}\right]=\frac{1}{2}\left[1-\mathbb{E}\left[D\hat{D}\right]\right], \\
	&\text{where}\ \hat{D}=\sgn(\theta_{var}^\top Z^e)=\sgn(((1-\mathbf{m})\circ\theta)^\top Z^e).
\end{align}
Similar to Equation~\eqref{proofeq:appendix_expected_error}, we can have:
\begin{align}
	\notag
	\Err^d&=\mathbb{P}\left[\frac{1}{M}(1-\mathbf{m})^\top Z^e\le0\mid D=1\right]\label{proofeq:appendix_expected_distribution_error}\\
	&=\mathbb{P}\left[\frac{1}{M_{inv}}(1-\mathbf{m})_{i\in\left[0, M_{inv}\right]}^\top Z^e_{inv}+\frac{1}{M_{var}}(1-\mathbf{m})_{i\in\left[M_{inv}, M\right]}^\top Z^e_{var}\le0\mid D=1\right].
\end{align}
Since all elements in $Z^e_{var}$ equal to $D$, we can have:
\begin{align}
	\Err^d&=\mathbb{P}\left[\frac{1}{M_{inv}}(1-\mathbf{m})_{i\in\left[0, M_{inv}\right]}^\top Z^e_{inv}\le0\mid D=1\right].
\end{align}
Therefore, for $i\in\left[0,M_{inv}\right]$, $\mathbb{P}\left[\mathbf{m}_i=1\right]=1-\frac{q^e}{2}$ and $\mathbb{P}\left[\mathbf{m}_i=0\right]=\frac{q^e}{2}$. As a result, the regularization can complement the mask by finding the invariant parameters with at least probability $1-\frac{q^e}{2}$. Moreover, based on a given sparsity ratio $R=\frac{M_{var}}{M}$, i.e., only $M_{inv}$ elements from $Z^e$ would be selected by $\mathbf{m}$, the erroneous mask that produces classification error can be further constrained from being too much. Particularly, from $\mathbb{P}\left[\mathbf{m}_i=1\right]=1-\frac{q^e}{2},  i\in\left[0,M_{inv}\right]$, we can have $\mathbb{P}\left[\mathbf{m}_i=1\right]=\frac{M_{inv}-(1-\frac{q^e}{2})M_{inv}}{M_{var}}=\frac{q^eM_{inv}}{2M_{var}},  i\in\left[M_{inv}, M\right]$ instead of $\frac{p^e}{2}$ which is calculated based on random initialization. Hence, we can again bound the classification error as:
\begin{align}
	\notag
	\mathbb{E}\left[\overline{\left[\mathbf{m}Z^e\right]}_{var}\right]&=2\mathbb{P}\left[\frac{1}{M_{var}}\mathbf{m}_{i\in\left[M_{inv}, M\right]}^\top Z^e_{var}>0\mid Y^e=1\right]-1\\
	&=p^e\frac{q^eM_{inv}}{M_{var}}-1,
\end{align}
\begin{align}
	\notag
	\Err&^e\le\mathbb{P}\left[\lvert\overline{\left[\mathbf{m}Z^e\right]}_{var}-\mathbb{E}\left[\overline{\left[\mathbf{m}Z^e\right]}_{var}\right]\rvert\ge\mathbb{E}\left[\overline{\left[\mathbf{m}Z^e\right]}_{var}\right]\mid Y^e=1\right]\\
	&\le2e^{-2(\frac{q^eM_{inv}}{M_{var}}p^e-1)^2M_{var}}=\mathcal{O}(e^{-(p^e)^2})
\end{align}

\section{Proof for MVT}
This section provides the proof of Theorem~\ref{MVT:theorem}.

\textbf{Pretraining distribution formulation.} We based on the in-context learning framework proposed by Xie et al.~\cite{xie2021explanation}. In this framework, a latent concept $\phi$ from a concept space $\Phi$ defines a pretraining distribution $p$ over prompt tokens $o$ observed from a vocabulary space $O$. To generate the desired content, we first sample a concept from a prior $p(\phi)$ and then sample the tokens conditioned on the concept. We denote the total length of the pretraining examples is $T$:
\begin{equation}
    p(o_1, \ldots, o_T)=\int_{\phi\in\Phi} p(o_1, \ldots, o_T|\phi)p(\phi)d\phi.
\end{equation}
The conditional probability $p(o_1, \ldots, o_T|\phi)$ is defined by a Hidden Markov Model (HMM). Based on the concept $\phi$, a transition matrix of the HMM hidden states $h_1, \ldots, h_T$ from a hidden state space $\mathcal{H}$ can be found.

\textbf{Prompt distribution formulation.} During the in-context learning process, we sample a prompt from a new distribution $p_{prompt}$, which contains $n$ independent exemplars and 1 query example, which are all conditioned on a shared prompt concept $\phi^*$. The goal is to predict the next token based on the exemplars and the query example. Specifically, the $i$-th exemplar $O_i$ consists of a tokenized image feature $x_i=O_i[1:k_x]$, a text description to claim the class of the image $y_i=O_i[k_x:k_x+k_y]$, and a binary prediction to judge whether the claim of the image is ``True'' or ``False'', which is denoted by $z_i=O_i[k_x+k_y:k_x+k_y+1]$ at the end of each exemplar. The generating process of the $i$-th exemplar is as follows: 
\begin{enumerate}
    \item First generate a start hidden state $h_i^{start}$ from prompt start distribution $p_{prompt}$;
    \item Given $h_i^{start}$, generate the exemplar sequence $O_i=[x_i, y_i, z_i]$ from $p(O_i|h_i^{start}, \phi^*)$, the generate exemplars are conditioned on a given prompt concept $\phi^*$.
\end{enumerate}

The query example $Q$ is sampled similarly without the binary prediction $z_q$, \textit{i.e.} $Q=[x_q, y_q]$. Between each exemplar and the query example, there is a special delimiter token $o^{delim}$ denoting the end of each exemplar sequence. The prompt can be formulated as follows:
\begin{equation}
    [S_n, Q] = [x_1, y_1, z_1, o^{delim}, x_2, y_2, z_2, o^{delim}, \ldots, x_n, y_n, z_n, o^{delim}, x_q, y_q]\sim p_{prompt},
\end{equation}
where $S_n$ denotes the $n$ independent exemplars for in-context demonstration.

\textbf{Denoising In-context learning task.} In our denoising in-context learning, the output prediction $z$ for each image and text pair $[x, y]$ is sampled based on the prompt distribution $p_{prompt}(z|x, y)$:
\begin{equation}
    z_q\sim p_{prompt}(z|x, y)=\mathbb{E}_{h_q^{start}\sim p_{prompt}(h_q^{start}|Q)}[p(z|Q, h_q^{start}, \phi^*)],
\end{equation}
where $h_q^{start}$ denotes the hidden state corresponding to the first token of $Q$, \textit{i.e.,} the first token of $x_q$.

Our goal is to analyze the in-context predictor $f_n(x_q, y_q) = \arg\max_{z}p(z|S_n, x_q, y_q)$ which outputs the most likely prediction over the pretraining distribution $p$ conditioned on the exemplars $S_n$ sampled from the prompt distribution $p_{prompt}$. We assume the in-context predictor is trained by $0-1$ error with $n$ training examples:
\begin{equation}
    \mathcal{L}_{0-1}(f_n)=\mathbb{E}_{x_q, y_q}\sim p_{prompt}[\mathbf{1}[f_n(x_q)\neq y_q]],
\end{equation}
and same for the prediction $z_q$:
\begin{equation}
    \mathcal{L}_{0-1}(f_n)=\mathbb{E}_{x_q, y_q, z_q}\sim p_{prompt}[\mathbf{1}[f_n(x_q, y_q)\neq z_q]].
\end{equation}

One major difference of our denoising in-context learning strategy is that we not only use positive exemplars that show exact image-text match, \textit{i.e.,} $(x, y)\sim p(x, y)=p_{prompt}(x, y)$, we also have negative exemplars where image and text are not corresponding to each other. To construct such a prompt, we have to first select the ideal $y$, based on the matching result of $x$ and $y$, we can further obtain the prediction $z$. Therefore, in the following theoretical proof, we propose to conduct two-step analyses on $y$ and $z$, respectively. 

\subsection{Assumptions}
Our theoretical framework is built upon Xie et al.~\cite{xie2021explanation}, whose assumptions are also applied to our analysis.

\begin{assumption}[Delimiter hidden states]
\label{ass:delimiterstates}
Let the delimiter hidden states $\mathcal{D}$ be a subset of $\mathcal{H}$. For any $h^{delim}\in \mathcal{D}$ and $\phi \in \Phi$, $p(o^{delim} \vert h^{delim}, \phi)=1$ and for any $h\notin \mathcal{D}$, $p(o^{delim} \vert h, \phi)=0$.
\end{assumption}

\begin{assumption}[Bound on delimiter transitions]
\label{ass:delimiterbound}
For any delimiter state $h^{delim} \in \mathcal{D}$ and any hidden state $h \in \mathcal{H}$, the probability of transitioning to a delimiter hidden state under $\phi$ is upper bounded $p(h^{delim} \vert h, \phi) < c_2$ for any $\phi \in \Phi \setminus \{\phi^*\}$, and is lower bounded $p(h^{delim} \vert h, \phi^*) > c_1 > 0$ for $\phi^*$.
Additionally, the start hidden state distribution for delimiter hidden states is bounded as $p(h^{delim} \vert \phi) \in [c_3, c_4]$.
\end{assumption}

\begin{assumption}[Distribution shift from prompt start distribution]
\label{ass:promptstartshift}
We assume that the prompt start distribution $p_{prompt}$ is close in TV distance to all hidden transition distributions (under $\phi^*$) starting from a delimiter hidden state:
\begin{equation}
    \max_{h^{delim} \in \mathcal{D}} TV(p_{prompt}(h) \| p(h \vert h^{delim}, \phi^*)) < \tau / 4.
\end{equation}
Here, $\tau = p_{prompt}(y_{max} \vert Q) - \max_{y \neq y_{max}}p_{prompt}(y \vert Q)$ is the margin between the most likely label $y_{max} = \arg\max_{y} p_{prompt}(y \vert Q)$ and the second most likely label.
\end{assumption}

\begin{assumption}[Well-specification]
\label{ass:wellspecified}
The prompt concept $\phi^*$ is in $\Phi$.
\end{assumption}

\begin{assumption}[Regularity]
\label{ass:regularity}
The pretraining distribution $p$ satisfies: (1) Lower bound on transition probability for the prompt concept $\phi^*$: for any pair of hidden states $h, h' \in \mathcal{H}$, $p(h \vert h', \phi^*) > c_5 > 0$. 
(2) Start hidden state is lower bounded: for any $h \in \mathcal{H}$, $p(h \vert \phi^*)\geq c_8 > 0$.
(3) All tokens can be emitted: for every symbol $o$, there is some hidden state $h\in \mathcal{H}$ such that $p(o \vert h, \phi^*) > c_6 > 0$,
(4) The prior $p(\phi)$ has support over the entire concept family $\Phi$ and is bounded above everywhere.
\end{assumption}

Except from the above five adapted assumptions from Xie et al.~\cite{xie2021explanation}, we have an another mild assumption:

\begin{assumption}[Distribution consistency]
\label{ass:consistency}
The pretraining distribution $p$ and prompt distribution $p_{prompt}$ satisfy $\forall (x_q, y_q)\sim p_{prompt}, p(x_q, y_q)=p_{prompt}(x_q, y_q)$.
\end{assumption}
This assumption indicates that the chosen prompt distribution is a sub-distribution of the pretraining distribution and the joint distribution of $x_q$ and $y_q$ is consistent across $p$ and $p_{prompt}$. This assumption avoids the situations where there are concept shifts between $p$ and $p_{prompt}$, \textit{i.e.}, all $y\sim p_{prompt}$ are known in $p$ and can find an exact match for each $x_q$ in $p$.

\subsection{Theoretical Proof}
We first show that given a query image $x_q$, when conditioned on a concept $\phi^*$ and prompt $S_n$, the most probable text output token for $y_q$ is the same as the class in the prompt distribution $p_{prompt}$ with maximum probability. Then, we show that: in our denoising in-context learning, when achieving the most likely prediction $z$ output by the MLLM predictor, the class text $y_q$ in the pretraining distribution $p$ is the same as the one found by the prompt distribution $p_{prompt}$, which is the exact match for the give image $x_q$.

Before we start analyzing the binary prediction $z$, we first investigate the most probable class $y$ given prompt and query image $\arg\max_y p(y\vert S_n, x_q)$. 
\begin{theorem}
    Assume that the above assumptions hold, if for all $\phi\in\Phi$, $\phi\neq\phi^*$, the concept $\phi^*$ satisfies the distinguishability condition: $\sum_{j=1}^k KL_j(\phi^*\|\phi) > \epsilon_{start}^{\phi} + \epsilon_{delim}^{\phi}$, then as $n\rightarrow\infty$, the prediction $y$ according to the pretraining distribution is
    \begin{equation}
        \arg\max_y p(y\vert S_n, x_q, \phi^*) \rightarrow \arg\max_y p_{prompt}(y\vert x_q).
    \end{equation}
    Thus, the in-context predictor $f_n$ achieves the optimal $0-1$ risk: $\lim_{n\rightarrow\infty}\mathcal{L}_{0-1}(f_n)=\inf_f\mathcal{L}_{0-1}(f)$.
    \label{theo:optimal_y_supp}
\end{theorem}
The detailed proof of this theorem is similar to Xie et al.~\cite{xie2021explanation}, please refer to the Section D of the original paper.

Under this assumption, the in-context predictor is guaranteed to have the highest probability of generating the class description $y$ that exactly matches the query image $x_q$. In another way, when $x_q$ does not belong to $y$, the probability $p(y\vert S_n, x_q)$ is less than the optimal value.

\begin{lemma}
    Under the same condiction of Theorem~\ref{theo:optimal_y_supp}, the prediction $z$ according to the pretraining distribution is
    \begin{equation}
        \arg\max_z p(z\vert S_n, x_q, y_q, \phi^*) \rightarrow \arg\max_z p_{prompt}(z\vert x_q, y_q).
    \end{equation}
    \label{prooflemm:optimal_z_supp}
\end{lemma}

Lemma~\ref{prooflemm:optimal_z_supp} can be easily derived based on Theorem~\ref{theo:optimal_y_supp} by considering $y$ as a fixed prompt.

\begin{theorem}
    Assume that the above assumptions hold, as $n\rightarrow\infty$, when achieving the largest prediction probability of $z$ given prompt under concept $\phi^*$, the corresponding class description $y$ follows the same $y$ obtained from the prompt distribution:
    \begin{equation}
        \arg\max_y p(z\vert S_n, x_q, y, \phi^*) \rightarrow \arg\max_y p_{prompt}(z\vert x_q, y)
    \end{equation}
    \label{theo:dicl_supp}
\end{theorem}
\begin{proof}
    Since we already have Theorem~\ref{theo:optimal_y_supp}, if we can prove that
    \begin{equation}
        \arg\max_y p(y\vert S_n, x_q, \phi^*)=\arg\max_y p(z\vert S_n, x_q, y, \phi^*),
    \end{equation}
    \begin{equation}
        \arg\max_y p_{prompt}(y\vert x_q)=\arg\max_y p_{prompt}(z\vert x_q, y),
    \end{equation}
    then we can complete the justification.
    \begin{align}
        p(z\vert S_n, x_q, y, \phi^*)=\sum_{h^{start}_q\in \mathcal{H}}p(z\vert h^{start}_q)p(h^{start}_q\vert S_n, x_q, y, \phi^*).
    \end{align}
    By expanding the last term, we have:
    \begin{align}
        p(h^{start}_q\vert S_n, x_q, y, \phi^*)=&\frac{p(x_q, y\vert h^{start}_q, S_n, \phi^*)p(h^{start}_q)}{p(x_q, y)}\\
        \propto& \frac{p(x_q, y\vert h^{start}_q, S_n, \phi^*)}{p(x_q, y)}
    \end{align}
    where $p(h^{start}_q)$ is considered as a constant. Moreover, based on Assumption~\ref{ass:consistency}, the joint distribution $p(x_q, y)$ is predefined by the pretraining distribution, which does not affect the marginal distribution of $z$, thus we can have
    \begin{align}
        \frac{p(x_q, y\vert h^{start}_q, S_n, \phi^*)}{p(x_q, y)}&=\frac{p(y\vert S_n, x_q, h^{start}_q, \phi^*)p(x_q\vert h^{start}_q)}{p(x_q, y)}\\
        &\propto p(y\vert S_n, x_q, h^{start}_q, \phi^*)p(x_q\vert h^{start}_q).
    \end{align}
    Since the change of $y$ does not affect the quantity of $p(z\vert h^{start}_q)$, therefore, applying argmax on both sides of the equation holds for finding the optimal $y$:
    \begin{align}
        \arg\max_y p(z\vert S_n, x_q, y, \phi^*)&=\arg\max_y \sum_{h^{start}_q\in \mathcal{H}}p(z\vert h^{start}_q)p(y\vert S_n, x_q, h^{start}_q, \phi^*)\\
        &=\arg\max_y p(y\vert S_n, x_q, h^{start}_q, \phi^*).
    \end{align}
    Similarly, we have
    \begin{align}
        p_{prompt}(z\vert x_q, y)&= \sum_{h^{start}_q\in \mathcal{H}}p(z\vert h^{start}_q, \phi^*)p_{prompt}(h^{start}_q\vert x_q, y),\\
        p_{prompt}(h^{start}_q\vert x_q, y)&= \frac{p_{prompt}(x_q, y\vert h^{start}_q)p_{prompt}(h^{start}_q)}{p_{prompt}(x_q, y)}\\
        &\propto p_{prompt}(x_q, y\vert h^{start}_q)\\
        &\propto p_{prompt}(y\vert x_q, h^{start}_q)p_{prompt}(x_q\vert h^{start}_q),\\
        \arg\max_y p_{prompt}(z\vert x_q, y) &= \arg\max_y \sum_{h^{start}_q\in \mathcal{H}}p_{prompt}(z\vert h^{start}_q, \phi^*)p_{prompt}(y\vert x_q, h^{start}_q)\\
        &=\arg\max_y p_{prompt}(y\vert x_q, h^{start}_q),
    \end{align}
    where the change of $y$ still does not affect the quantity of $p_{prompt}(z\vert h^{start}_q, \phi^*)$. Since
    \begin{align}
        p(y\vert S_n, x_q, \phi^*)&=\sum_{h^{start}_q\in \mathcal{H}}p(y\vert h^{start}_q, S_n, x_q, \phi^*)p(h^{start}_q\vert S_n, x_q, \phi^*),\\
        p_{prompt}(y\vert x_q)&=\sum_{h^{start}_q\in \mathcal{H}}p_{prompt}(y\vert h^{start}_q, x_q)p_{prompt}(h^{start}_q\vert x_q),
    \end{align}
    it is easy to find that
    \begin{align}
        \arg\max_y p_{prompt}(y\vert x_q, h^{start}_q) = \arg\max_y p_{prompt}(y\vert x_q),\\
        \arg\max_y p(y\vert S_n, x_q, h^{start}_q, \phi^*) = \arg\max_y p(y\vert S_n, x_q, \phi^*).
    \end{align}
    Thus, we have that as $n\rightarrow\infty$,
    \begin{equation}
        \arg\max_y p(z\vert S_n, x_q, y, \phi^*) \rightarrow \arg\max_y p_{prompt}(z\vert x_q, y).
    \end{equation}
\end{proof}

Lemma~\ref{prooflemm:optimal_z_supp} and Theorem~\ref{theo:dicl_supp} together show that when given a query image $x_q$, if the chosen query class description $y_q$ is the true class of $x_q$, then under the given assumptions, the binary prediction $z$ for judging the correctness of the image-text pair would be the maximum value compared to all other class descriptions $y\neq y_q, y\in \mathcal{Y}$. Therefore, we can justify that using an in-context predictor can help identify the true class label of a given image.

\section{Proof for OOM}
This section provides the proof of Theorem~\ref{OOM:theorem}.

\subsection{Lower Bound of Our VIB framework}
\label{sec:app_lower_bound}
Recall that we have the following factorization:
\begin{equation}
    p(X^{\mathrm{I}}, X^{\mathrm{O}}, V, Y) = p(V, Y|X^{\mathrm{O}}, X^{\mathrm{I}})p(X^{\mathrm{O}}|X^{\mathrm{I}})P(X^{\mathrm{I}}),
\end{equation}
with Markov chains $V\leftrightarrow X^{\mathrm{I}}\leftrightarrow X^{\mathrm{O}}$ and $X^{\mathrm{I}}\leftrightarrow Y \not\leftrightarrow X^{\mathrm{O}}$. Our goal is to maximize the information redundancy~\cite{liang2023multimodal, williams2010nonnegative}:
\begin{equation}
    \max I(X^{\mathrm{O}}; X^{\mathrm{I}}; Y)=I(X^{\mathrm{O}}; X^{\mathrm{I}}) - I(X^{\mathrm{O}}; X^{\mathrm{I}}|Y),
\end{equation}
where the first term is lower-bounded by:
\begin{align}
    \!\!\!I(X^{\mathrm{O}}; X^{\mathrm{I}})&\!\ge\!\int dx^{\mathrm{O}}\!dx^{\mathrm{I}}\!dv p(x^{\mathrm{I}})p(x^{\mathrm{O}}|x^{\mathrm{I}})p(v|x^{\mathrm{I}})\log q(x^{\mathrm{O}}|v)p(v|x^{\mathrm{I}}),
    \label{proofeq:app_first_lowerbound}
\end{align}

Then, we consider the second term $I(X^{\mathrm{O}}; X^{\mathrm{I}}|Y)$:
\begin{align}
    &I(X^{\mathrm{O}}; X^{\mathrm{I}}|Y)\!=\!\int\! dx^{\mathrm{O}}\!dx^{\mathrm{I}}\!dy p(x^{\mathrm{O}}, x^{\mathrm{I}}, y)\log\frac{p(x^{\mathrm{O}}, x^{\mathrm{I}}|y)}{p(x^{\mathrm{O}}|y)p(x^{\mathrm{I}}|y)}\\
    &\!=\!\int\! dx^{\mathrm{O}}\!dx^{\mathrm{I}}\!dy p(x^{\mathrm{O}}, x^{\mathrm{I}}, y)\log\frac{p(x^{\mathrm{O}}, x^{\mathrm{I}}, y)}{p(y|x^{\mathrm{O}})}\!-\!H(Y)\!+\!H(Y|X^{\mathrm{I}})\!+\!H(X^{\mathrm{O}})\!+\!H(X^{\mathrm{I}}).
\end{align}
Note that we use the factorization $p(x^{\mathrm{O}}\!, x^{\mathrm{I}}\!, y)=p(y|x^{\mathrm{I}})p(x^{\mathrm{O}}|x^{\mathrm{I}})p(x^{\mathrm{I}})$, and further ignore the entropy terms\footnote{We focus on the optimization of $p(Y|X^{\mathrm{O}})$, and $p(Y|X^{\mathrm{I}})$ is given and frozen in our setting.}, then we have:
\begin{align}
    \!\!\!I(X^{\mathrm{O}}\!; X^{\mathrm{I}}|Y)\!&\le\!\int\! dx^{\mathrm{O}}\!dx^{\mathrm{I}}\!dy p(y|x^{\mathrm{I}})p(x^{\mathrm{O}}|x^{\mathrm{I}})p(x^{\mathrm{I}})\log p(y|x^{\mathrm{I}})p(x^{\mathrm{O}}|x^{\mathrm{I}})p(x^{\mathrm{I}})\! -\! \log h(y|x^{\mathrm{O}}),
    \label{proofeq:app_second_upperbound}
\end{align}
which is based on the positivity of KL divergence between our classifier $h(y|x^{\mathrm{O}})$ and $p(y|x^{\mathrm{O}})$.

To this end, we can lower-bound our objective by combining Equations~\eqref{proofeq:app_first_lowerbound} and~\eqref{proofeq:app_second_upperbound}:
\begin{align}
    &I(X^{\mathrm{O}}; X^{\mathrm{I}}; Y)\ge\int\! dx^{\mathrm{O}}\!dx^{\mathrm{I}}\!dv p(x^{\mathrm{I}})p(x^{\mathrm{O}}|x^{\mathrm{I}})p(v|x^{\mathrm{I}})\log q(x^{\mathrm{O}}|v)p(v|x^{\mathrm{I}})\\
    &-\int\! dx^{\mathrm{O}}\!dx^{\mathrm{I}}\!dy p(y|x^{\mathrm{I}})p(x^{\mathrm{O}}|x^{\mathrm{I}})p(x^{\mathrm{I}})\log p(y|x^{\mathrm{I}})p(x^{\mathrm{O}}|x^{\mathrm{I}})p(x^{\mathrm{I}})\! +\! \log h(y|x^{\mathrm{O}})=\mathcal{L}_{con}.
    \label{proofeq:app_final_lowerbound}
\end{align}

\subsection{Proof of Theorem}
\label{sec:app_proof}
Now we start the proof of Theorem~\ref{OOM:theorem}.
\begin{proof}
\begin{assumption}[Relaxed triangle inequality]
    For the distance function $d:\mathcal{Y}\times\mathcal{Y}\rightarrow\mathbb{R^+}$, there exists $c_d\ge1$ such that $\forall \hat{y}_1, \hat{y}_2, \hat{y}_3\in \mathcal{\hat{Y}} d(\hat{y}_1, \hat{y}_2)\le c_d(d(\hat{y}_1, \hat{y}_3) + d(\hat{y}_2, \hat{y}_3))$.
\end{assumption}
\begin{assumption}[Inverse Lipschitz condition]
    For the function $d$, it holds that $\forall h$,
    \begin{equation}
        \mathbb{E}[d(h(x_1, x_2), h^*(x_1, x_2))]\le |\mathcal{L}(h)-\mathcal{L}(h^*)|,
    \end{equation}
    where $h^*$ is the Bayes optimal classifier on both $x_1$ and $x_2$; and
    \begin{equation}
        \mathbb{E}[d(h(x), h^*(x))]\le |\mathcal{L}(h)-\mathcal{L}(h^*)|,
    \end{equation}
    where $h^*$ is the Bayes optimal classifier on $x$. 
\end{assumption}
\begin{assumption}[Classifier optimality]
    For any classifiers $h$ in comparison to the Bayes' optimal classifier $h^*$, there exists constants $\epsilon>0$ such that $|\mathcal{L}(h)-\mathcal{L}(h^*)|^2\le\epsilon$.
\end{assumption}

To bridge $h_1^*$ and $h_2^*$, we use $h_{1,2}^*$ and $h^*$ to denote the Bayes' optimal classifier on both IM data and all data, respectively. Then, we capture the relationship between the uniqueness of OOM data given both IM data and the difference in their Bayes’ optimal prediction errors:
\begin{align}
    |\mathcal{L}(h_{1,2}^*) - \mathcal{L}(h^*)|^2 &= |\mathbb{E}_{X}\mathbb{E}_{Y|X_1^{\mathrm{I}}, X_2^{\mathrm{I}}, X^{\mathrm{O}}}\ell(h^*(x_1^{\mathrm{I}}, x_2^{\mathrm{I}}, x^{\mathrm{O}}), y)\notag\\
    &\quad- \mathbb{E}_{X_1^{\mathrm{I}}, X_2^{\mathrm{I}}}\mathbb{E}_{Y|X_1^{\mathrm{I}}, X_2^{\mathrm{I}}}\ell(h_1^*(x_1^{\mathrm{I}}, X_2^{\mathrm{I}}), y)|^2\\
    &\le |\mathbb{E}_{Y|X_1^{\mathrm{I}}, X_2^{\mathrm{I}}, X^{\mathrm{O}}}\ell(h^*(x_1^{\mathrm{I}}, x_2^{\mathrm{I}}, x^{\mathrm{O}}), y)-\mathbb{E}_{Y|X_1^{\mathrm{I}}, X_2^{\mathrm{I}}}\ell(h_1^*(x_1^{\mathrm{I}}, X_2^{\mathrm{I}}), y)|^2\\
    &\le \text{KL}(p(y|x_1^{\mathrm{I}}, x_2^{\mathrm{I}}, x^{\mathrm{O}})\parallel p(y|x_1^{\mathrm{I}}, x_2^{\mathrm{I}}))\\
    &\le \mathbb{E}_X\text{KL}(p(y|x_1^{\mathrm{I}}, x_2^{\mathrm{I}}, x^{\mathrm{O}})\parallel p(y|x_1^{\mathrm{I}}, x_2^{\mathrm{I}}))\\
    &=I(X^{\mathrm{O}}, Y|X_1^{\mathrm{I}}, X_2^{\mathrm{I}}).
\end{align}

Then, we first capture the redundancy between one IM and OOM data given another IM data:
\begin{align}
    |\mathcal{L}(h_1^*) - \mathcal{L}(h^*)|^2 &= |\mathbb{E}_{X}\mathbb{E}_{Y|X_1^{\mathrm{I}}, X_2^{\mathrm{I}}, X^{\mathrm{O}}}\ell(h^*(x_1^{\mathrm{I}}, x_2^{\mathrm{I}}, x^{\mathrm{O}}), y)- \mathbb{E}_{X_1^{\mathrm{I}}}\mathbb{E}_{Y|X_1^{\mathrm{I}}}\ell(h_1^*(x_1^{\mathrm{I}}), y)|^2\\
    &\le |\mathbb{E}_{Y|X_1^{\mathrm{I}}, X_2^{\mathrm{I}}, X^{\mathrm{O}}}\ell(h^*(x_1^{\mathrm{I}}, x_2^{\mathrm{I}}, x^{\mathrm{O}}), y)-\mathbb{E}_{Y|X_1^{\mathrm{I}}}\ell(h_1^*(x_1^{\mathrm{I}}), y)|^2\\
    &\le \text{KL}(p(y|x_1^{\mathrm{I}}, x_2^{\mathrm{I}}, x^{\mathrm{O}})\parallel p(y|x_1^{\mathrm{I}}))\\
    &\le \mathbb{E}_X\text{KL}(p(y|x_1^{\mathrm{I}}, x_2^{\mathrm{I}}, x^{\mathrm{O}})\parallel p(y|x_1^{\mathrm{I}}))\\
    &=I(X^{\mathrm{O}}, X_2^{\mathrm{I}}, Y|X_1^{\mathrm{I}}).
\end{align}

Further leveraging triangle inequality through the Bayes' optimal classifier $h^*$ and the inverse Lipschitz condition, we have:
\begin{align}
    \mathbb{E}_{p(x_1^{\mathrm{I}}, x_2^{\mathrm{I}}, x^{\mathrm{O}})}[d(h_1^*, h_{1,2}^*)]&\le \mathbb{E}_{p(x_1^{\mathrm{I}}, x_2^{\mathrm{I}}, x^{\mathrm{O}})}[d(h_1^*, h^*)]+\mathbb{E}_{p(x_1^{\mathrm{I}}, x_2^{\mathrm{I}}, x^{\mathrm{O}})}[d(h^*, h_{1,2}^*)]\\
    &\le |\mathcal{L}(h_1^*) - \mathcal{L}(h^*)|^2 + |\mathcal{L}(h^*)-\mathcal{L}(h_{1,2}^*)|^2 \\
    &\le I(X^{\mathrm{O}}, X_2^{\mathrm{I}}, Y|X_1^{\mathrm{I}})+I(X^{\mathrm{O}}, Y|X_1^{\mathrm{I}}, X_2^{\mathrm{I}}).
    \label{proofeq:first_disagreement}
\end{align}
Symmetrically, we can have $|\mathcal{L}(h_2^*) - \mathcal{L}(h^*)|^2\le I(X^{\mathrm{O}}, X_1^{\mathrm{I}}, Y|X_2^{\mathrm{I}})$ and further obtain:
\begin{equation}
    \mathbb{E}_{p(x_2^{\mathrm{I}}, x_2^{\mathrm{I}}, x^{\mathrm{O}})}[d(h_2^*, h_{1,2}^*)]\le I(X^{\mathrm{O}}, X_1^{\mathrm{I}}, Y|X_2^{\mathrm{I}})+I(X^{\mathrm{O}}, Y|X_1^{\mathrm{I}}, X_2^{\mathrm{I}}).
\end{equation}
Then combining with Equation~\eqref{proofeq:first_disagreement}:
\begin{equation}
    \mathbb{E}_{p(x_1^{\mathrm{I}}, x_2^{\mathrm{I}})}[d(h_1^*, h_2^*)]\le I(X^{\mathrm{O}}, X_2^{\mathrm{I}}, Y|X_1^{\mathrm{I}})+I(X^{\mathrm{O}}, X_1^{\mathrm{I}}, Y|X_2^{\mathrm{I}})+2I(X^{\mathrm{O}}, Y|X_1^{\mathrm{I}}, X_2^{\mathrm{I}})
\end{equation}
Finally, based on the decomposition of the task-related mutual information of $X^{\mathrm{O}}$:
\begin{equation}
    I(X^{\mathrm{O}}, Y)=I(X^{\mathrm{O}}, X_2^{\mathrm{I}}, Y|X_1^{\mathrm{I}})+I(X^{\mathrm{O}}, X_1^{\mathrm{I}}, Y|X_2^{\mathrm{I}})+I(X^{\mathrm{O}}, Y|X_1^{\mathrm{I}}, X_2^{\mathrm{I}})+I(X^{\mathrm{O}}, X_1^{\mathrm{I}}, X_2^{\mathrm{I}}, Y),
\end{equation}
as shown in Figure~\ref{OOMfig:MI_decomposition}, we can achieve:
\begin{equation}
    \alpha(h_1^*, h_2^*):=\mathbb{E}_{p(x_1^{\mathrm{I}}, x_2^{\mathrm{I}})}[d(h_1^*, h_2^*)]\le I(X^{\mathrm{O}}, Y)-I(X^{\mathrm{O}}, X_1^{\mathrm{I}}, X_2^{\mathrm{I}}, Y)+I(X^{\mathrm{O}}, Y|X_1^{\mathrm{I}}, X_2^{\mathrm{I}}),
    \label{proofeq:final_disagreement}
\end{equation}
\end{proof}

\chapter{Conclusion\label{cha:conclusion}}

In this thesis, we provide a systematic study on Trustworthy Machine Learning under Distribution Shifts by exploring capability and responsibility under various practical applications.

We first investigate perturbation shifts in Chapters~\ref{cha:HOOD} and~\ref{cha:SharpDRO}, where both the type and strength of distribution shift are studied. Specifically, different type refers to benigh OOD data with changed style and malign OOD data with unseen content; and strengths denotes the intensity of corruption applyed to the OOD data. To solve these problems, we proposed a causal framework to understand the data generation process and a robust optimization strategy to effectively enhance the generalization performance against different types of shifts with varied strengths, demonstrating both explainability, robustness, and adaptability for OOD data.

Then, we study domain shifts in Chapter~\ref{cha:SharpDRO}, Chapter~\ref{cha:EVIL}, and Chapter~\ref{cha:MVT}. Particularly, we develop multiple optimization strategies to handle OOD data from various domains, namely, worst-case optimization, sparse training, and machine-supervised training. For all these novel techniques, we aim to identify the critical knowledge, i.e., the invariant features in the data. In this way, our approaches can effectively generalize, meanwhile demonstrating great adaptability across domains.

Finally, we look into a novel distribution shift, modality shift, and study it in Chapters~\ref{cha:MVT} and \ref{cha:OOM}. We introduce this problem by revealing the modality gap between LLMs and vision models, which can be addressed by in-context learning to provide learnable signals across modalities. Further, we extend this problem to uncommon modalities and propose OOM generalization to uncover the hidden knowledge in novel modalities. We provide an information theoretic analysis to extract sharable knowledge, which enables generalize across modalities effectively.

This thesis inspires several extendable studies in the future. First of all, how to achieve general intelligence with alignment with human values is one of the most critical questions to be answered. We are in a pivotal period where there is a rise in general intelligence. When AI is taking over various tasks from different dimensions, i.e., visual tasks, language tasks, and physical tasks. It is urgent the develop a systematic, trustworthy protocol to uniformly regulate every possible dimension. Further, we observe various complexities of different distribution shifts; existing AI models lack universal capability, hindering their general practice. For example, LLMs are still suffering from slight perturbation or adversarial attacks. Therefore, future AIs are required to possess reliable capabilities across various shifts. Finally, AIs are driven by data. However, for some tasks, e.g., modality shift, rare modalities significantly lack data to generalize. Thus, it is critical to develop self-reflective and trustworthy AIs that can discover rules and patterns to explore an unknown knowledge scope.


\printbibliography
\end{document}